\newcounter{treecount}
\newcounter{branchcount}
\newsavebox{\parentbox}
\newsavebox{\treebox}
\newsavebox{\treeboxone}
\newsavebox{\treeboxtwo}
\newsavebox{\treeboxthree}
\newsavebox{\treeboxfour}
\newsavebox{\treeboxfive}
\newsavebox{\treeboxsix}
\newsavebox{\treeboxseven}
\newsavebox{\treeboxeight}
\newsavebox{\treeboxnine}
\newsavebox{\treeboxten}
\newsavebox{\treeboxeleven}
\newsavebox{\treeboxtwelve}
\newsavebox{\treeboxthirteen}
\newsavebox{\treeboxfourteen}
\newsavebox{\treeboxfifteen}
\newsavebox{\treeboxsixteen}
\newsavebox{\treeboxseventeen}
\newsavebox{\treeboxeighteen}
\newsavebox{\treeboxnineteen}
\newsavebox{\treeboxtwenty}
\newlength{\treeoffsetone}
\newlength{\treeoffsettwo}
\newlength{\treeoffsetthree}
\newlength{\treeoffsetfour}
\newlength{\treeoffsetfive}
\newlength{\treeoffsetsix}
\newlength{\treeoffsetseven}
\newlength{\treeoffseteight}
\newlength{\treeoffsetnine}
\newlength{\treeoffsetten}
\newlength{\treeoffseteleven}
\newlength{\treeoffsettwelve}
\newlength{\treeoffsetthirteen}
\newlength{\treeoffsetfourteen}
\newlength{\treeoffsetfifteen}
\newlength{\treeoffsetsixteen}
\newlength{\treeoffsetseventeen}
\newlength{\treeoffseteighteen}
\newlength{\treeoffsetnineteen}
\newlength{\treeoffsettwenty}

\newlength{\treeshiftone}
\newlength{\treeshifttwo}
\newlength{\treeshiftthree}
\newlength{\treeshiftfour}
\newlength{\treeshiftfive}
\newlength{\treeshiftsix}
\newlength{\treeshiftseven}
\newlength{\treeshifteight}
\newlength{\treeshiftnine}
\newlength{\treeshiftten}
\newlength{\treeshifteleven}
\newlength{\treeshifttwelve}
\newlength{\treeshiftthirteen}
\newlength{\treeshiftfourteen}
\newlength{\treeshiftfifteen}
\newlength{\treeshiftsixteen}
\newlength{\treeshiftseventeen}
\newlength{\treeshifteighteen}
\newlength{\treeshiftnineteen}
\newlength{\treeshifttwenty}
\newlength{\treewidthone}
\newlength{\treewidthtwo}
\newlength{\treewidththree}
\newlength{\treewidthfour}
\newlength{\treewidthfive}
\newlength{\treewidthsix}
\newlength{\treewidthseven}
\newlength{\treewidtheight}
\newlength{\treewidthnine}
\newlength{\treewidthten}
\newlength{\treewidtheleven}
\newlength{\treewidthtwelve}
\newlength{\treewidththirteen}
\newlength{\treewidthfourteen}
\newlength{\treewidthfifteen}
\newlength{\treewidthsixteen}
\newlength{\treewidthseventeen}
\newlength{\treewidtheighteen}
\newlength{\treewidthnineteen}
\newlength{\treewidthtwenty}
\newlength{\daughteroffsetone}
\newlength{\daughteroffsettwo}
\newlength{\daughteroffsetthree}
\newlength{\daughteroffsetfour}
\newlength{\branchwidthone}
\newlength{\branchwidthtwo}
\newlength{\branchwidththree}
\newlength{\branchwidthfour}
\newlength{\parentoffset}
\newlength{\treeoffset}
\newlength{\daughteroffset}
\newlength{\branchwidth}
\newlength{\parentwidth}
\newlength{\treewidth}
\newcommand{\ontop}[1]{\begin{tabular}{c}#1\end{tabular}}
\newcommand{\poptree}{%
\ifnum\value{treecount}=0\typeout{QobiTeX warning---Tree stack underflow}\fi%
\addtocounter{treecount}{-1}%
\setlength{\treeoffsettwo}{\treeoffsetthree}%
\setlength{\treeoffsetthree}{\treeoffsetfour}%
\setlength{\treeoffsetfour}{\treeoffsetfive}%
\setlength{\treeoffsetfive}{\treeoffsetsix}%
\setlength{\treeoffsetsix}{\treeoffsetseven}%
\setlength{\treeoffsetseven}{\treeoffseteight}%
\setlength{\treeoffseteight}{\treeoffsetnine}%
\setlength{\treeoffsetnine}{\treeoffsetten}%
\setlength{\treeoffsetten}{\treeoffseteleven}%
\setlength{\treeoffseteleven}{\treeoffsettwelve}%
\setlength{\treeoffsettwelve}{\treeoffsetthirteen}%
\setlength{\treeoffsetthirteen}{\treeoffsetfourteen}%
\setlength{\treeoffsetfourteen}{\treeoffsetfifteen}%
\setlength{\treeoffsetfifteen}{\treeoffsetsixteen}%
\setlength{\treeoffsetsixteen}{\treeoffsetseventeen}%
\setlength{\treeoffsetseventeen}{\treeoffseteighteen}%
\setlength{\treeoffseteighteen}{\treeoffsetnineteen}%
\setlength{\treeoffsetnineteen}{\treeoffsettwenty}%
\setlength{\treeshifttwo}{\treeshiftthree}%
\setlength{\treeshiftthree}{\treeshiftfour}%
\setlength{\treeshiftfour}{\treeshiftfive}%
\setlength{\treeshiftfive}{\treeshiftsix}%
\setlength{\treeshiftsix}{\treeshiftseven}%
\setlength{\treeshiftseven}{\treeshifteight}%
\setlength{\treeshifteight}{\treeshiftnine}%
\setlength{\treeshiftnine}{\treeshiftten}%
\setlength{\treeshiftten}{\treeshifteleven}%
\setlength{\treeshifteleven}{\treeshifttwelve}%
\setlength{\treeshifttwelve}{\treeshiftthirteen}%
\setlength{\treeshiftthirteen}{\treeshiftfourteen}%
\setlength{\treeshiftfourteen}{\treeshiftfifteen}%
\setlength{\treeshiftfifteen}{\treeshiftsixteen}%
\setlength{\treeshiftsixteen}{\treeshiftseventeen}%
\setlength{\treeshiftseventeen}{\treeshifteighteen}%
\setlength{\treeshifteighteen}{\treeshiftnineteen}%
\setlength{\treeshiftnineteen}{\treeshifttwenty}%
\setlength{\treewidthtwo}{\treewidththree}%
\setlength{\treewidththree}{\treewidthfour}%
\setlength{\treewidthfour}{\treewidthfive}%
\setlength{\treewidthfive}{\treewidthsix}%
\setlength{\treewidthsix}{\treewidthseven}%
\setlength{\treewidthseven}{\treewidtheight}%
\setlength{\treewidtheight}{\treewidthnine}%
\setlength{\treewidthnine}{\treewidthten}%
\setlength{\treewidthten}{\treewidtheleven}%
\setlength{\treewidtheleven}{\treewidthtwelve}%
\setlength{\treewidthtwelve}{\treewidththirteen}%
\setlength{\treewidththirteen}{\treewidthfourteen}%
\setlength{\treewidthfourteen}{\treewidthfifteen}%
\setlength{\treewidthfifteen}{\treewidthsixteen}%
\setlength{\treewidthsixteen}{\treewidthseventeen}%
\setlength{\treewidthseventeen}{\treewidtheighteen}%
\setlength{\treewidtheighteen}{\treewidthnineteen}%
\setlength{\treewidthnineteen}{\treewidthtwenty}%
\sbox{\treeboxtwo}{\usebox{\treeboxthree}}%
\sbox{\treeboxthree}{\usebox{\treeboxfour}}%
\sbox{\treeboxfour}{\usebox{\treeboxfive}}%
\sbox{\treeboxfive}{\usebox{\treeboxsix}}%
\sbox{\treeboxsix}{\usebox{\treeboxseven}}%
\sbox{\treeboxseven}{\usebox{\treeboxeight}}%
\sbox{\treeboxeight}{\usebox{\treeboxnine}}%
\sbox{\treeboxnine}{\usebox{\treeboxten}}%
\sbox{\treeboxten}{\usebox{\treeboxeleven}}%
\sbox{\treeboxeleven}{\usebox{\treeboxtwelve}}%
\sbox{\treeboxtwelve}{\usebox{\treeboxthirteen}}%
\sbox{\treeboxthirteen}{\usebox{\treeboxfourteen}}%
\sbox{\treeboxfourteen}{\usebox{\treeboxfifteen}}%
\sbox{\treeboxfifteen}{\usebox{\treeboxsixteen}}%
\sbox{\treeboxsixteen}{\usebox{\treeboxseventeen}}%
\sbox{\treeboxseventeen}{\usebox{\treeboxeighteen}}%
\sbox{\treeboxeighteen}{\usebox{\treeboxnineteen}}%
\sbox{\treeboxnineteen}{\usebox{\treeboxtwenty}}}
\newcommand{\leaf}[1]{%
\ifnum\value{treecount}=20\typeout{QobiTeX warning---Tree stack overflow}\fi%
\addtocounter{treecount}{1}%
\sbox{\treeboxtwenty}{\usebox{\treeboxnineteen}}%
\sbox{\treeboxnineteen}{\usebox{\treeboxeighteen}}%
\sbox{\treeboxeighteen}{\usebox{\treeboxseventeen}}%
\sbox{\treeboxseventeen}{\usebox{\treeboxsixteen}}%
\sbox{\treeboxsixteen}{\usebox{\treeboxfifteen}}%
\sbox{\treeboxfifteen}{\usebox{\treeboxfourteen}}%
\sbox{\treeboxfourteen}{\usebox{\treeboxthirteen}}%
\sbox{\treeboxthirteen}{\usebox{\treeboxtwelve}}%
\sbox{\treeboxtwelve}{\usebox{\treeboxeleven}}%
\sbox{\treeboxeleven}{\usebox{\treeboxten}}%
\sbox{\treeboxten}{\usebox{\treeboxnine}}%
\sbox{\treeboxnine}{\usebox{\treeboxeight}}%
\sbox{\treeboxeight}{\usebox{\treeboxseven}}%
\sbox{\treeboxseven}{\usebox{\treeboxsix}}%
\sbox{\treeboxsix}{\usebox{\treeboxfive}}%
\sbox{\treeboxfive}{\usebox{\treeboxfour}}%
\sbox{\treeboxfour}{\usebox{\treeboxthree}}%
\sbox{\treeboxthree}{\usebox{\treeboxtwo}}%
\sbox{\treeboxtwo}{\usebox{\treeboxone}}%
\sbox{\treeboxone}{\ontop{#1}}%
\sbox{\treeboxone}{\raisebox{-\ht\treeboxone}{\usebox{\treeboxone}}}%
\setlength{\treeoffsettwenty}{\treeoffsetnineteen}%
\setlength{\treeoffsetnineteen}{\treeoffseteighteen}%
\setlength{\treeoffseteighteen}{\treeoffsetseventeen}%
\setlength{\treeoffsetseventeen}{\treeoffsetsixteen}%
\setlength{\treeoffsetsixteen}{\treeoffsetfifteen}%
\setlength{\treeoffsetfifteen}{\treeoffsetfourteen}%
\setlength{\treeoffsetfourteen}{\treeoffsetthirteen}%
\setlength{\treeoffsetthirteen}{\treeoffsettwelve}%
\setlength{\treeoffsettwelve}{\treeoffseteleven}%
\setlength{\treeoffseteleven}{\treeoffsetten}%
\setlength{\treeoffsetten}{\treeoffsetnine}%
\setlength{\treeoffsetnine}{\treeoffseteight}%
\setlength{\treeoffseteight}{\treeoffsetseven}%
\setlength{\treeoffsetseven}{\treeoffsetsix}%
\setlength{\treeoffsetsix}{\treeoffsetfive}%
\setlength{\treeoffsetfive}{\treeoffsetfour}%
\setlength{\treeoffsetfour}{\treeoffsetthree}%
\setlength{\treeoffsetthree}{\treeoffsettwo}%
\setlength{\treeoffsettwo}{\treeoffsetone}%
\setlength{\treeoffsetone}{0.5\wd\treeboxone}%
\setlength{\treeshifttwenty}{\treeshiftnineteen}%
\setlength{\treeshiftnineteen}{\treeshifteighteen}%
\setlength{\treeshifteighteen}{\treeshiftseventeen}%
\setlength{\treeshiftseventeen}{\treeshiftsixteen}%
\setlength{\treeshiftsixteen}{\treeshiftfifteen}%
\setlength{\treeshiftfifteen}{\treeshiftfourteen}%
\setlength{\treeshiftfourteen}{\treeshiftthirteen}%
\setlength{\treeshiftthirteen}{\treeshifttwelve}%
\setlength{\treeshifttwelve}{\treeshifteleven}%
\setlength{\treeshifteleven}{\treeshiftten}%
\setlength{\treeshiftten}{\treeshiftnine}%
\setlength{\treeshiftnine}{\treeshifteight}%
\setlength{\treeshifteight}{\treeshiftseven}%
\setlength{\treeshiftseven}{\treeshiftsix}%
\setlength{\treeshiftsix}{\treeshiftfive}%
\setlength{\treeshiftfive}{\treeshiftfour}%
\setlength{\treeshiftfour}{\treeshiftthree}%
\setlength{\treeshiftthree}{\treeshifttwo}%
\setlength{\treeshifttwo}{\treeshiftone}%
\setlength{\treeshiftone}{0pt}%
\setlength{\treewidthtwenty}{\treewidthnineteen}%
\setlength{\treewidthnineteen}{\treewidtheighteen}%
\setlength{\treewidtheighteen}{\treewidthseventeen}%
\setlength{\treewidthseventeen}{\treewidthsixteen}%
\setlength{\treewidthsixteen}{\treewidthfifteen}%
\setlength{\treewidthfifteen}{\treewidthfourteen}%
\setlength{\treewidthfourteen}{\treewidththirteen}%
\setlength{\treewidththirteen}{\treewidthtwelve}%
\setlength{\treewidthtwelve}{\treewidtheleven}%
\setlength{\treewidtheleven}{\treewidthten}%
\setlength{\treewidthten}{\treewidthnine}%
\setlength{\treewidthnine}{\treewidtheight}%
\setlength{\treewidtheight}{\treewidthseven}%
\setlength{\treewidthseven}{\treewidthsix}%
\setlength{\treewidthsix}{\treewidthfive}%
\setlength{\treewidthfive}{\treewidthfour}%
\setlength{\treewidthfour}{\treewidththree}%
\setlength{\treewidththree}{\treewidthtwo}%
\setlength{\treewidthtwo}{\treewidthone}%
\setlength{\treewidthone}{\wd\treeboxone}}
\newcommand{\branch}[2]{%
\setcounter{branchcount}{#1}%
\ifnum\value{branchcount}=1\sbox{\parentbox}{\ontop{#2}}%
\setlength{\parentoffset}{\treeoffsetone}%
\addtolength{\parentoffset}{-0.5\wd\parentbox}%
\setlength{\daughteroffset}{0in}%
\ifdim\parentoffset<0in%
\setlength{\daughteroffset}{-\parentoffset}%
\setlength{\parentoffset}{0in}\fi%
\setlength{\parentwidth}{\parentoffset}%
\addtolength{\parentwidth}{\wd\parentbox}%
\setlength{\treeoffset}{\daughteroffset}%
\addtolength{\treeoffset}{\treeoffsetone}%
\setlength{\treewidth}{\wd\treeboxone}%
\addtolength{\treewidth}{\daughteroffset}%
\ifdim\treewidth<\parentwidth\setlength{\treewidth}{\parentwidth}\fi%
\sbox{\treebox}{\begin{minipage}{\treewidth}%
\begin{flushleft}%
\hspace*{\parentoffset}\usebox{\parentbox}\\
{\setlength{\unitlength}{2ex}%
\hspace*{\treeoffset}\begin{picture}(0,1)%
\put(0,0){\line(0,1){1}}%
\end{picture}}\\
\vspace{-\baselineskip}
\hspace*{\daughteroffset}%
\raisebox{-\ht\treeboxone}{\usebox{\treeboxone}}%
\end{flushleft}%
\end{minipage}}%
\setlength{\treeoffsetone}{\parentoffset}%
\addtolength{\treeoffsetone}{0.5\wd\parentbox}%
\setlength{\treeshiftone}{0pt}%
\setlength{\treewidthone}{\treewidth}%
\sbox{\treeboxone}{\usebox{\treebox}}%
\else\ifnum\value{branchcount}=2\sbox{\parentbox}{\ontop{#2}}%
\setlength{\branchwidthone}{\treewidthtwo}%
\addtolength{\branchwidthone}{\treeoffsetone}%
\addtolength{\branchwidthone}{-\treeshiftone}%
\addtolength{\branchwidthone}{-\treeoffsettwo}%
\setlength{\branchwidth}{\branchwidthone}%
\setlength{\daughteroffsetone}{\branchwidth}%
\addtolength{\daughteroffsetone}{-\branchwidthone}%
\addtolength{\daughteroffsetone}{-\treeshiftone}%
\setlength{\parentoffset}{-0.5\wd\parentbox}%
\addtolength{\parentoffset}{\treeoffsettwo}%
\addtolength{\parentoffset}{0.5\branchwidth}%
\setlength{\daughteroffset}{0in}%
\ifdim\parentoffset<0in%
\setlength{\daughteroffset}{-\parentoffset}%
\setlength{\parentoffset}{0in}\fi%
\setlength{\parentwidth}{\parentoffset}%
\addtolength{\parentwidth}{\wd\parentbox}%
\setlength{\treeoffset}{\daughteroffset}%
\addtolength{\treeoffset}{\treeoffsettwo}%
\setlength{\treewidth}{\wd\treeboxone}%
\addtolength{\treewidth}{\daughteroffsetone}%
\addtolength{\treewidth}{\treewidthtwo}%
\addtolength{\treewidth}{\daughteroffset}%
\ifdim\treewidth<\parentwidth\setlength{\treewidth}{\parentwidth}\fi%
\sbox{\treebox}{\begin{minipage}{\treewidth}%
\begin{flushleft}%
\hspace*{\parentoffset}\usebox{\parentbox}\\
{\setlength{\unitlength}{0.5\branchwidth}%
\hspace*{\treeoffset}\begin{picture}(2,0.5)%
\put(0,0){\line(2,1){1}}%
\put(2,0){\line(-2,1){1}}%
\end{picture}}\\
\vspace{-\baselineskip}
\hspace*{\daughteroffset}%
\makebox[\treewidthtwo][l]%
{\raisebox{-\ht\treeboxtwo}{\usebox{\treeboxtwo}}}%
\hspace*{\daughteroffsetone}%
\raisebox{-\ht\treeboxone}{\usebox{\treeboxone}}%
\end{flushleft}%
\end{minipage}}%
\setlength{\treeoffsetone}{\parentoffset}%
\addtolength{\treeoffsetone}{0.5\wd\parentbox}%
\setlength{\treeshiftone}{0pt}%
\setlength{\treewidthone}{\treewidth}%
\sbox{\treeboxone}{\usebox{\treebox}}\poptree%
\else\ifnum\value{branchcount}=3\sbox{\parentbox}{\ontop{#2}}%
\setlength{\branchwidthone}{\treewidthtwo}%
\addtolength{\branchwidthone}{\treeoffsetone}%
\addtolength{\branchwidthone}{-\treeshiftone}%
\addtolength{\branchwidthone}{-\treeoffsettwo}%
\setlength{\branchwidthtwo}{\treewidththree}%
\addtolength{\branchwidthtwo}{\treeoffsettwo}%
\addtolength{\branchwidthtwo}{-\treeshifttwo}%
\addtolength{\branchwidthtwo}{-\treeoffsetthree}%
\setlength{\branchwidth}{\branchwidthone}%
\ifdim\branchwidthtwo>\branchwidth%
\setlength{\branchwidth}{\branchwidthtwo}\fi%
\setlength{\daughteroffsetone}{\branchwidth}%
\addtolength{\daughteroffsetone}{-\branchwidthone}%
\addtolength{\daughteroffsetone}{-\treeshiftone}%
\setlength{\daughteroffsettwo}{\branchwidth}%
\addtolength{\daughteroffsettwo}{-\branchwidthtwo}%
\addtolength{\daughteroffsettwo}{-\treeshifttwo}%
\setlength{\parentoffset}{-0.5\wd\parentbox}%
\addtolength{\parentoffset}{\treeoffsetthree}%
\addtolength{\parentoffset}{\branchwidth}%
\setlength{\daughteroffset}{0in}%
\ifdim\parentoffset<0in%
\setlength{\daughteroffset}{-\parentoffset}%
\setlength{\parentoffset}{0in}\fi%
\setlength{\parentwidth}{\parentoffset}%
\addtolength{\parentwidth}{\wd\parentbox}%
\setlength{\treeoffset}{\daughteroffset}%
\addtolength{\treeoffset}{\treeoffsetthree}%
\setlength{\treewidth}{\wd\treeboxone}%
\addtolength{\treewidth}{\daughteroffsetone}%
\addtolength{\treewidth}{\treewidthtwo}%
\addtolength{\treewidth}{\daughteroffsettwo}%
\addtolength{\treewidth}{\treewidththree}%
\addtolength{\treewidth}{\daughteroffset}%
\ifdim\treewidth<\parentwidth\setlength{\treewidth}{\parentwidth}\fi%
\sbox{\treebox}{\begin{minipage}{\treewidth}%
\begin{flushleft}%
\hspace*{\parentoffset}\usebox{\parentbox}\\
{\setlength{\unitlength}{0.5\branchwidth}%
\hspace*{\treeoffset}\begin{picture}(4,1)%
\put(0,0){\line(2,1){2}}%
\put(2,0){\line(0,1){1}}%
\put(4,0){\line(-2,1){2}}%
\end{picture}}\\
\vspace{-\baselineskip}
\hspace*{\daughteroffset}%
\makebox[\treewidththree][l]%
{\raisebox{-\ht\treeboxthree}{\usebox{\treeboxthree}}}%
\hspace*{\daughteroffsettwo}%
\makebox[\treewidthtwo][l]%
{\raisebox{-\ht\treeboxtwo}{\usebox{\treeboxtwo}}}%
\hspace*{\daughteroffsetone}%
\raisebox{-\ht\treeboxone}{\usebox{\treeboxone}}%
\end{flushleft}%
\end{minipage}}%
\setlength{\treeoffsetone}{\parentoffset}%
\addtolength{\treeoffsetone}{0.5\wd\parentbox}%
\setlength{\treeshiftone}{0pt}%
\setlength{\treewidthone}{\treewidth}%
\sbox{\treeboxone}{\usebox{\treebox}}\poptree\poptree%
\else\ifnum\value{branchcount}=4\sbox{\parentbox}{\ontop{#2}}%
\setlength{\branchwidthone}{\treewidthtwo}%
\addtolength{\branchwidthone}{\treeoffsetone}%
\addtolength{\branchwidthone}{-\treeshiftone}%
\addtolength{\branchwidthone}{-\treeoffsettwo}%
\setlength{\branchwidthtwo}{\treewidththree}%
\addtolength{\branchwidthtwo}{\treeoffsettwo}%
\addtolength{\branchwidthtwo}{-\treeshifttwo}%
\addtolength{\branchwidthtwo}{-\treeoffsetthree}%
\setlength{\branchwidththree}{\treewidthfour}%
\addtolength{\branchwidththree}{\treeoffsetthree}%
\addtolength{\branchwidththree}{-\treeshiftthree}%
\addtolength{\branchwidththree}{-\treeoffsetfour}%
\setlength{\branchwidth}{\branchwidthone}%
\ifdim\branchwidthtwo>\branchwidth%
\setlength{\branchwidth}{\branchwidthtwo}\fi%
\ifdim\branchwidththree>\branchwidth%
\setlength{\branchwidth}{\branchwidththree}\fi%
\setlength{\daughteroffsetone}{\branchwidth}%
\addtolength{\daughteroffsetone}{-\branchwidthone}%
\addtolength{\daughteroffsetone}{-\treeshiftone}%
\setlength{\daughteroffsettwo}{\branchwidth}%
\addtolength{\daughteroffsettwo}{-\branchwidthtwo}%
\addtolength{\daughteroffsettwo}{-\treeshifttwo}%
\setlength{\daughteroffsetthree}{\branchwidth}%
\addtolength{\daughteroffsetthree}{-\branchwidththree}%
\addtolength{\daughteroffsetthree}{-\treeshiftthree}%
\setlength{\parentoffset}{-0.5\wd\parentbox}%
\addtolength{\parentoffset}{\treeoffsetfour}%
\addtolength{\parentoffset}{1.5\branchwidth}%
\setlength{\daughteroffset}{0in}%
\ifdim\parentoffset<0in%
\setlength{\daughteroffset}{-\parentoffset}%
\setlength{\parentoffset}{0in}\fi%
\setlength{\parentwidth}{\parentoffset}%
\addtolength{\parentwidth}{\wd\parentbox}%
\setlength{\treeoffset}{\daughteroffset}%
\addtolength{\treeoffset}{\treeoffsetfour}%
\setlength{\treewidth}{\wd\treeboxone}%
\addtolength{\treewidth}{\daughteroffsetone}%
\addtolength{\treewidth}{\treewidthtwo}%
\addtolength{\treewidth}{\daughteroffsettwo}%
\addtolength{\treewidth}{\treewidththree}%
\addtolength{\treewidth}{\daughteroffsetthree}%
\addtolength{\treewidth}{\treewidthfour}%
\addtolength{\treewidth}{\daughteroffset}%
\ifdim\treewidth<\parentwidth\setlength{\treewidth}{\parentwidth}\fi%
\sbox{\treebox}{\begin{minipage}{\treewidth}%
\begin{flushleft}%
\hspace*{\parentoffset}\usebox{\parentbox}\\
{\setlength{\unitlength}{0.5\branchwidth}%
\hspace*{\treeoffset}\begin{picture}(6,1)%
\put(0,0){\line(3,1){3}}%
\put(2,0){\line(1,1){1}}%
\put(4,0){\line(-1,1){1}}%
\put(6,0){\line(-3,1){3}}%
\end{picture}}\\
\vspace{-\baselineskip}
\hspace*{\daughteroffset}%
\makebox[\treewidthfour][l]%
{\raisebox{-\ht\treeboxfour}{\usebox{\treeboxfour}}}%
\hspace*{\daughteroffsetthree}%
\makebox[\treewidththree][l]%
{\raisebox{-\ht\treeboxthree}{\usebox{\treeboxthree}}}%
\hspace*{\daughteroffsettwo}%
\makebox[\treewidthtwo][l]%
{\raisebox{-\ht\treeboxtwo}{\usebox{\treeboxtwo}}}%
\hspace*{\daughteroffsetone}%
\raisebox{-\ht\treeboxone}{\usebox{\treeboxone}}%
\end{flushleft}%
\end{minipage}}%
\setlength{\treeoffsetone}{\parentoffset}%
\addtolength{\treeoffsetone}{0.5\wd\parentbox}%
\setlength{\treeshiftone}{0pt}%
\setlength{\treewidthone}{\treewidth}%
\sbox{\treeboxone}{\usebox{\treebox}}\poptree\poptree\poptree%
\else\ifnum\value{branchcount}=5\sbox{\parentbox}{\ontop{#2}}%
\setlength{\branchwidthone}{\treewidthtwo}%
\addtolength{\branchwidthone}{\treeoffsetone}%
\addtolength{\branchwidthone}{-\treeshiftone}%
\addtolength{\branchwidthone}{-\treeoffsettwo}%
\setlength{\branchwidthtwo}{\treewidththree}%
\addtolength{\branchwidthtwo}{\treeoffsettwo}%
\addtolength{\branchwidthtwo}{-\treeshifttwo}%
\addtolength{\branchwidthtwo}{-\treeoffsetthree}%
\setlength{\branchwidththree}{\treewidthfour}%
\addtolength{\branchwidththree}{\treeoffsetthree}%
\addtolength{\branchwidththree}{-\treeshiftthree}%
\addtolength{\branchwidththree}{-\treeoffsetfour}%
\setlength{\branchwidthfour}{\treewidthfive}%
\addtolength{\branchwidthfour}{\treeoffsetfour}%
\addtolength{\branchwidthfour}{-\treeshiftfour}%
\addtolength{\branchwidthfour}{-\treeoffsetfive}%
\setlength{\branchwidth}{\branchwidthone}%
\ifdim\branchwidthtwo>\branchwidth%
\setlength{\branchwidth}{\branchwidthtwo}\fi%
\ifdim\branchwidththree>\branchwidth%
\setlength{\branchwidth}{\branchwidththree}\fi%
\ifdim\branchwidthfour>\branchwidth%
\setlength{\branchwidth}{\branchwidthfour}\fi%
\setlength{\daughteroffsetone}{\branchwidth}%
\addtolength{\daughteroffsetone}{-\branchwidthone}%
\addtolength{\daughteroffsetone}{-\treeshiftone}%
\setlength{\daughteroffsettwo}{\branchwidth}%
\addtolength{\daughteroffsettwo}{-\branchwidthtwo}%
\addtolength{\daughteroffsettwo}{-\treeshifttwo}%
\setlength{\daughteroffsetthree}{\branchwidth}%
\addtolength{\daughteroffsetthree}{-\branchwidththree}%
\addtolength{\daughteroffsetthree}{-\treeshiftthree}%
\setlength{\daughteroffsetfour}{\branchwidth}%
\addtolength{\daughteroffsetfour}{-\branchwidthfour}%
\addtolength{\daughteroffsetfour}{-\treeshiftfour}%
\setlength{\parentoffset}{-0.5\wd\parentbox}%
\addtolength{\parentoffset}{\treeoffsetfive}%
\addtolength{\parentoffset}{2\branchwidth}%
\setlength{\daughteroffset}{0in}%
\ifdim\parentoffset<0in%
\setlength{\daughteroffset}{-\parentoffset}%
\setlength{\parentoffset}{0in}\fi%
\setlength{\parentwidth}{\parentoffset}%
\addtolength{\parentwidth}{\wd\parentbox}%
\setlength{\treeoffset}{\daughteroffset}%
\addtolength{\treeoffset}{\treeoffsetfive}%
\setlength{\treewidth}{\wd\treeboxone}%
\addtolength{\treewidth}{\daughteroffsetone}%
\addtolength{\treewidth}{\treewidthtwo}%
\addtolength{\treewidth}{\daughteroffsettwo}%
\addtolength{\treewidth}{\treewidththree}%
\addtolength{\treewidth}{\daughteroffsetthree}%
\addtolength{\treewidth}{\treewidthfour}%
\addtolength{\treewidth}{\daughteroffsetfour}%
\addtolength{\treewidth}{\treewidthfive}%
\addtolength{\treewidth}{\daughteroffset}%
\ifdim\treewidth<\parentwidth\setlength{\treewidth}{\parentwidth}\fi%
\sbox{\treebox}{\begin{minipage}{\treewidth}%
\begin{flushleft}%
\hspace*{\parentoffset}\usebox{\parentbox}\\
{\setlength{\unitlength}{0.5\branchwidth}%
\hspace*{\treeoffset}\begin{picture}(8,1)%
\put(0,0){\line(4,1){4}}%
\put(2,0){\line(2,1){2}}%
\put(4,0){\line(0,1){1}}%
\put(6,0){\line(-2,1){2}}%
\put(8,0){\line(-4,1){4}}%
\end{picture}}\\
\vspace{-\baselineskip}
\hspace*{\daughteroffset}%
\makebox[\treewidthfive][l]%
{\raisebox{-\ht\treeboxfour}{\usebox{\treeboxfive}}}%
\hspace*{\daughteroffsetfour}%
\makebox[\treewidthfour][l]%
{\raisebox{-\ht\treeboxfour}{\usebox{\treeboxfour}}}%
\hspace*{\daughteroffsetthree}%
\makebox[\treewidththree][l]%
{\raisebox{-\ht\treeboxthree}{\usebox{\treeboxthree}}}%
\hspace*{\daughteroffsettwo}%
\makebox[\treewidthtwo][l]%
{\raisebox{-\ht\treeboxtwo}{\usebox{\treeboxtwo}}}%
\hspace*{\daughteroffsetone}%
\raisebox{-\ht\treeboxone}{\usebox{\treeboxone}}%
\end{flushleft}%
\end{minipage}}%
\setlength{\treeoffsetone}{\parentoffset}%
\addtolength{\treeoffsetone}{0.5\wd\parentbox}%
\setlength{\treeshiftone}{0pt}%
\setlength{\treewidthone}{\treewidth}%
\sbox{\treeboxone}{\usebox{\treebox}}\poptree\poptree\poptree\poptree%
\else\typeout{QobiTeX warning--- Can't handle #1 branching}\fi\fi\fi\fi\fi}
\newcommand{\tree}{%
\usebox{\treeboxone}
\setlength{\treeoffsetone}{\treeoffsettwo}%
\sbox{\treeboxone}{\usebox{\treeboxtwo}}%
\poptree}

\input{psfig}

\def\a{\alpha}
\def\b{\beta}
\def\d{\delta}
\def\g{\gamma}
\def\e{\epsilon}
\def\l{\lambda}

\def\r{\rightarrow}

\def\t#1{\mbox{\it #1}}

\def\su#1{_{\mbox{\scriptsize #1}}}
\def\suu#1{^{\mbox{\scriptsize #1}}}
\def\suc#1{_{\mbox{\protect\scriptsize #1}}}

\def\etal{{\it et al.}}
\def\ie{{\it i.e.}}
\def\eg{{\it e.g.}}

\def\bi{\begin{itemize}}
\def\ei{\end{itemize}}
\def\i{\item}

\def\sec#1{\section{#1}}
\def\ssec#1{\subsection{#1}}
\def\sssec#1{\subsubsection{#1}}

\def\argmax{\mathop{\rm arg\,max}}
\def\argmin{\mathop{\rm arg\,min}}

\newcommand{\A}{\vec{A}}
\newcommand{\E}{\vec{E}}
\newcommand{\F}{\vec{F}}

\documentclass[11pt]{report}
\usepackage{thesis-tech}

\begin{document}

\pagenumbering{roman}

%
%

\thispagestyle{empty}

\mbox{}
\vfill

{\large
\begin{center}
{\bf \LARGE Building Probabilistic Models for Natural Language} \\
\mbox{} \\
A thesis presented \\
by \\
\bigskip
{\bf \Large Stanley F. Chen} \\
\bigskip
to \\
The Division of Applied Sciences \\
in partial fulfillment of the requirements \\
for the degree of \\
Doctor of Philosophy \\
in the subject of \\
Computer Science \\
\mbox{} \\
Harvard University \\
Cambridge, Massachusetts \\
\mbox{} \\
May 1996
\end{center}
}

\vfill

\clearpage

%
%

\mbox{}
\vfill

\begin{center}
\copyright 1996 by Stanley F. Chen\\
All rights reserved.
\end{center}

\vfill

%
%

%
\chapter*{Abstract}
%

Building models of language is a central task in natural
language processing.  Traditionally, language has been modeled
with manually-constructed grammars that describe which strings
are grammatical and which are not; however, with the recent
availability of massive amounts of on-line text, statistically-trained
models are an attractive alternative.  These models are generally
probabilistic, yielding a score reflecting sentence frequency instead
of a binary grammaticality judgement.  Probabilistic models of language
are a fundamental tool in speech recognition for resolving acoustically
ambiguous utterances.  For example,
we prefer the transcription {\it forbear\/}
to {\it four bear\/} as the former string is far more frequent
in English text.  Probabilistic models also have application in
optical character recognition, handwriting recognition, spelling
correction, part-of-speech tagging, and machine translation.

In this thesis, we investigate three problems involving the
probabilistic modeling of language: smoothing $n$-gram models,
statistical grammar induction, and bilingual sentence alignment.
These three problems employ models at three different levels
of language; they involve word-based, constituent-based,
and sentence-based models, respectively.  We describe techniques
for improving the modeling of language at each of these levels,
and surpass the performance of existing algorithms for each problem.
We approach the three problems using three
different frameworks.  We relate each of these frameworks
to the Bayesian paradigm, and show why each framework used
was appropriate for the given problem.  Finally, we show how
our research addresses two central issues in probabilistic modeling:
the sparse data problem and the problem of inducing hidden structure.

{ \singlespace

%
%

%
\chapter*{Acknowledgements}
%

{ 
\setlength{\parskip}{\smallskipamount}

\vskip-\baselineskip

I didn't realize graduate school was going to be this tough.  I
had a bad case of hubris coming in, and it took me about six years
to get it under control enough for me to graduate.  It
could have taken longer, but fortunately my advisor Stuart Shieber
was there to yell at me.  I think the turning point was
the beginning of my seventh year when Stuart said,
``You know this is your last year, don't you?''

I want to thank Stuart for giving me the freedom to try my crazy
ideas even though in hindsight they were pretty stupid,
for helping me finally learn how to do
research correctly, for teaching me how to write, and for teaching
me the importance of coming up with a ``story.''  I publicly apologize
to Stuart for not having listened to him earlier in my graduate career:
Yes, Professor Shieber, you were right all along.  Finally,
I want to thank Stuart for the vaunted ``full meal'' paradigm
of thesis writing, and the quote ``Thirty-five cents and
the big picture will get you a cup of coffee in Harvard Square.''

Next, I would like to thank Barbara Grosz for being my advisor
my first two years and for general support for the rest of them.
Again, I appreciate the freedom that Barbara gave me to pursue
my interests, and I don't hold a grudge for receiving the
fourth-lowest grade in CS 280 my year because I now realize
my final paper was crap.  (At the time, I thought it was pretty deep.)
I would also like to thank the rest of my committee,
Leslie Valiant and David Mumford, for their feedback on
my research and on my thesis, and for allowing me to graduate.

My summers at the IBM T.J.\ Watson Research Center were extremely
valuable to my graduate career.  Thanks to Peter Brown,
Stephen DellaPietra, Vincent DellaPietra, and Robert Mercer for
teaching me all I know about statistical natural language processing,
and for getting me that fellowship.  Thanks to Peter Brown for
being the best manager I've ever had.  Thanks to the rest
of the IBM crew for making it a blast:
Adam Berger, Eric Dunn, John Gillett (for quarters),
Meredith Goldsmith, Jan Hajic, Josh Koppelman (for nothing),
Ray Lau (for {\it rmon}),
David Magerman, Adwait Ratnaparkhi, Philip Resnik,
Jeff Reynar, Mike Schultz, and everybody else.

I would like to thank my undergraduate advisor, Fred Thompson,
for helping me get started on this whole artificial intelligence
thing.  He taught me how to be cynical, and was the first
to try to cure my hubris: he gave me a C in Artificial Intelligence,
but apparently I just didn't get it.

Without my Caltech friends, life would not have been nearly so bearable.
Thanks to Ray Sidney and Satomi Okazaki for inviting me over for dinner
all the time and for the R.\ incident, and thanks to Meera's
brother Bugoo for making such a fool of Ray.  Thanks to Donald Finnell
for being so easy to wail on, to Oscar Dur\`{a}n for being Guatamalan,
to Tracy Fu for having the edge, to Chris Tully and Paul
Rubinov for being hairy, and to Jared Bronski for the C.\ thing.

Thanks also go to my fellow graduate students for making
the department a great place to be:
Ellie Baker, Vanja Buvac, Rebecca Hwa, Lillian Lee (for messing
with my mind), Karen Lochbaum,
Christine Nakatani, Ted Nesson, Wheeler Ruml (for being Wheeler),
Kathy Ryall (for not washing pots), and
Nadia Shalaby (for choosing where to eat).
Thanks to Andrew Kehler,
the Golden Boy of NLP, for being my officemate and for
exposing me to fajitas, beer, ``Damn!'', and ``What's up with that?''.
Thanks to Joshua
Goodman for teaching me all I know about women, not!, for
pretending he knows everything about everything, and for
his theory of natural selection through the seat-belt law.

Of course, no acknowledgements could be complete without
mentioning my idol, Jon Christensen, better known as Slacker Boy.
Thanks for the couch and the best naps I've had in grad school,
for being so very charming and pretending to have integrity,
for teaching me about the
``I'd love to, but \ldots'' conversational macro, and for
revolutionizing the harem system through the ``Miss Tuesday Night''
concept.  Thanks also go to Jon for giving me the chance to whip
him in arm wrestling even though he outweighs me by over ten
pounds.  Last but not least, thanks to Jon for having given
me the honor of knowing him; these moments I will treasure forever.

I would like to thank my family: my sisters for giving me personal
advice which never turned out to be any good, and my parents
for their support and love.  Like Stuart, they were right all along,
and like Stuart, I didn't listen.

I am grateful for the financial support that I have received
over these years.  My research was supported in part by
a National Science Foundation Graduate Student Fellowship,
an IBM Graduate Student Fellowship, US West grant CS141212020,
the letter Q and the number 5, and National
Science Foundation grants IRI-91-57996, IRI-93-50192, and CDA-94-01024
along with matching grants from the Digital Equipment Corporation and the
Xerox Corporation.

Finally, I would like to thank myself.  If it weren't for me, I don't think
I would have made it.  I want to thank myself especially for
writing the {\it label\/} program for automatically placing
labels next to lines in graphs made by {\it gnuplot}, without
which many of the graphs in this thesis would not have been possible.
It uses neural nets and fuzzy logic.

}

%
\section*{Inspirational Quotes}
%

\begin{quote}

`` `... no one knows what's in your heart, only you, your heart and
your brain, that's all we have to battle Time...' ''
---{\it Marathon Man}, William Goldman

``... Babe replied, `It hurts too much, I'm burning up inside,' and
that made Nurmi angry, `Of course you're burning up inside, you're
supposed to burn up inside, and you keep going, you burst through
the pain barrier...' ''
---{\it Marathon Man}, William Goldman

``Don't hit the wall.  Run through it.'' --- Gatorade advertisement

``Test your faith daily.'' --- Nike advertisement

``... but for my own quite possibly perverse reasons I prefer those
scientists who drive toward daunting goals with nerves steeled against
failure and a readiness to accept pain, as much to test their own
character as to participate in the scientific culture.''
--- {\it Naturalist}, Edward O. Wilson

``Be your own pig.'' --- anonymous

\end{quote}

%
%

\tableofcontents
\listoffigures
\listoftables

}

\pagenumbering{arabic}

%
%

\chapter{Introduction} \label{ch:intro}

In this thesis, we describe novel techniques for building probabilistic
models of language.  We investigate three distinct problems involving
such models, and improve the state-of-the-art in each task.  In addition,
we show how the techniques developed in this work address two
central problems in probabilistic modeling.

In this chapter, we describe what probabilistic models of language are,
and demonstrate how such models play an
important role in many applications.  We introduce the
three problems examined and explain how they are related.  Finally, we
summarize the basic conclusions of this work.

Chapters \ref{ch:smooth}--\ref{ch:align} describe in detail the work on each
of the three tasks: smoothing $n$-gram models,
Bayesian grammar induction, and bilingual sentence alignment.
Chapter \ref{ch:concl} presents the conclusions of this thesis.

%
\sec{Models of Language}
%

A {\it model\/} of language is simply a description of language.
In its simplest form, it may just be a representation of the list
of the sentences belonging to a language; more complex models
may also try to describe the structure and meaning underlying
sentences in a language.  Historically, attempts to model language
have fallen in two general categories.  The older and more familiar
types of models are the grammars that were first developed in the
field of linguistics.  In more recent years, shallow probabilistic
models for use in applications such as speech recognition
have gained common usage.  It is these shallow probabilistic
models that we study in this thesis.  In this section, we introduce
and contrast these two types of models.

Traditionally, language has been modeled through grammars.
In linguistics, it was observed that language is structured in
a rather constrained hierarchical manner;
there seem to be a fairly small number of
primitive building blocks that can be combined together in a limited
number of ways to
create the widely diverse forms that are found in language.  For example,
at the very lowest level of written English we have the letter.
Letters can be combined to form words.  Words in turn can
be combined to form phrases, such as a {\it noun phrase}, \eg,
{\it John Smith\/} or {\it a boat}, or a {\it prepositional phrase},
\eg, {\it above the table}.  These phrases in turn can be
combined to create sentences, which in turn can be used to build paragraphs,
and so on.

{\it Grammars\/} can be used to describe such hierarchical
structure in a succinct manner \cite{Chomsky:64a}.  A grammar consists of
rules that describe allowable ways of combining structures at one
level to form structures at the next higher level.  For example,
we may have a grammar rule of the form:
$$ \begin{tabular}{ccl}
Noun-Phrase & $\r$ & Determiner Noun
\end{tabular} $$
which is generally abbreviated as
$$ \begin{tabular}{ccl}
NP & $\r$ & D N
\end{tabular} $$
This represents the observation that a determiner (\eg, {\it a\/} or {\it the})
followed by a noun can form a noun phrase.   By combining the
previous rule with the rules
$$ \begin{tabular}{ccl}
D & $\r$ & \t{a} $|$ \t{the} \\
N & $\r$ & \t{boat} $|$ \t{cat} $|$ \t{tree}
\end{tabular} $$
describing that a determiner can be formed by the words {\it a\/} or
{\it the\/} and a noun can be formed by the words {\it boat}, {\it cat},
or {\it tree}, we have that strings such as {\it a boat\/} or
{\it the tree\/} are noun phrases.

A grammar is a collection of rules like these that describe how
to form high-level structures such as sentences from low-level
structures such as words.  Using this representation, one can
attempt to describe the set of all sentences in a language,
and much work in linguistics is devoted to this goal,
though using grammar representations much richer
than the one described above.

Such grammars for language have wide application, most notably in
the field of {\it natural-language processing}.\footnote{
The term {\it natural language\/} is used to distinguish languages
used for human communication such as English or Urdu from languages
used with machines such as Basic or Lisp.  In this thesis,
we use the term {\it language\/} to mean only {\it natural languages}.}
The field of natural-language
processing deals with building automated systems that are able
to process language in some way.  For example, one of the goals
of the field is {\it natural-language understanding}, or being
able to build systems that can understand human-friendly
input such as {\it What is the capital of North Dakota\/} instead
of only computer-friendly input such as
{\it find X : capital(X, ``North Dakota'')}.

\begin{figure}
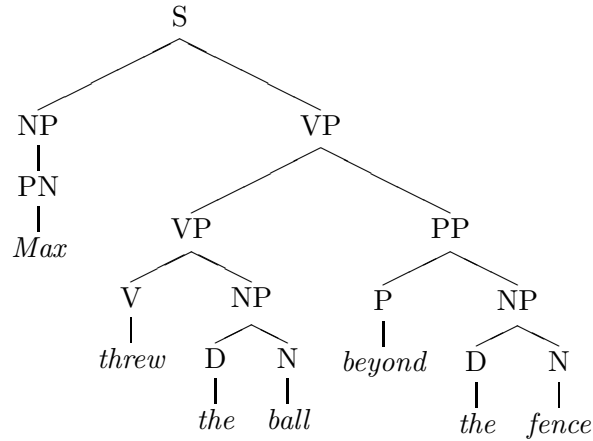


\leaf{\it Max}
\branch{1}{PN}
\branch{1}{NP}
\leaf{\it threw}
\branch{1}{V}
\leaf{\it the}
\branch{1}{D}
\leaf{\it ball}
\branch{1}{N}
\branch{2}{NP}
\branch{2}{VP}
\leaf{\it beyond}
\branch{1}{P}
\leaf{\it the}
\branch{1}{D}
\leaf{\it fence}
\branch{1}{N}
\branch{2}{NP}
\branch{2}{PP}
\branch{2}{VP}
\branch{2}{S}

$$ \tree $$
\caption{Parse tree for {\it Max threw the ball beyond the fence}}
	\label{fig:max}
\end{figure}

Grammars are useful models of language for natural language processing
because they provide insight into the structure of sentences,
which aids in determining their meanings.
For example, in most systems the first step in processing a sentence
is to {\it parse} the sentence to produce a {\it parse tree}.
We display the parse tree for {\it Max threw the ball
beyond the fence\/} in Figure \ref{fig:max}.  The parse tree
shows what rules in the grammar need to be applied
to form the top-level structure, in this case a sentence, from
the given lowest-level structures, in this case words.
In this parse tree, the top three nodes represent the applications of
the rules
$$ \begin{tabular}{ccl}
S & $\r$ & NP VP \\
NP & $\r$ & PN \\
VP & $\r$ & VP PP
\end{tabular} $$
stating that a noun phrase followed by a verb phrase can form
a sentence, a proper noun can form a noun phrase, and a verb phrase
followed by a prepositional phrase can form a verb phrase.

Substrings of the sentence that are exactly spanned by nodes
in the parse tree are intended to correspond to units that are
relevant in determining the meaning of the sentence, and
are called {\it constituents}.  For example,
the phrases {\it Max}, {\it the ball}, {\it threw the ball},
and {\it Max threw the ball beyond the fence\/} are all constituents,
while {\it the ball beyond\/} and {\it threw the\/} are not.
To give another example,
the phrase {\it the ball beyond the fence}, while meaningful,
is not a constituent because in this sentence the phrase
{\it beyond the fence\/} describes the throw, not the ball.
Thus, we see how grammars model not only which sentences belong
to a language, but also the structure that underlies the
meaning behind language.

While grammars have been the prevalent tool in modeling language
for a long time, it is generally accepted that building grammars
that can handle unrestricted language is at least many years away.
Instead, interest has shifted away from natural language understanding
toward applications that do not require such a rich model of language.
The most prominent of these applications is {\it speech recognition},
the task of constructing systems that can automatically transcribe
human speech.

In speech recognition, a model of language is used to help
disambiguate acoustically ambiguous utterances.  For example,
consider the task of transcribing an acoustic signal corresponding
to the string
$$ \t{he is too forbearing} $$
Possible transcriptions include the following:
$$ \begin{array}{ccl}
T_1 & = & \t{he is too forbearing} \\
T_2 & = & \t{he is two four baring}
\end{array} $$
While both strings have the same pronunciation, we prefer
the first transcription because it is the one that
is more likely to occur in language.  Hence, we see how models
that reflect the frequencies of different strings in language
are useful in speech recognition.

Such models typically take forms very different than linguistic grammars.
For example, a common shallow probabilistic model is the {\it bigram\/}
model.  With a bigram model, the probability
of the sentence {\it he is too forbearing\/} is expressed as
$$ p(\t{he}) p(\t{is}|\t{he}) p(\t{too}|\t{is}) p(\t{forbearing}|\t{too}) $$
Each probability $p(w_i|w_{i-1})$ attempts to reflect how often
the word $w_i$ follows the word $w_{i-1}$ in language, and these probabilities
are estimated by taking statistics on large amounts of text.

There are many significant differences between the shallow models
used in speech recognition and the grammatical models used
in linguistics and natural language processing besides their
disparate representations.
In linguistics, one attempts to build grammars that correspond exactly
to the set of {\it grammatical\/} sentences.
In speech recognition, one attempts to model
how frequently strings are spoken, regardless of grammaticality.
In linguistics and natural language processing, one is concerned
with building parse trees that reveal the meanings
of sentences.  In speech recognition, there is usually
no need for structural analysis or any other
deep processing of language.  In linguistics, models are not
probabilistic as one is only trying to express a binary grammaticality
judgement.  In speech recognition, models are almost exclusively
probabilistic in order to express frequencies.

Finally, linguistic grammars have traditionally been manually
constructed.  A linguist usually designs grammars without any
automated aid.  In contrast, models for speech recognition are
built by taking statistics on large corpora of text.  Such models
have a great number of probabilities that need to be estimated,
and this estimation is only practical
through the automated analysis of on-line text.

%
\sec{Applications for Probabilistic Models}
%

Probabilistic models of language are not only
valuable in speech recognition,
but they are also useful in applications as diverse as
spelling correction, machine translation, and part-of-speech tagging.
These and other applications can be placed in a single common
framework \cite{Bahl:83a},
the {\it source-channel\/} model used in information theory \cite{Shannon:48a}.
In this section, we explain how speech recognition
can be placed in this framework, and then explain how other applications
are just variations on this theme.

The task of speech recognition can
be framed as follows: for an acoustic signal $A$ corresponding to a sentence,
we want to find the most probable transcription $T$, \ie, to find
$$ T = \argmax_T p(T|A) $$
However, building accurate models
of $p(T|A)$ directly is beyond current know-how;\footnote{\newcite{Brown:91g}
present an explanation of why estimating $p(T|A)$ is difficult.}
instead, one applies Bayes' rule to get the relation:
\begin{equation}
T = \argmax_T \frac{p(T)p(A|T)}{p(A)} = \argmax_T p(T)p(A|T)
	\label{eqn:speech}
\end{equation}
The probability distribution $p(T)$ is called a {\it language model\/}
and describes how probable or frequent each sentence $T$ is in
language.  The distribution $p(A|T)$ is called
an {\it acoustic model\/} and describes which acoustic signals $A$
are likely realizations of a sentence $T$.
The language model $p(T)$ corresponds to the probabilistic model of
language for speech recognition discussed in the preceding section.

The {\it source-channel\/} model in information theory describes the problem
of recovering information that has been sent over a noisy channel.
One has a model of the information source, $p(I)$, and a model
of the noisy channel $p(O|I)$ describing the likely outputs $O$ of the channel
given an input $I$.  (For a perfect channel, we would just have that
$p(O|I) = 1$ for $O = I$ and $p(O|I) = 0$ otherwise.)
The task is to recover the original message $I$
sent over the channel given the noisy output $O$ received at the other
end.  This can be phrased as finding the message $I$
with highest probability given $O$, or finding
\begin{equation}
I = \argmax_I p(I|O) = \argmax_I \frac{p(I)p(O|I)}{p(O)} =
	\argmax_I p(I)p(O|I) \label{eqn:source}
\end{equation}
We can see an analogy with the task of speech recognition.
The information source in this case is a person generating the
text of a sentence according to the distribution $p(T)$.
The noisy channel corresponds to the process of a person converting
this sentence from text to speech according to $p(A|T)$.  Finally,
the goal is to recover the original text given the output
of this noisy channel.  While it may not be intuitive to refer to
a channel that converts text to speech as a noisy channel,
the mathematics are identical.

The source-channel model is a powerful paradigm because it
combines the model of the source and the model of the
channel in an elegant and efficacious manner.
For example, consider the previous example of
an acoustic utterance $A$ corresponding
to the sentence {\it he is too forbearing\/} and the possible
transcriptions:
$$ \begin{array}{ccl}
T_1 & = & \t{he is too forbearing} \\
T_2 & = & \t{he is two four baring}
\end{array} $$
Here we have two transcriptions with identical pronunciations
($p(A|T_1) \approx p(A|T_2)$), but because the former sentence
is much more common ($p(T_1) \gg p(T_2)$) we get
$p(T_1)p(A|T_1) \gg p(T_2)p(A|T_2)$ and thus prefer transcription $T_1$.
On the other hand, consider
$$ \begin{array}{ccl}
T_3 & = & \t{he is very forbearing}
\end{array} $$
In this case, we have two transcriptions with very similar
frequencies ($p(T_1) \approx p(T_3)$), but because $T_1$ has a much
higher acoustic score ($p(A|T_1) \gg p(A|T_3)$) we again
prefer $T_1$.  Thus, we see that the source-channel model
combines acoustic and language model information effectively
to prefer transcriptions that both are likely to occur
in language and match the acoustic signal well.

The source-channel model can be extended to many
other applications besides speech recognition by just varying
the channel model used \cite{Brown:90a}.  In optical character recognition
and handwriting recognition \cite{Hull:92a,Srihari:92a}, the channel can be
interpreted as converting from text to image data instead of from text
to speech, yielding the equation
$$ T = \argmax_T p(T)p(\t{image}|T). $$
In spelling correction \cite{Kernighan:90a}, the channel can
be interpreted as an imperfect typist that
converts perfect text $T$ to noisy text $T_n$ with spelling mistakes, yielding
$$ T = \argmax_T p(T)p(T_n|T). $$
In machine translation \cite{Brown:90b}, the channel can be interpreted as
a translator that converts text $T$ in one language into text $T_f$
in a foreign language, yielding
\begin{equation}
T = \argmax_T p(T)p(T_f|T).  \label{eqn:mt}
\end{equation}
In each of these cases, we try to recover the original text $T$ given
the output of a noisy channel, whether the noisy channel outputs image
data, text with spelling errors, or text in a foreign language.

By varying the source model, we can extend the source-channel model
to further applications.  In part-of-speech tagging \cite{Church:88a},
one attempts to label words in sentences with their part-of-speech.
We can apply the source-channel model by taking the source to generate
part-of-speech sequences $T\su{pos}$ corresponding to sentences,
and taking the channel to convert part-of-speech sequences $T\su{pos}$ to
sentences $T$ that are consistent with that part-of-speech sequence, yielding
$$ T\su{pos} = \argmax_{T\su{pos}} p(T\su{pos})p(T|T\su{pos}). $$
In this case, we try to recover the original part-of-speech
sequence $T\su{pos}$ given the text output of the noisy channel.
The same techniques used to build models $p(T)$ for regular text
can be used to build models $p(T\su{pos})$ for part-of-speech
sequences.

Notice that in all of these applications it is necessary to
build a source language model, either $p(T)$ or $p(T\su{pos})$.
Because of the importance of this task, this topic has become
its own field, {\it language modeling}.  Notice that
the term {\it language modeling\/} is used specifically to refer to
source language models such as $p(T)$; we use the
term {\it models of language\/} to include more general models
such as channel models or linguistic grammars.  The first two problems
we examine in this thesis are concerned with improving language modeling.  The
third problem is concerned with a model very similar to
a channel model, in particular the translation model
$p(T_f|T)$ in equation (\ref{eqn:mt}).

%
\sec{Problem Domains}
%

The three problems that we have selected investigate the task
of modeling language at three different levels: words, constituents,
and sentences.

First, we consider the problem of smoothing
$n$-gram language models \cite{Shannon:51a}.  Such models are dominant in
language modeling, yielding the best current performance.
In such models, the probability of a sentence is expressed
through the probability of each word in the sentence; such
models are {\it word-based\/} models.  The construction
of an $n$-gram model is straightforward, except for the
issue of {\it smoothing}, a technique used when there is
insufficient data to estimate probabilities accurately.
In this thesis, we introduce two novel smoothing methods
that outperform existing methods, and present an
extensive analysis of previous techniques.

Next, we consider the task of statistically inducing a grammatical
language model from text.  While it seems logical to use the
grammatical models developed in linguistics for probabilistic
language modeling, previous attempts at this have not yielded strong results.
Instead, we attempt to statistically induce a grammar from
a large corpus of text.
In grammatical language models, the probability of a sentence
is expressed through the probabilities of the constituents within
the sentence, and thus can be considered {\it constituent-based}.
Though yet to perform as well as word-based models, grammatical models
offer the best hope for significantly improving language modeling accuracy.
We introduce a novel grammar induction algorithm based on
the {\it minimum description length principle\/} \cite{Rissanen:78a}
that surpasses the performance of existing algorithms.

The third problem deals with the task of {\it bilingual sentence
alignment}.  There exist many corpora that contain equivalent text in multiple
languages.  For example, the Hansard corpus contains the Canadian parliament
proceedings in both English and French.  Multilingual corpora are useful
for automatically building tools for machine translation such as bilingual
dictionaries.  However, current algorithms for building such tools
require the specification of which sentence(s) in one language
translate to each sentence in the other language,
and this information is typically not included by human translators.
{\it Bilingual sentence alignment\/} is the
task of automatically producing this information.
This turns out to be a difficult problem as a sentence in one language
does not always correspond to a single sentence in the other language.
Sentence alignment can be approached within the source-channel
framework using equation (\ref{eqn:mt}) as in machine translation;
however, in this work we use a slightly different framework and
express the translation model $p(T_f|T)$ as a joint distribution
$p(T, T_f)$.  As sentence alignment is concerned
only with aligning text at the sentence
level, the models used are {\it sentence-based}.  We
design a sentence-based translation model that leads to
an efficient and accurate alignment algorithm that outperforms
previous algorithms.

Finally, we discuss how our work on these three problems forwards
research in probabilistic modeling.  We compare the strategies used
for building models in these three different domains from a
Bayesian perspective, and demonstrate why different strategies are
appropriate for different domains.  In addition, we show
how the techniques we have developed address two central issues in
probabilistic modeling.

%
\sec{Bayesian Modeling}
%

The Bayesian framework is an elegant and very general framework for
probabilistic modeling.  We explain the Bayesian framework
through an example: consider the task of inducing a grammar $G$
from some data or observations $O$.
In the Bayesian framework, one attempts to find
the grammar $G$ that has the highest probability given the data $O$, \ie,
to find
$$ G = \argmax_G p(G|O). $$
As it is difficult to estimate $p(G|O)$ directly, we apply Bayes' rule
to get
\begin{equation}
G = \argmax_G \frac{p(O|G)p(G)}{p(O)} = \argmax_G p(O|G) p(G).
	\label{eqn:ibayes}
\end{equation}
The term $p(O|G)$ describes the probability assigned to
the data by the grammar, and is a measure of how well the grammar models
the data.  The term $p(G)$ describes our {\it a priori\/} notion
of how likely a given grammar $G$ is.\footnote{While equation
(\ref{eqn:ibayes}) is very similar to equation (\ref{eqn:speech})
describing the source-channel model for speech recognition,
this equation differs in that
$p(G)$ is called a {\it prior\/} distribution and is built
using {\it a priori\/} information.  The analogous term $p(T)$
in equation (\ref{eqn:speech}) is called a {\it language model\/}
and is built using modeling techniques.}
This division between model accuracy and the prior belief
of model likelihood is a natural way of modularizing the
grammar induction problem.

While each of the three problems we investigated can be addressed
within the Bayesian framework, we instead selected three dissimilar
approaches.  For the grammar induction problem, we apply the
Bayesian framework in a straightforward manner.  For the
sentence alignment problem, we use {\it ad hoc\/} methods
that can be loosely interpreted as Bayesian in nature.
While the Bayesian framework is well-suited to sentence alignment,
the use of {\it ad hoc\/} methods greatly simplified
implementation at little or no cost in terms of performance.
Finally, for smoothing $n$-gram models we use non-Bayesian
methods.  It is unclear how to select a prior distribution
over smoothed $n$-gram models, and we have found that it is
more effective to optimize performance directly than to optimize
performance through examining different prior distributions.
We conclude that while the Bayesian framework
is elegant and general, in practice
less elegant methods are often effective.

%
\sec{Sparse Data and Inducing Hidden Structure}
%

Two issues that form a recurring theme in probabilistic modeling
are the {\it sparse data\/} problem and the problem of {\it inducing
hidden structure}; these are perhaps the two most important issues
in probabilistic modeling today.

The {\it sparse data\/} problem refers to the situation when there
is insufficient data to train one's model accurately.  This problem
is ubiquitous in statistical modeling; the models that perform
well tend to be very large and thus require a great deal of data
to train.  There are two main approaches to addressing this problem.
First, one can use the technique of {\it smoothing}, which describes
methods for accurately estimating probabilities in the presence
of sparse data.  Secondly, one can consider techniques for
building compact models.  Compact models have fewer parameters
to train and thus require less data.

The problem of {\it inducing hidden structure\/} describes the task
of building models that express structure not overtly present
in the training data.  To give an example, consider the bigram model
mentioned earlier, where the probability of the sentence
{\it he is too forbearing\/} is expressed as
$$ p(\t{he}) p(\t{is}|\t{he}) p(\t{too}|\t{is}) p(\t{forbearing}|\t{too}) $$
Expressing the probability of a sentence in terms of the probability
of each word in the sentence
conditioned on the immediately preceding word does not
seem particularly felicitous.  Intuitively, it seems likely that by
capturing the structure underlying language as is done
in linguistics, one may be able to build superior models.
We call this structure {\it hidden\/} as it is not explicitly
demarcated in text.\footnote{There is some data that has been
manually annotated with this information, \eg, the Penn Treebank.
However, manual annotation is expensive and thus only a limited amount
of such data is available.}
To date, bigram models and similar models greatly outperform
models that attempt to model hidden structure, but methods that induce
hidden structure offer perhaps the best hope for producing models
that significantly improve the current state-of-the-art.

In this thesis, we present several techniques that help address these
two central issues in probabilistic modeling.  For the sparse
data problem, we give novel techniques for both smoothing
and for constructing compact models.  In addition, we present novel
techniques for inducing hidden structure that are not only effective
but efficient as well.

%
%


\chapter{Smoothing $n$-Gram Models} \label{ch:smooth}

In this chapter, we describe
work on the task of smoothing $n$-gram models \cite{Chen:96a}.\footnote{
This research was joint work with Joshua Goodman.}
Of the three structural levels at which we model language
in this thesis, this represents work at the word level.
We introduce two novel smoothing techniques that significantly outperform
all existing techniques on trigram models, and that perform
competitively on bigram models.  We present an extensive
empirical comparison of existing smoothing techniques, which was
previously lacking in the literature.

%
\sec{Introduction}
%

As mentioned in Chapter \ref{ch:intro},
{\it language models\/} are a staple in many domains
including speech recognition, optical character recognition,
handwriting recognition, machine translation, and spelling correction.
A {\it language model\/} is a probability
distribution $p(s)$ over strings $s$ that attempts to
reflect how frequently a string $s$ occurs as a sentence.
For example, for a language model describing spoken
language, we might have $p(\t{hello}) \approx 0.01$ since
perhaps one out of every hundred sentences a person speaks
is {\it hello}.  On the other hand, we would have
$p(\t{chicken funky overload ketchup}) \approx 0$ and
$p(\t{asbestos gallops gallantly}) \approx 0$ since it
is extremely unlikely anyone would utter either string.
Notice that unlike in linguistics, grammaticality
is irrelevant in language modeling.  Even though
the string {\it asbestos gallops gallantly\/}
is grammatical, we still assign
it a near-zero probability.  Also, notice that in language modeling
we are only interested in the frequency
with which a string occurs as a {\it complete\/} sentence.  For instance,
we have $p(\t{you today}) \approx 0$ even though
the string {\it you today\/} occurs frequently in spoken
language, as in {\it how are you today}.

By far the most widely used language models are $n$-gram language
models.  We introduce these models by considering the case
$n=2$; these models are called {\it bigram\/} models.
First, we notice that for a sentence $s$ composed of
the words $w_1\cdots w_l$, without loss of generality
we can express $p(s)$ as
$$ p(s) = p(w_1) p(w_2 | w_1) p(w_3 | w_1w_2) \cdots p(w_l | w_1\cdots w_{l-1})
	= \prod_{i=1}^l p(w_i | w_1 \cdots w_{i-1}) $$
In bigram models, we make the approximation that the probability
of a word only depends on the identity of the immediately preceding
word, giving us
\begin{equation}
p(s) = \prod_{i=1}^l p(w_i | w_1\cdots w_{i-1})
	\approx \prod_{i=1}^l p(w_i | w_{i-1}) \label{eqn:bigram}
\end{equation}
To make $p(w_i | w_{i-1})$ meaningful for $i=1$, we can pad the
beginning of the sentence with a distinguished token $w\su{bos}$;
that is, we pretend $w_0$ is $w\su{bos}$.
In addition, to make the sum of the probabilities of all strings
$\sum_s p(s)$ equal 1, it is necessary to place a distinguished
token $w\su{eos}$ at the end of sentences and to include this
in the product in equation (\ref{eqn:bigram}).\footnote{Without
this, consider the
total probability associated with one-word strings.  We have
$$ \sum_{s=w_1} p(s) = \sum_{w_1} p(w_1 | w\su{bos}) = 1$$
That is, the probabilities associated with one-word strings alone sum to 1.
Similarly, without this device we would have that the total probability
of strings of exactly length $k$ is 1 for all $k > 0$, giving us
$$\sum_s p(s) = \sum_{l=1}^{\infty} \sum_{l(s) = l} p(s) =
	\sum_{l=1}^{\infty} 1 = \infty$$}
For example, to calculate $p(\t{John read a book})$ we would take
$$ p(\t{John read a book}) = p(\t{John} | w\su{bos})
	p(\t{read} | \t{John}) p(\t{a}|\t{read}) p(\t{book}|\t{a})
	p(w\su{eos} | \t{book}) $$

To estimate $p(w_i | w_{i-1})$, the frequency with which the
word $w_i$ occurs given that the last word is $w_{i-1}$,
we can simply count how often the bigram $w_{i-1} w_i$ occurs
in some text and normalize; that is, we can take
\begin{equation}
p(w_i | w_{i-1}) = \frac{c(w_{i-1} w_i)}{\sum_{w_i} c(w_{i-1}w_i)}
	\label{eqn:ngramml}
\end{equation}
where $c(w_{i-1} w_i)$ denotes the number of times the bigram
$w_{i-1}w_i$ occurs in the given text.\footnote{The expression
$\sum_{w_i} c(w_{i-1} w_i)$ in equation (\ref{eqn:ngramml}) can also
be expressed as simply $c(w_{i-1})$, the number of times the word $w_{i-1}$
occurs.  However, we generally use the summation form as this highlights
the fact that this expression is used for normalization.}
The text available
for building a model is called {\it training data}.  For $n$-gram models,
the amount of training data used is typically many millions
of words.  The estimate for $p(w_i | w_{i-1})$
given in equation (\ref{eqn:ngramml}) is called the
{\it maximum likelihood\/} (ML) estimate of $p(w_i | w_{i-1})$,
because this assignment of probabilities yields the bigram model that assigns
the highest probability to the training data of all possible bigram
models.\footnote{The probability of some data given
a language model is just the product of the probabilities of
each sentence in the data.  For training data $S$ composed
of the sentences $(s_1, \dots, s_{l_S})$, we have
$p(S) = \prod_{i=1}^{l_S} p(s_i)$.}

For $n$-gram models where $n > 2$, instead of conditioning the
probability of a word on the identity of just the preceding word,
we condition this probability on the identity of the last $n-1$
words.  Generalizing equation (\ref{eqn:bigram}) to $n > 2$, we get
$$ p(s) = \prod_{i=1}^l p(w_i | w_{i-n+1}^{i-1}) $$
where $w_i^j$ denotes the words $w_i\cdots w_j$.\footnote{
Instead of padding the beginning of sentences with a single $w\su{bos}$
as in a bigram model,
we need to pad sentences with $n-1$ $w\su{bos}$'s for an $n$-gram model.}
To estimate the probabilities $p(w_i | w_{i-n+1}^{i-1})$, the analogous
equation to equation (\ref{eqn:ngramml}) is
\begin{equation}
p(w_i | w_{i-n+1}^{i-1}) =
	\frac{c(w_{i-n+1}^i)}{\sum_{w_i} c(w_{i-n+1}^i)} \label{eqn:ngrammlb}
\end{equation}
In practice, the largest $n$ in wide use is $n=3$; this model is
referred to as a {\it trigram\/} model.

Let us consider a small example.  Let our training data $S$
be composed of the three sentences
$$ ( \t{John read Moby Dick}, \t{Mary read a different book},
\t{she read a book by Cher} ) $$
and let us calculate $p(\t{John read a book})$ for the
maximum likelihood bigram model.  We have
\begin{eqnarray*}
p(\t{John} | w\su{bos}) & = &
	\frac{c(w\su{bos} \t{John})}{\sum_w c(w\su{bos} w)} = \frac{1}{3} \\
p(\t{read} | \t{John}) & = &
	\frac{c(\t{John read})}{\sum_w c(\t{John}\ w)} = \frac{1}{1} \\
p(\t{a} | \t{read}) & = &
	\frac{c(\t{read a})}{\sum_w c(\t{read}\ w)} = \frac{2}{3} \\
p(\t{book} | \t{a}) & = &
	\frac{c(\t{a book})}{\sum_w c(\t{a}\ w)} = \frac{1}{2} \\
p(w\su{eos} | \t{book}) & = &
	\frac{c(\t{book}\ w\su{eos})}{\sum_w c(\t{book}\ w)} = \frac{1}{2}
\end{eqnarray*}
giving us
\begin{eqnarray*}
p(\t{John read a book})
	& = & p(\t{John} | w\su{bos}) p(\t{read} | \t{John})
		p(\t{a}|\t{read}) p(\t{book}|\t{a}) p(w\su{eos} | \t{book}) \\
	& = & \frac{1}{3} \times 1 \times \frac{2}{3}
		\times \frac{1}{2} \times \frac{1}{2} \approx 0.06
\end{eqnarray*}

Now, consider the sentence {\it Moby read a book}.  We have
$$ p(\t{read} | \t{Moby}) =
	\frac{c(\t{Moby read})}{\sum_w c(\t{Moby}\ w)} = \frac{0}{1} $$
so we will get $p(\t{Moby read a book}) = 0$.  Obviously,
this is an underestimate for the probability $p(\t{Moby read a book})$
as there is {\it some\/} probability
that the sentence occurs.  To show why it is important that
this probability should be given a nonzero value, we turn to
the primary application for language models, {\it speech recognition}.
As described in Chapter \ref{ch:intro}, in speech recognition
one attempts to find the sentence $s$ that maximizes
$p(A|s) p(s)$ for a given acoustic signal $A$.
If $p(s)$ is zero, then $p(A|s)p(s)$ will be zero and the string $s$
will never be considered as a transcription, regardless
of how unambiguous the acoustic signal is.  Thus, whenever a string $s$
such that $p(s) = 0$ occurs during a speech recognition task, an
error will be made.  Assigning
all strings a nonzero probability helps prevent errors
in speech recognition.

{\it Smoothing\/} is used to address this problem.
The term {\it smoothing\/} describes techniques for adjusting
the maximum likelihood estimate of probabilities (as
in equations (\ref{eqn:ngramml}) and (\ref{eqn:ngrammlb}))
to produce more
accurate probabilities.  Typically, smoothing methods
prevent any probability from being zero, but they also
attempt to improve the accuracy of the model as a whole.
Whenever a probability is estimated from few counts,
smoothing has the potential to significantly improve estimation.
For instance, from the three occurrences of the word {\it read} in the
above example we have that the maximum likelihood estimate of
the probability that the word {\it a\/} follows the word {\it read\/}
is $\frac{2}{3}$.  As this estimate is based on three counts, we
do not have great confidence in this estimate and intuitively,
it is a gross overestimate.  Smoothing would typically greatly
lower this estimate.

The name {\it smoothing\/} comes from the fact that these
techniques tend to make distributions more uniform, which can
be viewed as making them smoother.  Typically, very low probabilities
such as zero probabilities are adjusted upward, and high probabilities
are adjusted downward.

To give an example, one simple smoothing technique is to pretend each bigram
occurs once more than it actually does
\cite{Lidstone:20a,Johnson:32a,Jeffreys:48a}, yielding
\begin{equation}
p_{+1}(w_i | w_{i-1}) =
	\frac{c(w_{i-1} w_i) + 1}{\sum_w [c(w_{i-1} w_i) + 1]} =
	\frac{c(w_{i-1} w_i) + 1}{\sum_w c(w_{i-1} w_i) + |V|}
	\label{eqn:plusone}
\end{equation}
where $V$ is the vocabulary, the set of all words being
considered.\footnote{
Notice that if $V$ is taken to be infinite, the denominator
is infinite and all probabilities are set to zero.  In practice,
vocabularies are typically fixed to be tens
of thousands of words.  All words not in the vocabulary are
mapped to a single distinguished word, usually called the {\it unknown word}.}
Let us reconsider the previous example using this
new distribution, and let us take
our vocabulary $V$ to be the set of all words occurring
in the training data $S$, so that we have $|V| = 11$.

For the sentence {\it John read a book}, we have
\begin{eqnarray*}
p(\t{John read a book})
	& = & p(\t{John} | w\su{bos}) p(\t{read} | \t{John})
		p(\t{a}|\t{read}) p(\t{book}|\t{a}) p(w\su{eos} | \t{book}) \\
	& = & \frac{2}{14} \times \frac{2}{12} \times
		\frac{3}{14} \times \frac{2}{13} \times \frac{2}{13} \approx 0.0001
\end{eqnarray*}
In other words, we estimate that the
sentence {\it John read a book\/} occurs about
once every ten thousand sentences.  This is much more reasonable than the
maximum likelihood estimate of 0.06, or about once every seventeen
sentences.  For the sentence {\it Moby read a book}, we have
\begin{eqnarray*}
p(\t{Moby read a book})
	& = & p(\t{Moby} | w\su{bos}) p(\t{read} | \t{Moby})
		p(\t{a}|\t{read}) p(\t{book}|\t{a}) p(w\su{eos} | \t{book}) \\
	& = & \frac{1}{14} \times \frac{1}{12} \times
		\frac{3}{14} \times \frac{2}{13} \times \frac{2}{13} \approx 0.00003
\end{eqnarray*}
Again, this is more reasonable than the zero probability assigned
by the maximum likelihood model.

While smoothing is a central issue in language modeling, the
literature lacks a definitive comparison between the many
existing techniques.
Previous studies \cite{Nadas:84a,Katz:87a,Church:91a,MacKay:95a}
only compare a small number of methods (typically two) on
a single corpus and using a single training data size.  As a result,
it is currently difficult for a researcher to intelligently choose
between smoothing schemes.

In this work, we carry out an extensive empirical comparison of
the most widely used smoothing techniques, including those
described by
Jelinek and Mercer (1980), Katz(1987), and Church and Gale (1991).
\nocite{Jelinek:80a,Katz:87a,Church:91a}
We carry out
experiments over many training data sizes on varied
corpora using both bigram and trigram models.  We demonstrate that
the relative performance of techniques depends greatly on
training data size and $n$-gram order.  For example, for bigram models
produced from large training sets
Church-Gale smoothing has superior performance, while Katz smoothing
performs best on bigram models produced from smaller data.
For the methods with parameters that can be tuned to improve performance,
we perform an automated search for optimal values and show that
sub-optimal parameter selection can significantly decrease performance.
To our knowledge, this is the first smoothing work that systematically
investigates any of these issues.

In addition, we introduce two novel smoothing techniques:
the first belonging to
the class of smoothing models described by Jelinek and Mercer,
the second a very simple linear interpolation method.  While being
relatively simple to implement, we show
that these methods yield good performance in bigram models and
superior performance in trigram models.

We take the performance of a method $m$ to be its
{\it cross-entropy\/} on test data
$$ \frac{1}{N_T} \sum_{i=1}^{l_T} -\log_2 p_m(t_i) $$
where $p_m(t_i)$ denotes the language model produced with method $m$ and
where the test data $T$ is composed of sentences $(t_1, \ldots, t_{l_T})$
and contains a total of $N_T$ words.  The cross-entropy,
which is sometimes referred to as just {\it entropy}, is inversely
related to the average probability a model assigns to sentences in
the test data, and it is generally assumed
that lower entropy correlates with better performance in applications.
Sometimes the entropy is reported in terms of a {\it perplexity\/} value;
an entropy of $H$ is equivalent to a perplexity of $2^H$.
The perplexity can be interpreted as the inverse ($\frac{1}{p}$) of the
average probability ($p$) with which words are predicted by a model.
Typical perplexities yielded by $n$-gram models on English text
range from about 50 to several hundred, depending on the type of text.

In addition to evaluating the overall performance of various
smoothing techniques, we provide more detailed analyses
of performance.  We examine the performance of different algorithms
on $n$-grams with particular numbers of counts in the training data;
we find that Katz and Church-Gale smoothing
most accurately smooth $n$-grams with large counts,
while our two novel methods are best for small counts.
We calculate the relative impact on performance of small counts
and large counts
for different training set sizes and $n$-gram orders, and use this
data to explain the variation in performance of different algorithms in
different situations.  Finally, we discuss several
miscellaneous points including how
Church-Gale smoothing compares to linear interpolation,
and how deleted interpolation compares with held-out interpolation.

%
\sec{Previous Work}
%

\ssec{Additive Smoothing} \label{ssec:addintro}

The simplest type of smoothing used in practice is {\it additive\/}
smoothing \cite{Lidstone:20a,Johnson:32a,Jeffreys:48a}, which
is just a generalization of the smoothing given in
equation (\ref{eqn:plusone}).  Instead of pretending each $n$-gram
occurs once more than it does, we pretend it occurs $\d$ times
more than it does, where typically $0 < \d \leq 1$, \ie,
\begin{equation}
p\su{add}(w_i | w_{i-n+1}^{i-1}) =
	\frac{c(w_{i-n+1}^i) + \d}{\sum_{w_i} c(w_{i-n+1}^i) + \d |V|}
	\label{eqn:add}
\end{equation}
Lidstone and Jeffreys advocate taking $\d = 1$.
Gale and Church \shortcite{Gale:90a,Gale:94a} have argued that
this method generally performs poorly.

\ssec{Good-Turing Estimate} \label{ssec:gtintro}

The Good-Turing estimate \cite{Good:53a} is central
to many smoothing techniques.  The Good-Turing estimate
states that for any $n$-gram that occurs $r$ times, we
should pretend that it occurs $r^*$ times where
\begin{equation}
r^* = (r + 1) \frac{n_{r+1}}{n_r} \label{eqn:gt}
\end{equation}
and where $n_r$ is the number of
$n$-grams that occur exactly $r$ times in the training data.
To convert this count to a probability, we just normalize:
for an $n$-gram $\a$ with $r$ counts, we take
\begin{equation}
p\su{GT}(\a) = \frac{r^*}{N} \label{eqn:gtb}
\end{equation}
where $N$ is the total number of counts in the distribution.

To derive this estimate, assume that there are a total of $s$ different
$n$-grams $\a_1, \ldots, \a_s$ and that their
true probabilities or frequencies
are $p_1, \ldots, p_s$, respectively.  Let $c(\a_i)$ denote the number
of times the $n$-gram $\a_i$ occurs in the given training data.
Now, we wish to calculate the true probability of an $n$-gram $\a_i$
that occurs $r$ times; we can interpret this as calculating
$E(p_i | c(\a_i) = r)$, where $E$ denotes expected value.  This can be
expanded as
\begin{equation}
E(p_i | c(\a_i) = r) = \sum_{j=1}^s p(i = j | c(\a_i) = r) p_j
	\label{eqn:gtderiva}
\end{equation}
The probability $p(i=j|c(\a_i) = r)$ is the probability that
a randomly selected $n$-gram $\a_i$ with $r$ counts is actually the $j$th
$n$-gram $\a_j$.  This is just
$$ p(i = j | c(\a_i) = r) = \frac{p(c(\a_j) = r)}{\sum_{j=1}^s p(c(\a_j) = r)}
	= \frac{{N\choose r} p_j^r (1-p_j)^{N-r}}{\sum_{j=1}^s
		{N\choose r} p_j^r (1-p_j)^{N-r}}
	= \frac{p_j^r (1-p_j)^{N-r}}{\sum_{j=1}^s p_j^r (1-p_j)^{N-r}} $$
where $N = \sum_{i=1}^s c(\a_i)$, the total number of counts.
Substituting this into equation (\ref{eqn:gtderiva}), we get
\begin{equation}
E(p_i | c(\a_i) = r) = \frac{\sum_{j=1}^s p_j^{r+1} (1-p_j)^{N-r}}{
	\sum_{j=1}^s p_j^r (1-p_j)^{N-r}} \label{eqn:gta}
\end{equation}

Then, consider $E_N(n_r)$, the expected number of $n$-grams with exactly
$r$ counts given that there are a total of $N$ counts.  This is equal
to the sum of the probability that each $n$-gram has exactly $r$ counts:
$$ E_N(n_r) = \sum_{i=1}^s p(c(\a_i) = r)
	= \sum_{i=1}^s {N\choose r} p_i^r (1-p_i)^{N-r} $$
We can substitute this expression into equation (\ref{eqn:gta}) to yield
$$ E(p_i | c(\a_i) = r) = \frac{r+1}{N+1} \frac{E_{N+1}(n_{r+1})}{E_N(n_r)} $$
This is an estimate for the expected probability of an $n$-gram $\a_i$
with $r$ counts; to express this in terms of a corrected count $r^*$
we use equation (\ref{eqn:gtb}) to get
$$ r^* = Np(\a_i) = N \frac{r+1}{N+1} \frac{E_{N+1}(n_{r+1})}{E_N(n_r)}
	\approx (r + 1) \frac{n_{r+1}}{n_r} $$
Notice that the approximations
$E_N(n_r) \approx n_r$ and $\frac{N}{N+1} E_{N+1}(n_{r+1}) \approx n_{r+1}$
are used in the above equation.  In other words, we use the empirical values
of $n_r$ to estimate what their expected values are.

The Good-Turing estimate yields absurd values when
$n_r = 0$; it is generally necessary to ``smooth'' the $n_r$, \eg,
to adjust the $n_r$ so that they are all above zero.  Recently,
\newcite{Gale:95a} have proposed a simple and effective algorithm for smoothing
these values.

In practice, the Good-Turing estimate is not used by itself
for $n$-gram smoothing,
because it does not include the {\it interpolation\/} of higher-order models
with lower-order models necessary for good performance, as discussed
in the next section.  However, it is used as a tool in several
smoothing techniques.

\ssec{Jelinek-Mercer Smoothing} \label{ssec:jmintro}

Consider the case of constructing a bigram model on training data
where we have that $c(\t{burnish the}) = 0$ and
$c(\t{burnish thou}) = 0$.  Then, according to both additive
smoothing and the Good-Turing estimate, we will have
$$p(\t{the} | \t{burnish}) = p(\t{thou} | \t{burnish})$$
However, intuitively we should have
$$p(\t{the} | \t{burnish}) > p(\t{thou} | \t{burnish})$$
because the word {\it the\/} is much more common than
the word {\it thou}.  To capture this behavior,
we can {\it interpolate\/} the bigram model with a {\it unigram\/} model.
A {\it unigram\/} model is just a 1-gram model, which corresponds
to conditioning the probability of a word on no other words.  That is,
the unigram probability of a word just reflects its frequency in
text.  For example, the maximum likelihood unigram model is
$$ p\su{ML}(w_i) = \frac{c(w_i)}{\sum_{w_i} c(w_i)} $$

We can linearly interpolate a bigram model and unigram model as follows:
$$ p\su{interp}(w_i | w_{i-1}) =
	\l\; p\su{ML}(w_i | w_{i-1}) + (1 - \l)\; p\su{ML}(w_i) $$
where $0 \leq \l \leq 1$.  Because
$p\su{ML}(\t{the} | \t{burnish}) = p\su{ML}(\t{thou} | \t{burnish}) = 0$ while
$p\su{ML}(\t{the}) \gg p\su{ML}(\t{thou})$, we will have that
$$p\su{interp}(\t{the} | \t{burnish}) > p\su{interp}(\t{thou} | \t{burnish})$$
as desired.

In general, it is useful to linearly interpolate higher-order $n$-gram
models with lower-order $n$-gram models, because when there is
insufficient data to estimate a probability in the higher-order model,
the lower-order model can often provide useful information.
A general class of interpolated models is described
by \newcite{Jelinek:80a}.  An elegant way of performing
this interpolation is given by \newcite{Brown:91h} as follows
$$ p\su{interp}(w_i|w_{i-n+1}^{i-1}) =
	\l_{w_{i-n+1}^{i-1}}\; p\su{ML}(w_i | w_{i-n+1}^{i-1}) +
	(1 - \l_{w_{i-n+1}^{i-1}})\; p\su{interp}(w_i | w_{i-n+2}^{i-1}) $$
The $n$th-order smoothed model is defined recursively as
a linear interpolation between the $n$th-order maximum likelihood
model and the $(n-1)$th-order smoothed model.  To end
the recursion, we can take the smoothed 1st-order model to
be the maximum likelihood distribution, or we can
take the smoothed 0th-order model to be the uniform distribution
$$ p\su{unif}(w_i) = \frac{1}{|V|} $$

Given fixed $p\su{ML}$, it is possible to search
efficiently for the $\l_{w_{i-n+1}^{i-1}}$ that maximize the probability
of some data using the Baum-Welch algorithm \cite{Baum:72a}.
To yield meaningful results, the data used to estimate the
$\l_{w_{i-n+1}^{i-1}}$ need to be disjoint from the data used to
calculate the $p\su{ML}$.\footnote{When the same data is used to estimate
both, setting all $\l_{w_{i-n+1}^{i-1}}$ to one yields the optimal
result.}  In {\it held-out interpolation}, one reserves a section of
the training data for this
purpose.  Alternatively, Jelinek and Mercer describe a technique
called {\it deleted interpolation\/} where different parts of the
training data rotate in training either the $p\su{ML}$ or the
$\l_{w_{i-n+1}^{i-1}}$; the results are then averaged.

Training each parameter $\l_{w_{i-n+1}^{i-1}}$ independently is not
generally felicitous; we would need an enormous amount of data to
train so many independent parameters accurately.  Instead, Jelinek
and Mercer suggest dividing the $\l_{w_{i-n+1}^{i-1}}$ into
a moderate number of sets, and constraining all
$\l_{w_{i-n+1}^{i-1}}$ in the same set to be equal,
thereby reducing the number
of independent parameters to be estimated.  Ideally, we should
tie together those $\l_{w_{i-n+1}^{i-1}}$ that we have an {\it a priori\/}
reason to believe should have similar values.
\newcite{Bahl:83a} suggest choosing these sets of
$\l_{w_{i-n+1}^{i-1}}$ according to $\sum_{w_i} c(w_{i-n+1}^i)$, the
total number of counts in the higher-order distribution being interpolated.
The general idea is that this total count should correlate
with how strongly the higher-order distribution should be weighted.  That is,
the higher this count the higher $\l_{w_{i-n+1}^{i-1}}$ should be.
Distributions with the same number of total counts
should have similar interpolation constants.  More specifically,
Bahl \etal\ suggest dividing the range of possible total count
values into some number of partitions, and to constrain all
$\l_{w_{i-n+1}^{i-1}}$ associated with the same partition
to have the same value.  This process of dividing $n$-grams up into
partitions and training parameters independently for each partition
is referred to as {\it bucketing}.

\ssec{Katz Smoothing} \label{ssec:katzintro}

The other smoothing technique besides Jelinek-Mercer smoothing
used widely in speech recognition is due to \newcite{Katz:87a}.
Katz smoothing \shortcite{Katz:87a} extends the intuitions of Good-Turing
by adding the interpolation of higher-order models
with lower-order models.

We first describe Katz smoothing for bigram models.
In Katz smoothing, for every count $r > 0$ a {\it discount ratio\/} $d_r$
is calculated, and any bigram with $r > 0$ counts is assigned
a corrected count of $d_r r$ counts.
Then, to calculate a given conditional distribution
$p(w_i | w_{i-1})$, the nonzero counts are discounted according to $d_r$,
and the counts subtracted from the nonzero counts in that
distribution are assigned to the bigrams with zero counts.  These counts
assigned to the zero-count bigrams are distributed proportionally
to the next lower-order $n$-gram model, \ie, the unigram model.

In other words, if the original count of a bigram
$c(w_{i-1}^i)$ is $r$, we calculate its corrected count as follows:
\begin{equation}
c\su{katz}(w_{i-1}^i) = \left\{
\begin{array}{ll}
d_r r & \mbox{if $r > 0$} \\
\a\; p\su{katz}(w_i) & \mbox{if $r = 0$}
\end{array}
\right. \label{eqn:katz}
\end{equation}
where $\a$ is chosen such that the total number
of counts in the distribution $\sum_{w_i} c\su{katz}(w_{i-1}^i)$
is unchanged, \ie,
$\sum_{w_i} c\su{katz}(w_{i-1}^i) = \sum_{w_i} c(w_{i-1}^i)$.
To calculate $p\su{katz}(w_i | w_{i-1})$ from the corrected
count, we just normalize:
$$ p\su{katz}(w_i | w_{i-1}) =
	\frac{c\su{katz}(w_{i-1}^i)}{\sum_{w_i} c\su{katz}(w_{i-1}^i)} $$

The $d_r$ are calculated as follows: large counts are
taken to be reliable, so they are not discounted.  In particular, Katz
takes $d_r = 1$ for all $r > k$ for some $k$, where Katz suggests $k=5$.
The discount ratios for the lower counts $r \leq k$
are derived from the Good-Turing estimate applied to the global
bigram distribution; that is, the $n_r$ in equation (\ref{eqn:gt})
denote the total numbers of bigrams that occur exactly $r$ times
in the training data.  These $d_r$ are chosen in such a way
that the resulting discounts are proportional to the discounts
predicted by the Good-Turing estimate, and such that
the total number of counts discounted in the global bigram distribution
is equal to the total number of counts that should be assigned
to bigrams with zero counts according to the Good-Turing estimate.\footnote{
In the normal Good-Turing estimate, the number of counts
discounted from $n$-grams with nonzero counts happens to
be equal to the number of counts
assigned to $n$-grams with zero counts.  Thus, the normalization constant
for a smoothed distribution is identical to that of the original
distribution.
In Katz smoothing, Katz tries to achieve a similar effect except
through discounting only counts $r \leq k$.}
The former constraint corresponds to the equation
$$ 1 - d_r = \mu(1 - \frac{r^*}{r}) $$
for all $1 \leq r \leq k$ for some constant $\mu$.  Good-Turing estimates
that the total number of counts that should be assigned
to bigrams with zero counts is $n_0 0^* = n_0 \frac{n_1}{n_0} = n_1$,
so the second constraint corresponds to the equation
$$ \sum_{r=1}^k n_r (1 - d_r) r = n_1 $$
The unique solution to these equations is given by
$$ d_r = \frac{\frac{r*}{r} - \frac{(k+1)n_{k+1}}{n_1}}{1 -
	\frac{(k+1)n_{k+1}}{n_1}} $$

Katz smoothing for higher-order $n$-gram models is defined
analogously.  As we can see in equation (\ref{eqn:katz}), the
bigram model is defined in terms of the unigram model;
in general, the Katz $n$-gram model is defined in terms
of the Katz $(n-1)$-gram model, similar to Jelinek-Mercer
smoothing.  To end the recursion, the Katz unigram model
is taken to be the maximum likelihood unigram model:
$$ p\su{katz}(w_i) = p\su{ML}(w_i) = \frac{c(w_i)}{\sum_{w_i} c(w_i)} $$

Recall that we mentioned in Section \ref{ssec:gtintro} that
it is usually necessary to smooth $n_r$ when using the Good-Turing
estimate, \eg, for those $n_r$ that are very low.
However, in Katz smoothing
this is not essential because the Good-Turing estimate is
only used for small counts $r \leq k$, and $n_r$ is generally
fairly high for these values of $r$.

\begin{table}
$$ \begin{tabular}{|l|r|r|r|} \hline
& Jelinek-Mercer & N\'{a}das & Katz \\ \hline
bigram & 118 & 119 & 117 \\
trigram & 89 & 91 & 88 \\ \hline
\end{tabular} $$
\caption{Perplexity results reported by Katz and N\'{a}das on 100
	test sentences} \label{tab:katz}
\end{table}

Katz compares his algorithm with an unspecified version of
Jelinek-Mercer deleted estimation and with N\'{a}das smoothing \cite{Nadas:84a}
using 750,000 words of training data from an office correspondence
database.  The perplexities displayed in Table \ref{tab:katz}
are reported for a test set of 100 sentences.  (Recall that
smaller perplexities are desirable.)  Katz concludes that
his algorithm performs at least as well as Jelinek-Mercer smoothing and
N\'{a}das smoothing.

\ssec{Church-Gale Smoothing} \label{ssec:cgintro}

\newcite{Church:91a} describe a smoothing method that like Katz's,
combines the Good-Turing estimate with a method for merging
the information from lower-order models and higher-order models.

We describe this method for bigram models.  To motivate
this method, consider using the Good-Turing estimate directly
to build a bigram distribution.  For each bigram
with count $r$, we would assign a corrected count of
$r^* = (r + 1) \frac{n_{r+1}}{n_r}$.  As noted
in Section \ref{ssec:jmintro}, this has the undesirable
effect of giving all bigrams with zero count the same
corrected count; instead, unigram frequencies should
be taken into account.  Consider
the corrected count assigned by an interpolative model to
a bigram $w_{i-1}^i$ with zero counts.  In such a model,
we would have
$$ p(w_i | w_{i-1}) \propto p(w_i) $$
for a bigram with zero counts.
To convert this probability to a count, we multiply by the
total number of counts in the distribution to get
$$ p(w_i | w_{i-1}) \sum_{w_i} c(w_{i-1}^i) \propto
	p(w_i) \sum_{w_i} c(w_{i-1}^i) = p(w_i) c(w_{i-1}) \propto
	p(w_i) p(w_{i-1}) $$
Thus, $p(w_{i-1}) p(w_i)$ may be a good indicator of the
corrected count of a bigram $w_{i-1}^i$ with zero counts.

In Church-Gale smoothing, bigrams $w_{i-1}^i$ are partitioned or
{\it bucketed\/} according to the value of
$p\su{ML}(w_{i-1}) p\su{ML}(w_i)$.  That is, they divide
the range of possible $p\su{ML}(w_{i-1}) p\su{ML}(w_i)$ values
into a number of partitions, and all bigrams associated
with the same subrange are considered to be in the same bucket.
Then, each bucket is treated as a distinct probability
distribution and Good-Turing estimation is performed within
each.  For a bigram in bucket $b$ with $r_b$ counts, we
calculate its corrected count $r_b^*$ as
$$ r_b^* = (r_b + 1) \frac{n_{b,r+1}}{n_{b,r}} $$
where the counts $n_{b,r}$ include only those bigrams
within bucket $b$.

Church and Gale partition the range of
possible $p\su{ML}(w_{i-1}) p\su{ML}(w_i)$ values into about 35 buckets,
with three buckets in each factor of 10.
To smooth the $n_{b,r}$ for the Good-Turing estimate,
they use a smoother by Shirey and Hastie (1988).

While extensive empirical analysis is reported, they
present only a single entropy result, comparing the above smoothing
technique with another smoothing method introduced in their
paper, {\it extended deleted estimation}.

\ssec{Bayesian Smoothing}

Several smoothing techniques are motivated within a Bayesian framework.
A prior distribution over smoothed distributions is selected,
and this prior is used to somehow arrive at a final smoothed distribution.
For example, \newcite{Nadas:84a} selects smoothed probabilities
to be their mean {\it a posteriori\/} value given the prior distribution.

\newcite{Nadas:84a} hypothesizes a prior distribution from the
family of beta functions.  The reported experimental
results are presented in Table \ref{tab:katz}.
(The same results are reported in the Katz and N\'{a}das papers.)
These results indicate that N\'{a}das smoothing performs slightly
worse than Katz and Jelinek-Mercer smoothing.

\begin{table}
$$ \begin{tabular}{|c|ccc|c|} \hline
test set & \multicolumn{3}{c|}{Jelinek-Mercer} & MacKay-Peto \\
size (words) & 3 $\l$'s & 15 $\l$'s & 150 $\l$'s & \\ \hline
260,000 & & 79.60 & & 79.90 \\
243,000 & 89.57 & 88.47 & 88.91 & 89.06 \\
116,000 & & 91.82 & & 92.28 \\ \hline
\end{tabular} $$
\caption{Perplexity results reported by MacKay and Peto on three
	test sets} \label{tab:mackay}
\end{table}

\newcite{MacKay:95a} use Dirichlet priors in an attempt to motivate
the linear interpolation used in Jelinek-Mercer smoothing.
They compare their method with Jelinek-Mercer
smoothing for a single training set of about two million words.
For Jelinek-Mercer smoothing, deleted interpolation was used dividing
the corpus up into six sections.  The parameters $\l$ were
bucketed as suggested by \newcite{Bahl:83a}, and three different
bucketing granularities were tried.  They report results for three different
test sets; these results are displayed in Table \ref{tab:mackay}.
These results indicate that MacKay-Peto smoothing performs slightly worse
than Jelinek-Mercer smoothing.

%
\sec{Novel Smoothing Techniques}
%

Of the great many novel methods that we have tried, two techniques
have performed especially well.

\begin{figure}
$$ \psfig{figure=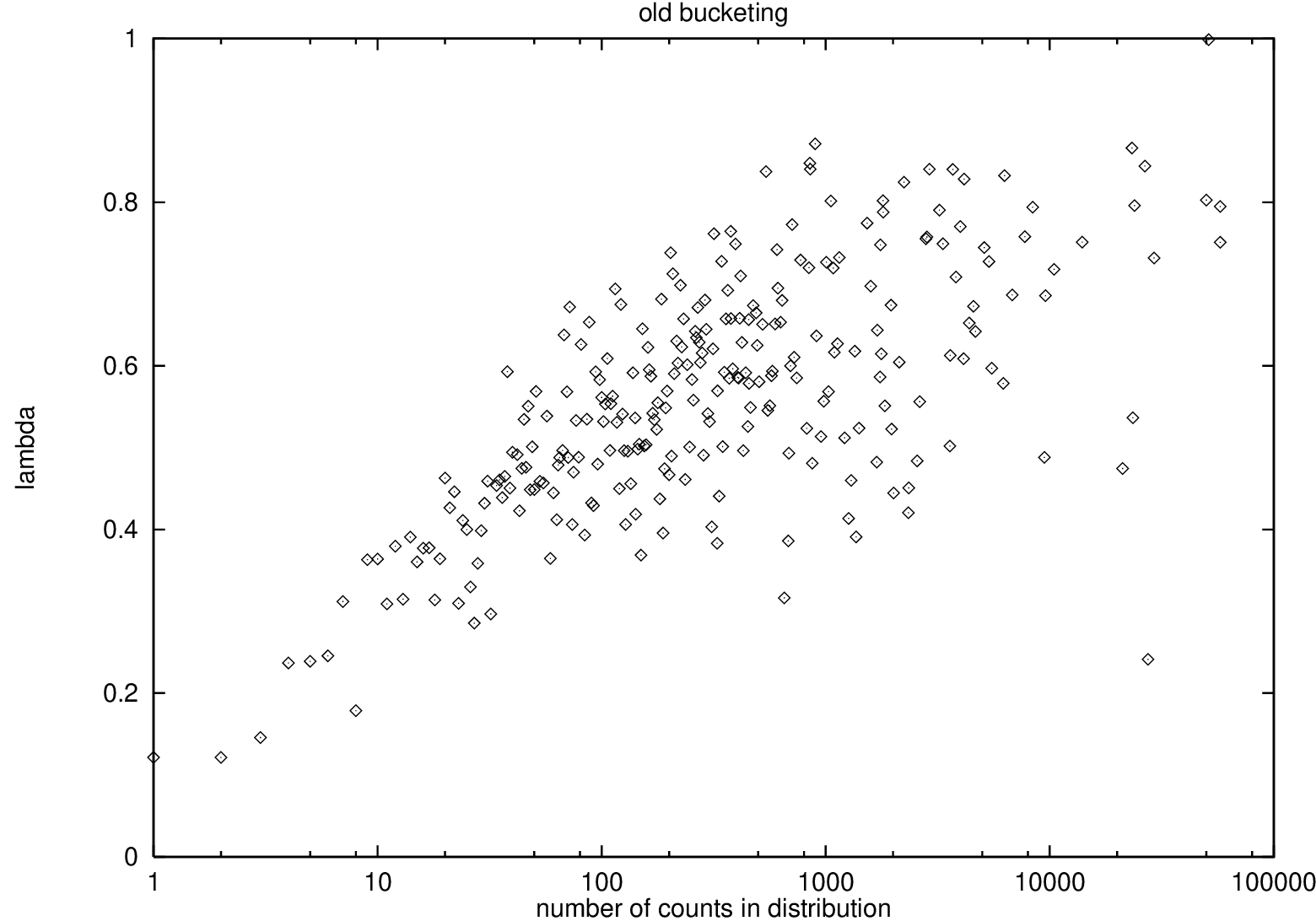,width=2.9in} \hspace{0.2in}
\psfig{figure=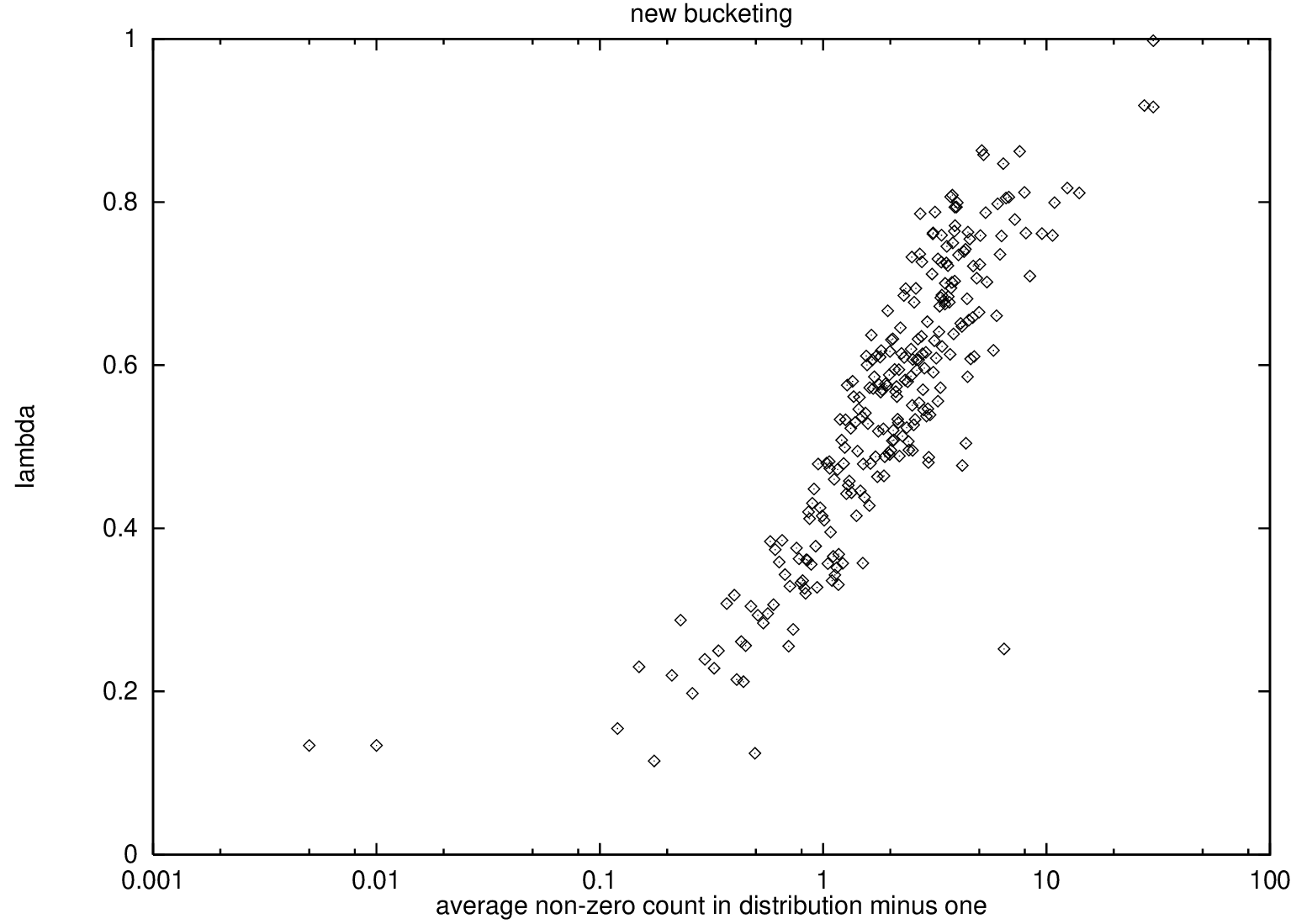,width=2.9in} $$
\caption[$\l$ values for old and new bucketing schemes for
	Jelinek-Mercer smoothing]{$\l$ values for old and new
	bucketing schemes for
	Jelinek-Mercer smoothing; each point represents a single bucket}
	\label{fig:lamb}
\end{figure}

\ssec{Method {\it average-count}} \label{ssec:methoda}

This scheme is an instance of Jelinek-Mercer smoothing.  Recall that
one takes
$$ p\su{interp}(w_i|w_{i-n+1}^{i-1}) =
	\l_{w_{i-n+1}^{i-1}}\; p\su{ML}(w_i | w_{i-n+1}^{i-1}) +
	(1 - \l_{w_{i-n+1}^{i-1}})\; p\su{interp}(w_i | w_{i-n+2}^{i-1}), $$
where Bahl \etal\ suggest that the $\l_{w_{i-n+1}^{i-1}}$ are bucketed
according to $\sum_{w_i} c(w_{i-n+1}^i)$, the total number of
counts in the higher-order distribution.  We have found that partitioning
the $\l_{w_{i-n+1}^{i-1}}$ according to the average number of counts
per nonzero element
$\frac{\sum_{w_i} c(w_{i-n+1}^i)}{|w_i : c(w_{i-n+1}^i) > 0|}$
yields better results.

Intuitively, the less sparse the data for estimating
$p\su{ML}(w_i | w_{i-n+1}^{i-1})$, the larger $\l_{w_{i-n+1}^{i-1}}$
should be.  While the larger the total number of counts in a distribution
the less sparse the distribution tends to be,
this measure ignores the allocation
of counts between words.  For example, we would consider a distribution
with ten counts distributed evenly among ten words to be much more sparse
than a distribution with ten counts all on a single word.  The
average number of counts per word seems to more directly express
the concept of sparseness.

In Figure \ref{fig:lamb}, we graph the value of $\l$
assigned to each bucket
under the original and new bucketing schemes on identical data.
The $x$-axis in each graph represents the criteria used for
bucketing.
Notice that the new bucketing scheme results in a much tighter plot,
indicating that it is better at grouping together distributions
with similar behavior.

One can use the Good-Turing estimate to partially explain this
behavior.  As mentioned in Section \ref{ssec:katzintro},
the Good-Turing estimate states that the number of counts
that should be devoted to $n$-grams with zero counts is $n_1$,
the number of $n$-grams in the distribution with exactly one count.
This is equivalent to assigning a total probability of $\frac{n_1}{N}$
to $n$-grams with zero counts, where $N$ is the total number of
counts in the distribution.  Notice that the value
$1 - \l_{w_{i-n+1}^{i-1}}$ in Jelinek-Mercer smoothing
is roughly proportional to the total probability assigned to $n$-grams
with zero counts: for $n$-grams $w_{i-n+1}^i$ with zero count we have
$p\su{ML}(w_i | w_{i-n+1}^{i-1}) = 0$ so
$$ p\su{interp}(w_i|w_{i-n+1}^{i-1}) =
	(1 - \l_{w_{i-n+1}^{i-1}})\; p\su{interp}(w_i | w_{i-n+2}^{i-1}) $$
Thus, it seems reasonable that we want to satisfy the relation
$$ 1 - \l_{w_{i-n+1}^{i-1}} \propto \frac{n_1}{N} $$
where in this case $n_1 = |w_i : c(w_{i-n+1}^i) = 1|$ and
$N = \sum_{w_i} c(w_{i-n+1}^i)$.\footnote{Notice that
$n_1$ and $N$ have different meanings in this context from those
found in
Katz smoothing and Church-Gale smoothing.  While in each of
these cases we use the Good-Turing estimate, we apply the
estimate to different distributions.  In Katz, we apply the
Good-Turing estimate to the {\it global\/} $n$-gram distribution,
so that $n_1$ represents the total number of $n$-grams with
exactly one count and $N$ represents the total count of $n$-grams.
Church-Gale is similar to Katz except that $n$-grams are
partitioned into a number of buckets.  However, in this context
we apply the Good-Turing estimate to a conditional distribution
$p(w_i | w_{i-n+1}^{i-1})$ for some fixed $w_{i-n+1}^{i-1}$.  Thus,
$n_1$ represents the number of $n$-grams $w_{i-n+1}^i$
beginning with $w_{i-n+1}^{i-1}$ with exactly one count, and
$N$ represents the total count of $n$-grams $w_{i-n+1}^i$
beginning with $w_{i-n+1}^{i-1}$.}

Our goal in choosing a bucketing scheme is to bucket $n$-grams
that should have similar $\l$ values.  Hence, given the
above analysis we should bucket $n$-grams $w_{i-n+1}^i$
according to the value of
\begin{equation}
\frac{n_1}{N} = \frac{|w_i : c(w_{i-n+1}^i) = 1|}{\sum_{w_i} c(w_{i-n+1}^i)}
	\label{eqn:newagt}
\end{equation}
This is very similar to the inverse of
$$\frac{\sum_{w_i} c(w_{i-n+1}^i)}{|w_i : c(w_{i-n+1}^i) > 0|},$$
the actual value we use to bucket with.  Instead of looking
at the number of $n$-grams with exactly one count, we use
the number of $n$-grams with nonzero counts.  Notice that it
does not matter much whether we bucket according to a value
or according to its inverse; the same $n$-grams are grouped together.

A natural experiment to try is to bucket according to the expression
given in equation (\ref{eqn:newagt}).  However, using this expression
and variations, we were unable to surpass the performance of
the given bucketing scheme.

\ssec{Method {\it one-count}}

This technique combines two intuitions.  First, \newcite{MacKay:95a}
show that by using a Dirichlet prior as a prior distribution
over possible smoothed distributions, we get (roughly speaking)
a model of the form
$$ p\su{one}(w_i | w_{i-n + 1}^{i-1}) = \frac{c(w_{i-n +1}^{i}) +
	\a p\su{one}(w_i | w_{i-n + 2}^{i-1})}{\sum_{w_i} c(w_{i-n+1}^i) + \a} $$
where $\a$ is constant across $n$-grams.  This is similar
to additive smoothing, except that instead of adding the same
number of counts to each $n$-gram, we add counts proportional
to the probability yielded by the next lower-order distribution.
The parameter $\a$ represents the total number of counts
that we add to the distribution.

Secondly, using a similar analysis as in the last section, the Good-Turing
estimate can be interpreted as stating that the number of extra
counts $\a$ should be proportional to $n_1$, the number of $n$-grams with
exactly one count in the given distribution.  Thus, instead of taking
$\a$ to be constant across $n$-grams $w_{i-n+1}^i$,
we take it to be a function of $n_1 = | w_i : c(w_{i-n+1}^i) = 1 |$.
We have found that taking
\begin{equation}
\a = \g\; (n_1 + \b) \label{eqn:newb}
\end{equation}
works well, where $\b$ and $\g$ are constants.
Notice that higher-order models are defined recursively
in terms of lower-order models.

Given the results mentioned in the last section, a natural
experiment to try is to take $\a$ to be a function of
the number of {\it nonzero\/} counts in the distribution, as opposed to
the number of {\it one\/} counts.  However, attempts in this vein
failed to yield superior results.

%
\sec{Experimental Methodology}
%

In our experiments, we compare our novel smoothing methods
with the most widely-used smoothing techniques in
language modeling: additive smoothing,
Jelinek-Mercer smoothing, and Katz smoothing.  For Jelinek-Mercer
smoothing, we try both held-out interpolation and deleted
interpolation.  In addition, we have also implemented Church-Gale
smoothing, as this has never been compared against popular techniques.
We do not consider N\'{a}das smoothing or MacKay-Peto smoothing
as they are not widely used and
as previous results indicate that they do not perform as well
as other methods.

As a baseline method, we choose a simple instance
of Jelinek-Mercer smoothing, one that uses much fewer parameters
than is typically used in real applications.

\ssec{Smoothing Implementations} \label{ssec:simpl}

In this section, we discuss the details of our implementations of
various smoothing techniques.  The titles
of the following sections include the mnemonic we use to refer
to the implementations in later sections.  We use the mnemonic
when we are referring to our specific implementation of a smoothing method,
as opposed to the algorithm in general.  For each method, we
mention the parameters that can be tuned to optimize performance;
in general, any variable mentioned is a tunable parameter.

\begin{table}

$$ \begin{tabular}{|l|r|} \hline
Method & Lines \\ \hline
{\tt plus-one} & 40 \\ \hline
{\tt plus-delta} & 40 \\ \hline
{\tt katz} & 300 \\ \hline
{\tt church-gale} & 1000 \\ \hline
{\tt interp-held-out} & 400 \\ \hline
{\tt interp-del-int} & 400 \\ \hline
{\tt new-avg-count} & 400 \\ \hline
{\tt new-one-count} & 50 \\ \hline
{\tt interp-baseline}\footnotemark & 400 \\ \hline
\end{tabular} $$
\caption{Implementation difficulty of various methods in terms
of lines of C++ code} \label{tab:diff}
\end{table}

\footnotetext{
For {\tt interp-baseline}, we used the {\tt interp-held-out} code as
it is just a special case.  Written anew, it probably would have been
about 50 lines.}

To give an informal estimate of the difficulty of implementation
of each method, in Table \ref{tab:diff}
we display the number of lines of C++ code in each
implementation excluding the core code common across techniques.

\sssec{Additive Smoothing ({\tt plus-one}, {\tt plus-delta})}

We consider two versions of additive smoothing.  Referring to
equation (\ref{eqn:add}) in Section \ref{ssec:addintro}, we fix $\d = 1$
in {\tt plus-one} smoothing.  In {\tt plus-delta}, we consider any $\d$.
(The values of parameters such as $\d$ are determined
through training on held-out data.)

\sssec{Jelinek-Mercer Smoothing ({\tt interp-held-out},
{\tt interp-del-int})} \label{sssec:jmimpl}

Recall that higher-order models are defined recursively in terms
of lower-order models.  We end the recursion by taking the 0th-order
distribution to be the uniform distribution $p\su{unif}(w_i) = 1 / |V|$.

We bucket the $\l_{w_{i-n+1}^{i-1}}$ according to $\sum_{w_i} c(w_{i-n+1}^i)$
as suggested by Bahl {\it et al}.  Intuitively, each bucket should be
made as small as possible, to only group together the most
similar $n$-grams, while remaining large enough to accurately estimate
the associated parameters.  We make the assumption
that whether a bucket is large enough for accurate parameter estimation
depends on how many $n$-grams that fall in that bucket occur
in the data used to train the $\l$'s.  We bucket in a such a way
that a minimum of $c\su{min}$ $n$-grams fall in each bucket.
We start from the lowest possible value of
$\sum_{w_i} c(w_{i-n+1}^i)$ (\ie, zero)
and put increasing values of $\sum_{w_i} c(w_{i-n+1}^i)$ into
the same bucket until this minimum count is reached.  We repeat
this process until all possible values of
$\sum_{w_i} c(w_{i-n+1}^i)$ are bucketed.  If the last bucket
has fewer than $c\su{min}$ counts, we merge it with the preceding bucket.
Historically, this process is called
the {\it wall of bricks} \cite{Magerman:94a}.
We use separate buckets for each $n$-gram model being interpolated.

In performing this bucketing, we create an array containing
how many $n$-grams occur for each value of
$\sum_{w_i} c(w_{i-n+1}^i)$ up to some maximum value of
$\sum_{w_i} c(w_{i-n+1}^i)$, which we call $c\su{top}$.
For $n$-grams $w_{i-n+1}^{i-1}$ with
$\sum_{w_i} c(w_{i-n+1}^i) > c\su{top}$, we pretend
$\sum_{w_i} c(w_{i-n+1}^i) = c\su{top}$ for bucketing purposes.

As mentioned in Section \ref{ssec:jmintro}, the $\l$'s can be
trained efficiently using the Baum-Welch algorithm.  Given
initial values for the $\l$'s, the Baum-Welch algorithm adjusts these
parameters iteratively to minimize the entropy of some data.
The algorithm generally decreases the entropy with each iteration,
and guarantees not to increase it.  We set all $\l$'s initially
to the value $\l_0$.  We terminate the algorithm when the entropy
per word changes less than $\d\su{stop}$ bits between iterations.

We implemented two versions of Jelinek-Mercer smoothing, one
using held-out interpolation and one using deleted interpolation.
In {\tt interp-held-out}, the $\l$'s are trained using held-out interpolation
on one of the development test sets.
In {\tt interp-del-int}, the $\l$'s are trained
using the {\it relaxed deleted interpolation\/} technique described by
Jelinek and Mercer, where one word is deleted at a time.
In {\tt interp-del-int}, we bucket
an $n$-gram according to its count before deletion, as this turned out
to significantly improve performance.  We hypothesize that this
is because this causes an $n$-gram to be placed in the same bucket
during training as in evaluation, allowing the $\l$'s to be meaningfully
geared toward individual $n$-grams.

\sssec{Katz Smoothing ({\tt katz})}

Referring to Section \ref{ssec:katzintro}, instead of a single $k$
we allow a different $k_n$ for each $n$-gram model being interpolated.

Recall that higher-order models are defined recursively in terms
of lower-order models, and that the recursion is ended by taking
the unigram distribution to be the maximum likelihood distribution.
While using the maximum likelihood unigram distribution works
well in practice, this choice is not well-suited to our work.
In practice, the vocabulary $V$ is
usually chosen to include only those words that occur
in the training data, so that
$p\su{ML}(w_i) > 0$ for all $w_i \in V$.  This assures that
the probabilities of all $n$-grams are nonzero.
However, in this work we do not satisfy the constraint that
all words in the vocabulary occur in the training data.  We
run experiments using many training set sizes, and we use
a fixed vocabulary across all runs so that results between
sizes are comparable.  Not all words in the vocabulary will
occur in the smaller training sets.  Thus, unless we smooth
the unigram distribution we may have $n$-gram probabilities that
are zero, which could lead to an infinite cross-entropy on test data.
To address this issue, we smooth the unigram distribution
in Katz smoothing using additive smoothing;
we call the additive constant $\d$.\footnote{In Jelinek-Mercer
smoothing, we address this issue by ending the model recursion
with a 0th-order model instead of a unigram model, and taking
the 0th-order model to be a uniform distribution.  We tried
a similar tack with Katz smoothing, but the natural way of
interpolating a unigram model with a uniform model in the Katzian
paradigm led to poor results.  We tried additive smoothing
instead, which is equivalent to interpolating with a uniform
distribution using the Jelinek-Mercer paradigm, and this
worked well.}

In the algorithm as described in the original paper,
no probability is assigned to $n$-grams with zero counts
in a conditional distribution $p(w_i|w_{i-n+1}^{i-1})$ if there are
no $n$-grams $w_{i-n+1}^i$ that occur between
1 and $k_n$ times in that distribution.
This can lead to an infinite cross-entropy on test data.
To address this, whenever there are no counts between 1 and $k_n$ in
a conditional distribution, we give the zero-count $n$-grams a total
of $\b$ counts, and increase the normalization constant appropriately.

\sssec{Church-Gale Smoothing ({\tt church-gale})} \label{sssec:cgimpl}

While Church and Gale use the maximum likelihood unigram
distribution, we instead smooth the unigram distribution using Good-Turing
(without bucketing) as this seems more consistent with the spirit
of the algorithm.  This should not affect performance much, as the
unigram probabilities are used only for bucketing purposes.\footnote{
We observed an interesting phenomenon when smoothing the unigram
distribution with Good-Turing.  We construct a vocabulary $V$ by collecting
all words in a corpus that occur at least $k$ times, for some value $k$.
For training sets that include the majority of a corpus,
there will be unnaturally few words occurring
fewer than $k$ times, since most of these words have been weeded out
of the vocabulary.  This results in $n_r$ that can yield odd
corrected counts $r^*$.  For example, for a given cutoff $k$ we
may get $n_k \gg n_{k-1}$ so that $(k-1)^*$ is overly high.
We have not found that this phenomenon significantly affects performance.
}

We use a different bucketing scheme than that described by Church and
Gale.  For a bigram model, they divide the range of possible
values of $p(w_{i-1}) p(w_i)$ into about 35 buckets, with three
buckets per factor of 10.  However, this bucketing strategy is not ideal
as bigrams are not distributed uniformly among different
orders of magnitude.
Furthermore, they provide analysis that
indicates that they had sufficient data to distinguish between
at least 1200 different probabilities to be assigned to
bigrams with zero counts; this is evidence that using significantly more
than 35 buckets might yield better performance.  Hence, we chose to
use {\it wall of bricks\/} bucketing as in our implementation
of Jelinek-Mercer smoothing.

We first do as Church and Gale do and partition the range of
possible $p(w_{i-1}) p(w_i)$ values using some constant
number of buckets per order of magnitude, except
instead of using a total of 35 buckets we use some very large
number of buckets, $c\su{mb}$.  Instead of calling
these partitions {\it buckets\/} we call them {\it minibuckets},
as we lump together these minibuckets to form our final buckets
using the wall of bricks technique.  We group together minibuckets
so that at least $c\su{min}$ $n$-grams with nonzero count fall
in each bucket.

To smooth the counts $n_r$ needed for the Good-Turing
estimate, we use the technique described by \newcite{Gale:95a}.
This technique assigns a total probability of
$\frac{n_1}{N}$ to $n$-grams with zero counts, as dictated
by the Good-Turing estimate.  However, it is possible
that $n_1 = N$ in which case no probability is assigned to
nonzero counts.  As this is unacceptable, we modify the algorithm so that
in this case, we assign a total probability
of $p_{n_1=N} < 1$ to zero counts.  In addition, it is possible that $n_1 = 0$
in which case no probability is assigned to zero counts.  In this
case, we instead assign a total probability of $p_{n_1=0} > 0$ to
zero counts.

Finally, the original paper describes only bigram smoothing in detail;
extending this method to trigram models is ambiguous.  In particular,
it is unclear whether to bucket trigrams according to
$p(w_{i-2}^{i-1}) p(w_i)$ or $p(w_{i-2}^{i-1}) p(w_i | w_{i-1})$.
We choose the former value; while the latter value may yield better
performance as it is a better estimate of $p(w_{i-2}^i)$,\footnote{
Referring to the analysis given in Section \ref{ssec:cgintro}, the
former choice roughly corresponds to interpolating the trigram model
with a unigram model, while the latter choice corresponds to
interpolating the trigram model with a bigram model. }
our belief is that it is much more difficult to implement and
it requires a great deal more computation.

\begin{figure}
{ \def\r#1{{\bf #1}}
\begin{tabbing}
mm \= mm \= mm \= mm \= \hspace{2in} \= \kill
; {\it count how many trigrams with nonzero counts fall in each minibucket} \\
\r{for} each trigram $w_{i-2}^i$ with $c(w_{i-2}^i) > 0$ \r{do} \\
\> increment the count for the minibucket $b_m$ that $w_{i-2}^i$ falls in \\
\mbox{} \\
group minibuckets $b_m$ into buckets $b$ using the wall of bricks technique \\
\mbox{} \\
; {\it calculate number of trigrams in each bucket by looping over
	all values of $p(w_{i-2}^{i-1}) p(w_i)$} \\
; {\it this is used later to calculate number of trigrams with zero counts
	in each bucket} \\
; {\it the first two loops loop over all possible values of $p(w_{i-2}^{i-1})$,
	the third loop is for $p(w_i)$} \\
\r{for} each bigram bucket $b_2$ \r{do} \\
\> \r{for} each count $r_2$ with $n_{b_2,r_2} > 0$ \r{do} \\
\> \> \r{for} each count $r_1$ with $n_{r_1} > 0$ in the unigram distribution
	\r{do} \\
\> \> \> \r{begin} \\
\> \> \> $b$ := the bucket a trigram $w_{i-2}^i$ falls in if
	$c(w_{i-2}^{i-1}) = r_2$ and $c(w_i) = r_1$ \\
\> \> \> increment the count for $b$ by the number of trigrams $w_{i-2}^i$
	such that \\
\> \> \> \> $c(w_{i-2}^{i-1}) = r_2$ and $c(w_i) = r_1$, \ie
	$|w_{i-2}^{i-1} : c(w_{i-2}^{i-1}) = r_2| \times |w_i : c(w_i) = r_1|$ \\
\> \> \> \r{end} \\
\mbox{} \\
; {\it calculate counts $n_{b,r}$ for each bucket $b$ and count $r > 0$} \\
\r{for} each trigram $w_{i-2}^i$ with $c(w_{i-2}^i) > 0$ \r{do} \\
\> calculate the bucket $b$ that $w_{i-2}^i$ falls in, and
	increase $n_{b,r}$ for $r = c(w_{i-2}^i)$ \\
\mbox{} \\
calculate $n_{b,0}$ by subtracting $\sum_{r=1}^{\infty} n_{b,r}$
	from the total number of trigrams in $b$ \\
\mbox{} \\
smooth the $n_{b,r}$ values using the Gale-Sampson algorithm \\
\mbox{} \\
; {\it calculate normalization constants $\sum_{w_i} c\su{GT}(w_{i-2}^i)$ for
	each $w_{i-2}^{i-1}$} \\
\r{for} each bigram bucket $b_2$ \r{do} \\
\> \r{for} each count $r_2$ with $n_{b_2,r_2} > 0$ \r{do} \\
\> \> calculate normalization constant $N_{b_2,r_2}$ for a
	bigram $w_{i-2}^{i-1}$ in bucket $b_2$ \\
\> \> \> with $c(w_{i-2}^{i-1}) = r_2$
	given that $c(w_{i-2}^i) = 0$ for all $w_i$ \\
\r{for} each bigram $w_{i-2}^{i-1}$ with $c(w_{i-2}^{i-1}) > 0$ \r{do} \\
\> calculate its normalization constant $\sum_{w_i} c\su{GT}(w_{i-2}^i)$
	by calculating its difference \\
\> \> from $N_{b_2,r_2}$ where $w_{i-2}^{i-1}$ falls in
	bucket $b_2$ and $c(w_{i-2}^{i-1}) = r_2$; this can be done \\
\> \> by looping through all trigrams $w_{i-2}^i$ with $c(w_{i-2}^i) > 0$ \\
\end{tabbing} }
\caption{Outline of our Church-Gale trigram implementation}
	\label{fig:cgtri}
\end{figure}

We outline the algorithm we use to construct a trigram model
in Figure \ref{fig:cgtri}.  The time complexity of the algorithm
is roughly $O(c\su{nz} + c\su{p})$, where $c\su{nz}$ denotes
the number of trigrams with nonzero
counts and $c\su{p}$ denotes the number of different possible
values of $p(w_{i-2}^{i-1}) p(w_i)$.  The term $c\su{p}$ comes
from the fact that to calculate the total number of $n$-grams in a
bucket $b$ (which is needed to efficiently calculate $n_{b,0}$),
it is necessary to loop over all possible values of
$p(w_{i-2}^{i-1}) p(w_i)$.
We take advantage of the fact that the number of possible
values for $p(w_{i-2}^{i-1})$ is at most the total number of different
bigram counts in each bigram bucket $|n_{b,r} : n_{b,r} > 0, \mbox{$b$ a
bigram bucket}|$, as each $(b, r)$ pair corresponds to a potentially different
corrected count
that can be assigned to a bigram.  Similarly, the number of possible
values for $p(w_i)$ is at most the total number of different unigram counts
$|n_r : n_r > 0, \mbox{for the unigram distribution}|$.

Now, consider the analogous algorithm except bucketing using
$p(w_{i-2}^{i-1}) p(w_i | w_{i-1})$.  The factor in $c\su{p}$
from $p(w_{i-2}^{i-1})$ remains the same, but the number of
different values for $p(w_i| w_{i-1})$ is much larger than the
number of different values for $p(w_i)$.  The number of
different values for $p(w_i| w_{i-1})$ is roughly equal to
the number of different bigrams with nonzero counts, while
the number of different values for $p(w_i)$
is at most the number of different unigram {\it counts}.  Thus,
this alternate bucketing scheme is much more expensive computationally.

\sssec{Novel Smoothing Methods ({\tt new-avg-count}, {\tt new-one-count})}

The implementation of smoothing method {\it average-count},
{\tt new-avg-count}, is identical to
{\tt interp-held-out} except that instead of bucketing the $\l_{w_{i-2}^{i-1}}$
according to $\sum_{w_i} c(w_{i-n+1}^i)$, we bucket according to
$\frac{\sum_{w_i} c(w_{i-n+1}^i)}{|w_i : c(w_{i-n+1}^i) > 0|}$
as described in Section \ref{ssec:methoda}.

In the implementation of smoothing method {\it one-count},
{\tt new-one-count}, we
have different parameters $\b_n$ and $\g_n$ in equation (\ref{eqn:newb})
for each $n$-gram model being interpolated.
Also, recall that higher-order models are defined recursively
in terms of lower-order models.  We end the recursion by taking the 0th-order
distribution to be the uniform distribution $p\su{unif}(w_i) = 1 / |V|$.

\sssec{Baseline Smoothing ({\tt interp-baseline})}

For our baseline smoothing method, we use
Jelinek-Mercer smoothing with held-out interpolation
where for each $n$-gram model being interpolated
we constrain all $\l_{w_{i-n+1}^{i-1}}$ in the model
to be equal to a single value $\l_n$, \ie,
$$p\su{base}(w_i | w_{i-n+1}^{i-1}) =
	\l_n\; p\su{ML}(w_i | w_{i-n+1}^{i-1}) +
	(1 - \l_n)\; p\su{base}(w_i | w_{i-n+2}^{i-1}).$$
This is identical to {\tt interp-held-out} where $c\su{min}$ is
set to $\infty$, so that there is only a single bucket
for each $n$-gram model.

\ssec{Implementation Architecture}

In this section, we give an overview of the entire implementation.
The coding was done in C++.
Each of the implementations of the individual smoothing techniques
were linked into a single program, to help ensure uniformity
in the methodology used with each smoothing technique.

For large training sets, it is difficult to fit an
entire bigram or trigram model into a moderate amount of memory.
Thus, we chose not to do the straightforward
implementation of first building an entire smoothed $n$-gram model in memory
and then evaluating it.

Instead, notice that for a given test set, it is only necessary to
build that part of the smoothed $n$-gram model that is applicable
to the test set.  To take advantage of this observation,
we first process the training set by taking
counts of all $n$-grams up to the target $n$, and we sort these
$n$-grams into an order suitable for future processing.  We then
iterate through these $n$-gram counts, extracting those counts
that are relevant for evaluating the test data (or the held-out data
used to optimize parameter values).  We use these extracted counts
to build the smoothed $n$-gram model on only the relevant data.
For some smoothing algorithms ({\tt katz} and {\tt church-gale}), it is
necessary to make additional passes through the $n$-gram counts
to collect other statistics.

In some experiments, we use very large test sets; in this case
the above algorithm is not practical as a too large fraction
of the total model is needed to evaluate the test data.
For these experiments, we process the test data in the
same way as the training data, by taking counts of all
relevant $n$-grams and sorting them.  We then iterate
through both the training $n$-gram counts and test $n$-gram
counts simultaneously, repeatedly building a small section of the
smoothed $n$-gram model, evaluating it on the associated
test data, and then discarding the partial model.

In our implementation, we include a general multidimensional search
engine for automatically searching for optimal parameter values
for each smoothing technique.  We use the implementation
of Powell's search algorithm \cite{Brent:73a} given
in {\it Numerical Recipes in C} \cite[pp. 309--317]{Press:88a}.
Powell's algorithm does not require the calculation of the gradient.  It
involves successive searches along vectors in the multidimensional
search space.

\ssec{Data}

We used the Penn treebank and TIPSTER corpora distributed
by the Linguistic Data Consortium.  From the treebank, we extracted
text from the tagged Brown corpus,
yielding about one million words.  From
TIPSTER, we used the Associated Press (AP), Wall Street Journal (WSJ), and
San Jose Mercury News (SJM) data, yielding 123, 84, and 43 million
words respectively.  We created two distinct vocabularies, one for the Brown
corpus and one for the TIPSTER data.  The former vocabulary
contains all 53,850 words occurring in Brown;
the latter vocabulary consists of the 65,173 words
occurring at least 70 times in TIPSTER.

For each experiment, we selected three segments of held-out data
along with the segment of training data.  These four segments
were chosen to be adjacent in the original corpus
and disjoint, the held-out segments
preceding the training to facilitate the use of common held-out data
with varying training data sizes.  The first held-out segment was
used as the test data for performance evaluation, and the other
two held-out segments were used as development test data for optimizing the
parameters of each smoothing method.  In experiments with multiple
runs on the same training data size, the data
segments of each run are completely disjoint.

Each piece of held-out data was chosen to be
roughly 50,000 words.  This decision does not
reflect practice well.  For example, if the training set size is
less than 50,000 words then it is not realistic to have this
much development test data available.
However, we made this choice to prevent
us having to optimize the training versus held-out data tradeoff for each
data size.  In addition, the development test
data is used to optimize typically
very few parameters, so in practice small held-out
sets are generally adequate, and perhaps can be avoided altogether
with techniques such as deleted estimation.

\ssec{Parameter Setting}

\begin{figure}
$$ \psfig{figure=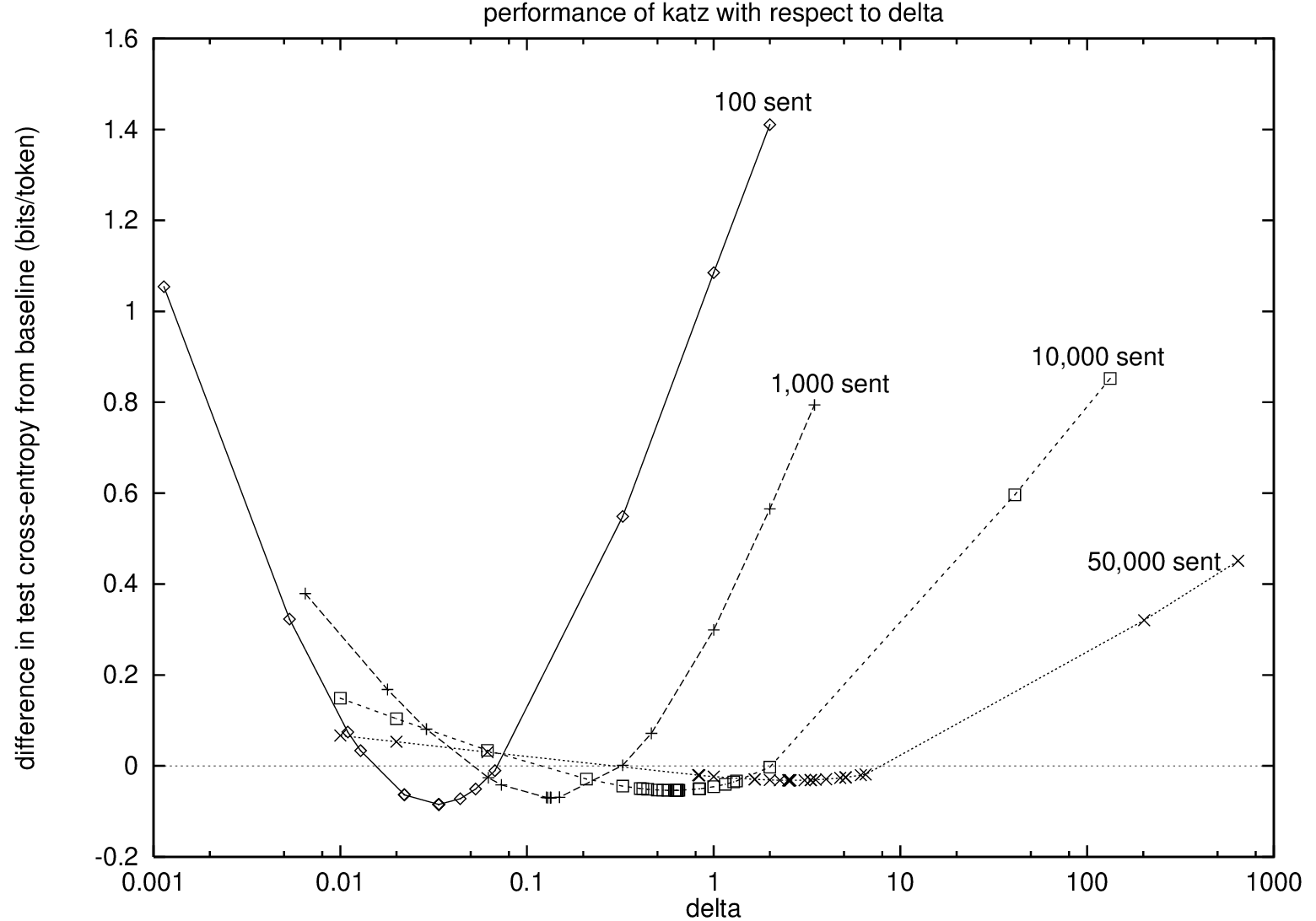,width=2.9in} \hspace{0.2in}
\psfig{figure=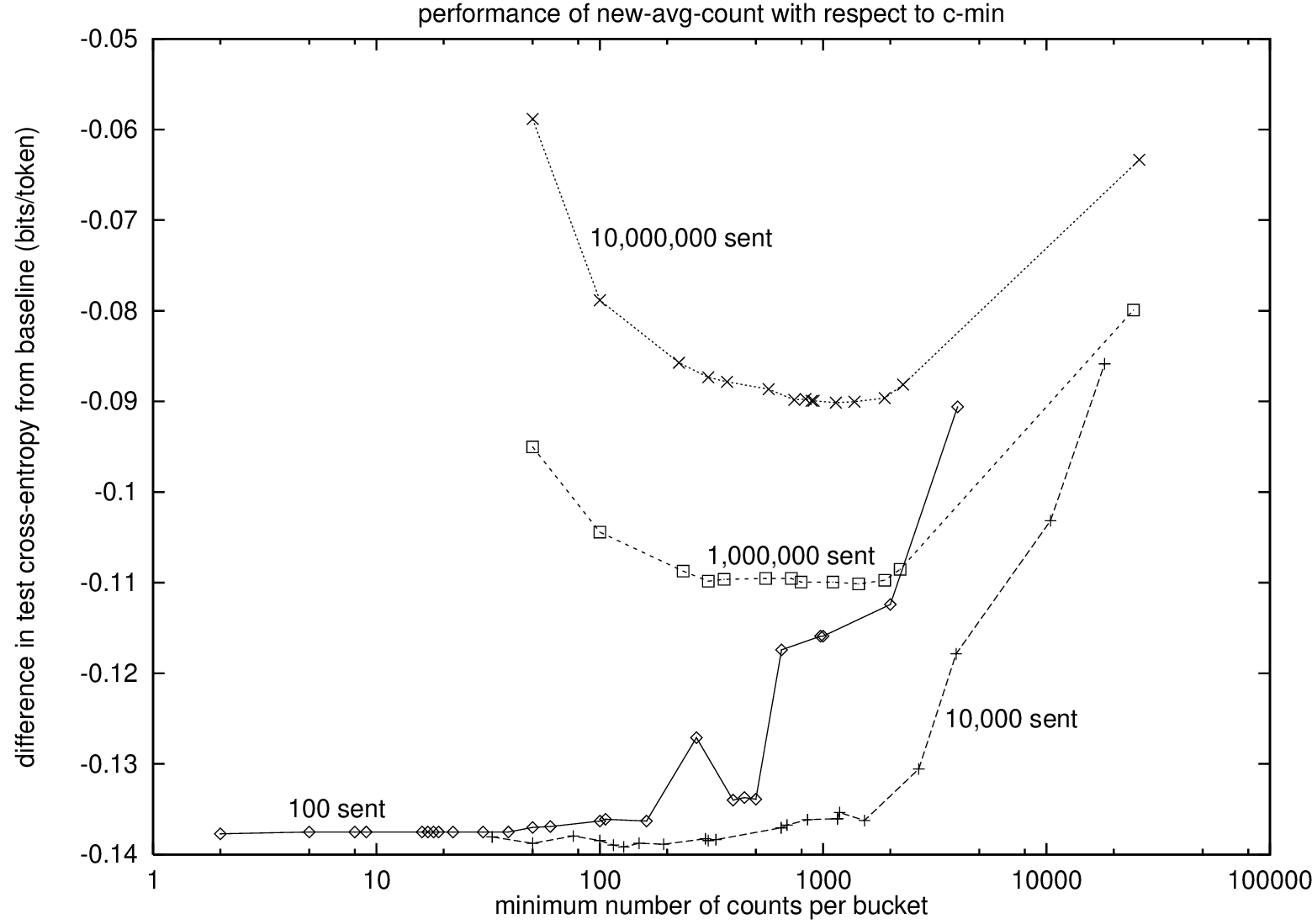,width=2.9in} $$
\caption{Performance relative to baseline method
	of {\tt katz} and {\tt new-avg-count} with respect to parameters
	$\delta$ and $c\suc{min}$, respectively, over several training set sizes}
	\label{fig:param}
\end{figure}

In Figure \ref{fig:param}, we show how the values of the parameters $\d$
and $c\su{min}$ affect the performance of methods {\tt katz} and
{\tt new-avg-count}, respectively, over several training data sizes.  Notice
that poor parameter setting can lead to very significant losses
in performance.  In Figure \ref{fig:param}, we see differences
in entropy from several hundredths of a bit to over a bit.
Also, we see that the optimal value of a parameter varies with
training set size.  Thus, it is important to optimize parameter
values to meaningfully compare smoothing techniques, and this
optimization should be specific to the given training set size.

In each experiment we ran except as noted below, optimal values for
the parameters of the given method
were searched for using Powell's search algorithm.
Parameters were chosen to optimize the cross-entropy of the
first of the two development
test sets associated with the given training set.
For {\tt katz} and {\tt church-gale}, we did not perform
the parameter search for training sets over 50,000 sentences due
to resource constraints, and instead manually extrapolated parameter
values from optimal values found on smaller data sizes.

For instances of Jelinek-Mercer smoothing, the $\l$'s were
trained using the Baum-Welch algorithm on the second development
test set; all other parameters were optimized using Powell's algorithm on
the first development test set.  More specifically, to evaluate
the entropy associated with a given set of (non-$\l$) parameters
in Powell's search, we first optimize the $\l$'s on the second
test set.

To constrain the parameter search in our main battery of experiments,
we searched only those parameters that were
found to affect performance
significantly, as indicated through preliminary experiments over several
data sizes.  In each run of these preliminary experiments, we
fixed all parameters but one to some reasonable value, and
used Powell's algorithm to search on the single free parameter.  We recorded
the entropy of the test data for each parameter value considered
by Powell's algorithm.  If the range of test data entropy over this search
was much smaller than the typical difference in entropies between
different algorithms, we considered it safe not to perform the
search over this parameter in the later experiments.  For each
parameter, we tried three different training sets: 20,000 words
from the WSJ corpus, 1M words from the Brown corpus, and 3M words
from the WSJ corpus.

We assumed that all parameters are significant for the
methods {\tt plus-one}, {\tt plus-lambda}, and {\tt new-one-count}.
We describe the results for the other algorithms below.

\sssec{Jelinek-Mercer Smoothing ({\tt interp-held-out}, {\tt interp-del-int},
	\sloppy {\tt new-avg-count}, {\tt interp-baseline})}

The parameter $\l_0$, the initial value of the $\l$'s in
the Baum-Welch search, affected entropy by less than 0.001 bits.
Thus, we decided not to search over this parameter in later
experiments.  We fix $\l_0$ to be 0.5.

The parameter $\d\su{stop}$, controlling when to terminate the
Baum-Welch search, affected entropy by less than 0.002 bits.
We fix $\d\su{stop}$ to be 0.001 bits.

The parameter $c\su{top}$, the top count considered in bucketing,
affected entropy by up to 0.006 bits, which is significant.  However,
we found that in general the entropy is lower for higher values
of $c\su{top}$; this range of 0.006 bits is mainly due to setting $c\su{top}$
too low.  We fix $c\su{top}$ to be a fairly large value, 100,000.

The parameter $c\su{min}$, the minimum number of counts in each
bucket, affected entropy by up to 0.07 bits, which is significant.
Thus, we search over this parameter in later experiments.

\sssec{Katz Smoothing ({\tt katz})}

\begin{figure}
$$ \psfig{figure=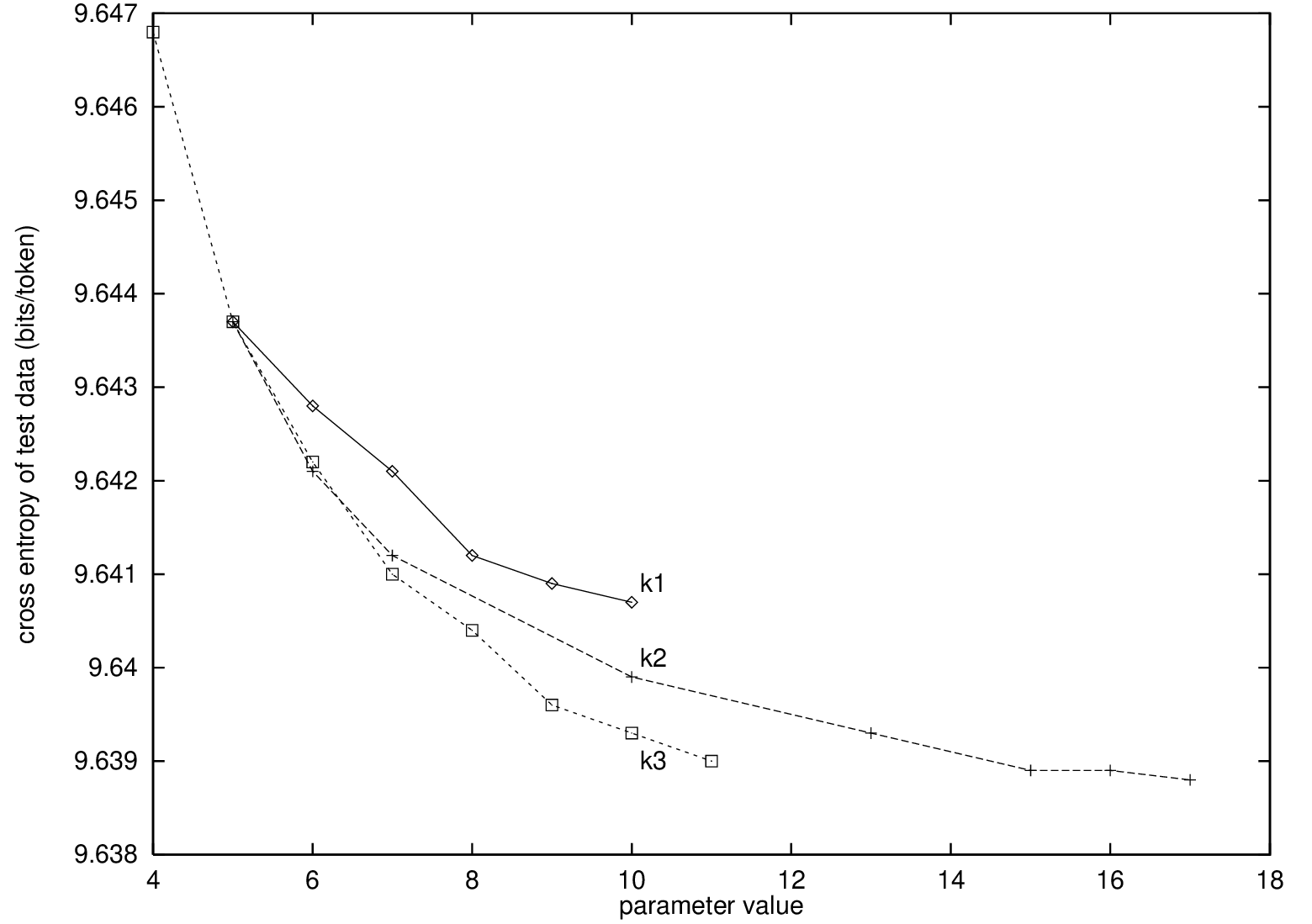,width=2.9in} $$
\caption{Effect of $k_n$ on Katz smoothing} \label{fig:kparam}
\end{figure}

The parameters $k_n$, specifying the count above which counts are
not discounted, affected entropy by up to 0.01 bits, which
is significant.  However, we found that the larger the $k_n$,
the better the performance.  In Figure \ref{fig:kparam},
we display the entropy on the Brown corpus
for different values of $k_1$, $k_2$, and $k_3$.
However, for large $k$ there will be counts $r$ such
that the associated discount ratio
$d_r$ takes on an unreasonable value, such as a nonpositive value
or a value above one.  We take $k_n$ to be
as large as possible such that the $d_r$ take on reasonable values.

The parameter $\b$, describing how many counts are given to $n$-grams with
zero counts if no counts in a distribution are discounted,
affected the entropy by less than 0.001 bits.  We fix $\b$ to be 1.

The parameter $\d$, the constant used for the additive smoothing
of the unigram distribution, affected entropy by up to 0.02 bits,
which is significant. Thus, we search over this parameter in
later experiments.  For large training sets (over 50,000 sentences),
we do not perform the search due to time constraints.  Instead,
we choose its value by manually extrapolating from the optimal values
found on smaller training sets.  For example, for TIPSTER we found that
$\d = 0.0011 \times l_S^{0.7}$ fits the optimal values found for smaller
training sets well, where $l_S$ denotes the number of sentences
in the training data.

\sssec{Church-Gale Smoothing ({\tt church-gale})}

The parameter $p_{n_1=0}$, the probability assigned to zero counts
if there are no one-counts in a distribution, affected the entropy
not at all.  We fix $p_{n_1=0}$ to be 0.01.

The parameter $p_{n_1=N}$, the probability assigned to zero counts
if all counts in a distribution are one-counts, affected the
entropy by up to 0.2 bits.  Thus, we search over this parameter
in later experiments.  However, for training sets over 50,000 sentences,
due to time constraints we do not perform parameter search
for {\tt church-gale}.  We noticed that for larger training sets
this parameter does not seem to have a large effect (0.002 bits
on the 3M words of WSJ), and the optimal value tends to be very close
to 1.  Thus, for large training sets we fix $p_{n_1=N}$ to be 0.995.

\begin{figure}
$$ \psfig{figure=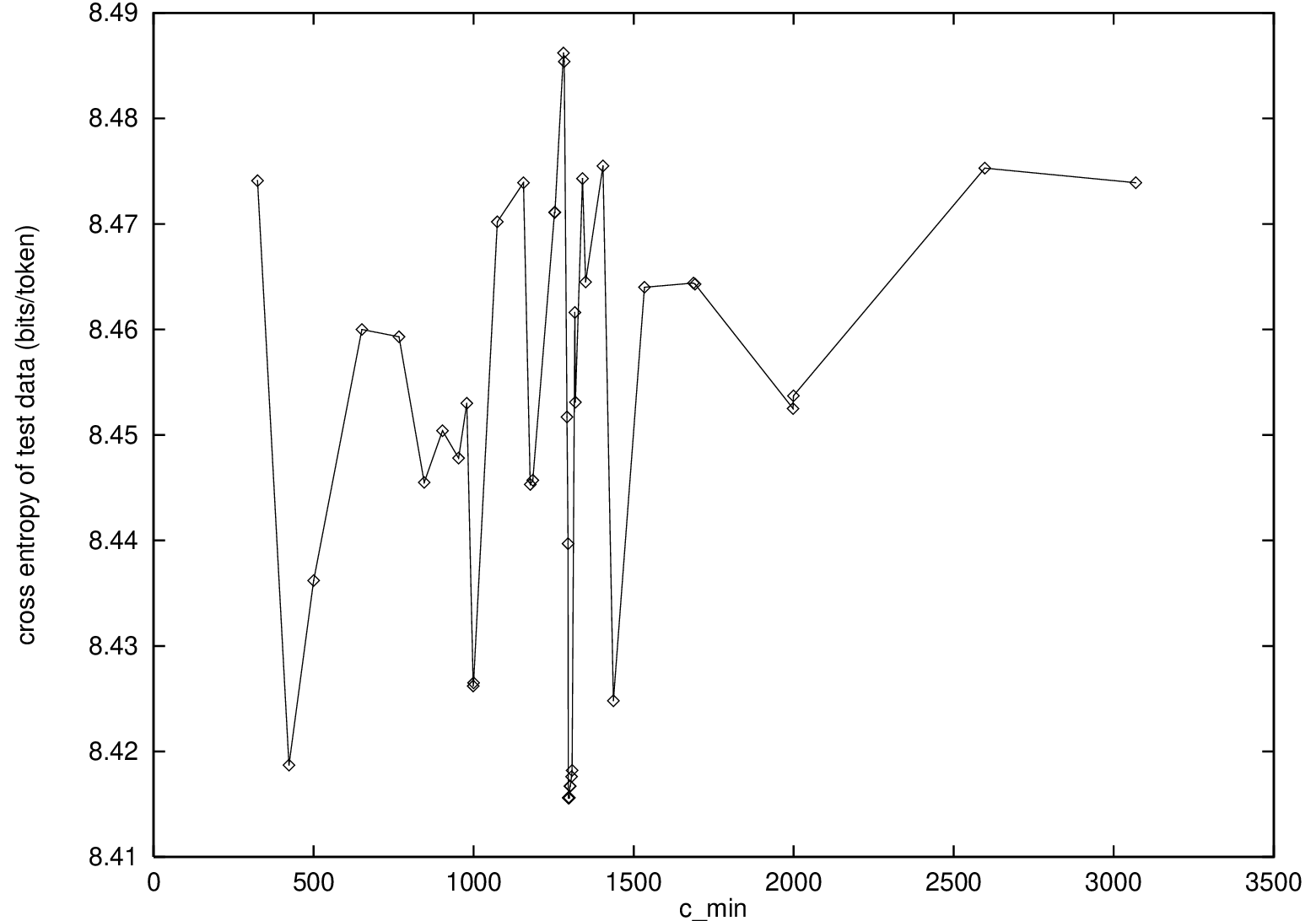,width=2.9in} \hspace{0.2in}
\psfig{figure=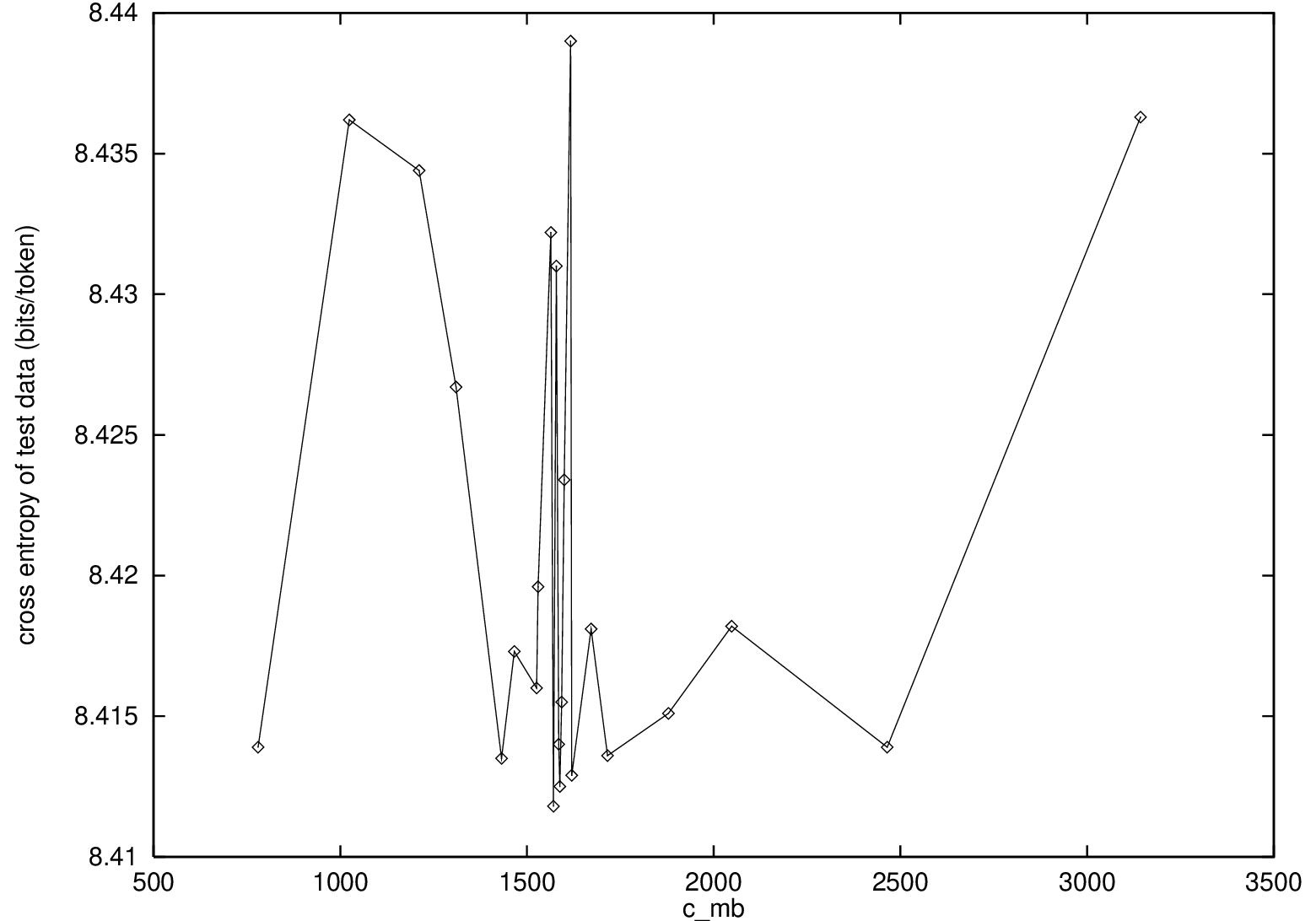,width=2.9in} $$
\caption{Effect of $c\suc{min}$ and $c\suc{mb}$ on Church-Gale smoothing}
	\label{fig:cgparam}
\end{figure}

The parameters $c\su{min}$, the minimum number of counts per bucket,
and $c\su{mb}$, the number of minibuckets, both affected the
entropy a great deal (over 0.5 bits).  Thus, we search over
these parameters in later experiments.  However, the search
space for both of these parameters is very bumpy, so it
is unclear how effective the search process is.  In
Figure \ref{fig:cgparam}, we display the entropy on test data
for various values of $c\su{min}$ and $c\su{mb}$ when training
on 3M words of WSJ.  The search
algorithm will find a local minimum, but we will have no guarantee on
the global quality of this minimum given the nature of the search space.

As mentioned above, for training sets over 50,000 sentences,
due to time constraints we do not perform parameter search
for {\tt church-gale}.  Fortunately, for larger training sets $c\su{min}$
and $c\su{mb}$ seem to have a smaller effect (0.07 and 0.03 bits,
respectively, on the 3M words of WSJ).  For training sets over 50,000
sentences, we just guess
reasonable values for these parameters:
we fix $c\su{min}$ to be 500 and $c\su{mb}$ to be 100,000.
For very large data sets, due to memory constraints
we take $c\su{min}$ to be $\frac{l_S}{200}$
to limit the number of buckets created.

%

\begin{figure}[p]
$$ \psfig{figure=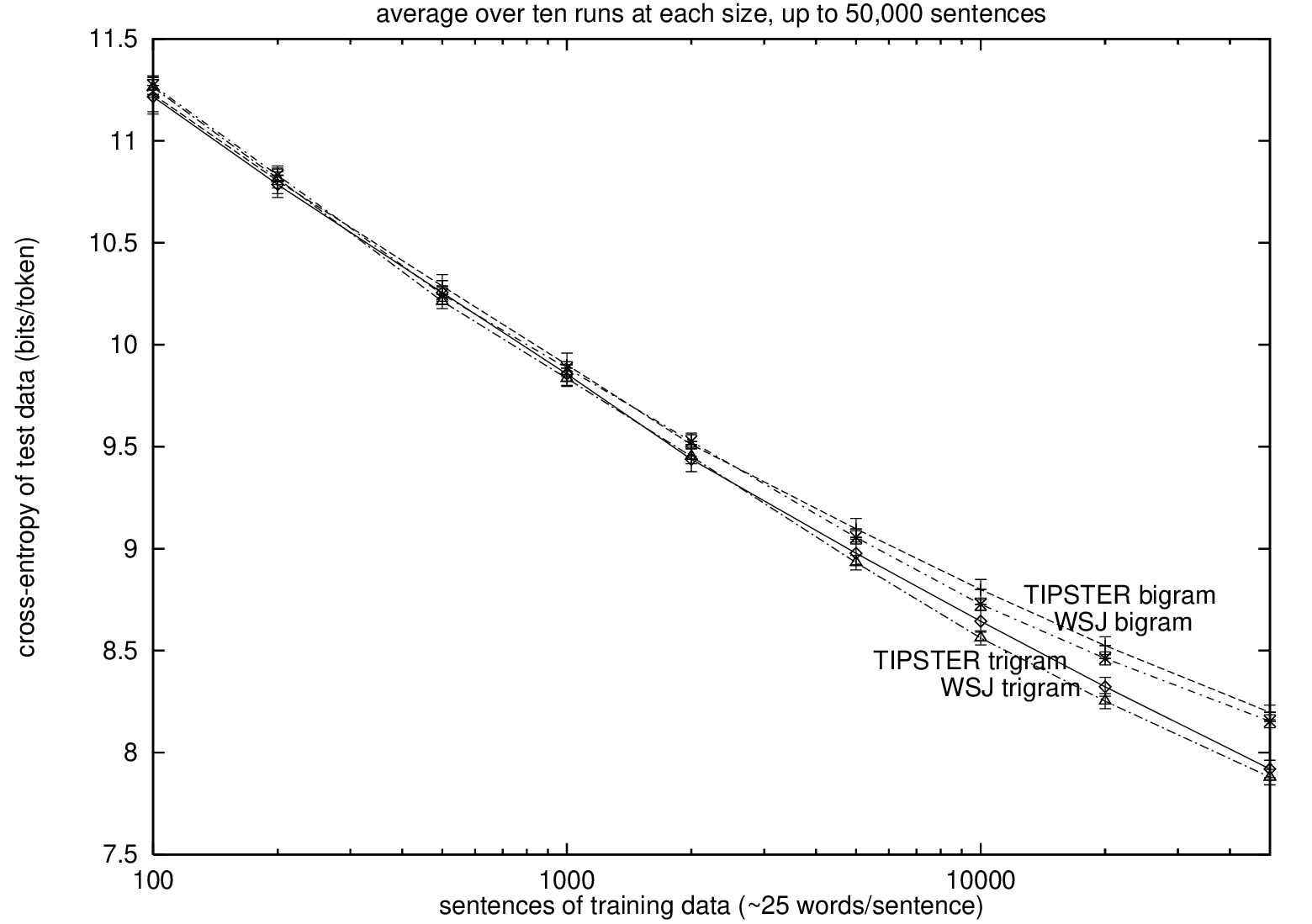,width=2.9in} \hspace{0.2in}
\psfig{figure=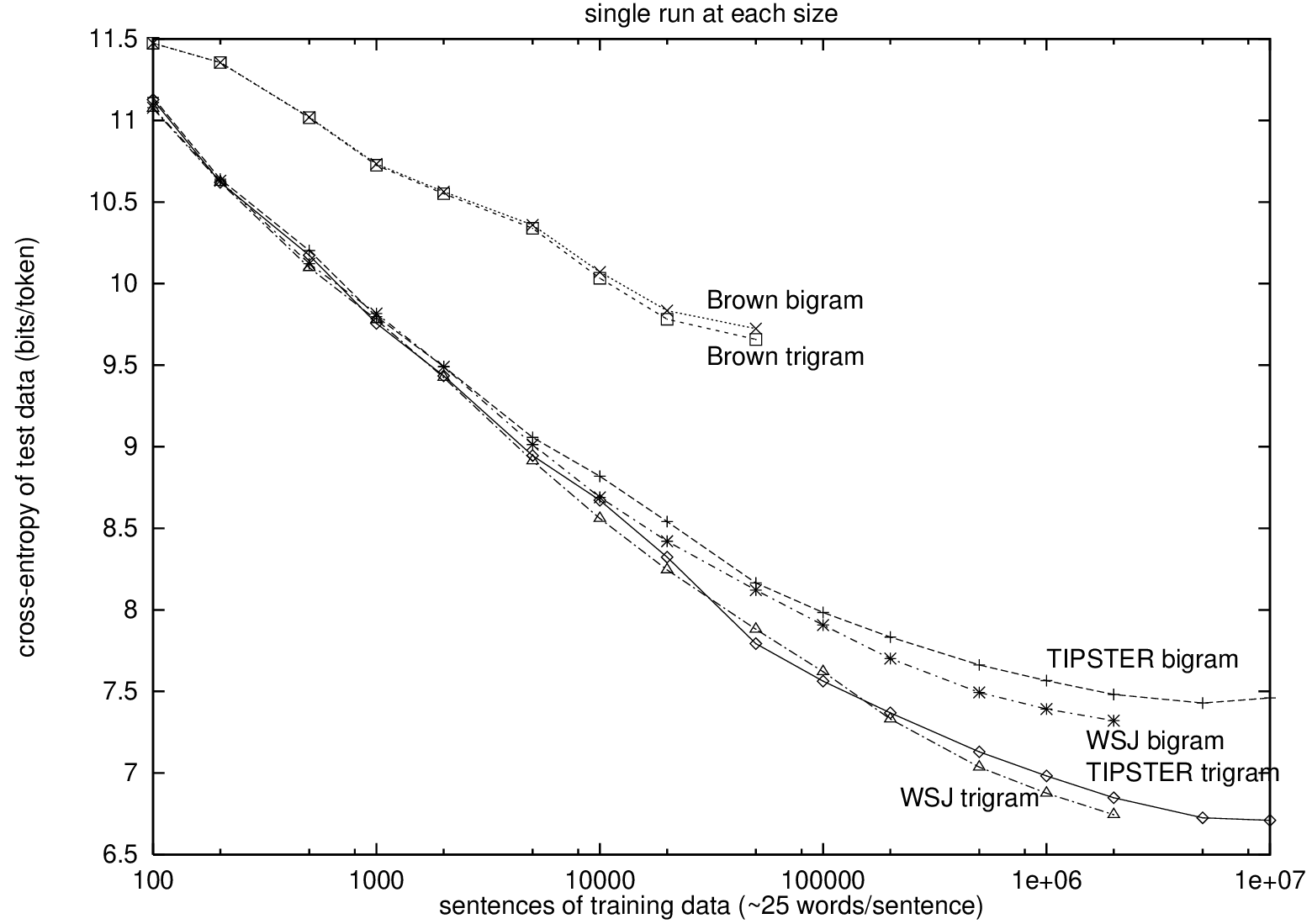,width=2.9in} $$
\caption[Baseline cross-entropy on test data]{Baseline cross-entropy
	on test data; graph on left displays
	averages over ten runs for training sets up to 50,000 sentences, graph
	on right displays single runs for training sets up to 10,000,000 sentences}
	\label{fig:base}
\end{figure}

\begin{figure}[p]
$$ \psfig{figure=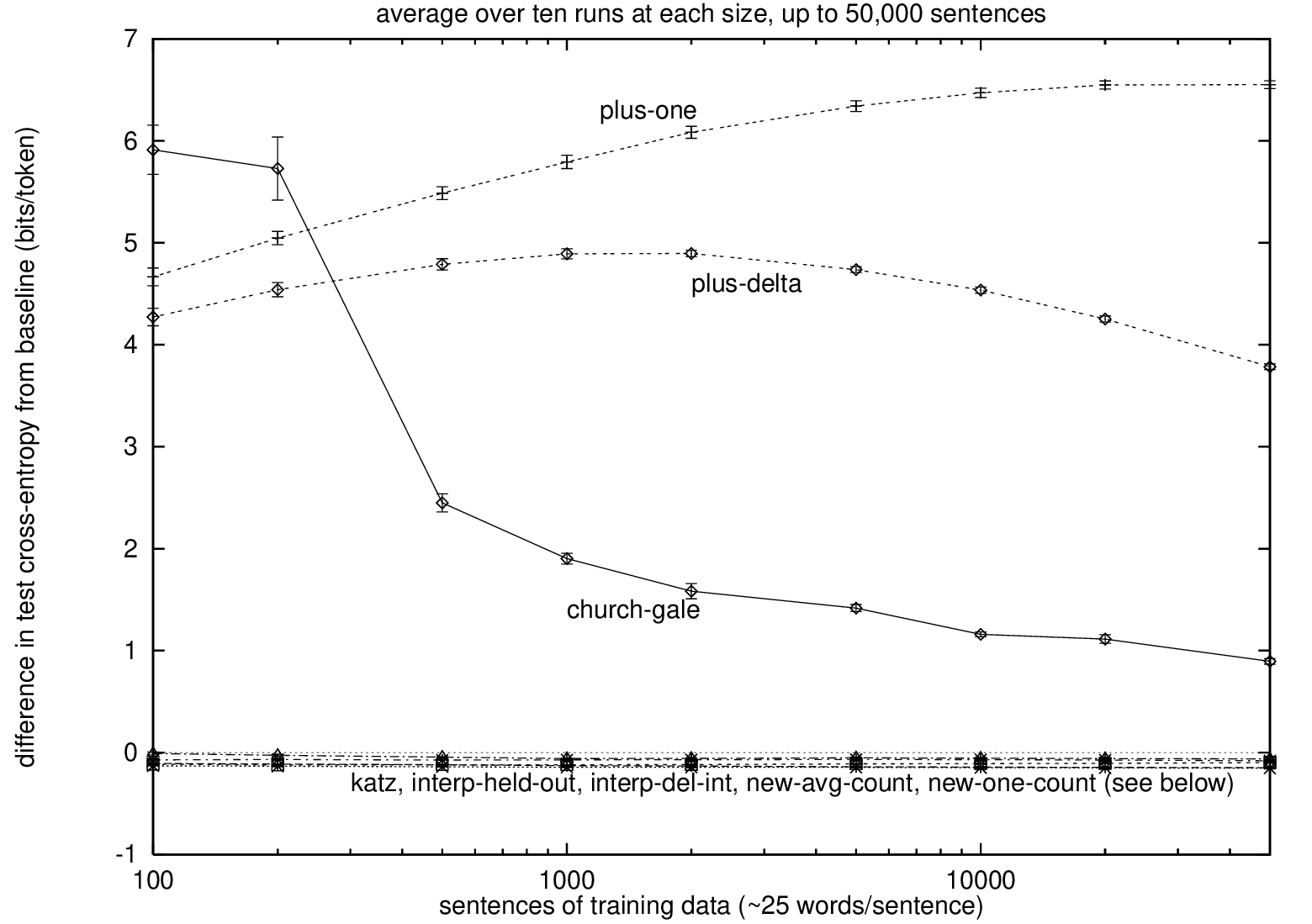,width=2.9in} \hspace{0.2in}
\psfig{figure=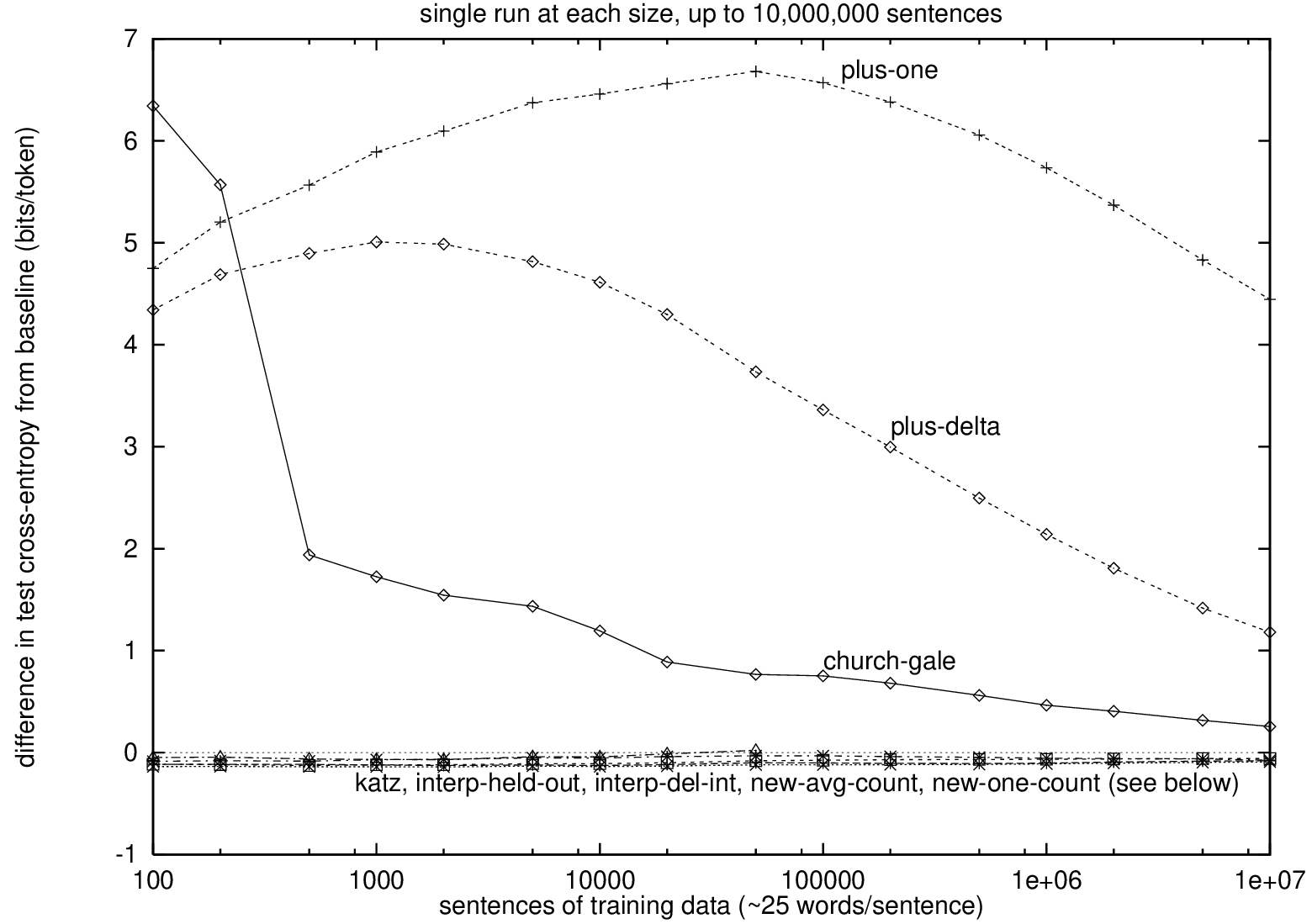,width=2.9in} $$
$$ \psfig{figure=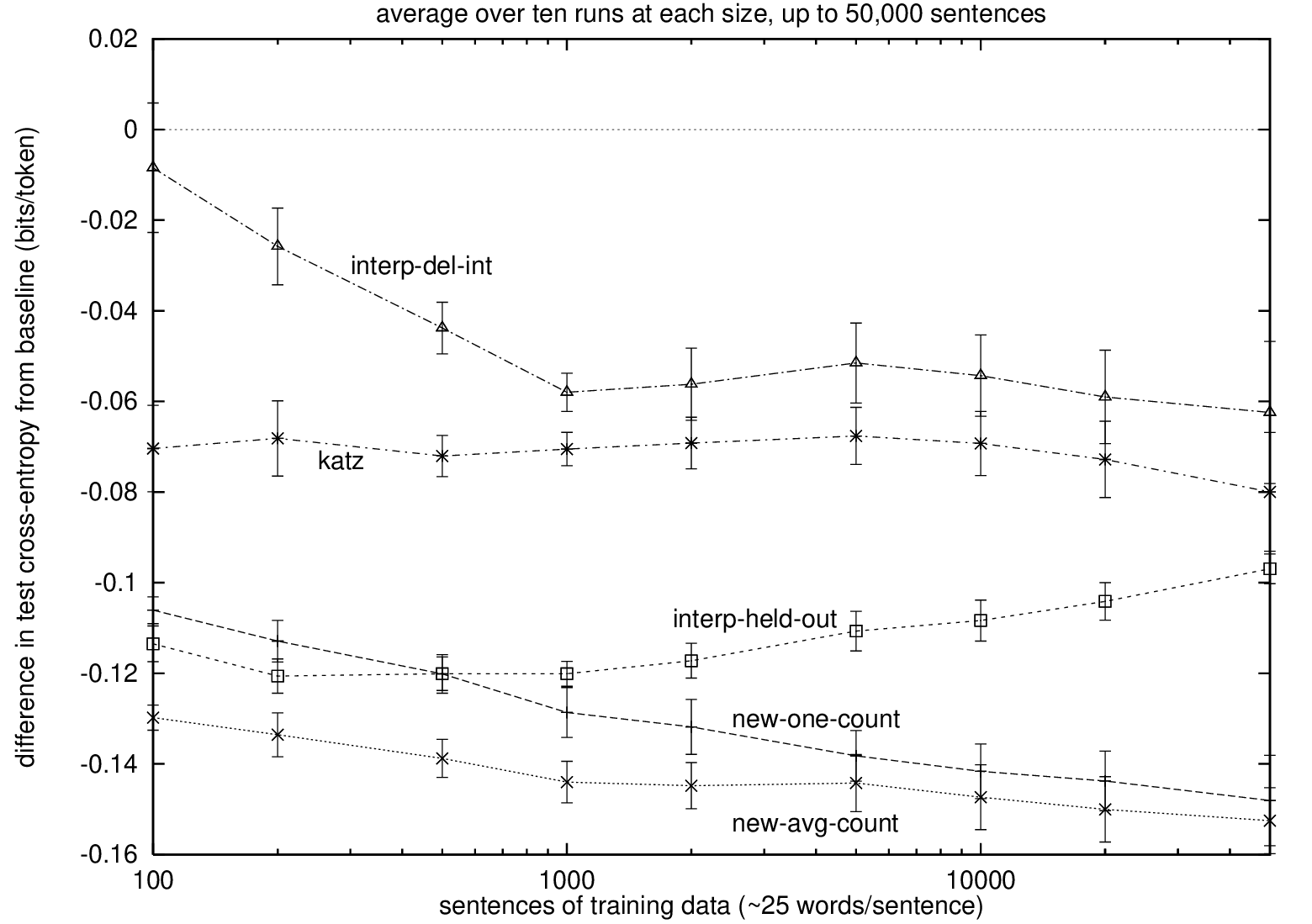,width=2.9in} \hspace{0.2in}
\psfig{figure=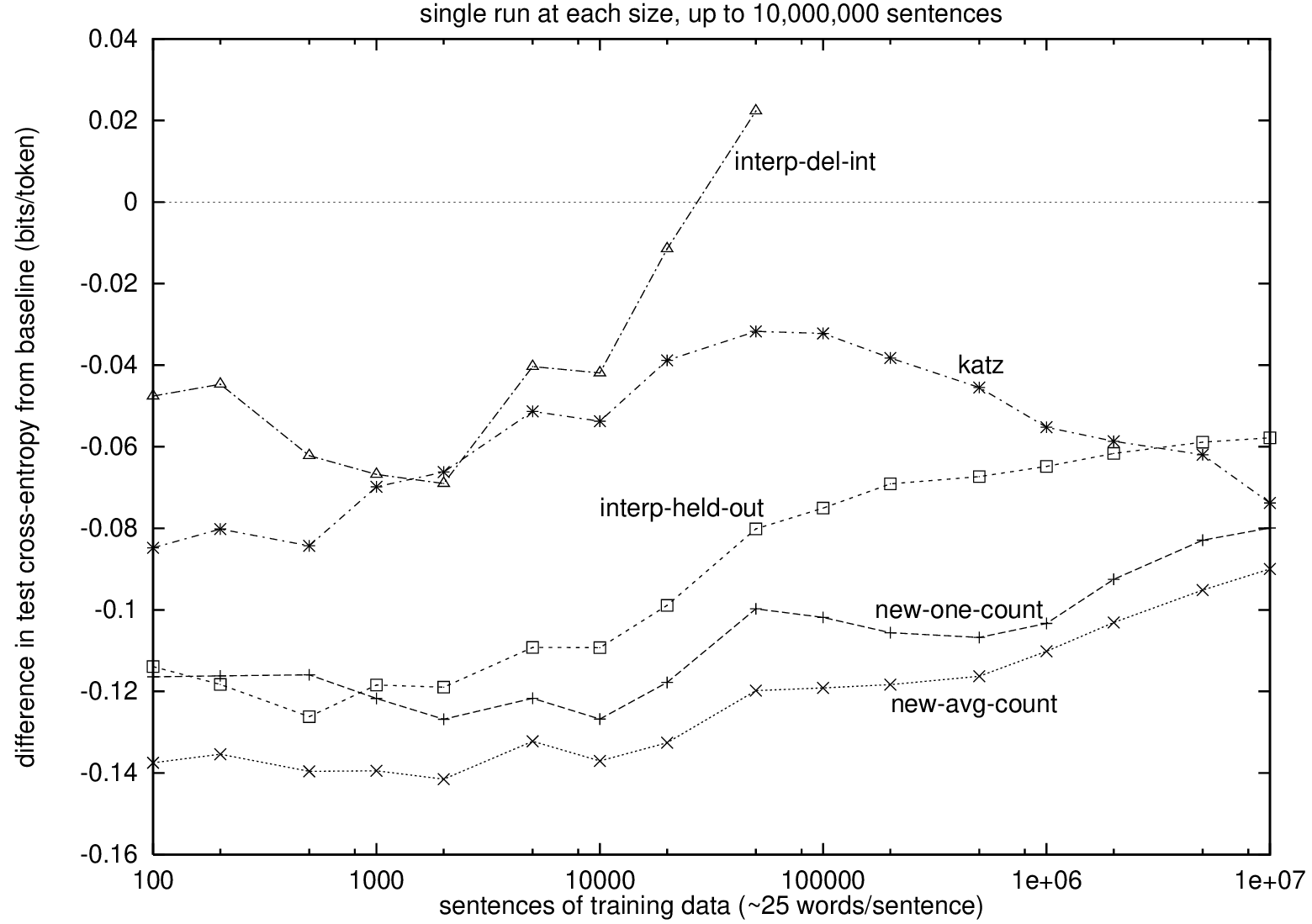,width=2.9in} $$
\caption[Trigram model on TIPSTER data;
	relative performance of various methods with respect
	to baseline]{Trigram model on TIPSTER data;
	relative performance of various methods with respect
	to baseline; graphs on left display
	averages over ten runs for training sets up to 50,000 sentences, graphs
	on right display single runs for training sets up to 10,000,000 sentences;
	top graphs show all algorithms, bottom graphs zoom in on those methods
	that perform better than the baseline method}
	\label{fig:tip3}
\end{figure}

\begin{figure}[p]
$$ \psfig{figure=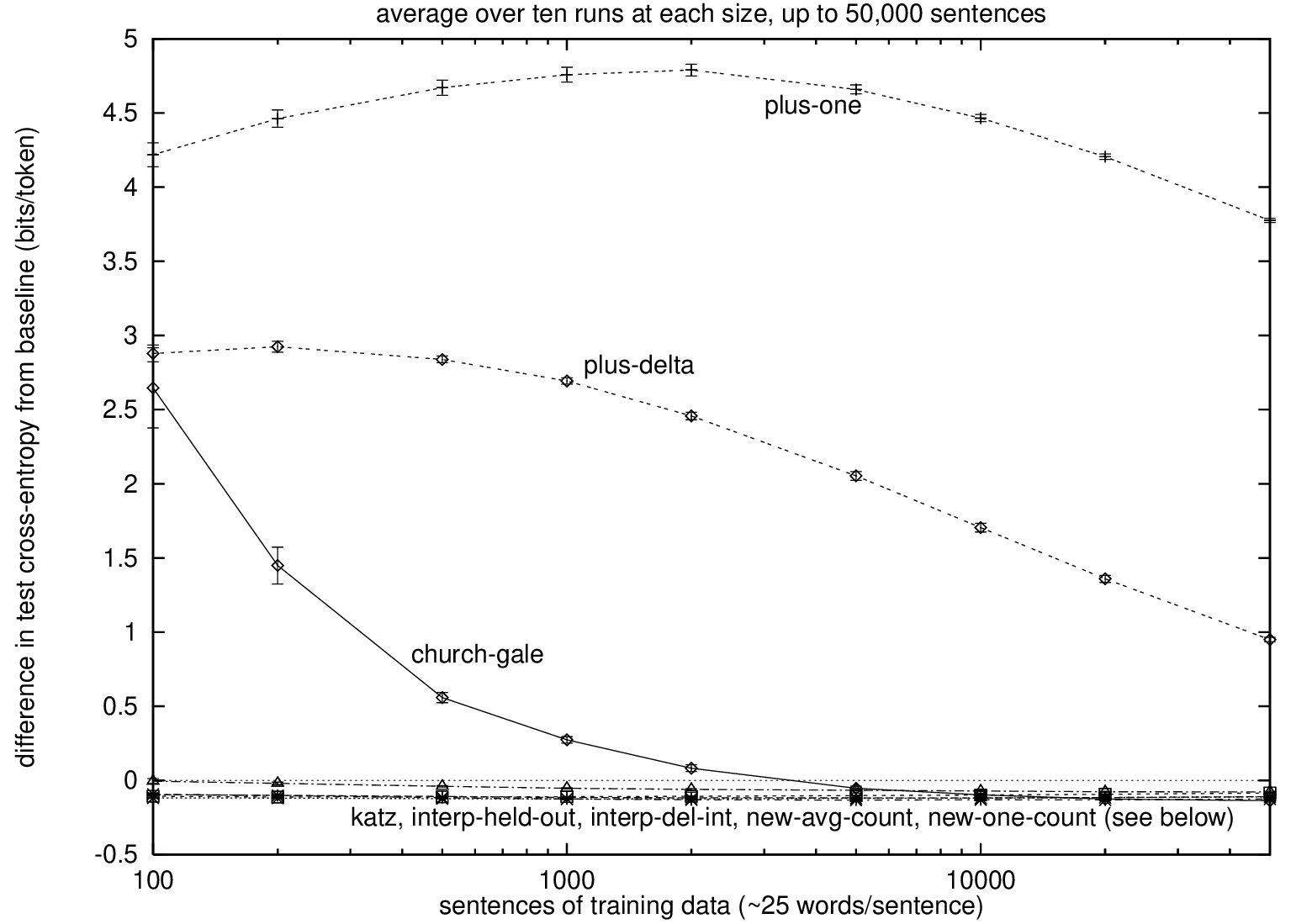,width=2.9in} \hspace{0.2in}
\psfig{figure=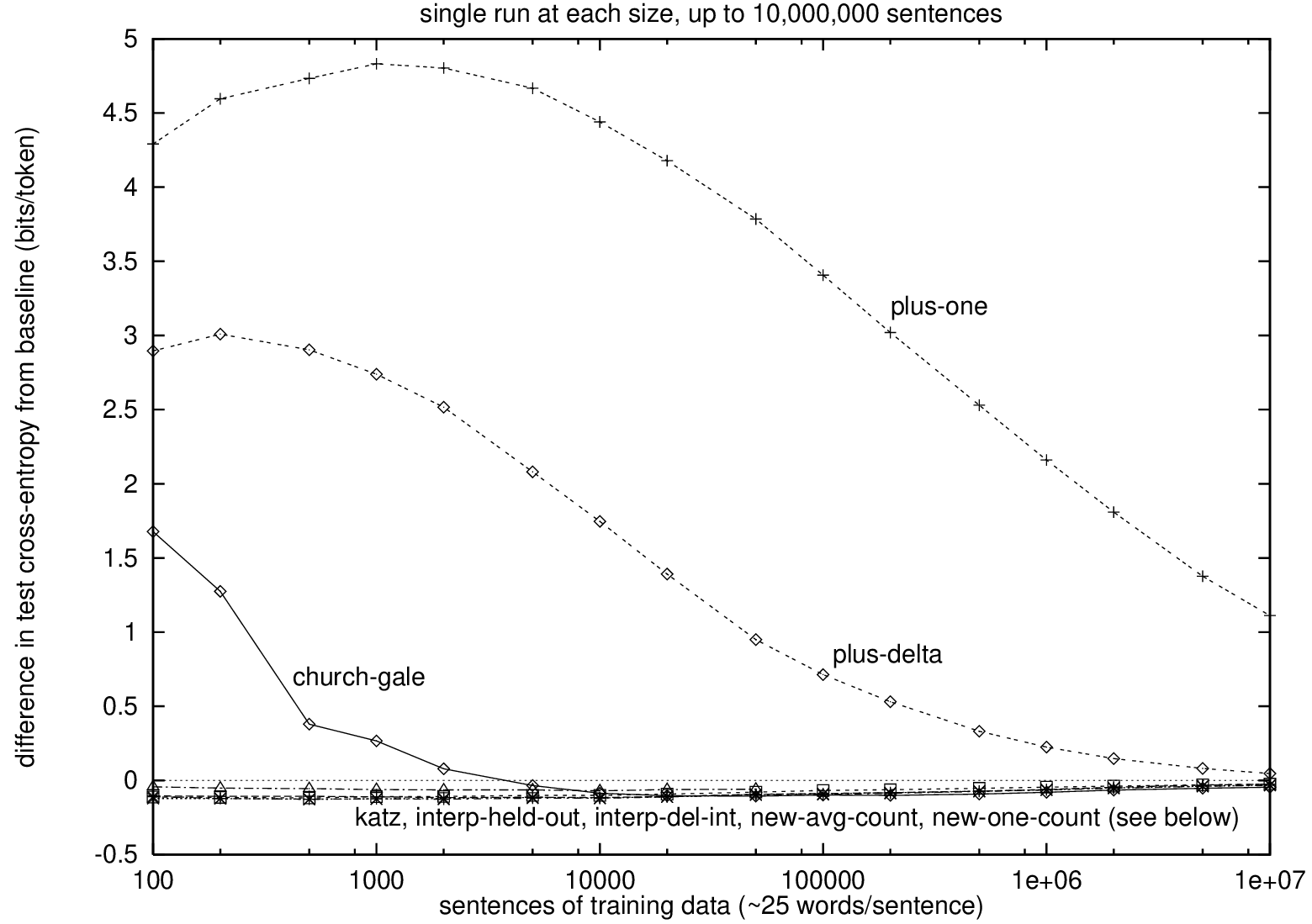,width=2.9in} $$
$$ \psfig{figure=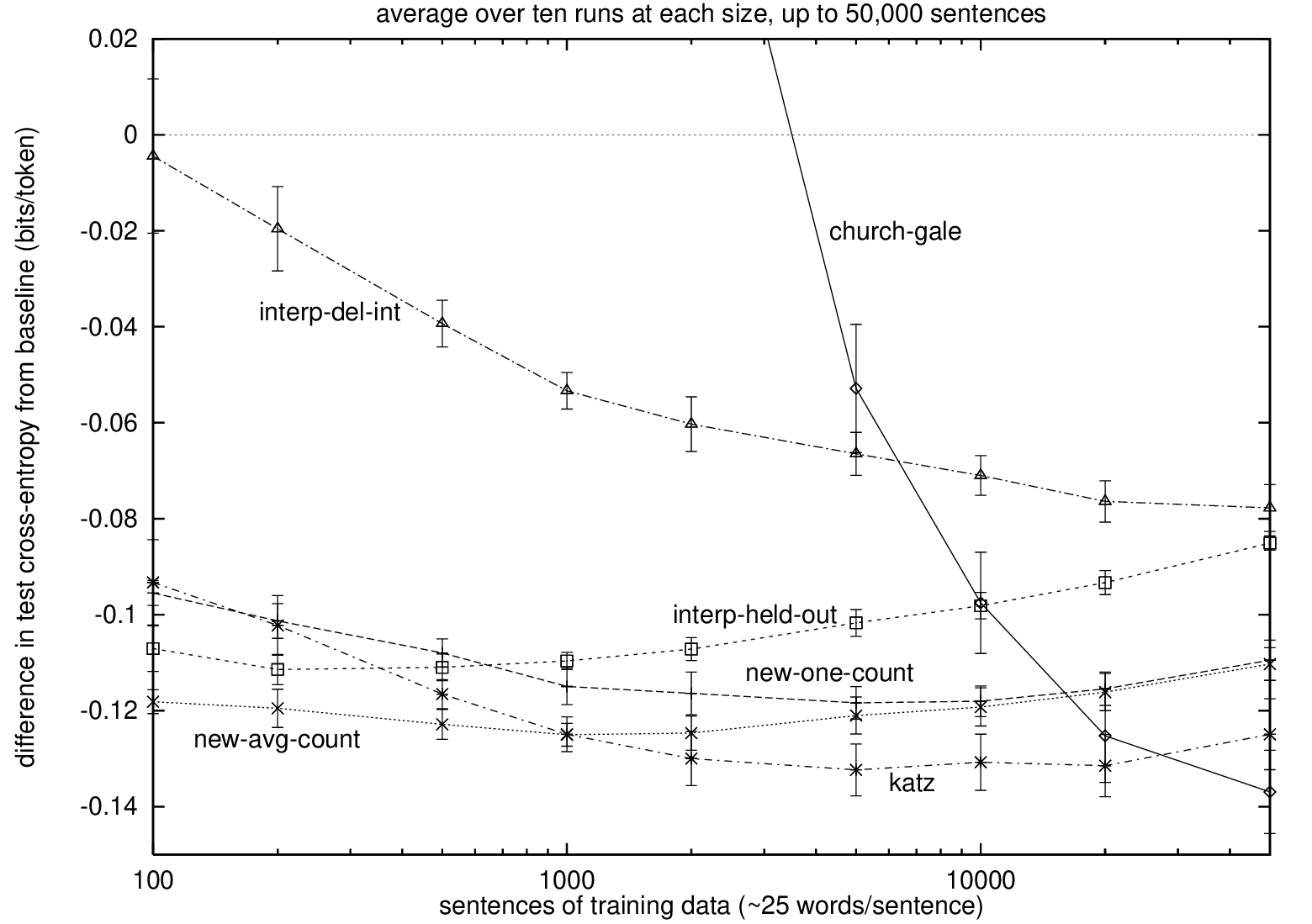,width=2.9in} \hspace{0.2in}
\psfig{figure=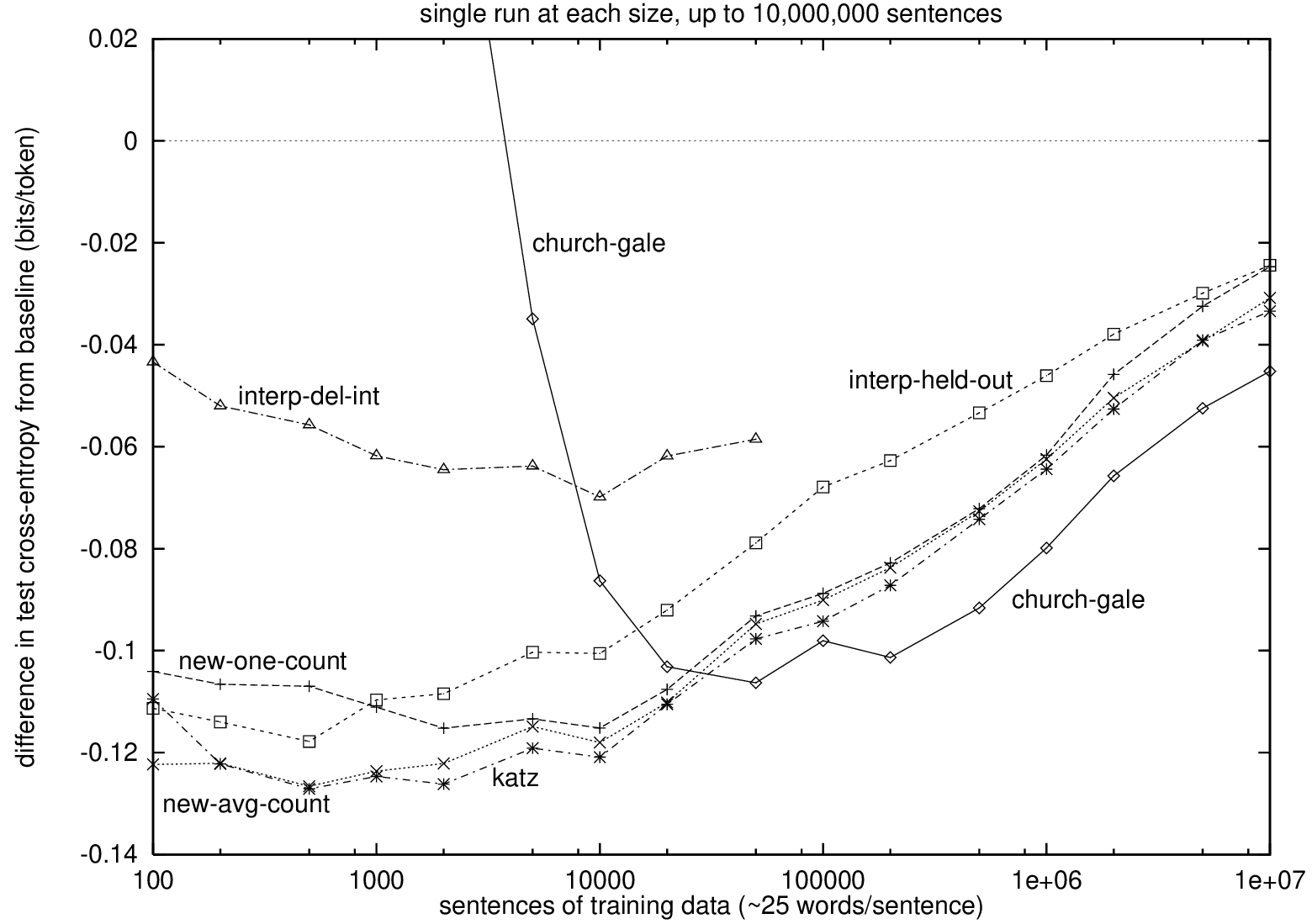,width=2.9in} $$
\caption[Bigram model on TIPSTER data;
	relative performance of various methods with respect
	to baseline]{Bigram model on TIPSTER data;
	relative performance of various methods with respect
	to baseline; graphs on left display
	averages over ten runs for training sets up to 50,000 sentences, graphs
	on right display single runs for training sets up to 10,000,000 sentences;
	top graphs show all algorithms, bottom graphs zoom in on those methods
	that perform better than the baseline method}
	\label{fig:tip2}
\end{figure}

\begin{figure}[p]
$$ \psfig{figure=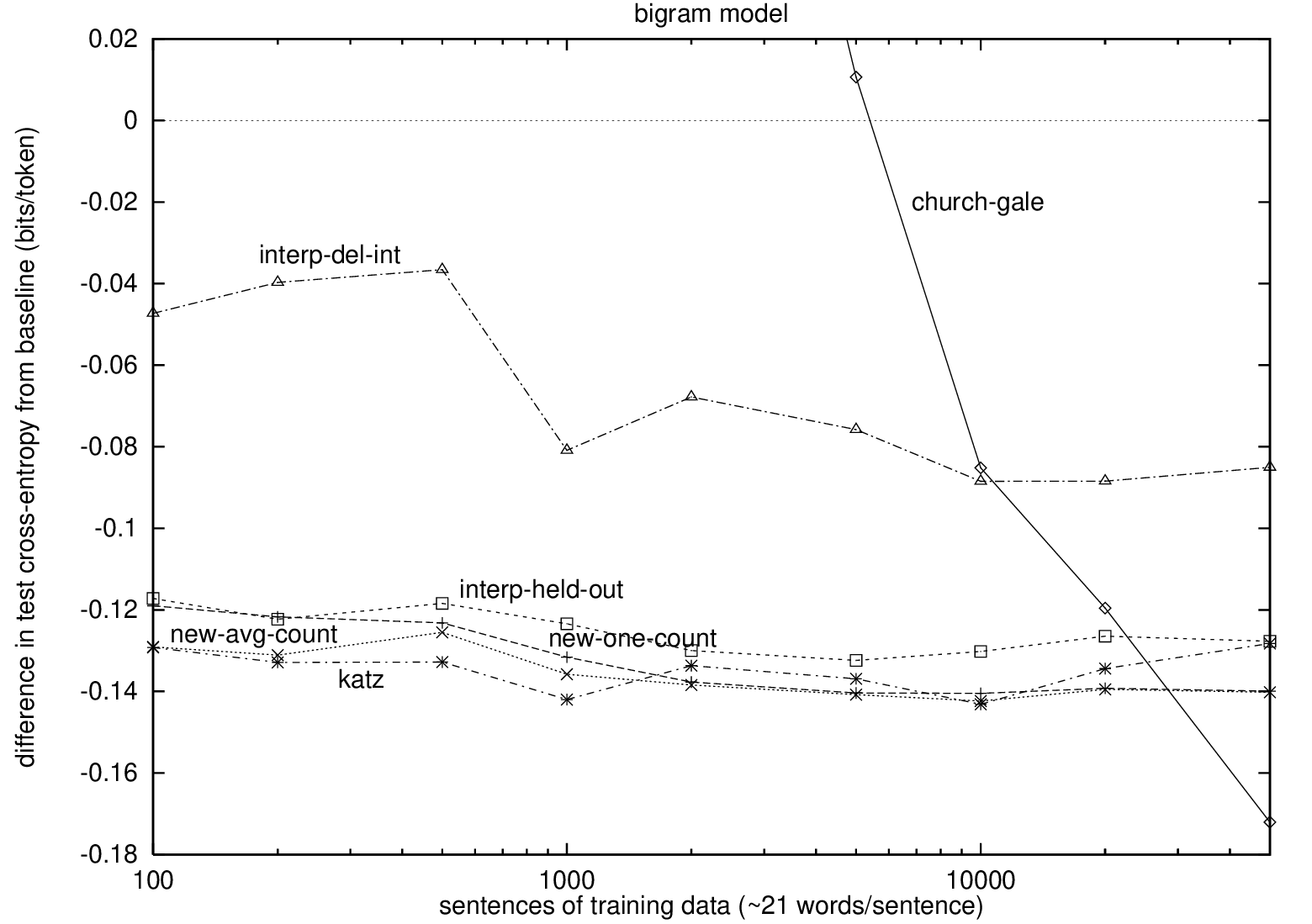,width=2.9in} \hspace{0.2in}
\psfig{figure=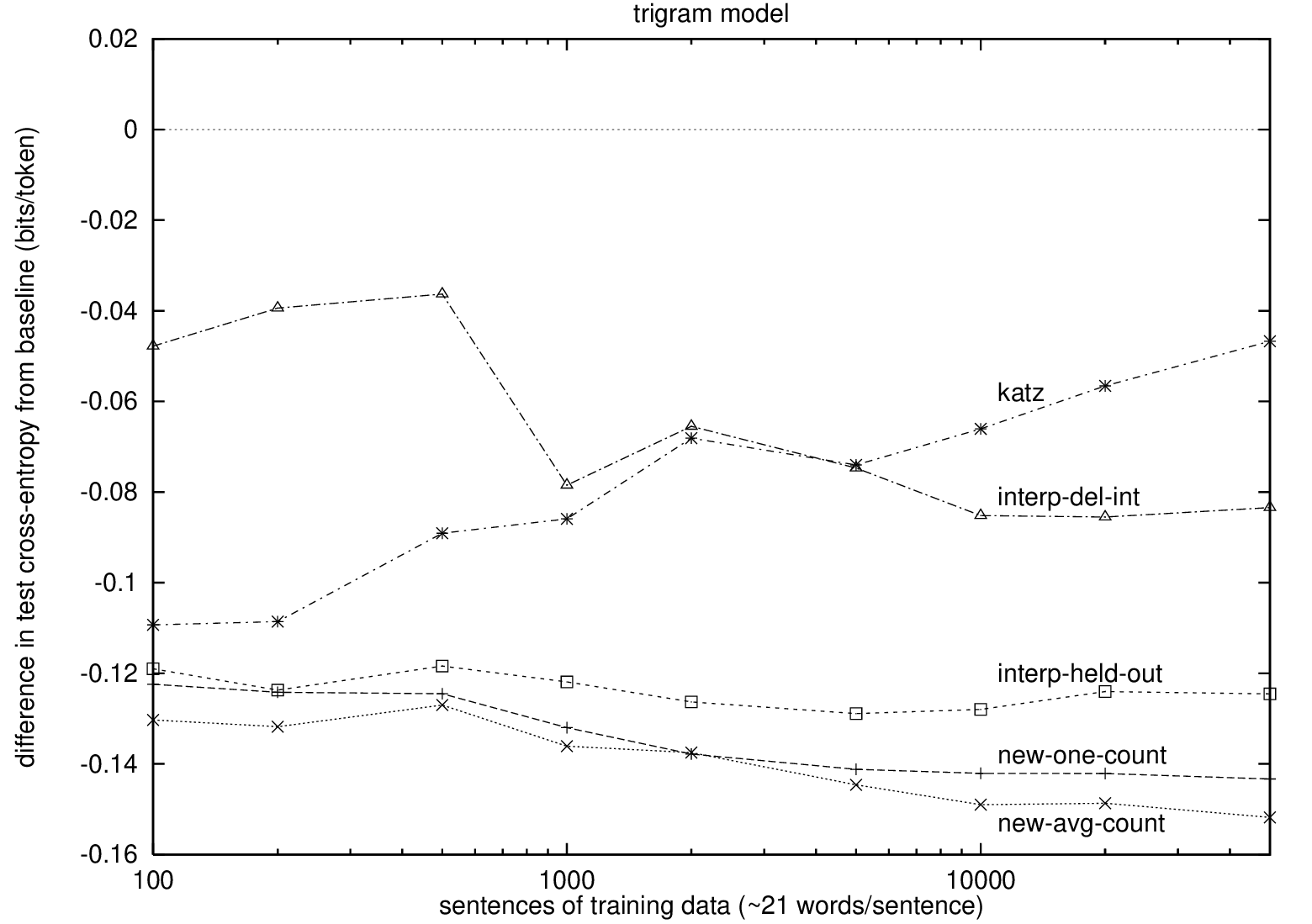,width=2.9in} $$
\caption{Bigram and trigram models on Brown corpus;
	relative performance of various methods with respect
	to baseline} \label{fig:brown}
\end{figure}

\begin{figure}
$$ \psfig{figure=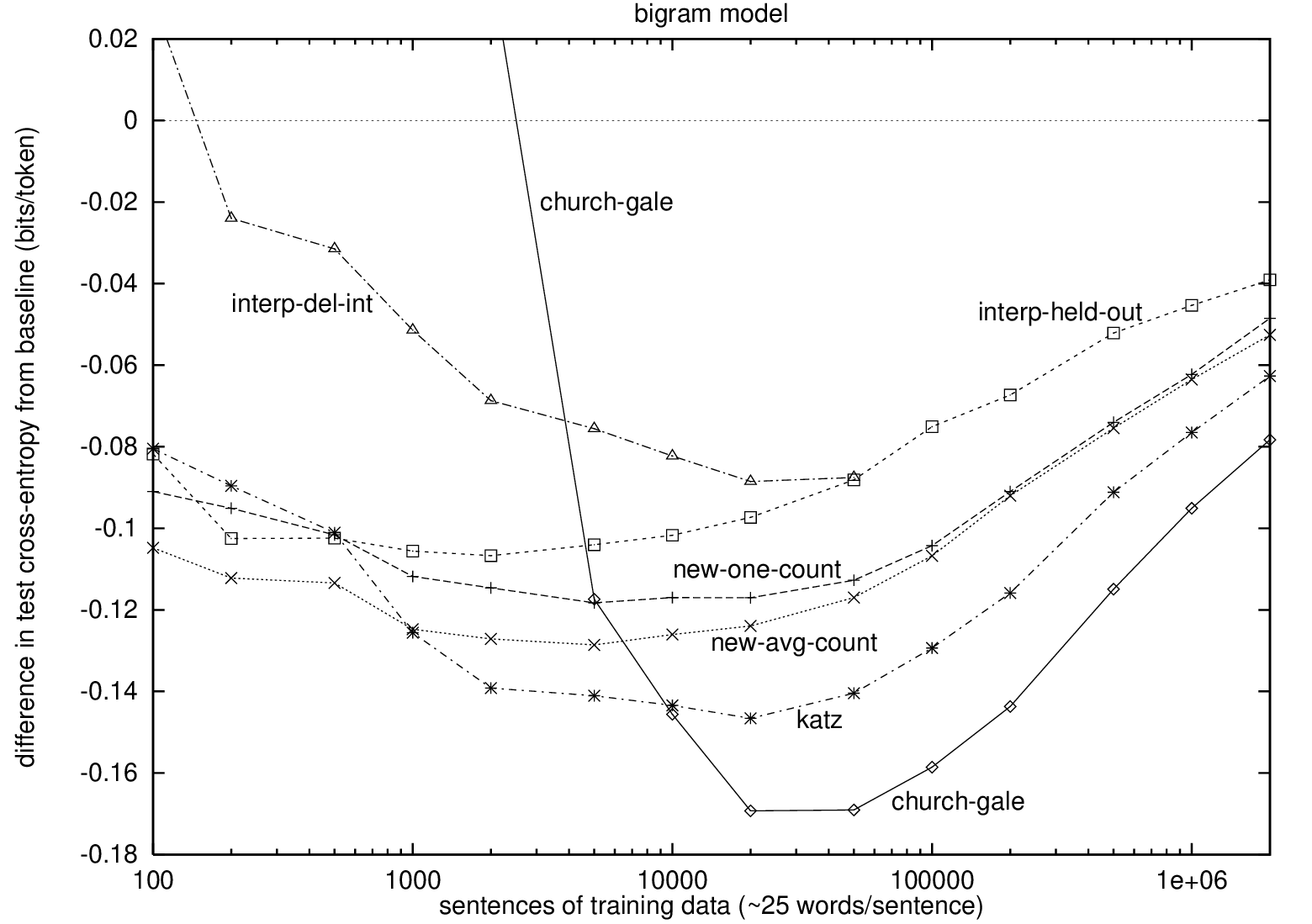,width=2.9in} \hspace{0.2in}
\psfig{figure=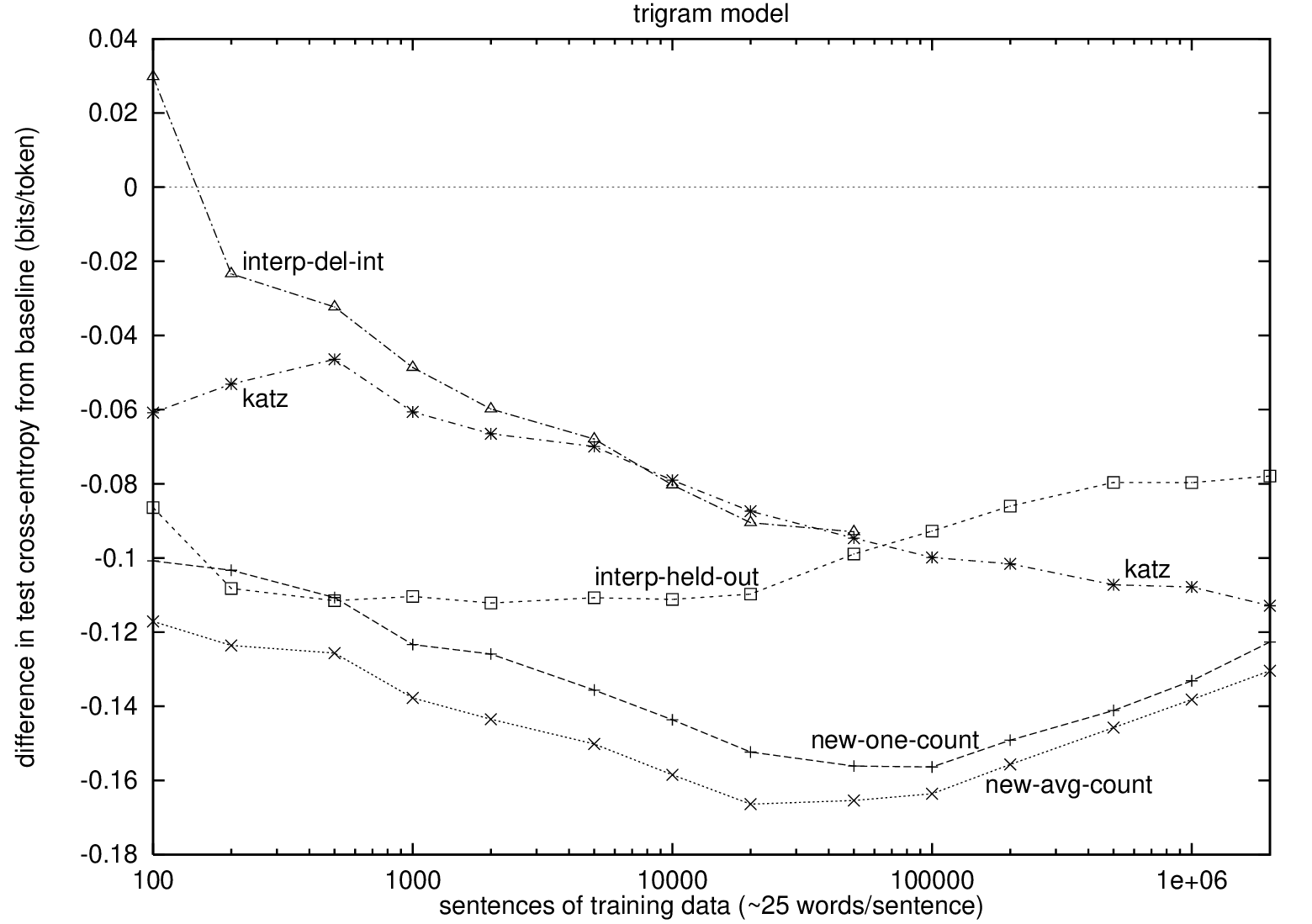,width=2.9in} $$
\caption{Bigram and trigram models on Wall Street Journal corpus;
	relative performance of various methods with respect
	to baseline} \label{fig:wsj}
\end{figure}

%
\sec{Results} \label{sec:sresults}
%

In this section, we present the results of our experiments.  First,
we present the performance of various algorithms for different training
set sizes on different corpora for both bigram and trigram models.
We demonstrate that the relative performance of smoothing methods
varies significantly over training sizes and $n$-gram order, and
we show which methods perform best in different situations.  We
find that {\tt katz} performs best for bigram models produced from
moderately-sized data sets, {\tt church-gale} performs best
for bigram models produced from large data sets, and
our novel methods {\tt new-avg-count} and {\tt new-one-count} perform
best for trigram models.

Then, we present a more detailed analysis of performance, rating different
techniques on how well they perform on $n$-grams with a particular
count in the training data, \eg, $n$-grams that have occurred exactly
once in the training data.  We find that {\tt katz} and
{\tt church-gale} most accurately smooth $n$-grams with large counts,
while {\tt new-avg-count} and {\tt new-one-count} are best for small counts.
We then show the relative impact on performance of small counts
and large counts
for different training set sizes and $n$-gram orders, and use this
data to explain the variation in performance of different algorithms in
different situations.

Finally, we examine three miscellaneous points: the accuracy of
the Good-Turing estimate in smoothing $n$-grams with zero counts,
how Church-Gale smoothing compares to linear interpolation,
and how deleted interpolation compares with held-out interpolation.

\ssec{Overall Results}

In Figure \ref{fig:base}, we display
the performance of the {\tt interp-baseline}
method for bigram and trigram models on TIPSTER, Brown, and the WSJ subset
of TIPSTER.  In Figures \ref{fig:tip3}--\ref{fig:wsj},
we display the
relative performance of various smoothing techniques with respect to
the baseline method on these corpora, as measured by difference
in entropy.  In the
graphs on the left of Figures \ref{fig:base}--\ref{fig:tip2}, each
point represents an average over ten runs; 
the error bars represent the empirical standard deviation over these runs.
Due to resource limitations, we only performed multiple runs for
data sets of 50,000 sentences or less.  Each point on the graphs
on the right represents a single run, but we consider
training set sizes up to the amount
of data available, \eg, up to 250M words on TIPSTER.
The graphs on the bottom of Figures
\ref{fig:tip3}--\ref{fig:tip2} are close-ups of the graphs above, focusing
on those algorithms that perform better than the baseline.  We ran
{\tt interp-del-int} only on training sets up to 50,000
sentences due to time constraints.  To give an idea of how these
cross-entropy differences translate to perplexity, each 0.014 bits
correspond roughly to a 1\% change in perplexity.

From these graphs, we see that additive smoothing performs poorly and that
the methods {\tt katz} and {\tt interp-held-out} consistently perform well,
with {\tt katz} performing the best of all algorithms on small
bigram training sets.
The implementation {\tt church-gale} performs poorly except on large
bigram training sets, where it performs the best.
The novel methods {\tt new-avg-count} and
{\tt new-one-count} perform well uniformly
across training data sizes, and are superior for trigram models.
Notice that while performance is relatively consistent across corpora,
it varies widely with respect to training set size and $n$-gram order.

\ssec{Count-by-Count Analysis} \label{ssec:scount}

To paint a more detailed picture of performance, we consider the
performance of different models on only those $n$-grams in
the test data that have
exactly $r$ counts in the training data, for small values of $r$.
This analysis provides information as to whether a model assigns the
correct amount of probability to categories such
as $n$-grams with zero counts,
$n$-grams with low counts, or $n$-grams with high counts.
In these experiments, we use about 10 million words of test data.

\begin{figure}
$$ \psfig{figure=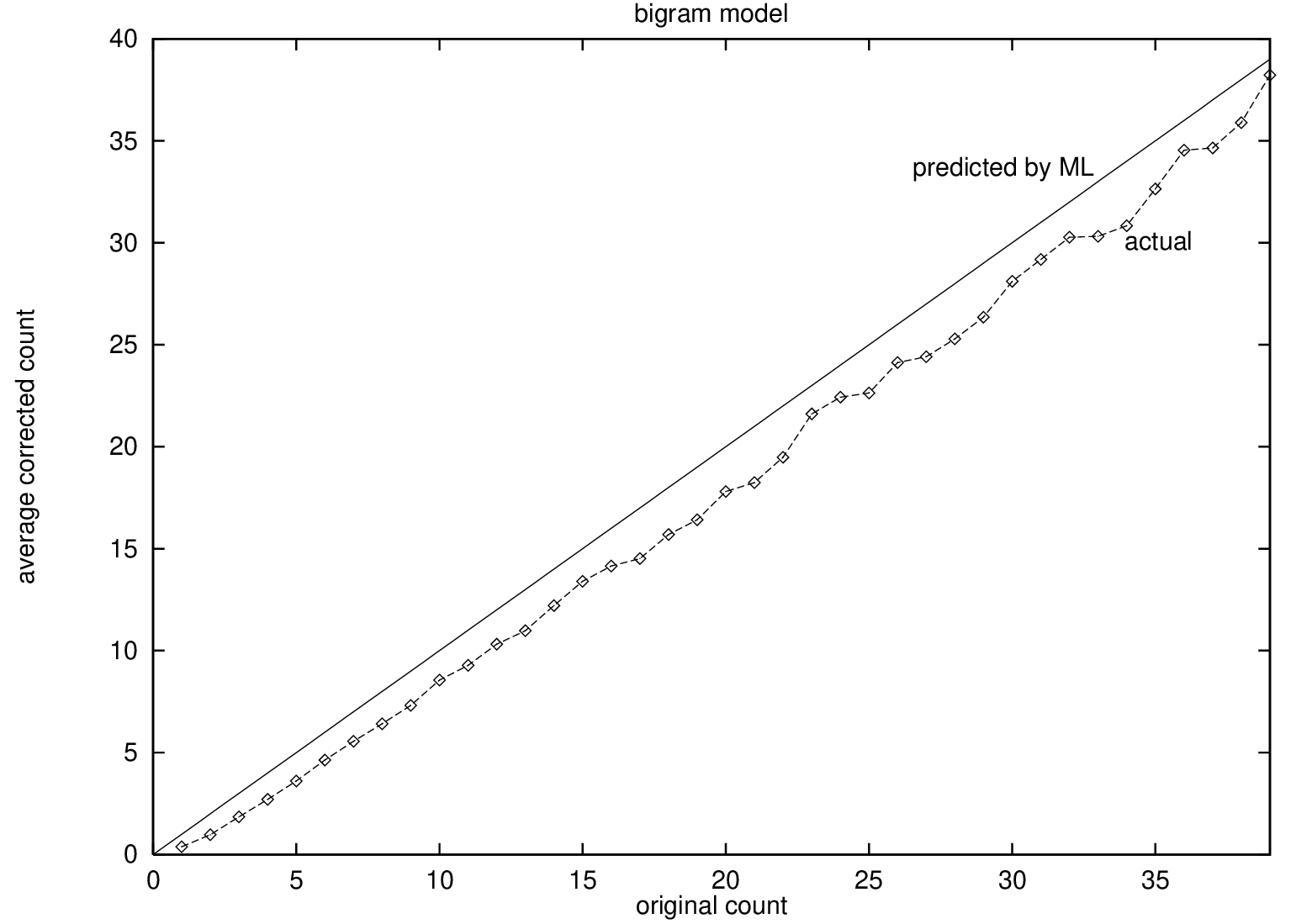,width=2.9in} \hspace{0.2in}
\psfig{figure=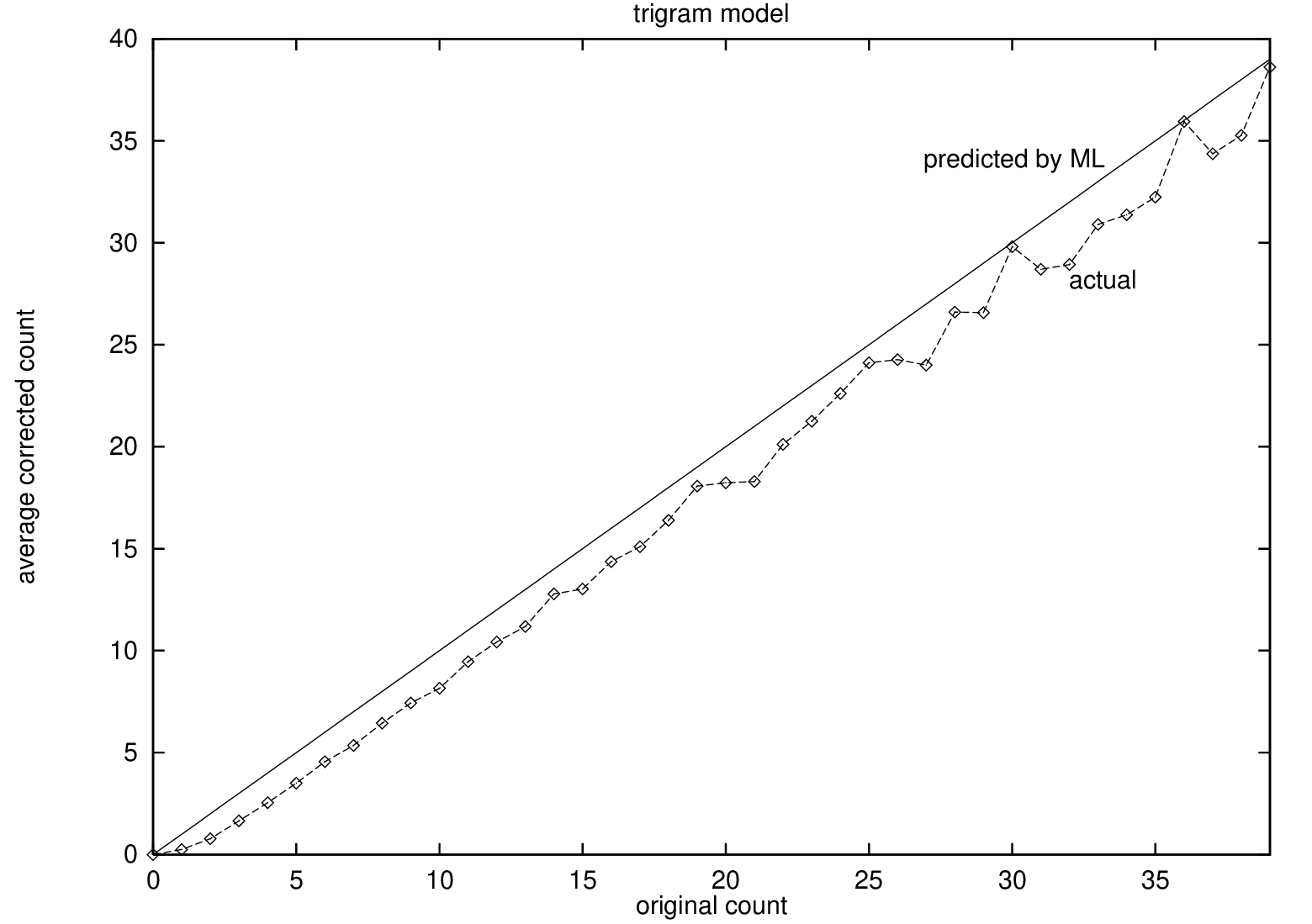,width=2.9in} $$
\caption{Average corrected counts for bigram and trigram models, 1M words
	training data}
	\label{fig:disca}

$$ \psfig{figure=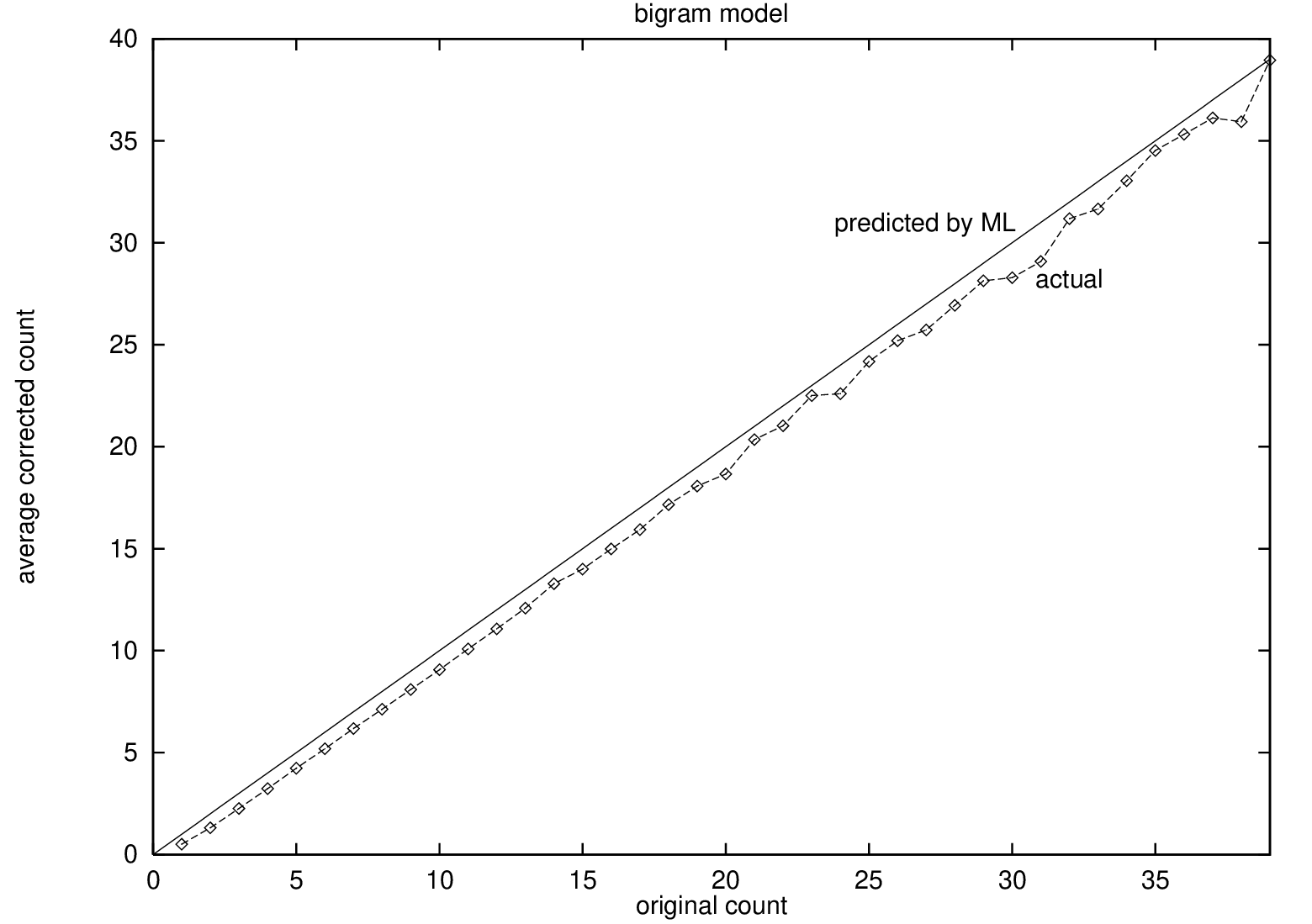,width=2.9in} \hspace{0.2in}
\psfig{figure=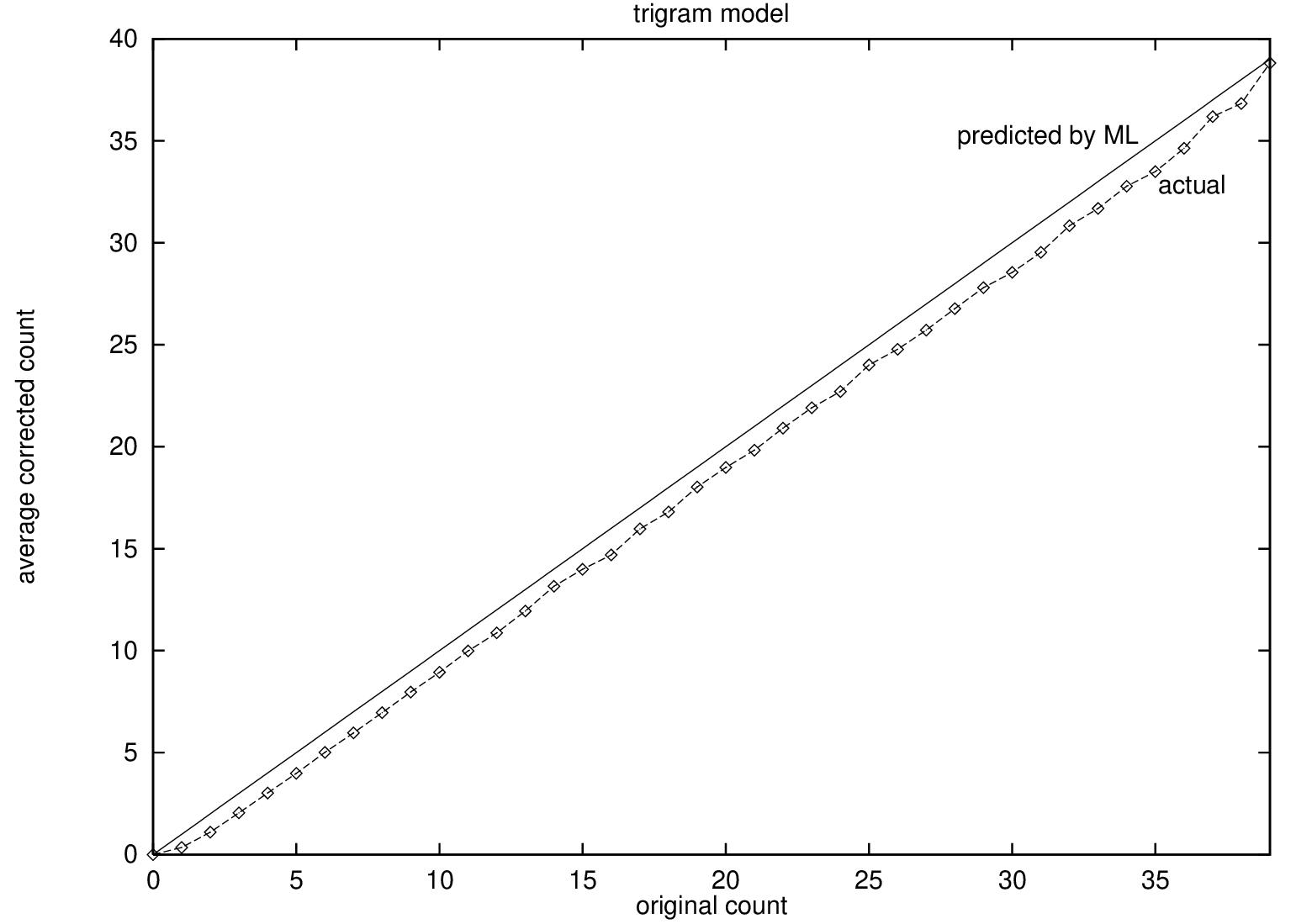,width=2.9in} $$
\caption{Average corrected counts for bigram and trigram models, 200M words
	training data}
	\label{fig:discb}
\end{figure}

First, we consider whether various smoothing methods assign
{\it on average\/} the correct discounted count $r^*$ for a
given count $r$.  The discounted count $r^*$ generally
varies for different $n$-grams; we only consider its average value here.
(Recall that the corrected count of an $n$-gram is proportional
to the probability assigned by a model to that $n$-gram; in particular,
in this discussion we assume that the probability assigned to an
$n$-gram $w_{i-n+1}^i$ with $r$ counts is just its normalized
corrected count $\frac{r^*}{N}$, where the normalization constant $N$
is equal to the original total count $\sum_{w_i} c(w_{i-n+1}^i)$.)
To calculate how closely a model comes to assigning the correct average $r^*$,
we compare the expected value of the number of times $n$-grams
with $r$ counts occur in the test data with
the actual number of times these $n$-grams occur.  When the
expected and actual counts agree, this corresponds to assigning the
correct average value of $r^*$.

We can estimate the actual correct average $r_0^*$ for a given count $r$ by
using the following formula:
$$ r_0^* = r \times \frac{\mbox{actual number of $n$-grams with
	$r$ counts in the test data}}{\mbox{expected number according to the
	maximum likelihood model}} $$
The maximum likelihood model represents the case where we take the
corrected count to just be the original count.
In Figure \ref{fig:disca}, we display the desired
average corrected count for each count less than 40, for 1M words of
training data from TIPSTER.

The last point in the graph corresponds to the
average discount for that count and all higher counts.  (This
property holds for later graphs, that the last point corresponds to that count
and all higher counts.)  The solid line corresponds to the maximum
likelihood model where the corrected count is taken to be equal
to the original count.
In Figure \ref{fig:discb}, we display the same graph except
for 200M words of training data from TIPSTER.

\begin{figure}
$$ \psfig{figure=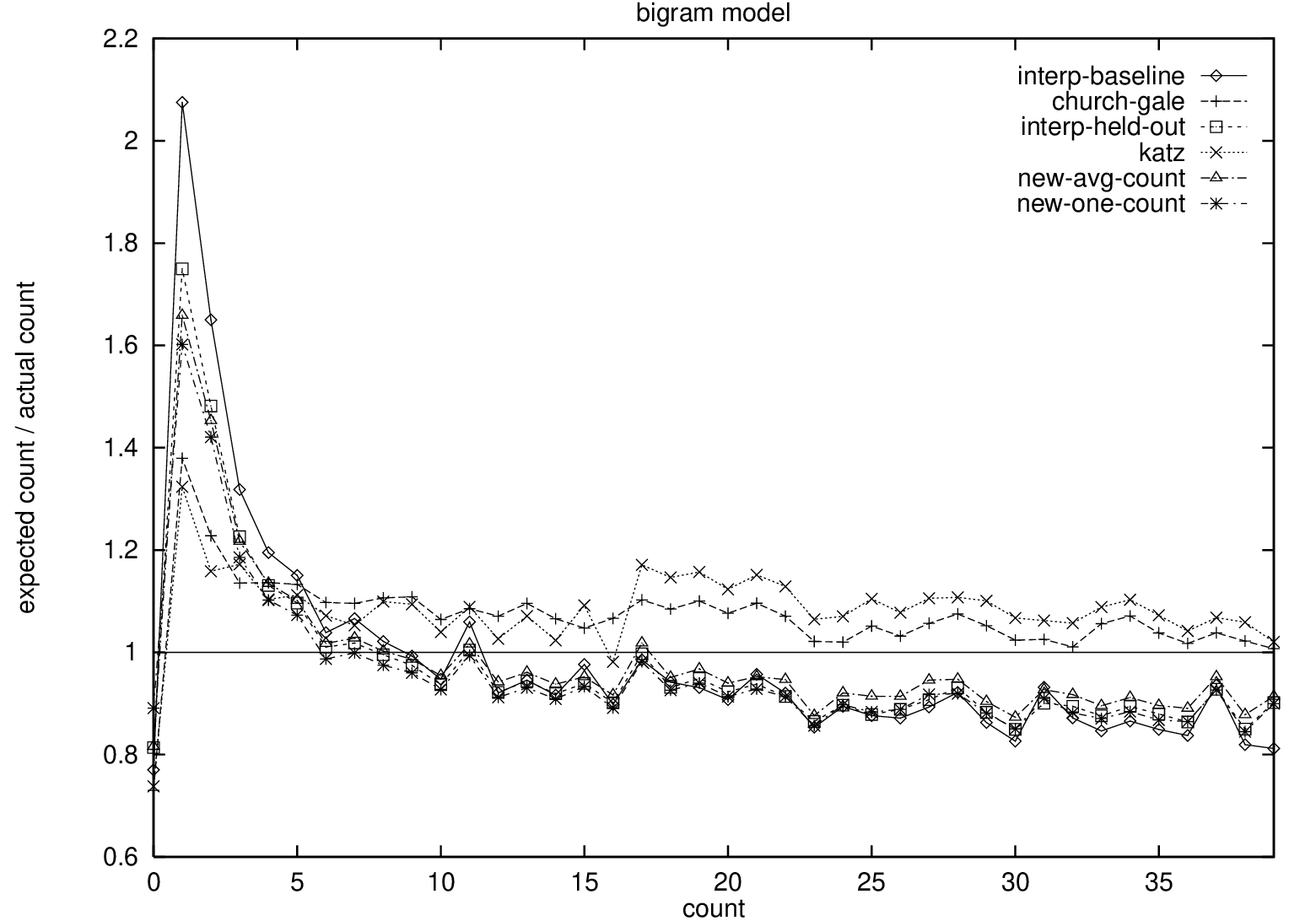,width=2.9in} \hspace{0.2in}
\psfig{figure=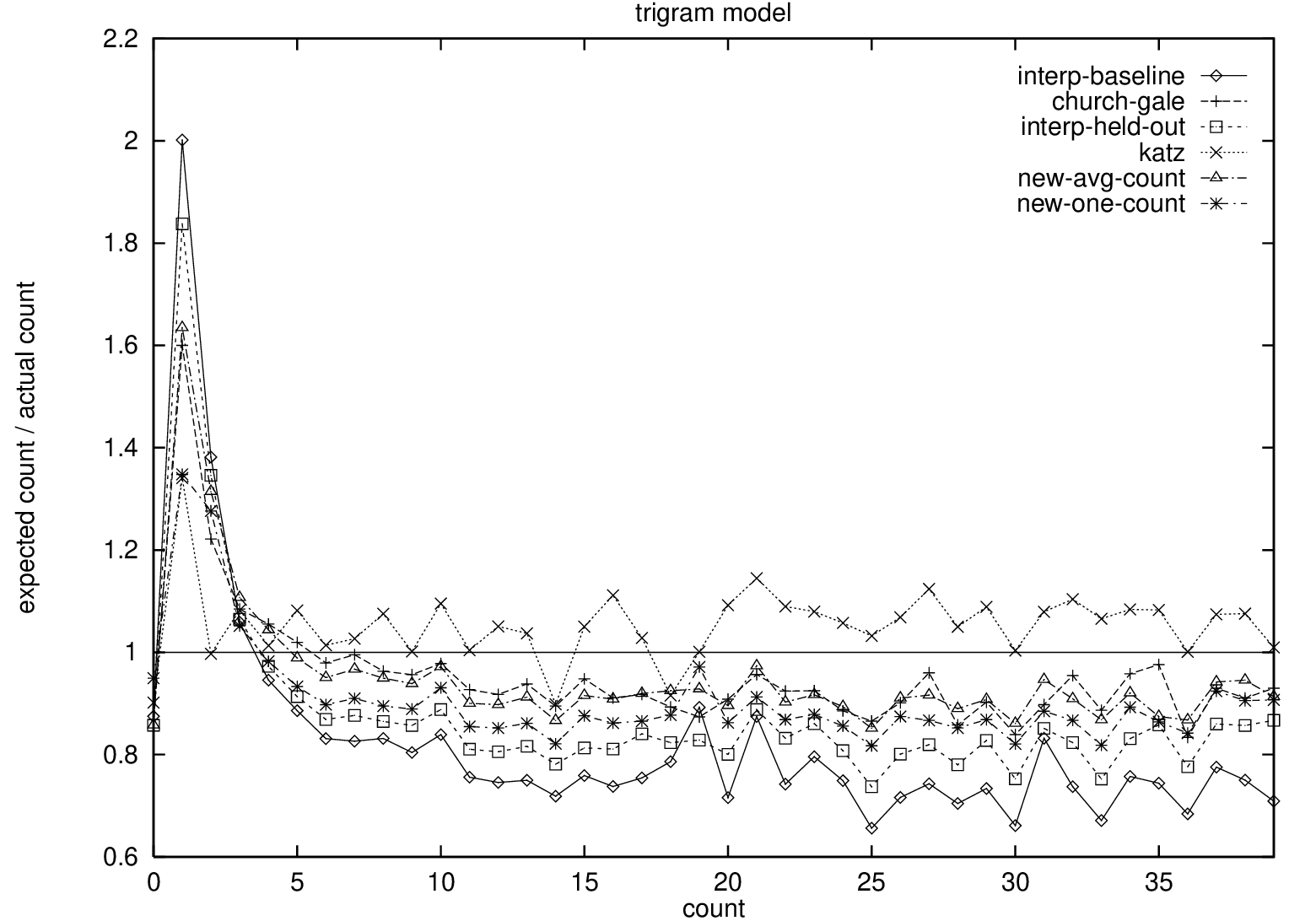,width=2.9in} $$
\caption{Expected over actual counts for various algorithms,
	bigram and trigram models, 1M words training data}
	\label{fig:expacta}

$$ \psfig{figure=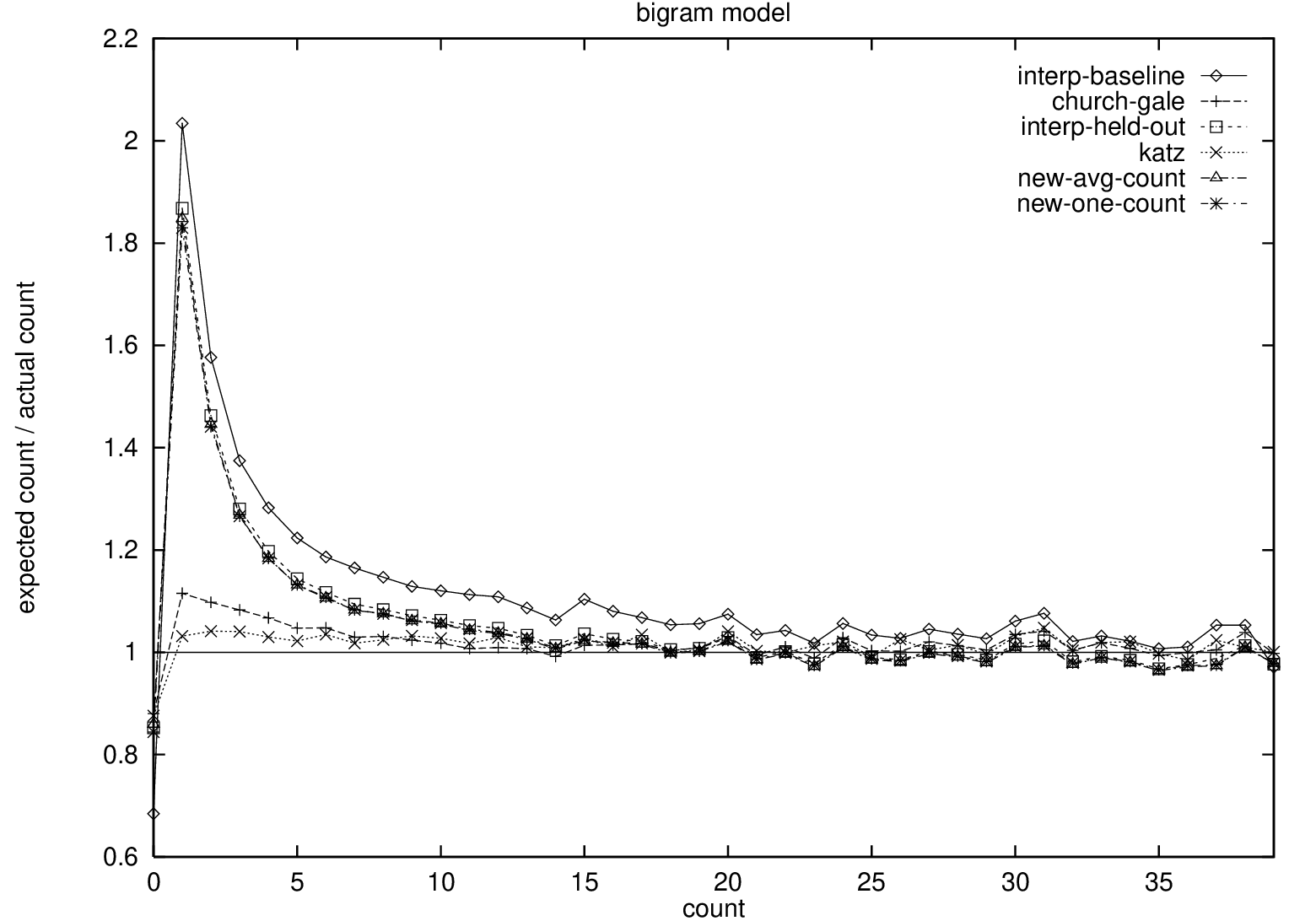,width=2.9in} \hspace{0.2in}
\psfig{figure=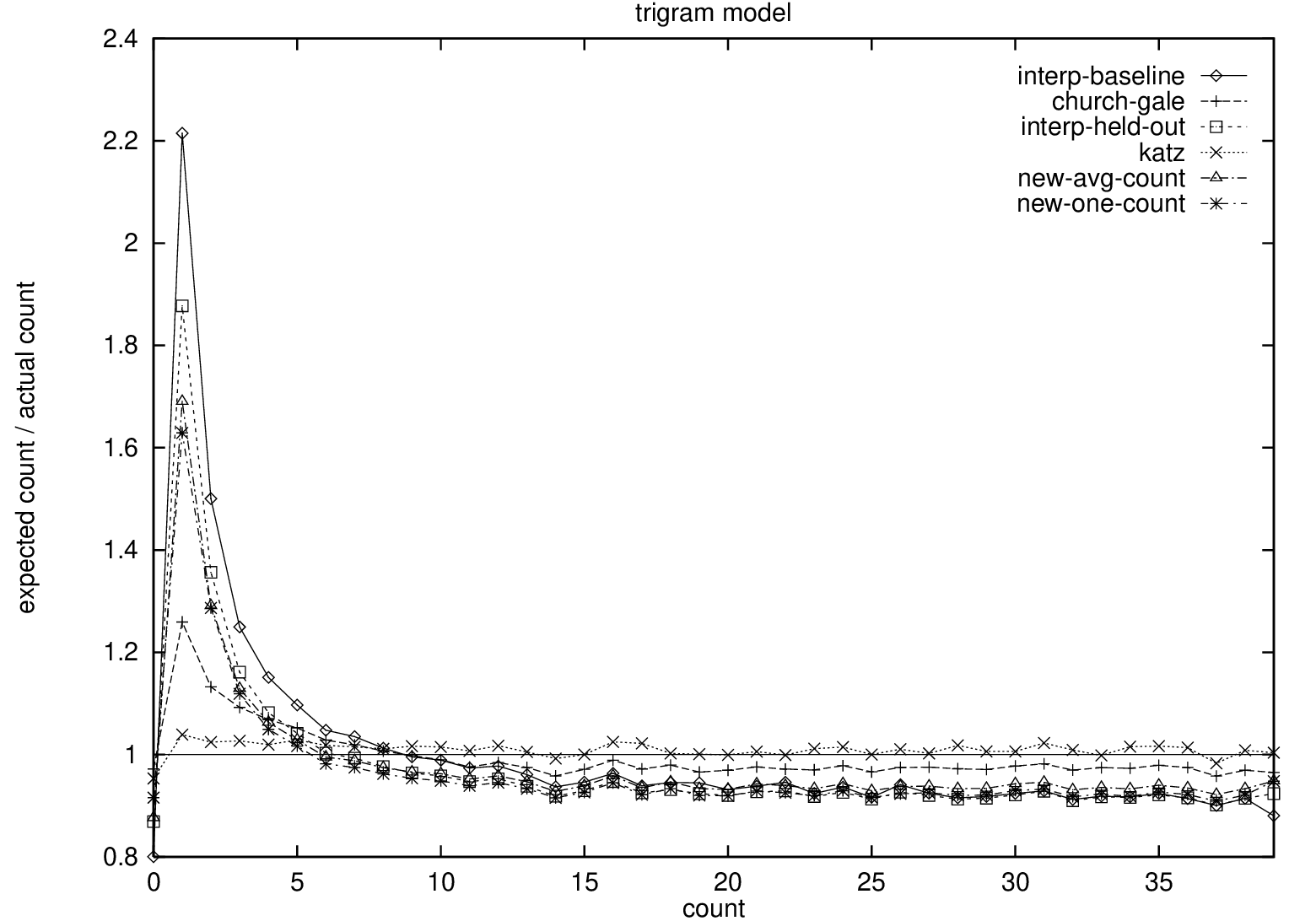,width=2.9in} $$
\caption{Expected over actual counts for various algorithms,
	bigram and trigram models, 200M words training data}
	\label{fig:expactb}
\end{figure}

In Figures \ref{fig:expacta} and \ref{fig:expactb}, we display how
close various smoothing methods came to the desired average corrected count,
again using 1M and 200M words of training data from TIPSTER.  For each
model, we graph the ratio of the actual average corrected count assigned
by the model to the ideal average corrected count.
For the zero count case, we exclude those
$n$-grams $w_{i-n+1}^i$  that occur in distributions that have
a {\it total\/} of zero counts, \ie, $\sum_{w_i} c(w_{i-n+1}^i) = 0$.
For these $n$-grams, the corrected count should be zero since
the total count is zero.  (These $n$-grams are also excluded
in later graphs.)

We see that all of the algorithms tested tend to assign slightly
too little probability to $n$-grams with zero counts, and
significantly too much probability to $n$-grams with one count.
For high counts, the algorithms tend to assign counts closer to
the correct average count on the larger training set than on
the smaller training set.  This effect did not hold for low counts.

The algorithms {\tt katz} and {\tt church-gale} consistently come
closest to assigning the correct average amount of probability to larger counts,
with {\tt katz} doing especially well.
Thus, we conclude that the Good-Turing estimate is a useful tool for
accurately estimating the desired average corrected count,
as both Katz and Church-Gale smoothing use this estimate.\footnote{This
only applies to Katz if a large $k$ is used, as counts above $k$ are
not discounted.}

In contrast, we see that methods involving linear interpolation
are not as accurate on large counts, overdiscounting large
counts in three of the experiments,
and underdiscounting large counts in the 1M word bigram experiment.
Roughly speaking, linear interpolation
corresponds to linear discounting; that is, a
corrected count $r^*$ is about $\l$
times the original count $r$ in Jelinek-Mercer smoothing.  Referring
to Figures \ref{fig:disca}
and \ref{fig:discb}, it is clear that the desired average corrected count is
not a constant multiplied by the original count; a more
accurate description is {\it fixed discounting}, that the corrected count
is the original count less a constant.

\begin{figure}
$$ \psfig{figure=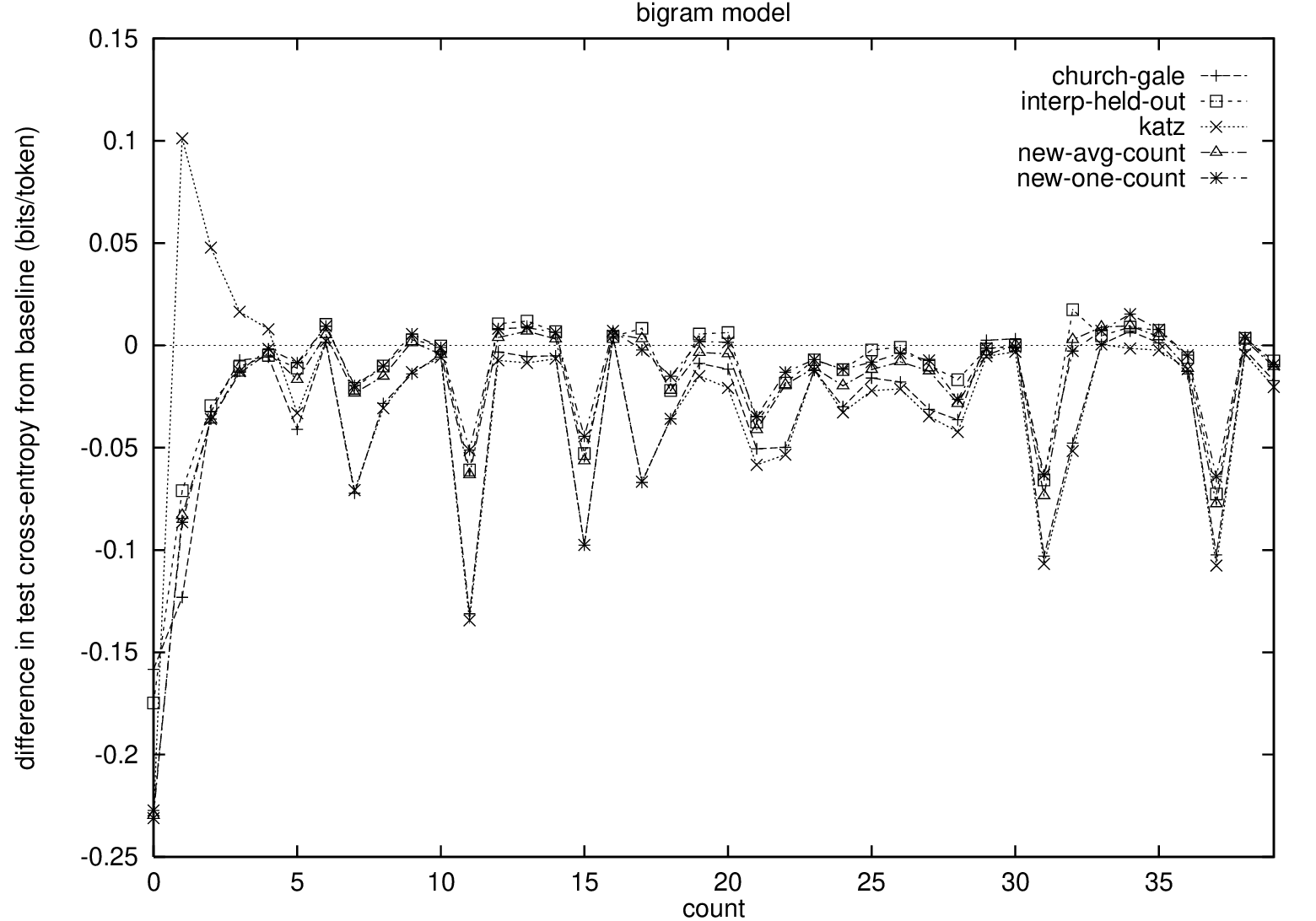,width=2.9in} \hspace{0.2in}
\psfig{figure=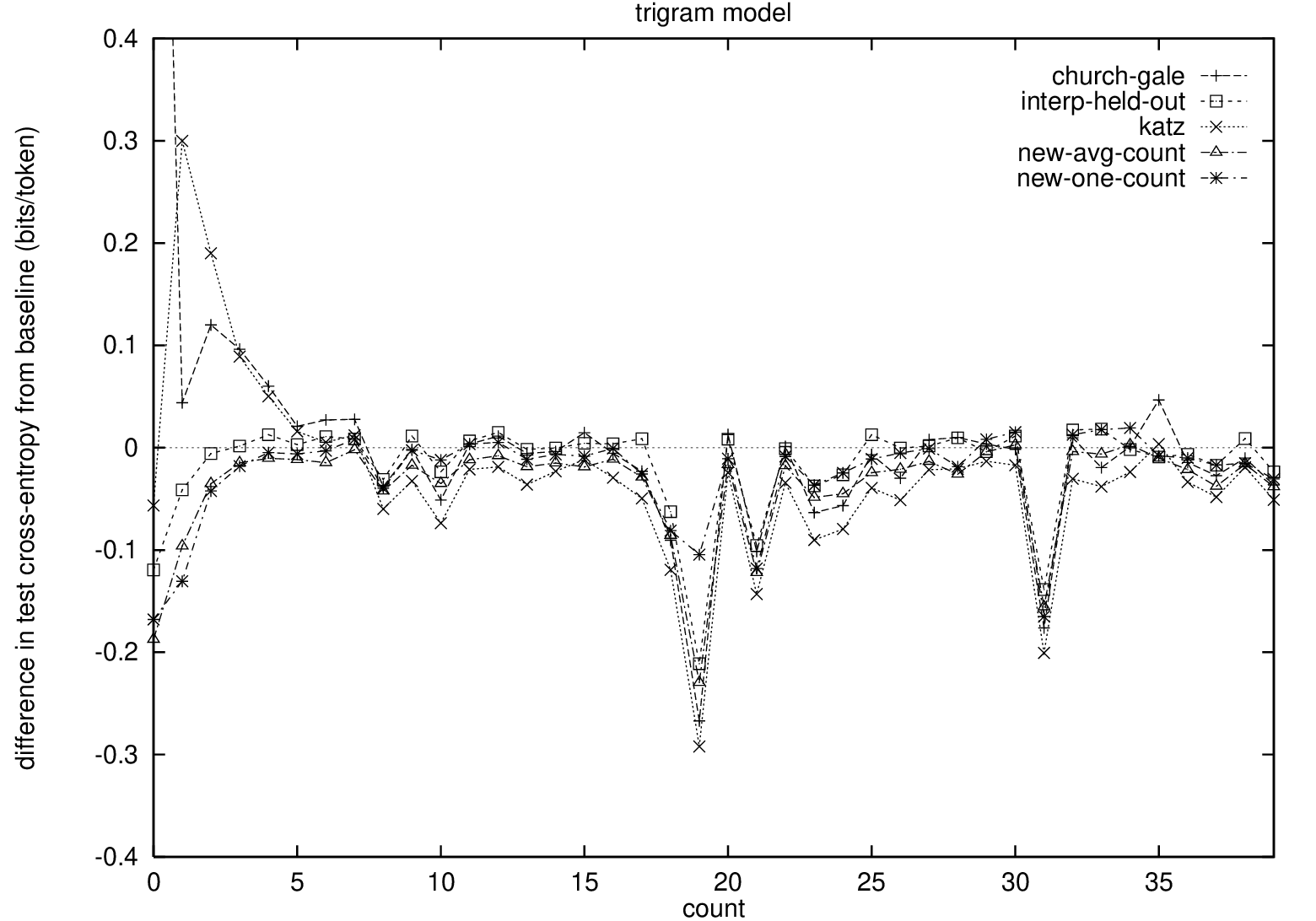,width=2.9in} $$
\caption{Relative performance at each count for various algorithms,
	bigram and trigram models, 1M words training data}
	\label{fig:banga}

$$ \psfig{figure=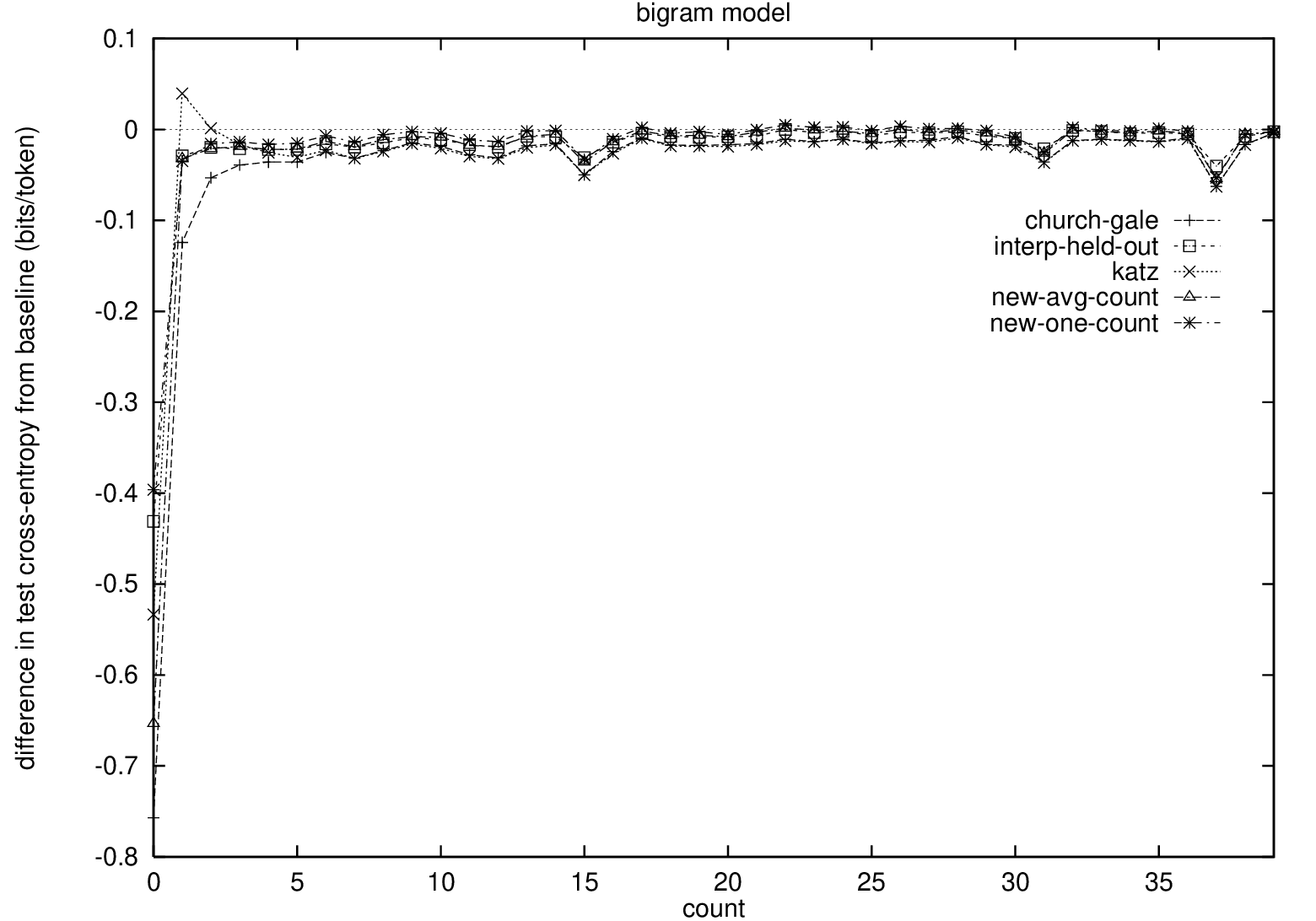,width=2.9in} \hspace{0.2in}
\psfig{figure=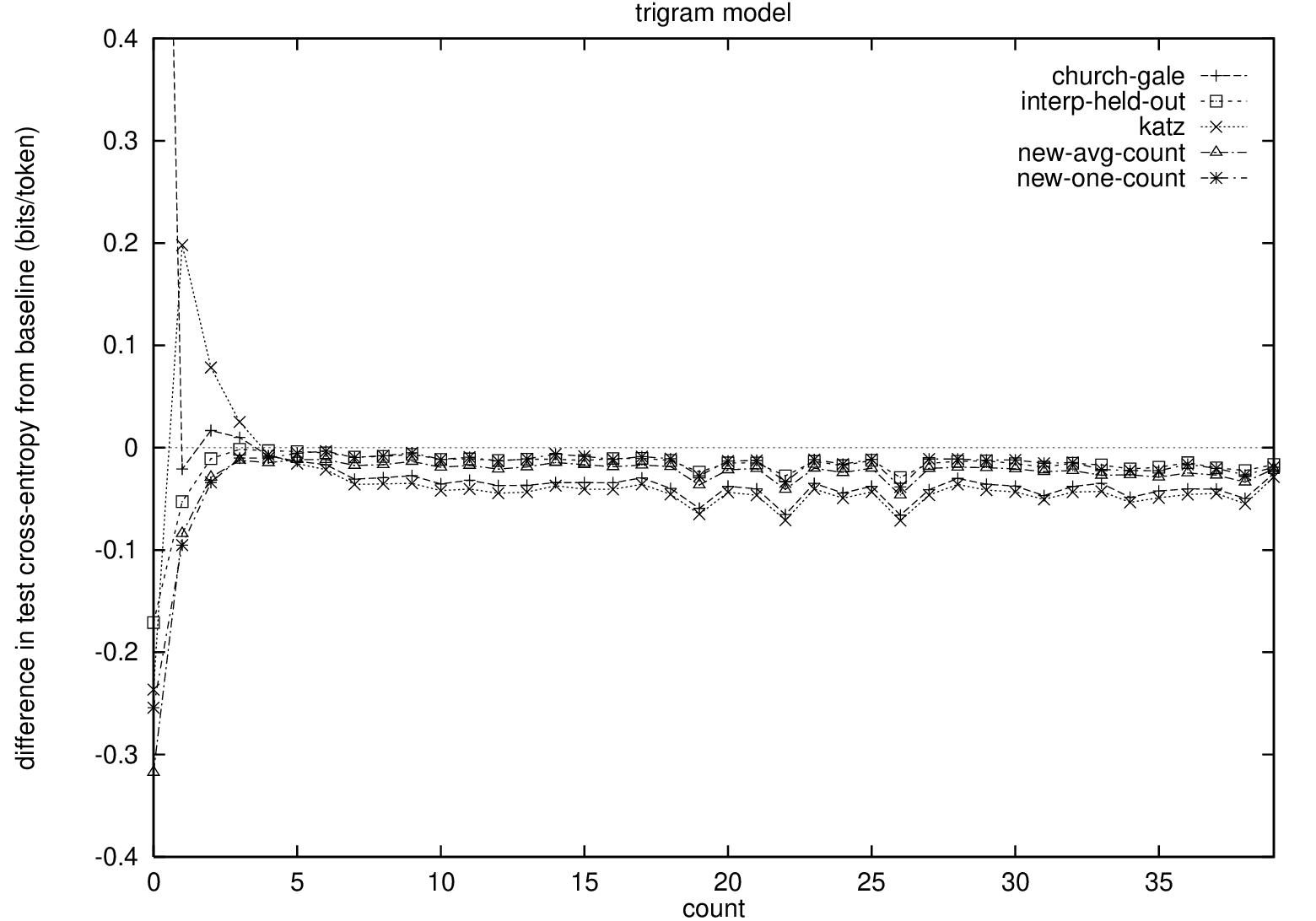,width=2.9in} $$
\caption{Relative performance at each count for various algorithms,
	bigram and trigram models, 200M words training data}
	\label{fig:bangb}
\end{figure}

The above analysis only considers whether an algorithm yields
the desired {\it average\/} corrected count; it does not provide insight into
whether an algorithm varies the corrected count in different distributions
in a felicitous manner.  For example, the Good-Turing estimate predicts
that one should assign a total probability of $\frac{n_1}{N}$ to
$n$-grams with zero counts; obviously, this value varies from
distribution to distribution.  In Figures \ref{fig:banga} and
\ref{fig:bangb}, we display a measure that we call {\it bang-for-the-buck\/}
that reflects how well a smoothing
algorithm varies the corrected count $r^*$ of a given count $r$ in different
distributions.  To explain this measure,
we first consider the measure of just taking the total entropy
assigned in the test data to $n$-grams with a given count $r$.  Presumably,
the smaller the entropy assigned to these $n$-grams, the better
a smoothing algorithm is at estimating these corrected counts.
However, an algorithm that assigns a higher average corrected
count will tend to have a lower entropy.  We want to factor
out this effect, and we do this by normalizing the average
corrected count of each algorithm to the same value before
calculating the entropy.  We call this measure {\it bang-for-the-buck\/}
as it reflects the relative performance of each algorithm given
that they all assign the same amount of probability to a given count.
In Figures \ref{fig:banga} and \ref{fig:bangb}, we display
the bang-for-the-buck per word of various algorithms relative to
the baseline method; as this score is an entropy value,
the lower the score, the better.

For larger counts, Katz and Church-Gale yield superior
bang-for-the-buck.  We hypothesize that this is because the linear discounting
used by other methods is a poor way to discount large counts.
On small nonzero counts, Katz smoothing does relatively
poorly.  We hypothesize that this is because
Katz smoothing does not perform any interpolation with lower-order
models for these counts, while other methods do.
It seems likely that lower-order
models still provide useful information if counts are low but nonzero.
The best methods for modeling zero counts are our two novel methods.

The method {\tt church-gale} performs especially poorly on zero counts
in trigram models.  This can be attributed to the implementation
choice discussed in Section \ref{sssec:cgimpl}.  We chose to
implement a version of the algorithm analogous to
interpolating the trigram model
directly with a unigram model, as opposed to a version analogous
to interpolating the trigram model with a bigram model.  (As discussed,
it is unclear whether the latter version is practical.)

\begin{figure}
$$ \psfig{figure=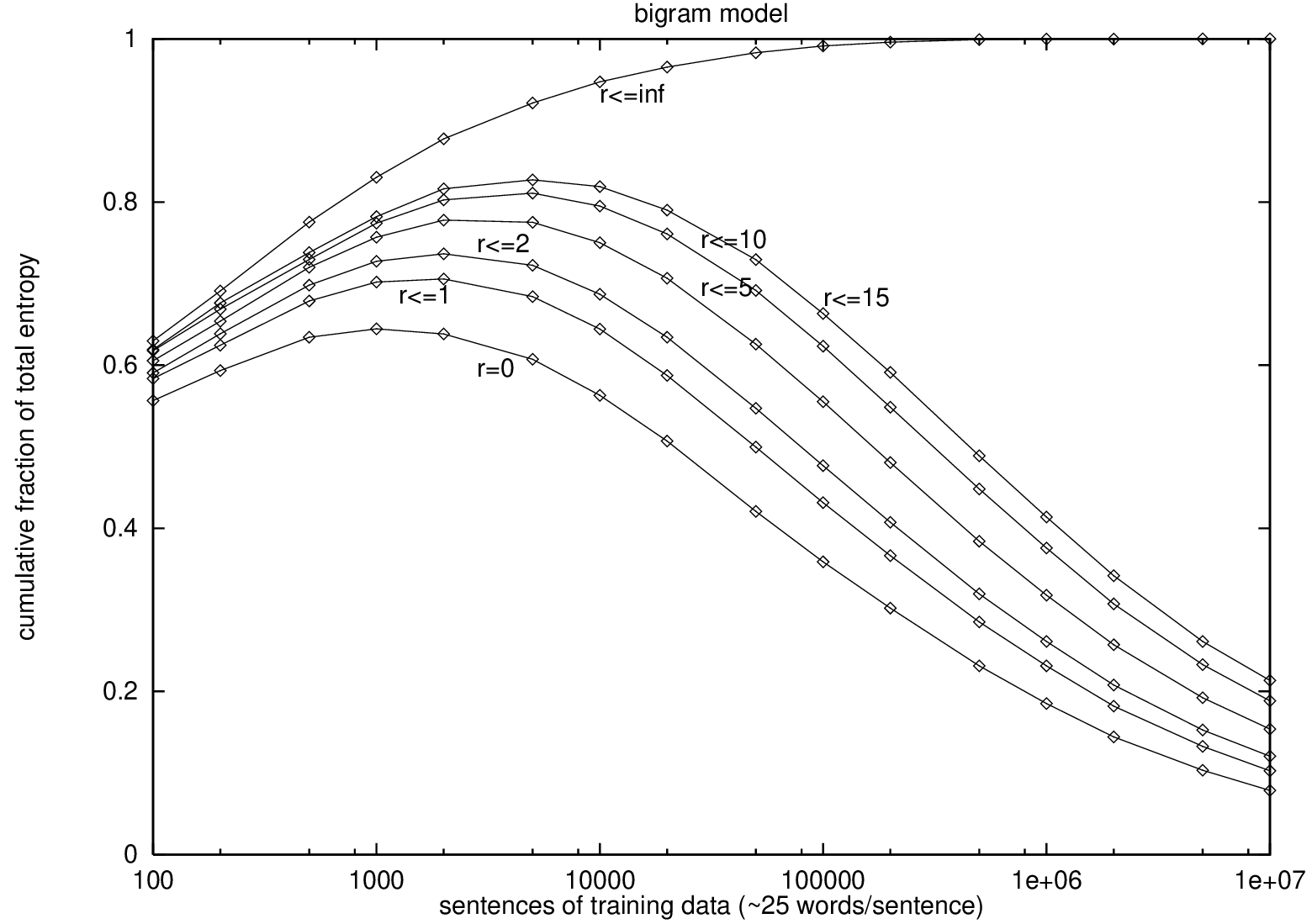,width=2.9in} \hspace{0.2in}
\psfig{figure=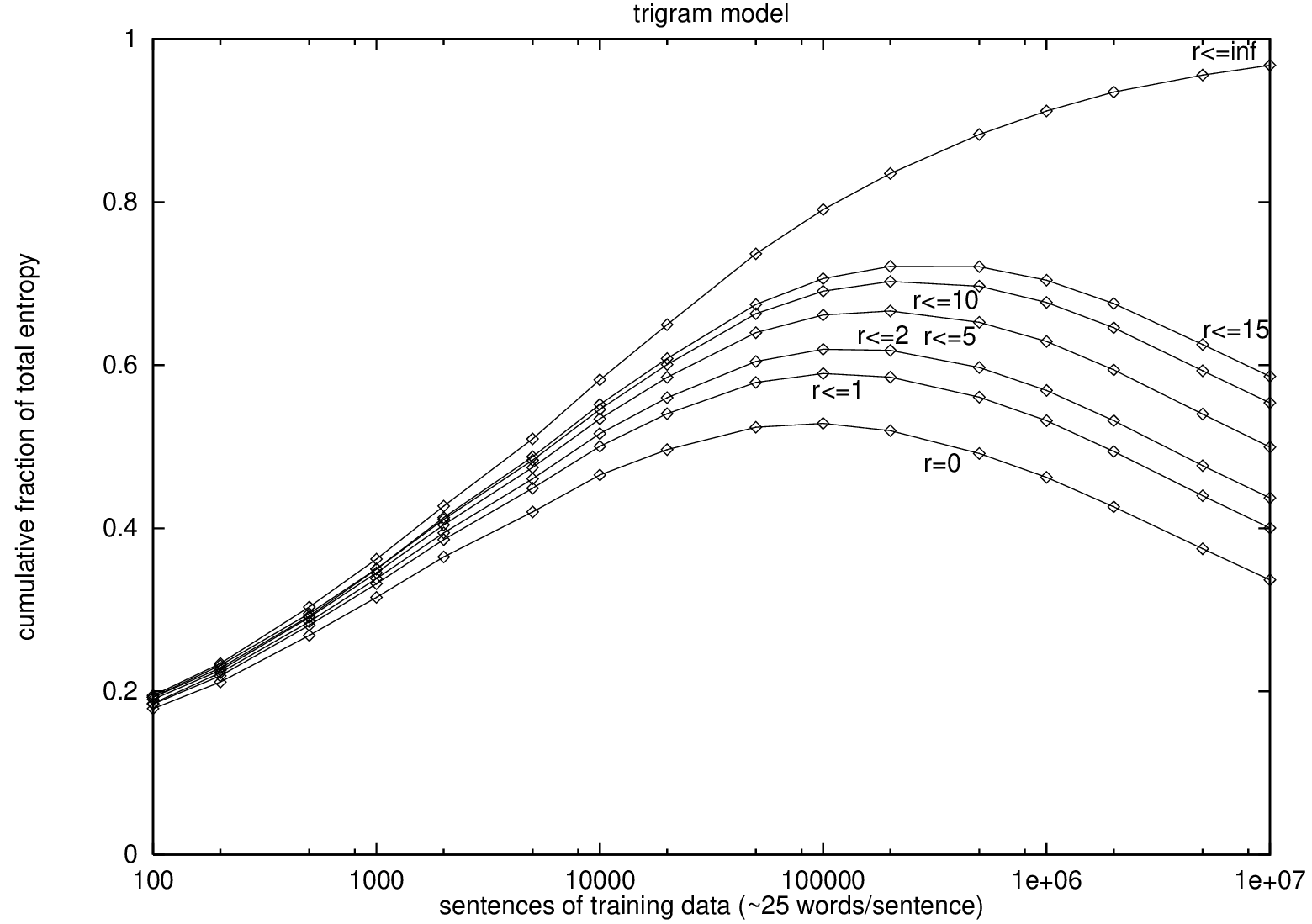,width=2.9in} $$
\caption{Fraction of entropy devoted to various counts over
	many training sizes, baseline smoothing, bigram and trigram models}
	\label{fig:entfrac}
\end{figure}

Given the above analysis, it is relevant to note what fraction
of the total entropy of the test data is associated with
$n$-grams of different counts.  In Figure \ref{fig:entfrac},
we display this information for different training set sizes
for bigram and trigram models.  A line labelled $r \leq k$
graphs the fraction of the entropy devoted to $n$-grams with up to
$k$ counts.  For instance, the region below the lowest line is the fraction
of the entropy devoted to zero counts (excluding those zero counts
that occur in distributions with a {\it total\/} of zero counts; as mentioned
before we treat these $n$-grams separately).  The fraction
of the entropy devoted to zero count $n$-grams occurring
in zero count distributions is represented by the region above
the top line in the graph.

This data explains some of the variation in the relative performance
of different algorithms over different training set sizes and
between bigram and trigram models.
Our novel methods get most of their performance gain
relative to other methods from their performance on zero counts.
Because zero counts are more frequent in trigram models,
our novel methods perform especially well on these models.
Furthermore, because zero counts are less frequent in large
training sets, our methods do not do as well from a relative
perspective on larger data.  On the other hand, Katz smoothing
and Church-Gale smoothing do especially well on large counts.
Thus, they yield better performance on bigram models and on
large training sets.

\ssec{Accuracy of the Good-Turing Estimate for Zero Counts}

\begin{figure}

$$ \psfig{figure=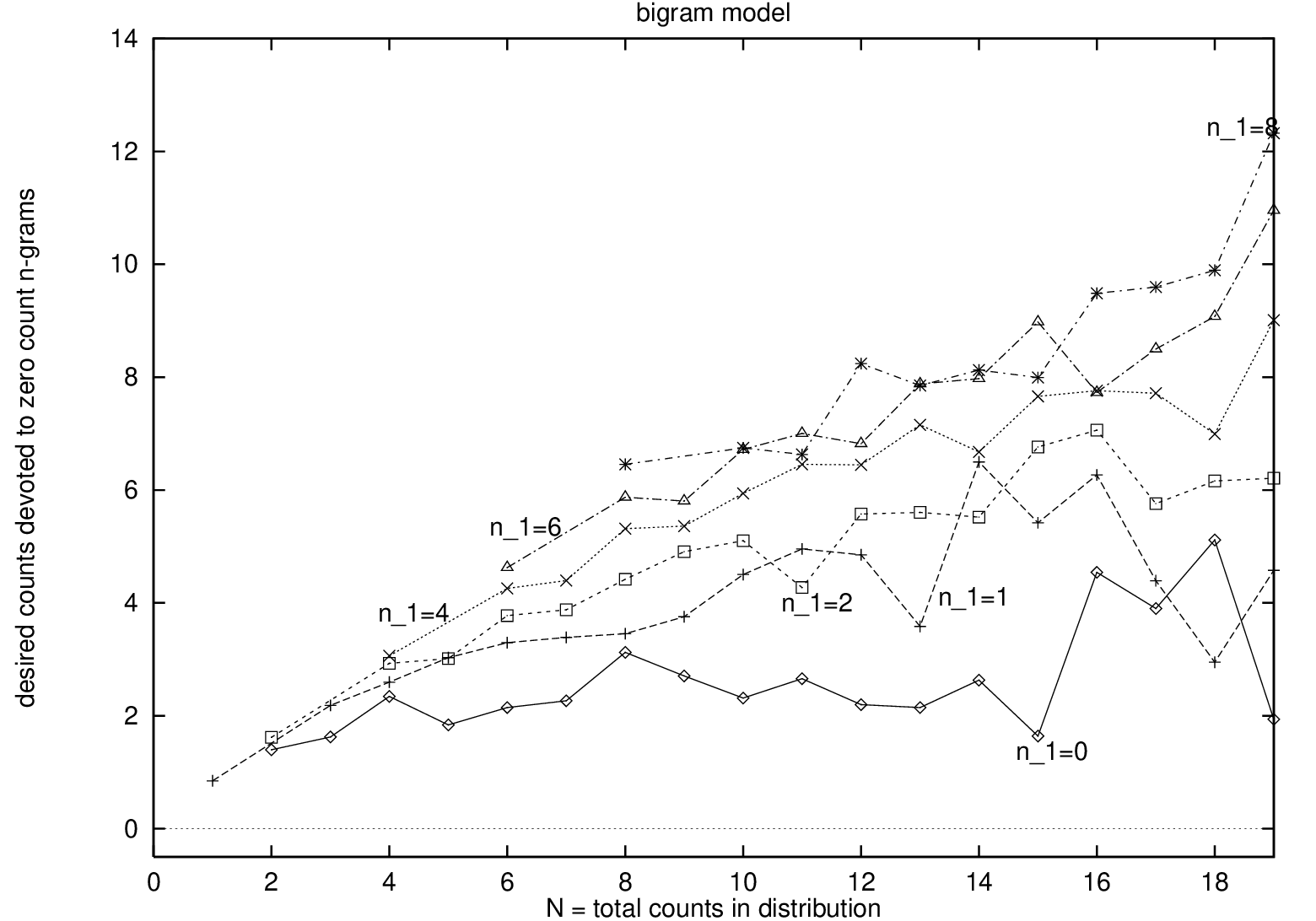,width=2.9in} \hspace{0.2in}
\psfig{figure=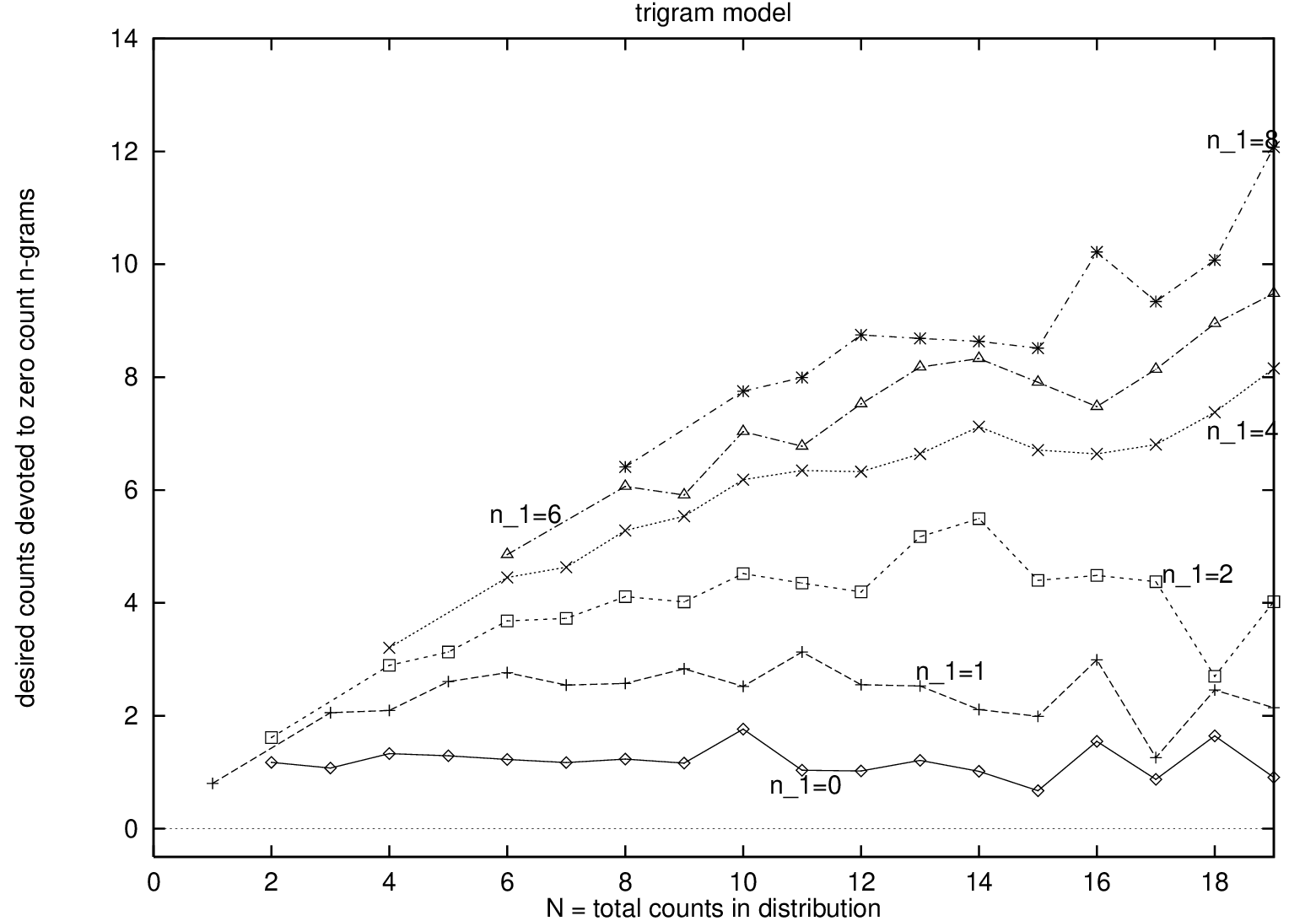,width=2.9in} $$
\caption{Average count assigned to $n$-grams with zero count for
	various $n_1$ and $N$, actual, bigram and trigram models}
	\label{fig:gtemp}

$$ \psfig{figure=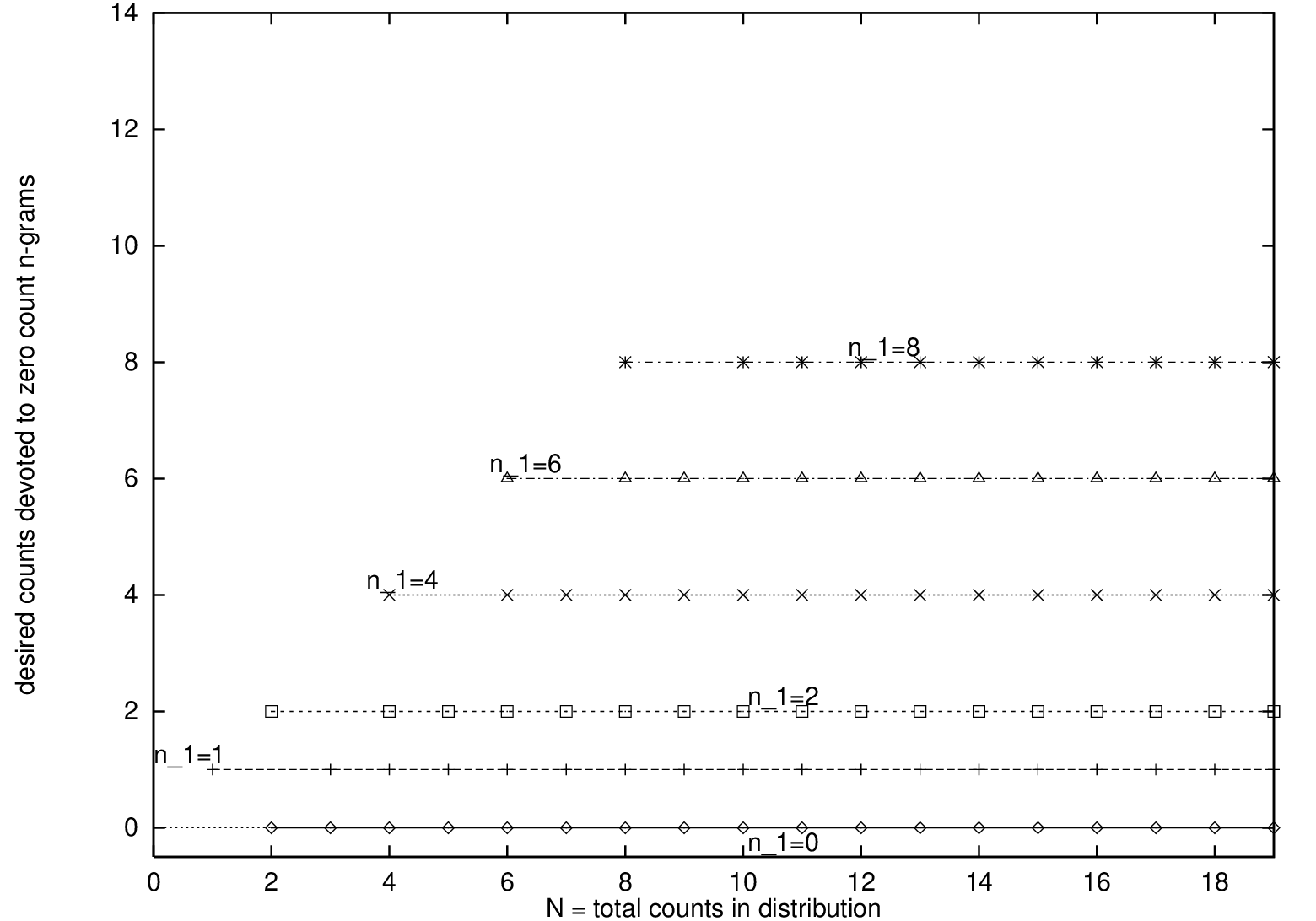,width=2.9in} $$
\caption{Average count assigned to $n$-grams with zero count for
	various $n_1$ and $N$, predicted by Good-Turing}
	\label{fig:gtpred}

\end{figure}

Because the Good-Turing estimate is a fundamental tool in
smoothing, it is interesting to test its accuracy empirically.
In this section, we describe experiments investigating
how well the Good-Turing estimate assigns probabilities
to $n$-grams with zero counts in conditional bigram and
trigram distributions.  We consider zero counts in particular because
zero-count $n$-grams contribute a very sizable fraction
of the total entropy, as shown in Figure \ref{fig:entfrac}.

The Good-Turing estimate
predicts that the total probability assigned to $n$-grams with zero
counts should be $\frac{n_1}{N}$, the number
of one-counts in a distribution divided by the total number of
counts in the distribution.  In terms of corrected counts, this
corresponds to assigning a total of $n_1$ counts to $n$-grams with
zero counts.\footnote{As in
Section \ref{ssec:methoda}, we use
$n_1$ and $N$ to refer to counts in a conditional
distribution $p(w_i|w_{i-n+1}^{i-1})$ for a fixed $w_{i-n+1}^{i-1}$,
as opposed to counts in the global $n$-gram distribution.}
We can calculate the {\it desired\/} average corrected count for
a given $n_1$ and $N$ by using a similar analysis as in
Section \ref{ssec:scount}, comparing the expected number and
actual number of zero-count $n$-grams in test data.
In Figure \ref{fig:gtemp}, we display the desired total number of
corrected counts assigned to zero counts for various values
of $n_1$ and $N$, for 1M words of TIPSTER training data.

Each line in the graph corresponds to a different
value of $n_1$.  The $x$-axis corresponds to $N$,
and the $y$-axis corresponds to the desired count to assign to $n$-grams
with zero counts.  If the Good-Turing estimate were exactly accurate,
then we would have a horizontal line for each $n_1$ at the level $y=n_1$,
as displayed in Figure \ref{fig:gtpred}.
However, we see that the lines are not very horizontal for
smaller $N$, and that asymptotically they seem to level out at
a value significantly larger than $n_1$.

We hypothesize that this is because the assumption made by Good-Turing
that successive $n$-grams are independent is incorrect.  The derivation
of the Good-Turing
estimate relies on the observation that for an event with probability $p$,
the probability that it will occur $r$ times in $N$ trials is
${N\choose r} p^r (1-p)^{N-r}$.  However, this only holds if
each trial is independent.  Clearly, language has decidedly
clumpy behavior.  For example, a given word has a higher chance of
occurring given that it has occurred recently.\footnote{This observation
is taken advantage of in {\it dynamic\/} language
modeling \cite{Kuhn:88a,Rosenfeld:92a,Rosenfeld:94a}.}
Thus, the actual number of one-counts is probably lower than what
would be expected if independence were to hold, so the probability given
to zero counts should be larger than $\frac{n_1}{N}$.

\ssec{Church-Gale Smoothing versus Linear Interpolation}

Church-Gale smoothing incorporates the information from lower-order models
into higher-order models
through its bucketing mechanism, unlike other smoothing methods
that use linear interpolation.  In this section, we present
empirical results on how these two different techniques compare.

\begin{figure}
$$ \psfig{figure=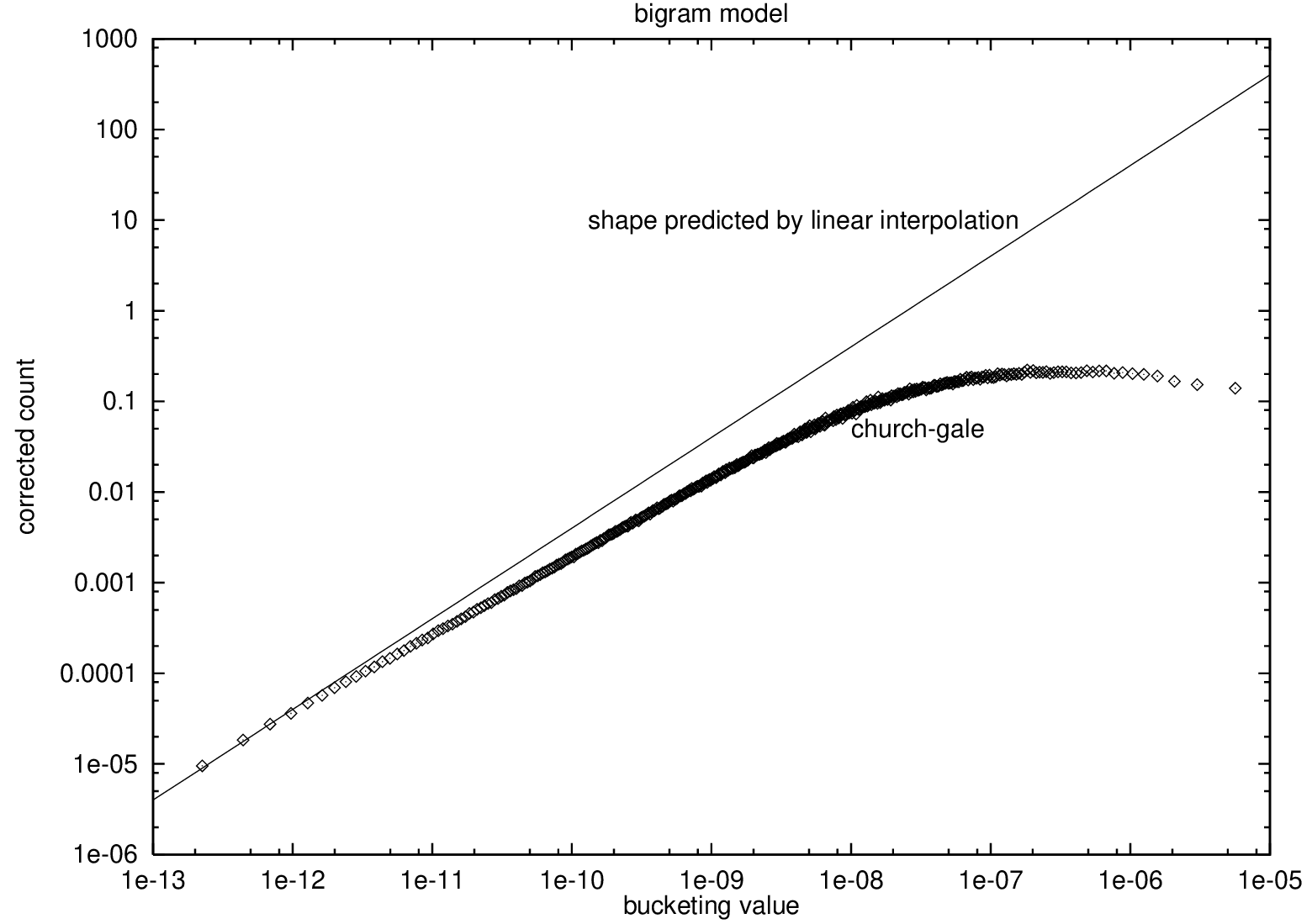,width=2.9in} \hspace{0.2in}
\psfig{figure=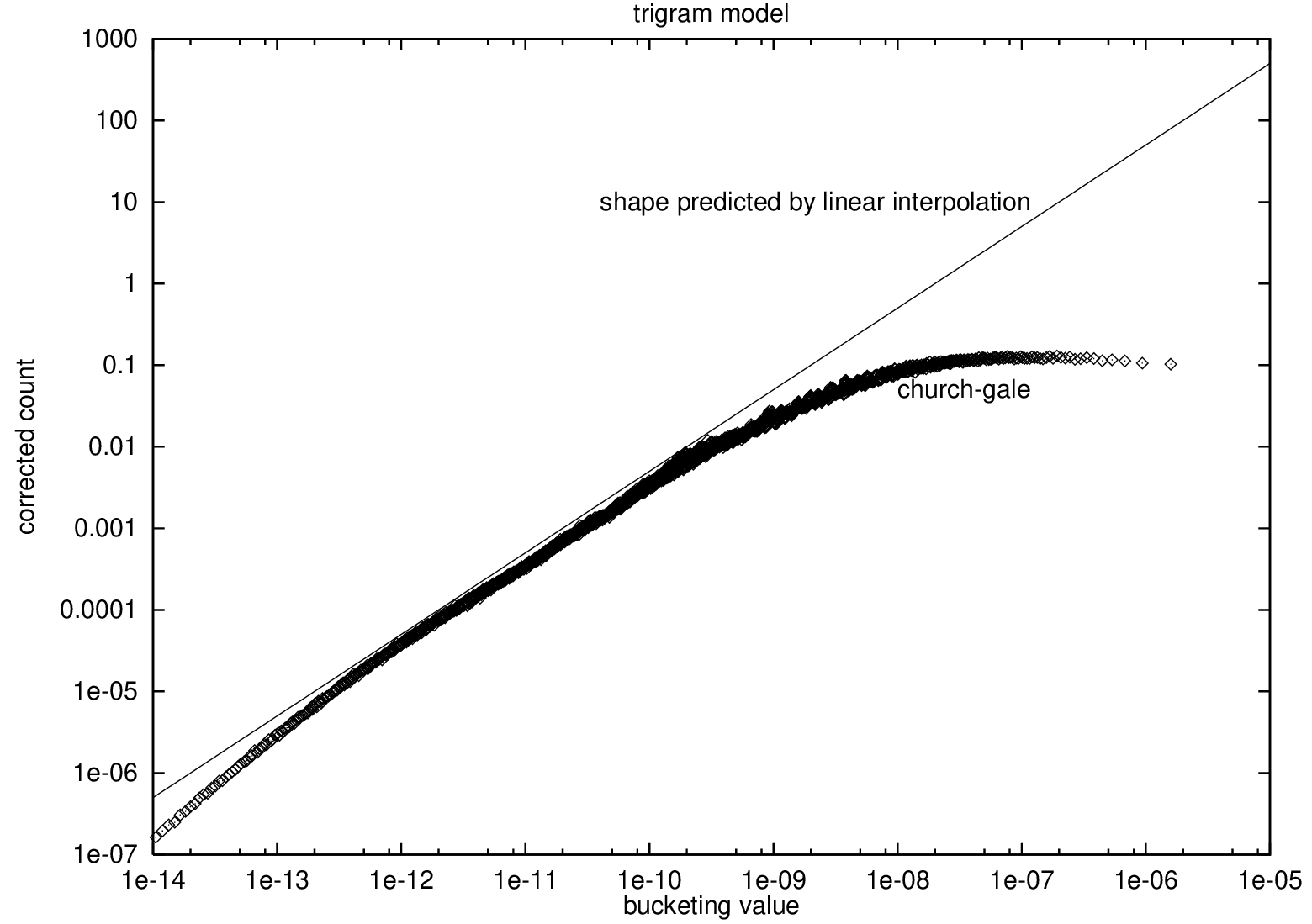,width=2.9in} $$
\caption{Corrected count assigned to zero counts by Church-Gale
	for all buckets, bigram and trigram models}
	\label{fig:cgparama}

$$ \psfig{figure=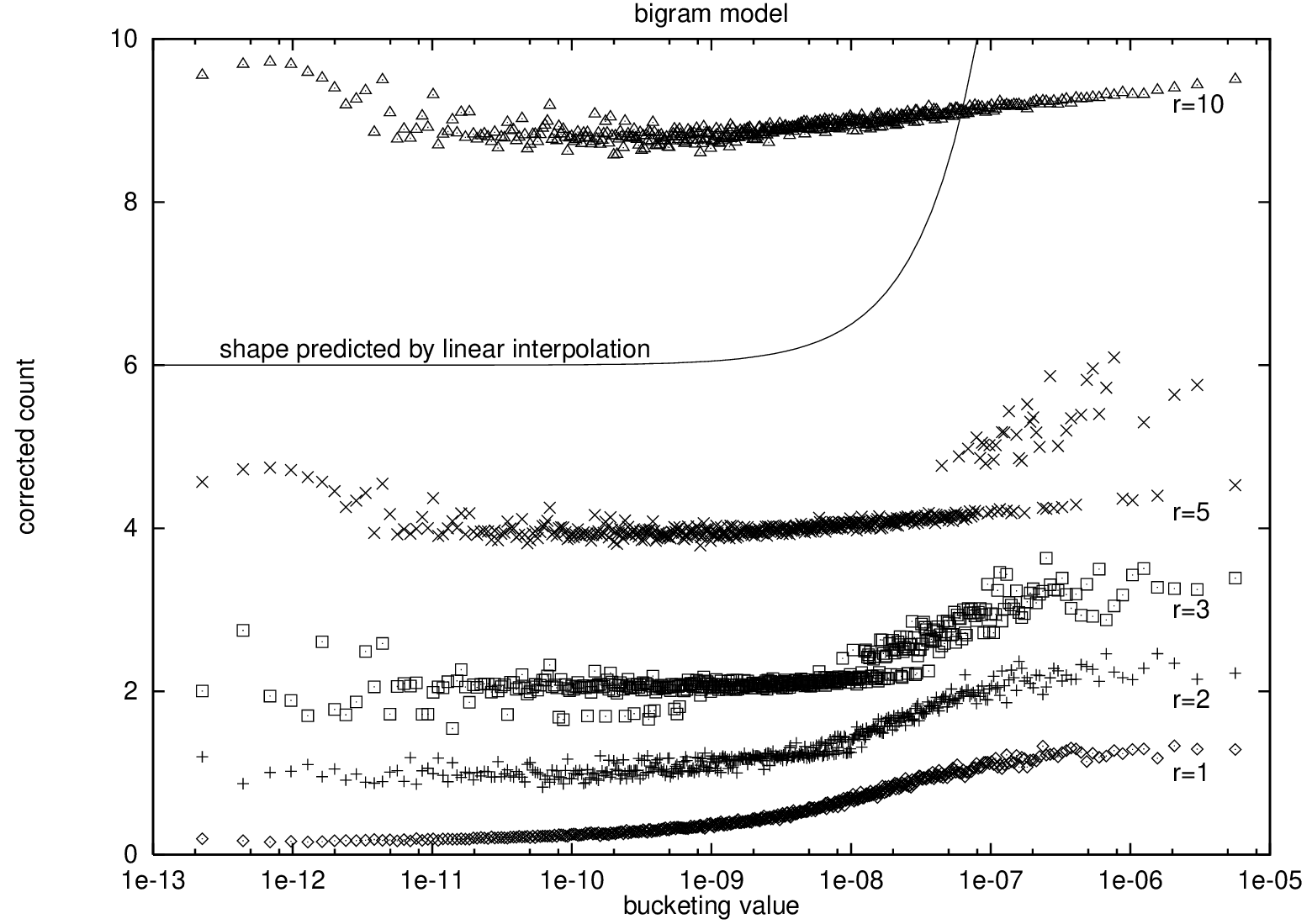,width=2.9in} \hspace{0.2in}
\psfig{figure=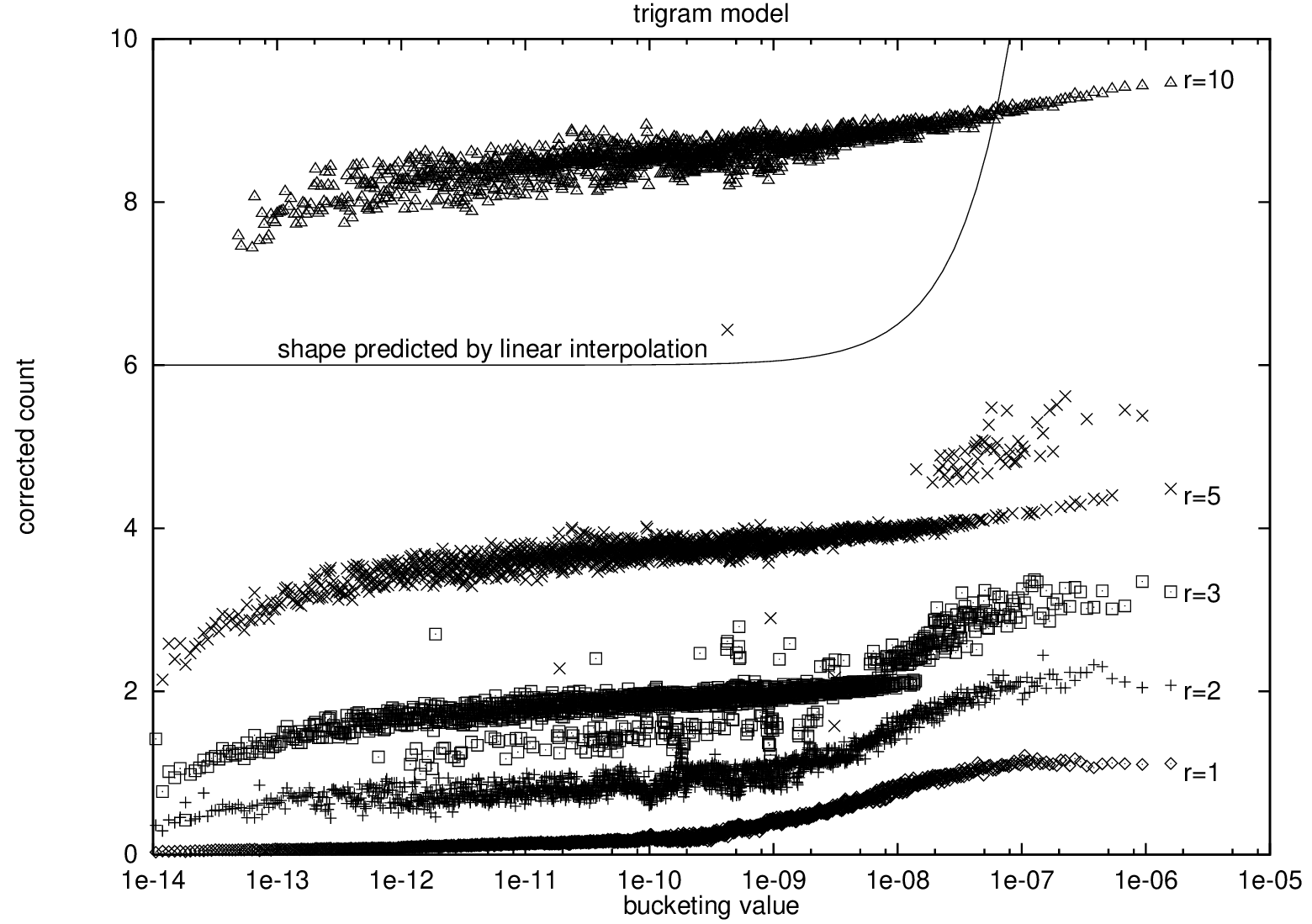,width=2.9in} $$
\caption{Corrected count assigned to various counts by Church-Gale
	for all buckets, bigram and trigram models}
	\label{fig:cgparamb}
\end{figure}

First, we compare how Church-Gale smoothing and linear interpolation
assign corrected counts to zero counts.
In Figure \ref{fig:cgparama}, we present the corrected count of
zero counts for each bucket in a Church-Gale run.  In linear
interpolation, the corrected count of an $n$-gram with
zero counts is proportional to the probability assigned to
that $n$-gram in the next lower-order model.  The $x$-axis
of the graph represents the value used for bucketing in Church-Gale,
which is proportional to the probability assigned to an $n$-gram by the
next lower-order model.  Thus, if Church-Gale smoothing
assigns corrected counts to $n$-grams with zero counts
similarly to linear interpolation, the graph will
be a line with slope 1 (given that both axes are logarithmic).
The actual graph is not far from this situation; the solid
lines in the graphs are lines with slope 1.  Thus, even
though Church-Gale smoothing is very far removed from
linear interpolation on the surface, for zero counts
their behaviors are rather similar.

In Figure \ref{fig:cgparamb}, we display the corrected counts for
Church-Gale smoothing
for $n$-grams with larger counts.  For these counts, the
curves are very different from what would be yielded with
linear interpolation.  (The shapes of the curves consistent
with linear interpolation are different from those found in the
previous figure because the $y$-axis is linear instead of
logarithmic scale.)

\ssec{Held-out versus Deleted Interpolation}

\begin{figure}

$$ \psfig{figure=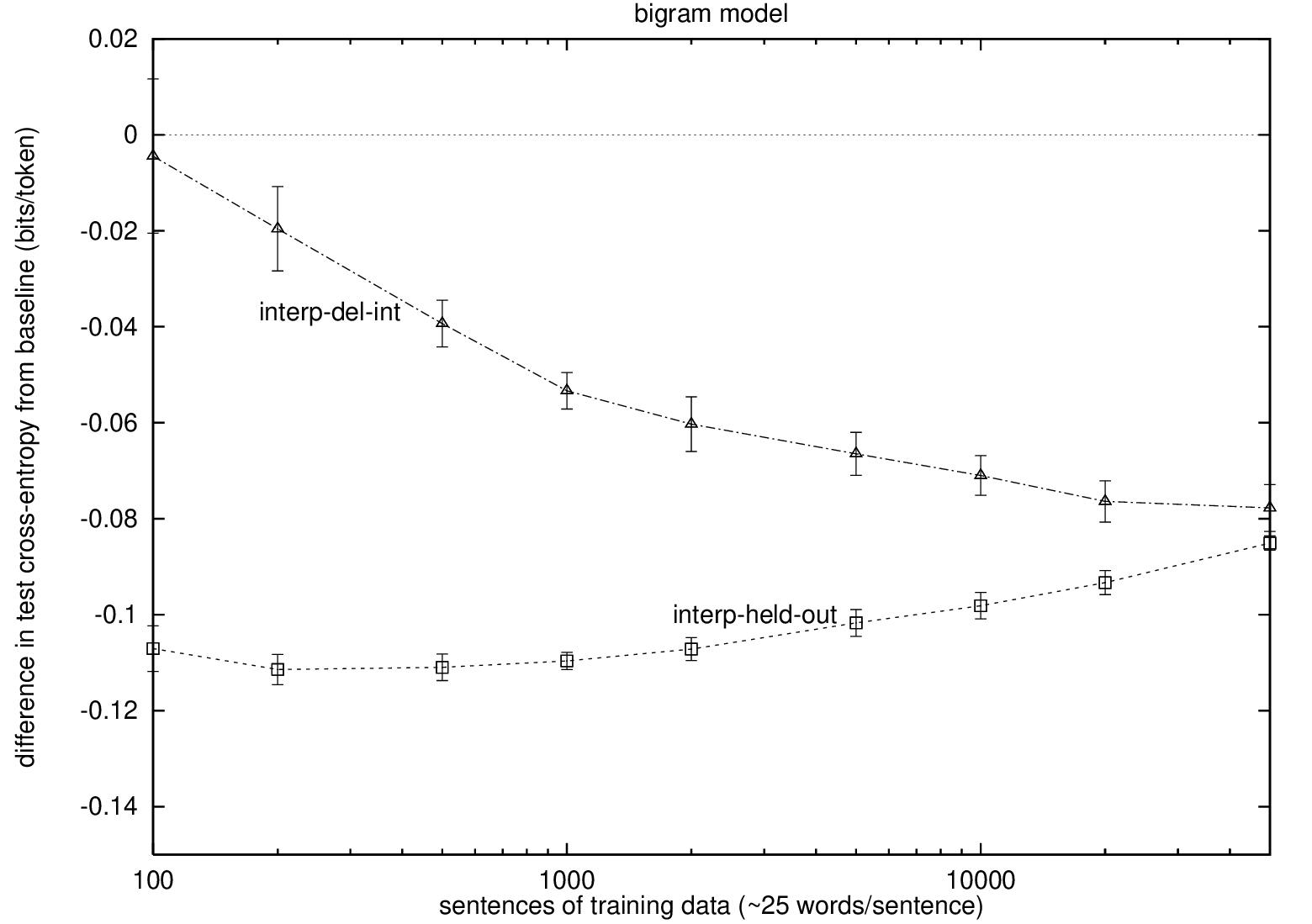,width=2.9in} \hspace{0.2in}
\psfig{figure=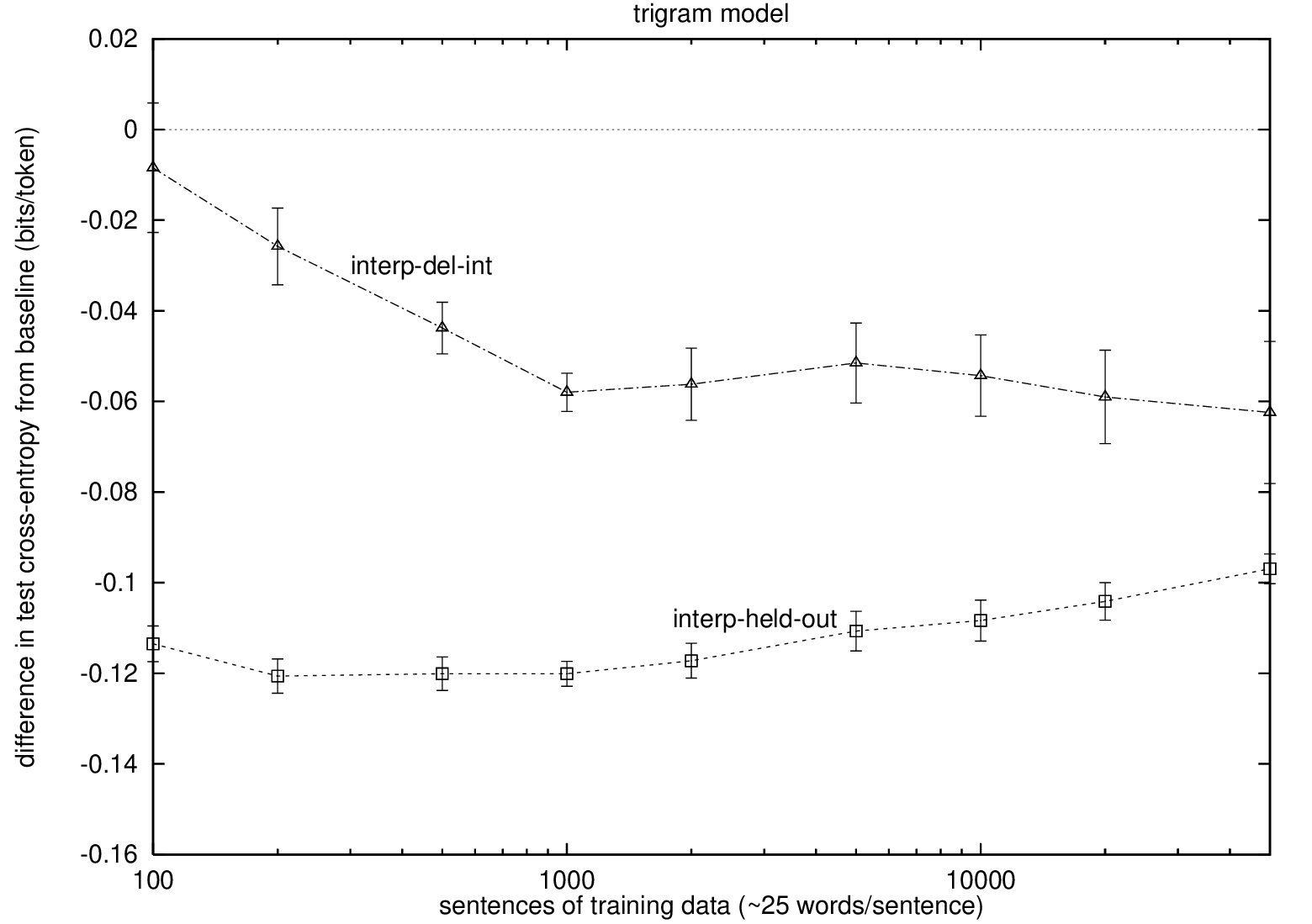,width=2.9in} $$
\caption{Held-out versus deleted interpolation on TIPSTER data,
	relative performance with respect to baseline,
	bigram and trigram models}
	\label{fig:delint}

\end{figure}

In this section, we compare the held-out and deleted interpolation
variations of Jelinek-Mercer smoothing.  Referring to
Figure \ref{fig:delint}, we notice that the method {\tt interp-del-int}
performs significantly
worse than {\tt interp-held-out} on TIPSTER data, though they differ
only in that the former method uses deleted interpolation while the latter
method uses held-out interpolation.  Similar results hold for
the other corpora, as shown in the earlier
Figures \ref{fig:tip3}--\ref{fig:wsj}.

However, the implementation
{\tt interp-del-int} does not completely characterize the
technique of deleted interpolation
as we do not vary the size of the chunks that are deleted.
In particular, we made the choice of deleting only a single
word at a time for implementation ease; we hypothesize that
deleting larger chunks would lead to more similar performance to
{\tt interp-held-out}.

As mentioned earlier, language tends to have clumpy behavior.
Held-out data external to the training data will tend to be more
different from the training data than data that is deleted
from the middle of the training data.
As our evaluation test data is also external to the training data (as is
the case in applications), $\l$'s trained from held-out data should
better characterize the evaluation test data.  However, the larger the
chunks deleted in deleted interpolation, the more the deleted
data behaves like held-out data.  For example, if we delete half
of the data at a time, this is very similar to the held-out data
situation.  Thus, larger chunks should yield better performance
than that achieved by deleting one word at a time.

However, for large training sets the computational expense of
deleted interpolation becomes a factor.  In particular, the
computation required is linear in the {\it training\/} data
size.  For held-out interpolation, the computation is linear
in the size of the {\it held-out\/} data.  Because there are relatively
few $\l$'s, these parameters can be trained reliably using
a fairly small amount of data.  Furthermore, for large
training sets it matters little that held-out interpolation requires
some data to be reserved for training $\l$'s while in
deleted interpolation no data needs to be reserved for this purpose.
Thus, for large training sets,
held-out interpolation seems the sensible choice.

%
\sec{Discussion}
%

Smoothing is a fundamental technique for statistical modeling,
important not only for language modeling but for many other applications
as well, \eg, prepositional phrase attachment
\cite{Collins:95a}, part-of-speech tagging \cite{Church:88a},
and stochastic parsing \cite{Magerman:94a}.  Whenever data sparsity
is an issue (and it always is), smoothing has the potential
to improve performance with moderate effort.  Thus, thorough studies
of smoothing can benefit the research community a great deal.

To our knowledge, this is the first empirical comparison of
smoothing techniques in language modeling of such scope: no other
study has systematically examined multiple training data sizes, corpora, or has
performed parameter optimization.  We show that in order to completely
characterize the relative performance of two techniques, it
is necessary to consider multiple training set sizes and to try both
bigram and trigram models.  We show that sub-optimal
parameter selection can also significantly affect relative performance.

Multiple runs should be performed whenever possible to
discover whether any calculated differences are
statistically significant; it is unclear whether previously reported
results in the literature are reliable given that they are
based on single runs and given the variances found in this work.
For example, we found that the standard deviation of the average
performance of Katz smoothing
relative to the baseline method is about 0.005 bits
for ten runs.  Extrapolating to a single run, we expect
a standard deviation of about $\sqrt{10} \times 0.005 \approx 0.016$ bits,
which translates to about a 1\% difference in perplexity.
In the N\'{a}das and Katz papers, differences in perplexity between algorithms
of about 1\% are reported for a single test set of 100 sentences.
MacKay and Peto present perplexity differences between algorithms
of significantly less than 1\%.

Of the techniques studied, we have found that Katz smoothing performs
best for bigram models produced from small training sets, while Church-Gale
performs best for bigram models produced from large training sets.
This is a new result; Church-Gale smoothing has never previously
been empirically compared with any of the popular smoothing
techniques for language modeling.  Our novel methods
{\it average-count\/} and {\it one-count\/} are superior for trigram models
and perform well in bigram models;
method {\it one-count\/} yields
marginally worse performance but is extremely easy
to implement.

Furthermore, we provide a count-by-count analysis of the performance
of different smoothing techniques.  By analyzing how
frequently different counts occur in a given domain, we can make rough
predictions on the relative performance of different algorithms
in that domain.  For example, this analysis lends insight
into how different algorithms will perform on training sizes
and $n$-gram orders other than those we tested.

However, it is extremely important to note that in this work
performance is measured solely through the cross-entropy of test data.
This choice was made because it is fairly inexpensive to
evaluate cross-entropy, which enabled us to run experiments of such scale.
Yet it is unclear how entropy differences translate to
differences in performance in real-world applications
such as speech recognition.  While entropy generally correlates
with performance in applications, small differences in entropy
have an unpredictable effect; sometimes a reduction in entropy
can lead to an {\it increase\/} in application error-rate,
\eg, as reported by \newcite{Iyer:94a}.  In other words,
entropy by no means completely characterizes application performance.
Furthermore, it is not unlikely that relative smoothing performance results
found in one application will not translate to other applications.
Thus, to accurately estimate the effect of smoothing in a given
application, it is probably necessary to run experiments using
that particular application.

\begin{table}

\begin{center} Isotani and Matsunaga \end{center}
$$ \begin{tabular}{|l|c|c|} \hline
entropy & sentence error rate & decrease in error rate \\ \hline \hline
original & 48.7 & \\ \hline
-0.05 bits & 48.2 & 1.0\% \\ \hline
-0.10 bits & 47.6 & 2.3\% \\ \hline
-0.15 bits & 46.7 & 4.1\% \\ \hline
\end{tabular} $$

\begin{center} Rosenfeld \end{center}
$$ \begin{tabular}{|l|c|c|} \hline
entropy & word error rate & decrease in error rate \\ \hline \hline
original & 19.9 & \\ \hline
-0.05 bits & 19.7 & 1.0\% \\ \hline
-0.10 bits & 19.5 & 2.0\% \\ \hline
-0.15 bits & 19.3 & 3.0\% \\ \hline
\end{tabular} $$

$$ \begin{tabular}{lcl}
0.05 bits & $\geq$ & typical entropy difference between best methods \\
0.10 bits & $\approx$ & maximum entropy difference between best methods \\
0.15 bits & $\approx$ & typical entropy difference between best methods
	and baseline method \\
\end{tabular} $$

\caption{Effect on speech recognition performance
	of typical entropy differences found between smoothing methods}
	\label{tab:appperf}
\end{table}

However, we can guess how smoothing might affect application
performance by extrapolating from existing results.  For example,
\newcite{Isotani:94a}
present the error rate of a speech recognition system using
three different language models.  As they also report the entropies
of these models, we can linearly extrapolate to estimate
how much the differences in entropy typically found between
smoothing methods affect speech recognition performance.
In Table \ref{tab:appperf}, we list typical entropy differences
found between smoothing methods, where the ``best'' methods
refer to {\tt interp-held-out}, {\tt katz}, {\tt new-avg-count},
and {\tt new-one-count}.  We also display how these
entropy differences affect application performance as extrapolated
from the Isotani data.  The row labelled {\it original\/} lists
the error rate of the model tested by Isotani
and Matsunaga with the highest entropy; the lower rows list the
extrapolated error rate
if the model entropy were decreased by the prescribed amount.
In Table \ref{tab:appperf}, we also display a similar analysis using data
given by \newcite{Rosenfeld:94b}.  This analysis suggests
that smoothing does not matter much as long as one uses
a ``good'' implementation of one of the better algorithms,
\eg, those algorithms that perform significantly better than the baseline;
it is more likely that the differences between the best and worst algorithms
are significant.

We have found that it is surprisingly difficult to design
a ``good'' implementation of an existing algorithm.  Given
the description of our implementations, it is clear that
there are usually many choices that need to be made in implementing
a given algorithm; most smoothing techniques are incompletely
specified in the literature.  For example, as pointed out
in Section \ref{ssec:simpl}, in certain cases Katz smoothing
as originally described can assign probabilities of zero,
which is undesirable as this leads to an infinite entropy.
We needed to perform a fair amount
of tuning for each algorithm before we guessed our implementation
was a reasonable representative of the algorithm.  Poor choices
often led to very significant differences in performance.

Finally, we point out that because of the variation in the
performance of different smoothing methods and the variation
in the performance of different
implementations of the same smoothing method (\eg, from parameter
setting), it is vital to specify the exact smoothing technique and
implementation of that technique used when referencing
the performance of an $n$-gram model.  For example, the Katz
and N\'{a}das papers describe comparisons of their algorithms
with ``Jelinek-Mercer'' smoothing, but they do not specify
the bucketing scheme used or the granularity used in deleted interpolation.
Without this information, it is impossible to determine
whether their comparisons are meaningful.  More generally,
there has been much work comparing the performance of various models
with that of $n$-gram models where the type
of smoothing used is not specified, \eg, work
by \newcite{McCandless:93a} and \newcite{Carroll:95a}.
Again, without this information we cannot tell if the comparisons are
significant.

\ssec{Future Work}

Perhaps the most important work that needs to be done is to
see how different smoothing techniques perform in actual
applications.  This would reveal how entropy differences
relate to performance in different applications, and would
indicate whether it is worthwhile to continue work in smoothing, given the
largest entropy differences we are likely to achieve.
Also, as mentioned before smoothing is
used in other language tasks such as prepositional phrase attachment,
part-of-speech tagging, and stochastic parsing.
It would be interesting to see whether our results
extend to domains other than language modeling.

Some smoothing algorithms that we did not consider that
would be interesting
to compare against are those from the field of {\it data compression},
which includes the subfield of {\it text compression} \cite{Bell:90a}.
However, smoothing
algorithms for data compression have different requirements from
those used for language modeling.  In data compression, it is
essential that smoothed models can be built extremely quickly
and using a minimum of memory.  In language modeling, these
requirements are not nearly as strict.

As far as designing additional smoothing methods that surpass
existing techniques, there were many avenues that we did not
pursue.  Hybrid smoothing methods
look especially promising.  As we found different methods to be superior
for bigram and trigram models, it may be advantageous to
use different smoothing methods in the different $n$-gram models
that are interpolated together.  Furthermore, in our count-by-count
analysis we found that different algorithms were superior on
low versus high counts.  Using different algorithms for
low and high counts may be another way to improve performance.

%

%
%

\def\P{{\cal P}}

\chapter{Bayesian Grammar Induction for Language Modeling} \label{ch:gram}

In this chapter, we describe a corpus-based induction algorithm for
probabilistic context-free grammars \cite{Chen:95c}
that significantly outperforms
the grammar induction algorithm introduced
by Lari and Young \shortcite{Lari:90a}, the most widely-used
algorithm for probabilistic grammar induction.  In addition,
it outperforms $n$-gram models on data generated with medium-sized
probabilistic context-free grammars, though not on naturally-occurring data.
Of the three structural levels at which we model language
in this thesis, this represents work at the constituent level.

%
\sec{Introduction}
%

While $n$-gram models currently yield the best performance
in language modeling, they seem to have obvious deficiencies.
For instance, $n$-gram language models can only
capture dependencies within
an $n$-word window, where currently the largest practical $n$ for
natural language is three, and many dependencies in natural
language occur beyond a three-word window.  In addition, $n$-gram
models are extremely large, thus making them difficult
to implement efficiently in memory-constrained applications.

An appealing alternative is grammar-based language models.  Grammar
has long been the representation of language used in linguistics
and natural language processing, and intuitively such models
capture properties of language that $n$-gram models cannot.
For example, it has been shown that
grammatical language models can express long-distance
dependencies \cite{Lari:90a,Resnik:92a,Schabes:92a}. Furthermore,
grammatical models have the potential to be more compact while
achieving equivalent performance as $n$-gram models \cite{Brown:90a}.
To demonstrate these points, we
introduce the grammar formalism we use, {\it probabilistic
context-free grammars} (PCFG).

\ssec{Probabilistic Context-Free Grammars}

We first give a brief introduction to (non-probabilistic) context-free
grammars \cite{Chomsky:64a}.  As mentioned in the
introduction, grammars consist of rules that describe how
structures at one level of language combine to form structures
at the next higher level.  For example, consider the following
grammar:\footnote{The following abbreviations are used in this work:
$$ \begin{tabular}{ccl}
S & = & sentence \\
VP & = & verb phrase \\
NP & = & noun phrase \\
D & = & determiner \\
N & = & noun \\
V & = & verb
\end{tabular} $$ }
$$ \begin{tabular}{ccl}
S & $\r$ & NP VP \\
VP & $\r$ & V NP \\
NP & $\r$ & D N \\
D & $\r$ & \t{a} $|$ \t{the} \\
N & $\r$ & \t{boat} $|$ \t{cat} $|$ \t{tree} \\
V & $\r$ & \t{hit} $|$ \t{missed}
\end{tabular} $$
This third through fifth rules
state that a noun phrase can be composed of a determiner
followed by a noun, a determiner may be formed by
the words {\it a\/} or {\it the\/}, and a noun may be formed by
the words {\it boat}, {\it cat}, or {\it tree}.  Thus, we
have that strings such as {\it a cat\/} or {\it the boat\/}
are noun phrases.  Applying the other rules in the grammar,
we see that strings such as {\it a boat missed the tree\/}
or {\it the cat hit the boat\/} are sentences.  Grammars
provide a compact and elegant way for representing a set of strings.

The above grammar is considered {\it context-free\/} because
there is a single symbol on the left-hand side of each rule; there
are grammar formalisms that allow multiple symbols.
The symbols at the lowest level of the grammar such
as {\it a}, {\it hit}, and {\it tree\/} are called the
{\it terminal symbols\/} of the grammar.  In all of the
grammars we consider, the terminal symbols correspond to words.
The other symbols in the grammar such as S, NP, and D
are called {\it nonterminal\/} symbols.

For every grammar, a particular nonterminal
symbol is chosen to be the {\it sentential symbol}.  The sentential
symbol determines the set of strings the grammar is
intended to describe; a grammar is said to {\it accept\/}
a string if the string forms an instance of the sentential symbol.
For example, if in the above example we take the sentential
symbol to be S, then the grammar accepts the string
{\it a cat hit the tree\/} but not the string {\it the tree}.
The sentential symbol is usually taken to be the symbol corresponding to
the highest level of structure
in the grammar, and in language domains it usually
corresponds to the linguistic concept of a sentence.  In this
work, we always name the sentential symbol S and it is always
meant to correspond to a sentence
(as opposed to a lower- or higher-level linguistic structure).

A {\it probabilistic\/} context-free grammar \cite{Solomonoff:59a}
is a context-free
grammar that not just describes a set of strings, but also
assigns probabilities to these strings.\footnote{Actually,
a probabilistic context-free grammar also assigns probabilities to strings
not accepted by the grammar; this probability is just zero.}
A probability is associated with each rule in the grammar, such
that the sum of the probabilities of all rules expanding a given
symbol is equal to one.\footnote{This assures (except for some pathological
cases) that the probabilities assigned to strings sum to one.}
This probability represents the frequency with which the rule is applied
to expand the symbol on its left-hand side.  For example, the
following is a probabilistic context-free grammar:
$$ \begin{tabular}{cclc}
S & $\r$ & NP VP & (1.0) \\
VP & $\r$ & V NP & (1.0) \\
NP & $\r$ & D N & (1.0) \\
D & $\r$ & \t{a} & (0.6) \\
D & $\r$ & \t{the} & (0.4) \\
N & $\r$ & \t{boat} & (0.5) \\
N & $\r$ & \t{cat} & (0.3) \\
N & $\r$ & \t{tree} & (0.2) \\
V & $\r$ & \t{hit} & (0.7) \\
V & $\r$ & \t{missed} & (0.3)
\end{tabular} $$

\begin{figure}
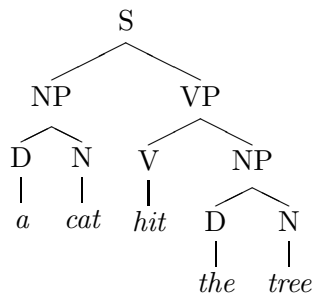


\leaf{\it a}
\branch{1}{D}
\leaf{\it cat}
\branch{1}{N}
\branch{2}{NP}
\leaf{\it hit}
\branch{1}{V}
\leaf{\it the}
\branch{1}{D}
\leaf{\it tree}
\branch{1}{N}
\branch{2}{NP}
\branch{2}{VP}
\branch{2}{S}

$$ \tree $$
\caption{Parse tree for {\it a cat hit the tree}} \label{fig:sampparse}
\end{figure}

To explain how a probabilistic context-free grammar assigns probabilities
to strings, we first need to describe how such a grammar assigns
probabilities to {\it parse trees}.  A parse tree of a string
displays the grammar rules that are applied to form the sentential symbol
from the string.  For example, a parse tree of {\it a cat hit the
tree\/} is displayed in Figure \ref{fig:sampparse}.  Each non-leaf
node in the tree represents the application of a grammar rule.  For instance,
the top node represents the application of the S $\r$ NP VP rule.
The probability assigned to a parse is simply the product of
the probabilities associated with each rule in the parse.
The probability assigned to the parse in Figure \ref{fig:sampparse}
is $0.6 \times 0.3 \times 0.7 \times 0.4 \times 0.2 = 0.01008$, the
terms corresponding to the rules D $\r \t{a}$, N $\r \t{cat}$,
V $\r \t{hit}$, D $\r \t{the}$, and N $\r \t{tree}$, respectively.
All other rules used in the parse have probability 1.  The probability
assigned to a string is the sum of the probabilities of all
of its parses; it is possible for a sentence to have more than
a single parse.\footnote{A good introduction to probabilistic
context-free grammars has been written by \newcite{Jelinek:92a}.}

\ssec{Probabilistic Context-Free Grammars and $n$-Gram Models}
	\label{ssec:pcfgngram}

\begin{figure}
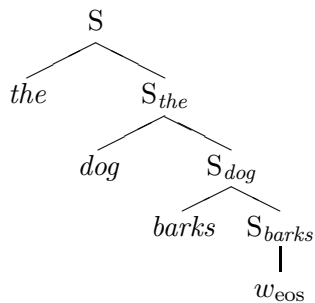


\leaf{\it the}
\leaf{\it dog}
\leaf{\it barks}
\leaf{$w\su{eos}$}
\branch{1}{S$\su{\it barks}$}
\branch{2}{S$\su{\it dog}$}
\branch{2}{S$\su{\it the}$}
\branch{2}{S}

$$ \tree $$
\caption{Parse of {\it the dog barks\/} using a bigram-equivalent grammar}
	\label{fig:ngramparse}
\end{figure}

In this section, we discuss the relationship between
probabilistic context-free grammars and $n$-gram models.  First,
we note that $n$-gram models are
actually instances of probabilistic context-free grammars.
For example, consider a bigram model with probabilities
$p(w_i | w_{i-1})$.  This model can be expressed
using a grammar with $|T| + 1$ nonterminal symbols, where $T$ is
the set of all terminal symbols, \ie, the set of all words.
We have the sentential symbol S and a nonterminal symbol S$_w$ for each word
$w \in T$.  A symbol S$_w$ can be interpreted as representing the
state of having the word $w$ immediately to the left.
The grammar consists of all rules
$$ \begin{array}{ccll}
{\rm S} & \r & w_i \; {\rm S}_{w_i} & (p(w_i | w\su{bos})) \\
{\rm S}_{w_{i-1}} & \r & w_i \; {\rm S}_{w_i} & (p(w_i | w_{i-1})) \\
{\rm S}_{w_{i-1}} & \r & w\su{eos} & (p(w\su{eos} | w_{i-1}))
\end{array} $$
for $w_{i-1}, w_i \in T$ where $w\su{bos}$ and $w\su{eos}$ are the
beginning- and end-of-sentence tokens.  The values in
parentheses are the probabilities associated with each rule
expressed in terms of the probabilities of the corresponding bigram model.
This grammar assigns the identical
probabilities to strings as the original bigram model.
For example, consider the sentence {\it the dog barks}.  The only
parse of this sentence under the above grammar is displayed
in Figure \ref{fig:ngramparse}.  The probability of a
parse is the product of the probabilities of each rule used,
and going from top to bottom we get
$$ p(\t{the} | w\su{bos}) p(\t{dog} | \t{the}) p(\t{barks} | \t{dog})
	p(w\su{eos} | \t{barks}) $$
which is identical to the probability assigned by a bigram model.

Not only can probabilistic context-free grammars 
model the same local dependencies as $n$-gram models,
but they have the potential to model long-distances dependencies beyond the
scope of $n$-gram models.  To demonstrate this, consider the sentence
$$ \t{John read the boy a {\bf story}.} $$
In a trigram model, the word {\it story\/} is assumed to depend
only on the phrase {\it boy a}.  However, there is a strong
dependence between the words {\it read\/} and {\it story}.
We can model this using the following grammar fragment:
$$ \begin{tabular}{lclll}
S$\su{\it read}$ & $\r$ & NP$\su{\it read}\suu{subj}$ & VP$\su{\it read}$ \\
VP$\su{\it read}$ & $\r$ & V$\su{\it read}$ & NP$\su{\it read}\suu{i-obj}$ &
	NP$\su{\it read}\suu{d-obj}$ \\
V$\su{\it read}$ & $\r$ & {\it read} \\
\\
NP$\su{\it story}$ & $\r$ & D$\su{\it story}$ & N$\su{\it story}$ \\
N$\su{\it story}$ & $\r$ & {\it story} \\
\\
NP$\su{\it read}\suu{d-obj}$ & $\r$ & NP$\su{\it story}$
\end{tabular} $$
The symbols with subscript {\it read\/} are symbols that
we restrict to only occur in sentences with the main verb {\it read}.
The symbols with subscript {\it story\/} are symbols that
we restrict to only occur in noun phrases whose head word is {\it story}.
The superscripts on the NP's represent the different roles a noun phrase
can play in a sentence.  The probability associated with the
last rule represents the probability that the word {\it story\/} is
the head of the
direct object of the verb {\it read}; this captures the long-distance
dependency present between these two words.  Probabilistic
context-free grammars can express dependencies between words
arbitrarily far apart.

Thus, we see that probabilistic context-free
grammars are a more powerful formalism than
$n$-gram models, and thus have the potential for superior
performance.  Furthermore, grammars also have the potential
to be more compact than $n$-gram models while achieving equivalent
performance, because grammars can express {\it classing}, or
the grouping together of similar words.
For example, consider the words {\it corporal\/} and {\it sergeant}.
These words have very similar bigram behaviors: for
most words $w$ we have $p(w| \t{corporal}) \approx p(w|\t{sergeant})$
and $p(\t{corporal}|w) \approx p(\t{sergeant}|w)$.  However,
in bigram models the probabilities associated with these two
words are estimated completely independently.  In a grammatical
representation, it is possible instead to introduce a symbol, say $A$,
that corresponds to both words, \ie, to have
$$ A \r \t{sergeant}\; |\; \t{corporal}. $$
We can then have a single set of bigram probabilities $p(w|A)$ and $p(A|w)$
for the symbol $A$,
instead of a separate set for each word.
Notice that this does not preclude having some bigram probabilities specific
to either \t{corporal} or \t{sergeant}, in the cases their behavior
differ.  Because grammars can {\it class\/} together similar words,
equivalent performance to $n$-gram models can be achieved with
much smaller models \cite{Brown:90a}.


Notice that when we use the term {\it grammar},
we are talking of its formal meaning, \ie, a collection
of rules that describe how to build
sentences from words.  This
contrasts with the connotation of the term {\it grammar\/}
in linguistics, of a representation that describes
linguistically meaningful concepts such as noun phrases
and verb phrases.  The symbols in the grammars we consider
do not generally have any relation to linguistic constituents.
Thus, the grammars we consider would not be applicable to
the task of {\it parsing\/} for natural language processing,
where grammars are used to build structure useful
for determining sentence meaning.\footnote{
Probabilistic context-free grammars have been argued to be
inappropriate for modeling natural language because they cannot
model lexical dependencies as do $n$-gram
models \cite{Resnik:92a,Schabes:92a}.  As we have shown that $n$-gram
models are instances of probabilistic context-free grammars, this is obviously
not strictly true.  A more accurate statement is that the context-free
grammars traditionally used for parsing are not appropriate
for language modeling.  These grammars typically have a small
set of nonterminal symbols, \eg, \{ S, NP, VP, \ldots \}, and
grammars with few nonterminal symbols cannot express many lexical dependencies.
However, with expanded symbol sets it is possible to
express these dependencies, \eg, as in the example given earlier in
this section where we qualify nonterminal symbols with their head words.}

There have been attempts to use linguistic grammars for
language modeling \cite{Newell:73a,Woods:76a}.  However, such attempts
have been unsuccessful.  These manually-designed grammars
cannot approach the coverage achieved by algorithms
that statistically analyze millions of words of text.
Furthermore, linguistic grammars are geared toward compactly
describing language; in language modeling the
goal is to describe language in a probabilistically accurate manner.
Models with large numbers of parameters like $n$-gram models
are better suited to this task.

In this work, our goal was to design an algorithm that
induces grammars somewhere between
the rich grammars of linguistics and the flat grammars corresponding
to $n$-gram models, grammars that have the structure for modeling
long-distance dependencies as well as the size for modeling
specific $n$-gram-like dependencies.  In addition, we desired the grammars
to still be significantly more compact than comparable $n$-gram models.

We produced a grammar induction algorithm that largely satisfied these goals.
In experiments, it significantly outperforms the most widely-used grammar
induction algorithm, the Lari and Young algorithm, and on
artificially-generated corpora it outperforms $n$-gram models.
However, on naturally occurring data $n$-gram models are still superior.
The algorithm induces a probabilistic context-free grammar through a greedy
heuristic search within a Bayesian framework, and it refines
this grammar with a post-pass using the Inside-Outside algorithm.
The algorithm does not require the training data to be manually
annotated in any way.\footnote{Some grammar induction algorithms
require that the training data be annotated with parse tree information
\cite{Pereira:92a,Magerman:94a}.  However, these algorithms tend
to be geared toward parsing instead of language modeling.  It is
expensive to manually annotate data, and it is not practical to
annotate the amount of data typically used in language modeling. }

%
\sec{Grammar Induction as Search} \label{sec:assearch}
%

Grammar induction can be framed as a search problem, and has been framed as
such almost without exception in past research \cite{Angluin:83a}.
The search space is taken to be some class of grammars; for example,
in our work we search within the space of probabilistic context-free
grammars.  We search for a grammar that optimizes some quantity,
referred to as the {\it objective function}.  In grammar induction,
the objective function generally contains a factor that reflects how
accurately the grammar models the training data.

Most work in language modeling, including $n$-gram models and the
Inside-Outside algorithm, falls under the {\it maximum likelihood\/} paradigm.
In this paradigm, the objective function is taken to be the
likelihood or probability of the training data given the grammar.  That is,
we try to find the grammar
$$ G = \argmax_G p(O|G) $$
where $O$ denotes the training data or {\it observations}.  The probability
of the training data is the product of the probability of each sentence in
the training data, \ie,
$$ p(O|G) = \prod_{i=1}^n p(o_i|G) $$
if the training data $O$ is composed of the sentences $\{o_1, \ldots, o_n\}$.
The probability of a sentence $p(o_i|G)$ is straightforward to
calculate for a probabilistic grammar $G$.

However, the optimal grammar under
this objective function is one that generates
only sentences in the training data and no other sentences.  In particular,
the optimal grammar consists exactly of all rules of the form
S $\r o_i$, each such rule having probability $c(o_i) / n$ where $c(o_i)$
is the number of times the sentence $o_i$ occurs in the training data.
Obviously, this grammar is a poor model of language at large even
though it assigns a high probability to the training data; this
phenomenon is called {\it overfitting\/} the training data.

In $n$-gram models and work with the
Inside-Outside algorithm \cite{Lari:90a,Lari:91a,Pereira:92a},
this issue is evaded because all of the models considered
are of a fixed size, so that the
``optimal'' grammar cannot be expressed.\footnote{As seen in Chapter
\ref{ch:smooth}, even though $n$-gram models cannot express the
optimal grammar there is still a grave overfitting problem,
which is addressed through smoothing.}
However, in our work we do not wish to limit the size of the
grammars considered.

We can address this issue elegantly by using a Bayesian framework
instead of a maximum likelihood framework.
As touched on in Chapter \ref{ch:intro}, in the Bayesian framework
one attempts to find the grammar $G$ with highest probability
given the data $p(G|O)$, as opposed to the grammar that
yields the highest probability of the data
$p(O|G)$ as in maximum likelihood.  Intuitively, finding the most
probable grammar is more correct than finding the grammar
that maximizes the probability of the data.

Looking at the mathematics, in the Bayesian framework we try to find
$$ G = \argmax_G p(G | O). $$
As it is unclear how to estimate $p(G|O)$ directly, we apply
Bayes' Rule and get
\begin{equation}
G = \argmax_G \frac{p(O|G)p(G)}{p(O)} = \argmax_G p(O|G)p(G)
	\label{eqn:objfn}
\end{equation}
where $p(G)$ denotes the {\it prior\/} probability of a grammar $G$.
The prior probability $p(G)$ is supposed to reflect our {\it a priori\/}
notion of how frequently the grammar $G$ appears in the given domain.

Notice that the Bayesian framework is equivalent to the
maximum likelihood framework
if we take $p(G)$ to be a uniform distribution.  However,
it is mathematically improper to have a uniform
distribution over a countably
infinite set, such as the set of all context-free
grammars.  We give an informal argument describing its mathematical
impossibility and relate this to why the maximum likelihood approach
tends to overfit training data.

Consider selecting a context-free grammar randomly using a uniform
distribution over all context-free grammars.
Now, let us define a {\it size} for each
grammar; for example, we can take the number of characters in the
textual description of a grammar to be its size.  Then,
notice that for any value $k$, there is a zero probability of
choosing a grammar of size less than $k$, since there are
an infinite number of grammars of size larger than $k$ but
only a finite number of grammars smaller than $k$.  Hence, in some sense
the ``average'' grammar according to the uniform distribution
is infinite in size, and this relates to why a uniform distribution
is mathematically improper.  In addition, this is related to why
the maximum likelihood approach prefers overlarge, overfitting grammars,
as the uniform prior assigns far too much probability to large grammars.

Instead, we argue that taking a
{\it minimum description length\/} (MDL) principle \cite{Rissanen:78a}
prior is desirable.  The minimum description length principle
states that one should select a grammar $G$ that minimizes
the sum of $l(G)$, the length of the description of the grammar,
and $l(O|G)$, the length
of the description of the data given the grammar.  We will later
give a detailed description of what these lengths mean; for now, suffice
it to say that this corresponds to taking a prior of the form
$$ p(G) = 2^{-l(G)} $$
where $l(G)$ is the length of the grammar $G$ in bits.  For example,
we can take $l(G)$ to be the length of a textual description of
the grammar.

Intuitively, this prior is appealing because it captures the intuition
behind Occam's Razor, that simpler (or smaller) grammars are preferable over
complex (or larger) grammars.  Clearly, the prior $p(G)$ assigns
higher probabilities to smaller grammars.  However, this prior
extends Occam's Razor by providing a concrete way to trade off
the size of a grammar with how accurately the grammar models
the data.  In particular, we try to find the grammar that
maximizes $p(G) p(O|G)$.  The term $p(G)$ favors
grammars that are small, and the term $p(O|G)$ favors
grammars that model the training data well.

This preference for small grammars over large addresses the
problem of overfitting.  The optimal grammar under
the maximum likelihood paradigm will be given a poor score
because its prior probability $p(G)$ will be very small, given its vast size.
Instead, the optimal grammar under MDL will be a compromise
between size and modeling accuracy.

Because {\it coding theory\/} plays a key role in the future discussion,
we digress at this point to introduce some basic concepts in the field.
Coding theory forms the basis of
the {\it descriptions\/} used in the minimum description
length principle, and it
can be used to tie together the MDL principle with the Bayesian framework.

\ssec{Coding}

{\it Coding\/} can be thought of as the study of storing
information compactly.  In particular, we are interested
in representing information just using a binary alphabet, 0's and 1's,
as is necessary for storing information in a computer.
Coding just describes ways of mapping information into
strings of binary digits or bits in a one-to-one manner, so
that the original information can be reconstructed from the associated
bit string.

For example, consider the task of coding the outcome of a coin flip.
A sensible code is to map tails to the bit string 0, and to map
heads to the bit string 1 (or vice versa).  Another possibility
is to map both heads and tails to 0.  This is an invalid code,
because it is impossible to reconstruct whether a
coin flip was heads or tails from the yielded bit string.
In general, distinct outcomes must be mapped to distinct bit strings; that is,
mappings need to be one-to-one.
Another possible code is to map heads to the bit string 00,
and to map tails to the bit string 11.  This is a valid code,
but it is inefficient as it codes the information using two bits
when one will do.

Codes can be used to store arbitrarily complex information.  For
example, the ASCII convention maps letters of the alphabet to
eight-bit strings.  In this convention, the text {\it hi\/} would be mapped to
the sixteen-bit string 0110100001101001.  By using the ASCII
convention, any data that can be expressed through text can be
mapped to bit strings.

For obvious reasons, coding theory is concerned with finding ways to
code information with as few bits as possible.  For example, coding theory
is at the core of the field of {\it data compression}.  We now describe how
to code data optimally, moving from simple examples to more complex ones.

\sssec{Fixed-Length Coding}

First, consider the case of coding the outcome of a single coin flip,
which we showed earlier can be coded using a single bit.
There are two possible outcomes to a coin flip, and there are two
possible values for a single bit, so it is possible to find
a one-to-one mapping from outcomes to bit values.
Now, consider coding the outcome of $k$ coin flips.  Intuitively,
this should be codable using $k$ bits, and in fact it can.
There are $2^k$ possible outcomes to $k$ coin flips, and there
are $2^k$ possible values of a $k$-bit string, so again we
can find a one-to-one mapping.  In general, to code any
information that has exactly $2^k$ possible values, we need at most $k$ bits.
Alternatively, we can phrase this as:
to code information with $n$ possible values,
we need at most $\lceil \log_2 n \rceil$ bits.
For example, in New York Lotto, which involves
picking 6 distinct numbers from the values 1 through 48, there are
${48\choose 6} = 12,271,512$ possible combinations.  Then,
we need at most $\lceil log_2 12,271,512 \rceil = 24$
bits to code a New York Lotto ticket.  Notice that
in this discussion we ignore how difficult it is to construct
the coder and decoder; to write a program that maps Lotto
tickets to and from distinct 24-bit strings is not trivial.
It usually possible to find inefficient
codes that are much easier to code and decode.  For example, we could
just store a Lotto ticket as text using the ASCII convention.

\sssec{Variable-Length Coding} \label{sssec:varcode}

While the preceding analysis is optimal if we require that
all of the bit strings mapped to are of the same length,
in most cases one can do better on average if outcomes can
be mapped to bit strings of different lengths.  In particular,
if certain outcomes are more frequent than
others, these should be mapped to shorter bit strings.
While this may cause infrequent strings to be mapped to
longer bit strings than in a fixed-length coding, this
is more than made up by the savings from shorter bit strings
since the shorter strings correspond to more frequent outcomes.\footnote{
We see this principle followed in language: most common words have
short spellings, and long expressions that are used frequently
in some context are often abbreviated.}

For example, let us consider the coding of the information
of which of three consecutive coin flips, if any, is the first one to be heads.
There are four possible outcomes: the first flip, the second
flip, the third flip, or none of them.  Thus, we can code
the outcome using two bits with a fixed-length code.  Now,
let us consider a different coding, where we just use three bits
to code the outcome of each of the three coin flips in order, using 0 to
mean tails and 1 heads.  This is a valid coding, since we can
still recover which of the flips yields the first head, but
obviously this coding is less efficient
than the previous one because it uses three bits instead of two.
However, notice that in some cases some of the bits in this
coding are superfluous.  For example,
if the first bit is 1, then we know the earliest flip to be heads
is the first one regardless of the later flips, so there is no
need to include the last two bits.  Likewise, if the
first bit is 0 and the second bit is 1, we do not need to
include the last bit because we know the earliest flip to be heads is
the second one.  Instead of a fixed-length code, we can
assign the bit strings 1, 01, 001, and 000 to the four outcomes.
Notice that the probabilities of these four outcomes are
0.5, 0.25, 0.125, and 0.125, if the coin is fair.  Thus, on average
we expect to use
$0.5 \times 1 + 0.25 \times 2 + 0.125 \times 3 + 0.125 \times 3 =  1.75$ bits
to code an outcome, taking into account the relative frequencies
of each outcome.  This is superior to the average of two bits yielded by
the optimal fixed-length code.\footnote{One may ask why we could not
use an even shorter code, \eg, the bit strings 0, 1, 00, and 01.
The reason is that it is required that codes are unambiguous even
when used to code multiple trials consecutively.
For example, if we coded two of the above trials
consecutively with this new code and yielded
the bit string 000, we would not be able to tell whether this should
be interpreted as (0)(00) or (00)(0).  One way to assure
unambiguity is to require that no codeword is a prefix of another,
as is satisfied by the original code given in the example.}

In general, if each outcome has a probability of the form $2^{-k}$ for
$k$ integer, then it is provably optimal to assign an outcome with
probability $2^{-k}$ a bit string of length $k$.  Alternatively phrased:
given that the probabilities of all outcomes are
of the form $2^{-k}, k \in {\cal N}$,
an outcome with probability $p$ is optimally coded using
$\log_2 \frac{1}{p}$ bits.  Thus, the code
described in the last example is optimal, as $\log_2 \frac{1}{0.5} = 1$,
$\log_2 \frac{1}{0.25} = 2$, and $\log_2 \frac{1}{0.125} = 3$.
Notice that in the case there are $2^k$ equiprobable outcomes, this
formula just comes out to a fixed-length code of length $k$.

Now, consider the case of coding probabilities that are not negative
powers of two.  For example, let us code the outcome of a single toss
of a fair 3-sided die.
Intuitively, we want to assign codeword lengths that are appropriate
for probabilities that are
negative powers of two that are near to the actual probabilities
of each outcome.  In fact, there is an algorithm for performing
this assignment in an optimal way, namely Huffman coding \cite{Huffman:52a}.
In this case, Huffman coding yields the codewords 0, 10, and 11.
Clearly, the codeword lengths do not follow the relation that
an outcome with probability $p$ has codeword length $\log_2 \frac{1}{p}$ as
in this case these values are not integers.\footnote{However,
Huffman coding does guarantee that on average outcomes are assigned
codewords at most one bit longer than what is dictated by the
$\log_2 \frac{1}{p}$ relation.  To see how this bound can be
achieved simply, we can just round down each probability to the next
lower negative power of two, and assign codeword lengths as described
earlier.}

However, consider the case where instead of coding a single toss,
we are coding $k$ tosses of a fair 3-sided die.  Notice that
there are $3^k$ possible outcomes, and as mentioned earlier we
can code this using $\lceil \log_2 3^k \rceil$ bits using a fixed-length
code.  Hence, on average each coin toss requires
$\frac{\lceil \log_2 3^k \rceil}{k} = \frac{\lceil k \log_2 3 \rceil}{k}$ bits.
As $k$ grows large, this approaches the value $\log_2 3$.  By
coding multiple outcomes jointly, we approach the
limit where each individual outcome (of probability $p = \frac{1}{3}$)
can be coded on average using
$log_2 \frac{1}{p} = \log_2 \frac{1}{\frac{1}{3}} = log_2 3$
bits; this is the same relation we found when all probabilities were
negative powers of two.

This result extends to the case where not all outcomes are equiprobable;
instead of fixed-length coding for the joint trials, we can use
Huffman coding of the joint trials to approach this limit.
In general, if a particular outcome has probability $p$,
in the limit of coding a large number of trials, each of those
outcomes will take on average $\log_2 \frac{1}{p}$ bits to code
in the optimal coding \cite{Shannon:48a,Cover:78a}.  In fact, this limit
can be realized in practice with an efficient algorithm called
{\it arithmetic coding} \cite{Pasco:76a,Rissanen:76a}.

\ssec{Description Lengths} \label{ssec:desclen}

Now, let us return to the minimum description length principle and
the meaning of a {\it description}.  Recall that MDL states that one should
minimize the sum of $l(G)$, the length of the description of the
grammar, and $l(O|G)$, the length
of the description of the data given the grammar.
A description simply refers to the bit string that is used to code
the given information.  Notice that we do not care what
the actual bit string that composes a description is; we are only
concerned with its length.

First, let us consider $l(O|G)$.  Typically, $G$ is a probabilistic
grammar that assigns probabilities to sentences $p(o_i|G)$, and
we can calculate the probability of the training data as
$p(O|G) = \prod_{i=1}^n p(o_i|G)$ where the training data $O$
is composed of the sentences $\{ o_i, \ldots, o_n\}$.  Then,
using the result that an outcome with probability $p$ can be coded
optimally with $\log_2 \frac{1}{p}$ bits, we get that taking
$l(O|G)$ to be $\log_2 \frac{1}{p(O|G)}$
should yield the lowest lengths on average.

Notice that for the MDL principle to be meaningful,
we need to use an optimal coding as opposed
to some arbitrary inefficient coding.  There are many
descriptions of a given piece of data.  For example,
for any description of some data given a grammar, we can create
additional descriptions of the same data by just padding
the end of the original description with 0's.  Clearly,
it is easy to make descriptions arbitrarily long.  However,
it is not possible to make descriptions arbitrarily compact.
There is a lower bound to the length of the description
of any piece of data \cite{Solomonoff:60a,Solomonoff:64a,Kolmogorov:65a},
and we can use this lower bound
to define a meaningful description length for a piece of data.
This is why we choose an optimal coding for calculating $l(O|G)$.
This dictum of optimal coding extends as well
to calculating $l(G)$, the length of the description of a
grammar.

Thus, it is not appropriate to use textual descriptions of grammars
as mentioned in Section \ref{sec:assearch}, as this is
rather inefficient.  For example, consider the following textual
description of a grammar segment:
\begin{verbatim}
NP->D N
D->a|the
N->boat|cat|tree
\end{verbatim}
This textual description is 34 characters long (including carriage
returns), which translates to
272 bits under the ASCII convention of eight bits per character.
We can achieve a significantly smaller description using
a more complex encoding, where the grammar is coded in three
distinct sections:
\bi
\i We code the list of terminal symbols as text:
\begin{verbatim}
a the boat cat tree
\end{verbatim}
which comes to 20 characters including the carriage return.

\i We code the number of nonterminal symbols and the number of grammar rules
also as text:
\begin{verbatim}
3 3
\end{verbatim}
which comes to 4 characters including the carriage return.  Notice
that the names of nonterminal symbols are not relevant in describing
a grammar; these symbols can be renamed arbitrarily without
affecting what strings the grammar generates.

\i Finally, we code the list of grammar rules, where each grammar rule
is coded in several parts:
	\bi
	\i The nonterminal symbol on the left-hand side of a rule can be
	coded using two bits, as there are a total of three nonterminal symbols.
	\i To code whether a rule is of the form $A \r a_1 a_2 \cdots$
	or of the form $A \r a_1 | a_2 | \cdots$, we use a single bit.
	(In this example, we do not consider rules combining both forms.)
	\i To code how many symbols are on the right-hand side
	of a rule, we use three bits.  With three bits we can code up
	to a length of eight;
	if a rule is longer it can be split into multiple rules.
	\i To code each symbol on the right-hand side of a rule, we use three bits
	to code which of the eight possible symbols it is (three nonterminal,
	five terminal).
	\ei
Under this coding, the first two rules each take 12 bits, and the
third takes 15 bits.
\ei

This comes to a total of 24 characters for the first two sections,
or 192 bits under the ASCII convention, and 39 bits for the last section,
yielding a total of 231 bits.
This is significantly less than the 272 bits of a naive textual encoding.
Using more advanced techniques that will be described later,
grammar descriptions can be made even more compact.  In addition,
the grammars we will be using later will be probabilistic, so we
will also have to code probability values.

Just as we used $p(O|G)$ to calculate $l(O|G)$, we can use
the prior probability $p(G)$ mentioned in equation (\ref{eqn:objfn})
to give us insight into $l(G)$.  According to coding theory,
to calculate the optimal length $l(G)$ of a grammar $G$ we need to know the
probability of the grammar $p(G)$.  An alternative approach
to explicitly designing encodings like above is instead to
design a prior probability $p(G)$ and to
define an encoding such that $l(G) = \log_2 \frac{1}{p(G)}$ just
as we did for $l(O|G)$.  However, unlike $p(O|G)$ the distribution
$p(G)$ is not straightforward to estimate.  Furthermore,
it is important to note that in order for a coding to be optimal
(\ie, produce the shortest descriptions on average), the
underlying probability distribution must be accurate.
For some distribution $p(G)$, we know that using $\log_2 \frac{1}{p(G)}$ bits
to code a grammar $G$ is optimal {\it only if\/} $p(G)$ is
the {\it correct\/} underlying distribution on grammars.

For instance, consider the example given in Section \ref{sssec:varcode}
of coding which of three consecutive coin flips is the first to turn
up heads.  A fixed-length code requires two bits to code this,
and we showed that by assigning the codewords 1, 01, 001, and 000
to the outcomes: first flip, second flip, third flip, and no flip,
respectively, we can achieve an improved average of 1.75 bits, assuming
the coin is fair.  However, consider a biased coin whose probability
of heads is $\frac{1}{4}$.  Then,
the frequencies of the four outcomes become 1/4, 3/16, 9/64, and 27/64,
respectively, and this yields an average codeword length of
$(1/4 \times 1) + (3/16 \times 2) + (9/64 \times 3) + (27/64 \times 3) =
2.3125$ bits, which is significantly worse than the fixed-length code.
Thus, we see that for a coding to be efficient we must have an
accurate model of the data.

Applying this observation to coding grammars, we see that deriving
the lengths $l(G)$ of grammars from a prior $p(G)$ is no better
than estimating $l(G)$ directly; we have no guarantee that the
prior $p(G)$ we choose is at all accurate.  However, this
relationship does provide us with another perspective
with which to view grammar encodings.  For every grammar encoding
describing grammar lengths $l(G)$ there is an associated
prior $p(G) = 2^{-l(G)}$, and we should choose encodings that
lead to priors $p(G)$ that are good models of grammar frequency.
For example, for
grammars $G$ we perceive to be typical, \ie, to have high probability $p(G)$,
we want $l(G)$ to be low.  In other words, we want typical grammars
to have short descriptions.  Hence, referring to the two
grammar encodings given earlier, as the latter grammar encoding
assigns shorter descriptions to typical grammars than the naive
encoding,\footnote{Actually, this is not clear.  For smaller grammars,
the complex encoding should be more efficient since, for example,
it can code symbol identities using a small number of bits while
in a text representation a symbol is represented using a minimum of
one character, or eight bits.  For large grammars, text encodings
may be more efficient since they can express variable-length encodings
of symbol identities, while the complex encoding assumes fixed-length
encodings of symbol identities.
} we conclude that in some sense the latter encoding corresponds
to a more accurate prior probability on grammars.


\ssec{The Minimum Description Length Principle} \label{ssec:mdl}

As touched on in the last section,
the observation that an object with probability $p$ should be
coded using $\log_2 \frac{1}{p}$ bits gives us a way
to equate probabilities and description lengths, and this
is the key in showing the relation between the minimum description length
principle and the Bayesian framework.
Under the Bayesian framework, we want to find the grammar
$$ G = \argmax_G p(O|G) p(G). $$
Under the minimum description length principle, we want to find the grammar
$$ G = \argmin_G [l(O|G) + l(G)]. $$
Then, we get that
\begin{eqnarray*}
G & = & \argmax_G p(O|G) p(G) \\
	& = & \argmin_G [-\log_2 p(O|G) p(G)] \\
	& = & \argmin_G [\log_2 \frac{1}{p(O|G)} + \log_2 \frac{1}{p(G)}] \\
	& = & \argmin_G [l_p(O|G) + l_p(G)]
\end{eqnarray*}
where $l_p(\a)$ denotes the length of $\a$ under the optimal coding
given $p$.  Thus, any problem framed in the Bayesian framework
can be converted to an equivalent problem under MDL, by just
taking the description lengths to be the optimal ones dictated by
the given probabilities.  Likewise, any problem framed under MDL can
be converted to an equivalent one in the Bayesian framework, by
choosing the probability distributions that would yield the given
description lengths.  For example, it is easy to see that
the Bayesian prior corresponding to the MDL principle is
$p(G) = 2^{-l(G)}$, as touched on earlier.

Thus, from a mathematical point of view, the minimum description
length principle does not give us anything above the Bayesian
framework.  However, from
a paradigmatic perspective, MDL provides two important ideas.

Firstly, MDL gives us a new perspective for creating prior
distributions on grammars.
By noticing that any grammar encoding scheme implicitly describes
a probability distribution $p(G) = 2^{-l(G)}$, we can create
priors by just designing encoding schemes.  For example,
both of the grammar encoding schemes used in Section \ref{ssec:desclen}
lead to prior distributions rather different from those usually
found in probability theory.  Viewing prior distributions
in terms of encodings extends the toolbox one has for designing
prior distributions.  In addition, one can mix and match conventional
prior distributions from probability theory with those stemming
from an encoding perspective.

Secondly, it has been observed that ``MDL-style'' priors of
the form $p(G) = 2^{-l(G)}$ can be
good models of the real world
\cite{Solomonoff:64a,Rissanen:78a,Li:93a}.\footnote{Closely related
to the minimum description length principle is
the {\it universal a priori probability}.  The universal {\it a priori\/}
probability can be shown to dominate all enumerable prior
distributions by a constant.  The minimum description length principle
can be thought of as a simplification of this elegant but
incomputable universal distribution.  A thorough discussion of this topic
is given by \newcite{Li:93a}.
} To demonstrate this, let
us consider some examples of real-world data.
Let us say you see one hundred flips of a coin,
and each time it turns up heads.  Clearly,
you expect a head with very high probability on the next toss.\footnote{
If you {\it knew\/} the coin was fair, then you would still expect
the next toss to be heads with probability 0.5, as in the
canonical grade school example.  However, it is rare that you
know with absolute certainty that a coin is fair. }
Or, let us say you peek at someone's computer terminal
and see the following numbers output: 2, 3, 5, 7, \ldots, 83, 89.
Then, you expect the next number to be output to be 97 with
very high probability.  Or, let us say you look at some text
and notice that after each of the ten occurrences of
the word {\it Gizzard's\/} the word {\it Gulch\/} appears immediately
afterwards.  Then, if you see the word {\it Gizzard's\/} again you expect
the word {\it Gulch} will follow with high probability.
In general, when you notice a pattern in some data in the real world,
you expect the pattern to continue in later samples
of the same type of data.

This behavior can be captured with an MDL-style prior.  In
particular, we can capture this behavior by choosing a prior
that assigns high probabilities to data that can be described
with short {\it programs}.  By {\it programs}, we mean programs
written in a computer language such as Pascal or Lisp.  For
example, let us take our programming language to be a Pascal-like
pseudo-code.  Now, consider estimating the probability that a
coin turns up heads on the next toss, given that
all hundred previous tosses of the coin yielded heads.  That is,
we want to estimate
$$ p(h| 100\:h\mbox{'s}) =
	\frac{p(100\:h\mbox{'s}, h)}{\sum_{x=\{h, t\}} p(100\:h\mbox{'s}, x)} =
	\frac{p(101\:h\mbox{'s})}{p(101\:h\mbox{'s}) +
	p(100\:h\mbox{'s}, t)}. $$
Intuitively, this probability should be high, so we want
$$ p(101 h\mbox{'s}) > p(100 h\mbox{'s}, 1 t). $$
A program that outputs 101 $h$'s is significantly
shorter than a program that outputs 100 $h$'s and a $t$.  For example,
for the former we might have
\begin{verbatim}
for i := 1 to 101 do
  print "h";
\end{verbatim}
while for the latter we might have
\begin{verbatim}
for i := 1 to 100 do
  print "h";
print "t";
\end{verbatim}
Thus, by assigning higher probabilities to data that can be
generated with shorter programs, we get the desired behavior on
this example.

Similarly, for the case of predicting the next output given
the preceding outputs 2, 3, 5, 7, \ldots, 89, we want
that
$$ p( 2,3,5,\ldots, 89, 97) > p(2, 3, 5, \ldots, 89, x) $$
for $x \neq 97$.  Again, a program that generates the former
will generally be shorter than one that generates the latter.  For example,
we might have
\begin{verbatim}
for i := 2 to 97 do
  <code for printing out i if it is prime>
\end{verbatim}
as opposed to
\begin{verbatim}
for i := 2 to 89 do
  <code for printing out i if it is prime>
print x;
\end{verbatim}

For the example where the word {\it Gulch\/} always follows the word
{\it Gizzard's\/} the ten times the word {\it Gizzard's\/} occurs, and
where we want to estimate the probability that the word {\it Gulch\/}
follows {\it Gizzard's\/} in its next occurrence,
consider a program that encodes text using a bigram-like model.
Assume that for efficiency, the program only explicitly
codes those bigram probabilities
that are non-zero, as only a small fraction of all bigrams
occur in practice.  To model the case
where {\it Gulch\/} does follow {\it Gizzard's\/} in
its next occurrence, we only need to code a single nonzero probability
of the form $p(x | \t{Gizzard's})$, \ie, for $x = \t{Gulch}$.
However, if a different
word follows {\it Gizzard's}, to model this new data we need
an additional nonzero probability of the form $p(x | \t{Gizzard's})$.
Thus, presumably the program (including the description of its bigram
model) coding this latter case will be larger than the one coding
the former case.  Thus, by assigning higher probability to
data generated with smaller programs, we get the desired
behavior of predicting {\it Gulch\/} with high probability.

Now, notice that using a prior of the form $p(G) = 2^{-l(G)}$
results in this behavior if we just replace grammars $G$
with programs $G_p$.  That is, we can express the
probability $p(O)$ of some data or observations $O$ as
$$ p(O) = \sum_{G_p} p(O, G_p) = \sum_{G_p} p(G_p) p(O|G_p)
	= \sum\su{output($G_p$) = $O$} p(G_p) $$
where we have $p(O|G_p) = 1$ if the output of program $G_p$ is $O$
and $p(O|G_p) = 0$ otherwise.  Substituting in the prior on
programs $p(G_p) = 2^{-l(G_p)}$, we get
$$ p(O) = \sum\su{output($G_p$) = $O$} 2^{-l(G_p)} $$
which gives us that data that can be described with shorter
programs have higher probability.

While the MDL-style prior $p(G_p) = 2^{-l(G_p})$ yields
this nice behavior, there are several provisos.  First
of all, notice that this prior is not appropriate for
making precise predictions.  For example, while in the above
examples we make arguments about the relative magnitude
of different probabilities, it would be folly to try
to nail down actual probabilities and expect them to be accurate.
Also, notice that we used data sets of non-trivial
size; this is because the inaccuracy of this type of prior is especially
marked for small data sets.  For example, consider the case
of predicting the next value in the sequence 2, 3, 5, 7.
In this case, it is unlikely that the shortest program that outputs
this sequence is of the form
\begin{verbatim}
for i := 2 to 7 do
  <code for printing out i if it is prime>
\end{verbatim}
and the argument given earlier for the longer sequence of primes does
not hold.  Instead, a shorter program would be
\begin{verbatim}
print "2, 3, 5, 7";
\end{verbatim}
Hence, for this short sequence of primes it is unclear whether
the MDL-style prior would predict 11 with high probability,
even though intuitively this is the correct prediction.

Both of these issues are related to the fact that there are many different
programming languages we could use to describe programs, and that
the same program in different languages may have very different
lengths.  Thus, the specific behavior of the prior depends greatly
on the language used.  However, for large pieces of data the
relative differences in program length between programming languages
becomes smaller.  For example, if a program is 10 lines in Lisp
and 1,000 lines in Basic, this is a relatively large difference.
However, a 100,010-line Lisp program and a 101,000-line Basic program
are nearly the same length from a relative perspective.\footnote{
For any two Turing-machine-equivalent languages, there exists a constant $c$
such that any program in one language, say of length $l$ bits,
can be duplicated in the other language using at most $l + c$ bits.
The general idea behind the proof is that you can just write
an interpreter (of length $c$ bits) for the former language in the
latter language.}  Thus, for large pieces of data the prior
will yield qualitatively similar results independent of programming
language.

In any case, we choose an MDL-style prior in this work because of
the observation that by assigning higher probabilities to smaller
programs we get a very rich behavior that seems to model the
real-world fairly well.  However, instead of considering
a general programming language, we tailor our description
language to one that describes only probabilistic context-free grammars.
Considering a restricted language simplifies the search problem a great deal,
and context-free grammars are able to
express many of the important properties of language.  Furthermore,
we observed above that a general MDL prior cannot make
quantitatively accurate predictions.  In this work, we attempt
to tailor the prior so that meaningful quantitative predictions can be
made in the language domain.

To summarize, we treat grammar induction as a search for
the grammar $G$ with the highest probability given the data
or observations $O$,
which is equivalent to finding the grammar $G$ that maximizes
the objective function $p(O|G) p(G)$, the likelihood of the training data
multiplied by the prior probability of the grammar.
We take the prior $p(G)$ to be $2^{-l(G)}$ as dictated by
the minimum description length principle.
While this framework does not restrict us
to a particular grammar formalism, in this work we consider only
probabilistic context-free grammars, as it is a fairly
simple, yet expressive, representation.
We describe our search strategy in Section \ref{sec:goutline}.
We describe what encoding scheme we use to calculate $l(G)$ in
Section \ref{sec:gdetails}.

%
\sec{Algorithm Outline} \label{sec:goutline}
%

We assume a simple greedy search strategy.\footnote{
While searches that maintain a population of hypotheses can yield
better performance, it is unclear how to efficiently
maintain multiple hypotheses in this domain because each hypothesis
is a grammar that can potentially be very large.  However,
stochastic searches such as simulated annealing could be practical,
though we have not tested them.}
We maintain a single
hypothesis grammar that is initialized to a small, trivial grammar.
We then try to find a modification to the hypothesis grammar, such
as the addition of a grammar rule, that results in a grammar
with a higher score on the objective function.  When we find a superior
grammar, we make this the new hypothesis grammar.  We repeat
this process until we can no longer find a modification that improves
the current hypothesis grammar.

For our initial grammar, we choose a grammar that can generate
any string, to assure that the grammar
assigns a nonzero probability to the training data.\footnote{
Otherwise, the objective function will be zero, and unless
there is a single move that would cause the objective function
to be nonzero, the gradient will also be zero, thus making it
difficult to search intelligently.}
At the highest level of the grammar, we have the rules
$$ \begin{array}{ccll}
{\rm S} & \r & {\rm SX} \hspace{0.3in} & (1 - \e) \\
{\rm S} & \r & {\rm X}  & (\e)
\end{array} $$
expressing that a sentence S is a sequence of X's.
The quantities in parentheses are the probabilities associated
with the given rules; we describe $\e$ and other
rule probability parameters in detail in Section \ref{ssec:rules}.

Then, we have rules
$$ \begin{array}{ccll}
{\rm X} & \r & A \hspace{0.3in} & (p(A))
\end{array} $$
for every nonterminal symbol $A \neq$ S,X in the grammar.  Combined
with the earlier rules, we have that a sentence is composed of a
sequence of independently
generated nonterminal symbols.  We maintain this property throughout
the search process; that is, for every symbol $A$ that we add to the grammar,
we also add a rule X $\r A$.  This assures that
the sentential symbol can expand to every symbol; otherwise, adding
a symbol will not affect the probabilities that a grammar assigns
to strings.

\begin{table}

$$\begin{array}{l}
\begin{array}{ccllcl}
{\rm S} & \r & {\rm SX} \hspace{0.3in} & (1 - \e) \\
{\rm S} & \r & {\rm X}  & (\e) \\
{\rm X} & \r & A & (p(A)) & \;\;\;\;\; & \forall \; A \in N - \{{\rm S, X}\} \\
A_{\a} & \r & \a & (1) & & \forall \; \a \in T
\end{array} \\
\\
\begin{array}{ccl}
N & = & \mbox{the set of all nonterminal symbols} \\
T & = & \mbox{the set of all terminal symbols} \\
\\
\multicolumn{3}{l}{\mbox{Probabilities for each rule are in parentheses.}}
\end{array}
\end{array} $$
\caption{Initial hypothesis grammar \label{tab:init}}
\end{table}

To complete the initial grammar, we have rules
$$ \begin{array}{ccll}
A_{\a} & \r & \a \hspace{0.3in} & (1)
\end{array} $$
for every terminal symbol or word $\a$.  That is, we have a nonterminal
symbol expanding exclusively to each terminal symbol.  With
the above rules, the sentential
symbol can expand to every possible sequence of words.
(For every symbol $A_{\a}$, there will be an accompanying rule
X $\r A_{\a}$.)
The initial grammar is summarized in Table \ref{tab:init}.

We use the term {\it move set\/} to describe the set of modifications
we consider to the current hypothesis grammar to hopefully produce
a superior grammar.  Our move set includes the following moves:
\begin{description}
\item[Move 1:] Create a rule of the form $A \r BC$ ({\it concatenation})
\item[Move 2:] Create a rule of the form $A \r B|C$ ({\it classing})
\end{description}
For any context-free grammar, it is possible to express a weakly equivalent
grammar using only rules of these forms.
As mentioned before, with each new symbol $A$ we also create a rule X $\r A$.
We describe the move set in more detail in
Section \ref{ssec:moveset}.\footnote{In this chapter, we will
use the symbols $A, B, \ldots$ and symbols of the form
$A_{\a}, B_{\a}, \ldots$ to denote general nonterminal symbols, \ie,
nonterminal symbols other than S and X.}

\ssec{Evaluating the Objective Function}

Consider the task of calculating the objective function
$p(O|G)p(G)$ for some grammar $G$.  Calculating $p(G) = 2^{-l(G)}$ turns out
to be inexpensive; however, calculating $p(O|G)$ requires evaluating
the probability $p(o_i|G)$ for each sentence $o_i$ in the training data,
which entails parsing each sentence in the
training data.
We cannot afford to parse the training data for each
grammar considered; indeed, to ever be practical for large data sets,
it seems likely that we can only afford to parse the data once.

To achieve this goal, we employ several approximations.  First, notice that
we do not ever need to calculate the actual value of the objective function;
we need only to be able to distinguish when a move applied to the current
hypothesis grammar produces a grammar that has a higher score on the
objective function.  That is, we need only to be able to calculate
the {\it difference\/} in the objective function resulting from a move.
This can be done efficiently if we can quickly approximate how the
probability of the training data changes when a move is applied.

\begin{figure}
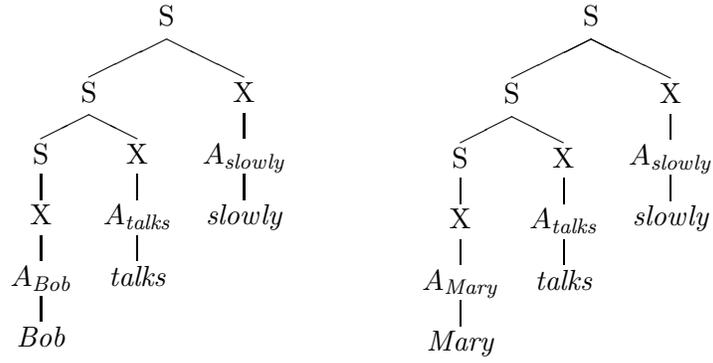


$$
\leaf{\it Bob}
\branch{1}{$A\su{\it Bob}$}
\branch{1}{X}
\branch{1}{S}
\leaf{\it talks}
\branch{1}{$A\su{\it talks}$}
\branch{1}{X}
\branch{2}{S}
\leaf{\it slowly}
\branch{1}{$A\su{\it slowly}$}
\branch{1}{X}
\branch{2}{S}
\vtop{\hbox{}\hbox{\tree}} \hspace{0.5in}
\leaf{\it Mary}
\branch{1}{$A\su{\it Mary}$}
\branch{1}{X}
\branch{1}{S}
\leaf{\it talks}
\branch{1}{$A\su{\it talks}$}
\branch{1}{X}
\branch{2}{S}
\leaf{\it slowly}
\branch{1}{$A\su{\it slowly}$}
\branch{1}{X}
\branch{2}{S}
\vtop{\hbox{}\hbox{\tree}} $$

\caption{Initial Viterbi parse} \label{fig:before}
\end{figure}

\begin{figure}
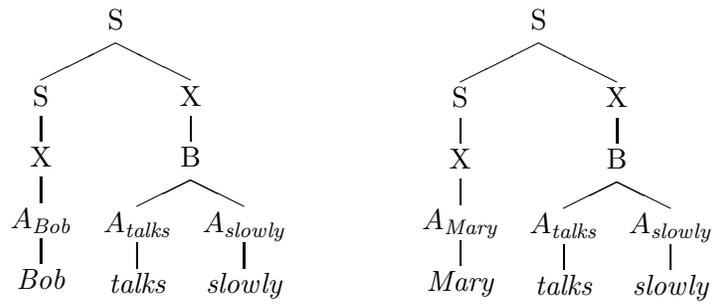


$$
\leaf{\it Bob}
\branch{1}{$A\su{\it Bob}$}
\branch{1}{X}
\branch{1}{S}
\leaf{\it talks}
\branch{1}{$A\su{\it talks}$}
\leaf{\it slowly}
\branch{1}{$A\su{\it slowly}$}
\branch{2}{B}
\branch{1}{X}
\branch{2}{S}
\vtop{\hbox{}\hbox{\tree}} \hspace{0.5in}
\leaf{\it Mary}
\branch{1}{$A\su{\it Mary}$}
\branch{1}{X}
\branch{1}{S}
\leaf{\it talks}
\branch{1}{$A\su{\it talks}$}
\leaf{\it slowly}
\branch{1}{$A\su{\it slowly}$}
\branch{2}{B}
\branch{1}{X}
\branch{2}{S}
\vtop{\hbox{}\hbox{\tree}} $$

\caption{Predicted Viterbi parse} \label{fig:after}
\end{figure}

To make this possible, we approximate the probability of the
training data $p(O|G)$ by the probability of the single most probable
parse, or {\it Viterbi\/} parse, of the training data.  Furthermore,
instead of recalculating the Viterbi parse of the training data from
scratch when a move is applied, we use heuristics to predict how a move
will change the Viterbi parse.
For example, consider the case where the training
data consists of the two sentences
$$ O = \{ \t{Bob talks slowly}, \t{Mary talks slowly} \} $$
In Figure \ref{fig:before}, we display the Viterbi parse of this
data under the initial hypothesis grammar of our algorithm.

Now, let us consider the move of adding the rule
$$ B \r A\su{\it talks} \; A\su{\it slowly} $$
to the initial grammar
(as well as the concomitant rule X $\r B$).  A reasonable heuristic
for predicting how the Viterbi parse will change is to replace
adjacent X's that expand to $A\su{\it talks}$ and
$A\su{\it slowly}$
respectively with a single X that expands to $B$, as displayed
in Figure \ref{fig:after}.  This is the actual heuristic we use for moves
of the form $A \r BC$, and we have analogous heuristics for each move
in our move set.  By predicting the differences in the Viterbi parse resulting
from a move, we can quickly estimate the change in the probability of
the training data.

Notice that our predicted Viterbi parse can stray a great deal from
the actual Viterbi parse, as errors can accumulate as move after move
is applied.  To minimize these effects,
we process the training data incrementally.
Using our initial hypothesis grammar, we parse the first sentence
of the training data and search for the optimal
grammar over just that one sentence using the described
search framework.  We use the resulting grammar to parse the
second sentence, and then search for the optimal grammar over the
first two sentences using the last grammar as the starting point.
We repeat this process, parsing the next
sentence using the best grammar found on the previous sentences
and then searching for the best grammar taking into account this
new sentence, until the entire training corpus is covered.

Delaying the parsing of a sentence
until all of the previous sentences are processed should yield
more accurate Viterbi parses during the search process than if
we simply parse the whole corpus with the initial hypothesis grammar.
In addition, we still achieve the goal of parsing each sentence
but once.

\ssec{Parameter Training}

In this section, we describe how the parameters of our grammar,
the probabilities associated with each grammar rule, are set.
Ideally, in evaluating the objective function for a particular grammar
we should use its optimal parameter settings given the training data,
as this is the full score that the given grammar can achieve.
However, searching for optimal parameter values is extremely expensive
computationally.  Instead, we grossly approximate the optimal values by
deterministically setting parameters based on
the Viterbi parse of the training data parsed so far.
We rely on the post-pass, described later, to refine parameter values.

Referring to the rules in Table \ref{tab:init}, the parameter $\e$ is set to
an arbitrary small constant.  Roughly speaking,
the values of the parameters $p(A)$ are
set to the frequency of the X $\r A$ reduction in the
Viterbi parse of the data seen so far and the remaining symbols
are set to expand uniformly among their possible expansions.
This issue is discussed in more detail in Section \ref{ssec:rules}.

\ssec{Constraining Moves} \label{ssec:constrain}

Consider the move of creating a rule of the form $A \r BC$.
This corresponds to $k^3$ different specific rules that might be created,
where $k$ is the current number of nonterminal
symbols in the grammar.  As it is too
computationally expensive to consider each of these rules at every point in
the search, we use heuristics to constrain which moves are appraised.

For the left-hand side of a rule, we always create a new symbol.
This heuristic selects the optimal choice the vast majority of
the time; however, under this constraint the
moves described earlier in this section cannot yield arbitrary context-free
languages.  A symbol can only be defined in terms of symbols created
earlier in the search process, so recursion cannot be
introduced into the grammar.  To partially address this, we add the move
\begin{description}
\item[Move 3:] Create a rule of the form $A \r AB | B$
\end{description}
This creates a symbol that expands to an arbitrary number of $B$'s.
With this iteration move, we can construct grammars that generate
arbitrary regular languages.  In Section \ref{ssec:extend}, we
discuss moves that extend our coverage to arbitrary context-free
languages.

To constrain the symbols we consider on the right-hand side of a new rule,
we use what we call {\it triggers}.\footnote{This is not to be confused with
the use of the term {\it triggers\/} in dynamic language modeling.}
A {\it trigger\/} is a configuration in the Viterbi parse of a sentence that is
indicative that a particular move might lead to a better grammar.
For example, in Figure \ref{fig:before} the fact that
the symbols $A\su{\it talks}$ and
$A\su{\it slowly}$ occur
adjacently is indicative that it could be profitable to create
a rule $B \r A\su{\it talks} A\su{\it slowly}$.
We have developed a set of triggers for each move in our move set,
and only consider a specific move if it is triggered somewhere
in the current Viterbi parse.

\ssec{Post-Pass}

A conspicuous shortcoming in our search framework is
that the grammars in our search space are fairly unexpressive.
Firstly, recall that our grammars model a sentence as a sequence of
independently generated symbols; however, in language there is a large
dependence between adjacent constituents.  Furthermore, the only free
parameters in our search are the parameters $p(A)$; all symbols besides
S and X are fixed to expand uniformly.  These choices were necessary to
make the search tractable.

To address these issues, we use an Inside-Outside algorithm post-pass.
Our methodology is derived from that
described by \newcite{Lari:90a}.  We create $n$ new nonterminal symbols
$\{X_1, \ldots, X_n\}$, and create all rules of the form:
$$ \begin{array}{cccll}
X_i & \r & X_j & X_k \hspace{0.3in} & i, j, k \in \{ 1, \ldots, n \} \\
X_i & \r & A & & i \in \{1, \ldots, n\}, A \in N\su{\it old} - \{{\rm S, X} \}
\end{array} $$
$N\su{\it old}$ denotes the set of nonterminal symbols
acquired in the initial grammar induction phase, and $X_1$ is taken to be the
new sentential symbol.  These new rules replace the first three rules
listed in Table \ref{tab:init}.
The parameters of these rules are initialized
randomly.  Using this grammar as the starting point, we
run the Inside-Outside algorithm on the training data until convergence.

In other words, instead of using the naive S $\r$ SX $|$ X rule to attach
symbols together in parsing data, we now use
the $X_i$ rules and depend on the Inside-Outside algorithm to
train these randomly initialized rules intelligently.
This post-pass allows us to express dependencies between adjacent
symbols.  In addition, it allows us to train parameters that were
fixed during the initial grammar induction phase.

\ssec{Algorithm Summary} \label{ssec:galgsum}

\begin{figure}
{ \def\r#1{{\bf #1}}
\begin{tabbing}
mm \= mm \= mm \= \hspace{2in} \= \kill
; {\it $G$ holds current hypothesis grammar} \\
; {\it $\P$ holds best parse for each sentence seen so far} \\
$G := \t{initial hypothesis grammar}$ \\
$\P := \e$ \\
\mbox{} \\
; {\it training data is composed of sentences $(o_1, \ldots, o_n)$} \\
\r{for} $i:=1$ \r{to} $n$ \r{do} \\
\> \r{begin} \\
\> ; {\it calculate best parse $\P_i$ of current sentence and append} \\
\> ; {\it to $\P$, the list of best parses} \\
\> $\P_i$ := best parse of sentence $o_i$ under grammar $G$ \\
\> $\P$ := {\it append}($\P$, $\P_i$) \\
\mbox{} \\
\> ; {\it $T$ holds the list of triggers yet to be checked} \\
\> $T$ := set of triggers in $\P_i$ \\
\> \r{while} $T \neq \e$ \r{do} \\
\> \> \r{begin} \\
\> \> ; {\it pick the first trigger $t$ in $T$ and remove from $T$} \\
\> \> $t$ := {\it first}($T$) \\
\> \> $T$ := {\it remove}($T$, $t$) \\
\mbox{} \\
\> \> ; {\it check if associated move is profitable, if so, apply} \\
\> \> $m$ := move associated with trigger $t$ \\
\> \> $G_m$ := grammar yielded if move $m$ is applied to the grammar $G$, \\
\> \> \> including parameter re-estimation \\
\> \> $\P_m$ := best parse yielded if move $m$ is applied to $G$ \\
\> \> $\Delta$ := change in objective function if $G$ becomes $G_m$
	and $\P$ becomes $\P_m$ \\
\> \> \r{if} $\Delta > 0$ \r{then} \\
\> \> \> \r{begin} \\
\> \> \> $G := G_m$ \\
\> \> \> $\P := \P_m$ \\
\> \> \> $T$ := {\it append}($T$, new triggers in $\P$) \\
\> \> \> \r{end} \\
\> \> \r{end} \\
\> \r{end}
\end{tabbing} }
\caption{Outline of search algorithm} \label{fig:galgsum}
\end{figure}

We summarize the algorithm, excluding the post-pass, in
Figure \ref{fig:galgsum}.  In Section \ref{sec:gprevwork},
we relate our algorithm to previous work on grammar induction.
In Section \ref{sec:gdetails}, we flesh out
the details of the algorithm, including the move set, the
encoding of the grammar used to calculate the objective function,
and the parsing algorithm used.  In addition, we describe
extensions to the basic algorithm that we have implemented.

%
\sec{Previous Work} \label{sec:gprevwork}
%

\ssec{Bayesian Grammar Induction}

Work by Solomonoff \shortcite{Solomonoff:60a,Solomonoff:64a} is
the first to lay out the general Bayesian grammar induction
framework that we use.  Solomonoff points out the relation
between encodings and prior probabilities, and using this relation
describes objective functions for several induction problems
including probabilistic context-free grammar induction.  Solomonoff evaluates
the objective function manually for a few example grammars
to demonstrate the viability of the given objective function,
but does not specify an algorithm for automatically searching
for good grammars.  Solomonoff's work can be seen as a precursor
to the minimum description length principle, and would in fact lead to
the closely related {\it universal a priori
probability} \cite{Solomonoff:60a,Solomonoff:64a,Li:93a}.

\newcite{Cook:76a} present a probabilistic context-free
grammar induction algorithm that employs a similar
framework.  While not formally Bayesian, their objective
function strongly resembles a Bayesian objective function.
In particular, their objective function is a weighted sum of
the {\it complexity\/} of a grammar and the {\it discrepancy\/}
between the grammar and the training data.  The first term is analogous to
a prior on grammars, and the second is analogous to the probability
of the data given the grammar.  However, the actual measures
used for complexity and discrepancy are rather dissimilar from
those used in this work.

Their initial hypothesis grammar consists of the sentential symbol expanding
to every sentence in the training data and only those strings.
This contrasts to the simple overgenerating grammar we use
for our initial grammar.  For Cook \etal, initially the
length of the grammar is large and the length of the data given the grammar
is small, while the converse is true for our approach.  We hypothesize that
neither approach is inherently superior; in the
former approach one just chooses moves that tend to compact the
grammar, while in the latter approach one chooses moves that tend to
compact the data.

The move set used by Cook \etal\ includes: {\it substitution}, which is
analogous to our concatenation move except that instead of a new symbol
being created on the left-hand side existing symbols can be used as well;
{\it disjunction}, which is analogous to our classing rule; a move
for removing inaccessible productions; and a move for merging two symbols into
one.  They describe a greedy heuristic search strategy, and present results on
small data sets, the largest being tens of sentences.
Their work is geared toward finding elegant grammars as opposed to finding
good language models, and thus they do not present any language
modeling results.

\newcite{Stolcke:94b} also present a similar algorithm.
Again, there are several significant differences from our approach.
They adhere to the Bayesian framework as we do; however,
their prior on grammars is divided into two different terms: a prior on
grammar rules, and a prior on the probabilities associated
with grammar rules.  For the former, they use an MDL-like
prior $p(G) = c^{-l(G)}$ where $c$ is varied during the
search process.  For the latter, they use a Dirichlet prior.
In our work, both grammar rules and rule probabilities
are expressed within the MDL framework.
In addition, like Cook \etal, Stolcke and
Omohundro choose an initial grammar where
the sentential symbol expands
to every sentence in the training data and only those strings.

The most important differences between this work and ours concern the
move set and search strategy.
Stolcke and Omohundro describe only two moves: a move for
merging two nonterminal
symbols into one, and a move named {\it chunking\/} that is analogous
to our concatenation move.  As their search strategy, they use a {\it beam
search\/}, which requires maintaining multiple hypothesis grammars.
They describe how this is necessary because with their move set,
often several moves must be made in conjunction to improve the
objective function.  We have addressed this problem in our work
by using a rich move set (see Section \ref{ssec:moveset});
we have complex moves that
hopefully correspond to those move tuples of Stolcke and Omohundro
that often lead to improvements in the objective function.
In addition, at each point in the
search they consider every possible move, as opposed to using
the triggering heuristics we use to constrain the moves considered.
Because of these differences, we assume our algorithm is significantly
more efficient than theirs.  They do not present any results
on data sets approaching the sizes that we used, and like Cook \etal,
they do not present any language modeling results.

\ssec{Other Approaches}

The most widely-used tool in probabilistic grammar induction
is the Inside-Outside algorithm \cite{Baker:79b},
a special case of the Expectation-Maximization algorithm \cite{Dempster:77a}.
The Inside-Outside algorithm takes a probabilistic context-free
grammar and adjusts its probabilities iteratively to attempt to maximize
the probability the grammar assigns to some training data.
It is a hill-climbing search; it generally improves
the probability of the training data in each iteration and is guaranteed
not to lower the probability.

Lari and Young
\shortcite{Lari:90a,Lari:91a} have devised a grammar induction algorithm
centered on the Inside-Outside algorithm.  In this approach,
the initial grammar consists of a very
large set of rules over some fixed number of nonterminal
symbols.  Probabilities are initialized randomly, and the Inside-Outside
algorithm is used to prune away extraneous rules
by setting their probabilities to near zero; the intention is that
this process reveals the correct grammar.
In Section \ref{sec:gresults}, we give a more detailed description.
Lari and Young present results on various training corpora,
with some success.  In our experiments, we replicate the Lari
and Young algorithm for comparison purposes.

\newcite{Pereira:92a} extend the Lari and Young work by training
on corpora that have been manually parsed.  They use the
manual annotation to constrain the Inside-Outside training.
However, their goal was parsing as opposed to language modeling,
so no language modeling results are reported.

\newcite{Carroll:95a} describes a heuristic algorithm for
grammar induction that employs the Inside-Outside algorithm extensively.
Carroll restricts the grammars he considers to a type of
probabilistic {\it dependency grammars}, which are a subset of
probabilistic context-free grammars.  In particular, he only considers
grammars where
there is one nonterminal symbol $\bar{A}$ associated with each
terminal symbol $A$ and no other nonterminal symbols.  Furthermore, all rules
expanding a nonterminal symbol $\bar{A}$ must have the corresponding
terminal symbol $A$ somewhere on the right-hand side.

Carroll begins with a seed grammar that is manually constructed.
The training corpus is parsed a sentence at a time,
and he has heuristics for adding new grammar rules if a
sentence is unparsable with the current grammar.  The Inside-Outside
algorithm is used during this process as well as afterwards to
refine rule probabilities.  In addition, there are manually-constructed
constraints on the new rules that can be created.

Carroll reports results for building language models for part-of-speech
sequences corresponding to sentences.  Training on 300,000
words/part-of-speech tags
from the Brown Corpus, he reports slightly better perplexities on test
data than trigram part-of-speech tag models on the $\sim99\%$ of the
sentences the grammar can parse.  In addition, by linearly
interpolating the grammatical model and the trigram model,
he achieves a better perplexity on the entire test
set than the trigram model alone.

\newcite{McCandless:93a} present a heuristic
grammar induction algorithm that does not use the Inside-Outside
algorithm.  They begin with a grammar consisting of the sentential
symbol expanding to every sentence, and they have a single move
for improving the grammar, a move that combines classing and
concatenation.  However, they do not take a Bayesian approach
in determining which moves to take.  Instead, classing is
based on how similar the bigram distributions of two symbols are,
and concatenation is based on how frequently symbols occur
adjacently.  For evaluation, they
build an $n$-gram {\it symbol\/} model using the symbols induced.  That is,
instead of predicting the next word based on the last $n-1$ {\it words},
they predict the next word based on the last $n-1$ {\it symbols}.  With
these $n$-gram symbol models, they achieve slightly better
perplexity on test data than the corresponding $n$-gram word models.

%
\sec{Algorithm Details} \label{sec:gdetails}
%

\ssec{Grammar Specification} \label{ssec:rules}

In this section, we describe in detail the forms of the grammar rules we
consider and we discuss how rules are assigned probabilities.

Recall the structure of the grammar we use as described in
Section \ref{sec:goutline}.  We have rules expressing that
a sentence S is a sequence of X's
$$ \begin{array}{ccllcl}
{\rm S} & \r & {\rm SX} \hspace{0.3in} & (1 - \e) \\
{\rm S} & \r & {\rm X}  & (\e)
\end{array} $$
where the quantity in parentheses is the probability associated
with the given rule.
We take $\e$ to be an arbitrarily small constant so that it
can be safely ignored in the objective function calculation.\footnote{
In a parse tree, the latter rule can be applied at most once
while the former rule can be applied many times.  Thus, the probability
contributed to a parse by these rules is of the form $(1-\e)^k \e$.
For small $\e$, this expression is very nearly equal to just $\e$,
a constant.  As we are only concerned with {\it changes\/} in the
objective function, constant expressions can be ignored.}

Then, we have a rule of the form
$$ \begin{array}{ccllcl}
{\rm X} & \r & A \hspace{0.3in} & (p(A))
\end{array} $$
for each nonterminal symbol $A \neq$ S, X.  To calculate $p(A)$,
we use the frequency with which the associated rule has been
used in past parses.
We keep track of $c({\rm X} \r A)$, the number of times the rule X $\r A$
is used in the current best parse $\P$ (see Figure \ref{fig:galgsum}), and
we just normalize this value to yield $p(A)$ as follows:
$$ p(A) = \frac{c({\rm X} \r A)}{\sum_A c({\rm X} \r A)} $$

Finally, we have rules that define the expansions of
the nonterminal symbols besides S and X.
We restrict such nonterminal symbols to expand in exactly one
of four ways:
\bi
\i expansion to a terminal symbol: $A \r a$
\i concatenation of two nonterminal symbols: $A \r B C$
\i classing of two nonterminal symbols: $A \r B | C$
\i repetition of a nonterminal symbol: $A \r A B | B$
\ei
For instance, we do not allow a symbol $A \r a | B C$ that expands to both
a terminal symbol and a concatenation of two nonterminal symbols.

While composing rules of the first three forms is sufficient to
describe any context-free language, the move set we use
cannot introduce recursion into the grammar.
We add the fourth form to model a simple but common instance
of recursion.  Even with this extra form, we can still only model
regular languages; in Section \ref{ssec:extend} we describe
extensions that release this restriction.

For rules of the first two forms, the probability associated
with the rule is $1 - p_s$, where $p_s$ is the probability associated
with {\it smoothing\/} rules, which will be discussed in the
next section.

\begin{figure}
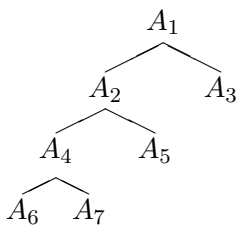


\leaf{$A_6$}
\leaf{$A_7$}
\branch{2}{$A_4$}
\leaf{$A_5$}
\branch{2}{$A_2$}
\leaf{$A_3$}
\branch{2}{$A_1$}

$$ \tree $$
\caption{Example class hierarchy} \label{fig:classtree}
\end{figure}

For classing rules, we choose probabilities to form
a uniform distribution over all symbols the
class can expand to, as defined as follows.
First, notice that while a given classing rule only classes
two symbols, by composing several of these rules you can effectively
class an arbitrary number of symbols.  For example, consider the rules
$$ \begin{array}{ccl}
A_1 & \r & A_2 | A_3 \\
A_2 & \r & A_4 | A_5 \\
A_4 & \r & A_6 | A_7
\end{array} $$
which we can express using a tree as in Figure \ref{fig:classtree}.
We can see that $A_1$ classes together the symbols
$A_3$, $A_5$, $A_6$, and $A_7$ at its leaves.
Then, instead of assigning 0.5 probability to $A_1$ expanding to each of $A_2$
and $A_3$, we assign 0.25 probability to $A_1$ expanding
to each of $A_3$, $A_5$, $A_6$,
and $A_7$.  That is, we assign a uniform distribution on all symbols
the class recursively expands to at its leaves,
not a uniform distribution on the
symbols that the class immediately expands to.  (In this example,
we assume that $A_3$, $A_5$, $A_6$, and $A_7$ are not classing
symbols themselves.)  Then, to satisfy these leaf expansion
probabilities, we have that $A_2$ expands to
$A_4$ with probability 2/3 and $A_5$ with probability 1/3, and
$A_1$ expands to $A_2$ with probability 3/4 and $A_3$ with probability 1/4.
Notice that this uniform leaf probability
constraint assigns consistent probabilities
among different symbols.  The probabilities in the class hierarchy
we set to satisfy
uniform leaf probabilities for $A_1$ are the same we use to satisfy
the uniform leaf probabilities for $A_2$.  In actuality, the preceding
discussion is not quite accurate as we multiply
each of the probabilities by $1-p_s$ as in the non-classing rules,
to allow for smoothing.

For the repetition rule, while the rule $A \r A B | B$ is accurate
in terms of the strings expanded to, it is inaccurate in terms
of the way we assign probabilities to expansions.  In particular,
the probability we assign to the symbol $A$ expanding to exactly $n$ $B$'s is
$$ p(A \Rightarrow^* B^n) = p\su{MDL}(n) = \frac{6}{\pi^2}
	\frac{1}{n [\log_2 (n+1)]^2} $$
where $p\su{MDL}$ is the universal MDL prior over the natural
numbers \cite{Rissanen:89a}.  We choose this parameterization because
it prevents us from needing to estimate the probability of
the $A \r AB$ expansion versus the $A \r B$ expansion, and because
in some sense it is the most conservative distribution one can
take, in that it asymptotically assigns as high probabilities to large $n$ as
possible.  Again, in actuality the above probabilities should be
multiplied by $1-p_s$ for smoothing.

Notice that we have minimized the number of parameters we need to
estimate; this is to simplify the search task.  So far, the
only parameters we have described are the $p(A)$ associated with
rules of the form X $\r A$, which are estimated deterministically
from the best parse $\P$, and $p_s$, the probability assigned
to smoothing rules.  In Section \ref{ssec:extend}, we describe
extensions where we consider richer parameterizations.

\ssec{Move Set and Triggers} \label{ssec:moveset}

In this section, we list the moves that we use to adjust
the current hypothesis grammar, and the patterns in the
current best parse that trigger the consideration of a particular
instance of a move form.  These moves all take on the form of
adding a rule to the grammar; in Section \ref{ssec:extend} we
consider rules of different forms.  For the rule added to the grammar,
we generate a new, unique symbol $A$ for the left-hand side of the rule
and also create a concomitant rule of the form X $\r A$ as mentioned
in Section \ref{sec:goutline}.  Recall that whether a move
is taken depends on whether it improves the objective function, and
that in order to estimate this effect we approximate how the best
parse of previous data changes.  In this section, we also describe
for each move how we approximate its effect on the best parse if the
move is applied.

For future reference, we use the notation $A_{\a}$
to refer to a nonterminal symbol that expands exclusively
to the string $\a$.  For example, the symbol $A\su{\it Bob}$
expands to the word {\it Bob\/} and no other strings.


\begin{figure}

$$
\leaf{\it Bob}
\branch{1}{$A\su{\it Bob}$}
\branch{1}{X}
\branch{1}{S}
\leaf{\it talks}
\branch{1}{$A\su{\it talks}$}
\branch{1}{X}
\branch{2}{S}
\leaf{\it slowly}
\branch{1}{$A\su{\it slowly}$}
\branch{1}{X}
\branch{2}{S}
\vtop{\hbox{}\hbox{\tree}} \hspace{0.5in}
\leaf{\it Mary}
\branch{1}{$A\su{\it Mary}$}
\branch{1}{X}
\branch{1}{S}
\leaf{\it talks}
\branch{1}{$A\su{\it talks}$}
\branch{1}{X}
\branch{2}{S}
\leaf{\it slowly}
\branch{1}{$A\su{\it slowly}$}
\branch{1}{X}
\branch{2}{S}
\vtop{\hbox{}\hbox{\tree}} $$

\caption{Triggering concatenation} \label{fig:concata}

$$
\leaf{\it Bob}
\branch{1}{$A\su{\it Bob}$}
\branch{1}{X}
\branch{1}{S}
\leaf{\it talks}
\branch{1}{$A\su{\it talks}$}
\leaf{\it slowly}
\branch{1}{$A\su{\it slowly}$}
\branch{2}{$A_1$}
\branch{1}{X}
\branch{2}{S}
\vtop{\hbox{}\hbox{\tree}} \hspace{0.5in}
\leaf{\it Mary}
\branch{1}{$A\su{\it Mary}$}
\branch{1}{X}
\branch{1}{S}
\leaf{\it talks}
\branch{1}{$A\su{\it talks}$}
\leaf{\it slowly}
\branch{1}{$A\su{\it slowly}$}
\branch{2}{$A_1$}
\branch{1}{X}
\branch{2}{S}
\vtop{\hbox{}\hbox{\tree}} $$

\caption{After concatenation/triggering classing} \label{fig:concatb}

$$
\leaf{\it Bob}
\branch{1}{$A\su{\it Bob}$}
\branch{1}{$A_2$}
\branch{1}{X}
\branch{1}{S}
\leaf{\it talks}
\branch{1}{$A\su{\it talks}$}
\leaf{\it slowly}
\branch{1}{$A\su{\it slowly}$}
\branch{2}{$A_1$}
\branch{1}{X}
\branch{2}{S}
\vtop{\hbox{}\hbox{\tree}} \hspace{0.5in}
\leaf{\it Mary}
\branch{1}{$A\su{\it Mary}$}
\branch{1}{$A_2$}
\branch{1}{X}
\branch{1}{S}
\leaf{\it talks}
\branch{1}{$A\su{\it talks}$}
\leaf{\it slowly}
\branch{1}{$A\su{\it slowly}$}
\branch{2}{$A_1$}
\branch{1}{X}
\branch{2}{S}
\vtop{\hbox{}\hbox{\tree}} $$

\caption{After classing} \label{fig:classa}

\end{figure}

\sssec{Concatenation Rules}

To trigger the creation of rules of the
form $A \r B C$, we look for two adjacent
instances of the symbol X expanding to the symbols $B$ and $C$
respectively.  (Recall that the sentential symbol S expands
to a sequence of X's.)  For example, in Figure \ref{fig:concata},
the following rules are triggered: $A \r A\su{\it Bob} A\su{\it talks}$,
$A \r A\su{\it talks} A\su{\it slowly}$, and
$A \r A\su{\it Mary} A\su{\it talks}$.
To approximate how the best parse changes with the creation of a rule
$A \r B C$, we simply replace all adjacent pairs of X's expanding
to $B$ and $C$ with a single X expanding to $A$.
In this example, if the
rule $A_1 \r A\su{\it talks} A\su{\it slowly}$ is actually created, then we
would estimate the best parse to be as in Figure \ref{fig:concatb}.

\sssec{Classing Rules} \label{sssec:clrules}

To trigger the creation of
rules of the form $A \r B | C$, we look for cases where
by forming the classing rule, we can more compactly express the
grammar.  For example, consider Figure \ref{fig:concatb}.  These parses
trigger concatenation rules of the form $A \r A\su{\it Bob} A_1$ and
$A \r A\su{\it Mary} A_1$.  Now, if we create a rule of the
form $A_2 \r A\su{\it Bob} | A\su{\it Mary}$, instead of creating two different
concatenation rules to model these two sentences, we can create a single
rule of the form $A \r A_2 A_1$.  Thus, adding a classing rule can
enable us to create fewer rules to model a given piece of data.
In this instance, the cost of creating the classing rule may
offset the gain in creating one fewer concatenation rule; however,
a classing rule can be profitable if it saves rules in more
than one place in the grammar.

In particular, whenever there are two triggered rules
that differ by only one symbol, we consider classing the symbols
that they differ by.  Triggered rules that do not improve the
objective function by themselves may become
profitable after classing, as a class can reduce the number of
new rules that need to be created.  Recall that the prior on grammars
assigns lower probabilities to larger grammars, so the fewer
rules created, the better.

In the above example, we consider building the class
$A_2 \r A\su{\it Bob} | A\su{\it Mary}$ because we have two triggered
concatenation rules that only differ in that $A\su{\it Bob}$
is replaced with $A\su{\it Mary}$ in the latter rule.  To approximate
how the best parse changes if a classing rule is created, we
apply the classing rule wherever the associated triggering
rules would be applied.  In this example, if we create the classing rule
the current parse would be updated to be as in Figure \ref{fig:classa}.

However, notice that if there exists a rule
$A_3 \r A\su{\it Bob} | A\su{\it John}$,
then by classing together $A_3$ and $A\su{\it Mary}$ we can get the
same affect as classing together $A\su{\it Bob}$ and $A\su{\it Mary}$:
we will only need to create a single concatenation rule to describe the two
sentences in the previous example.  Similarly, if $A_3$ belongs to a
class $A_4$, then we can get the same effect by classing together $A_4$ and
$A\su{\it Mary}$.  Thus, when two triggered rules
differ by a symbol, instead of considering classing just those
two symbols, we should consider classing each class that each of
those two symbols recursively belong two.  However, this is too expensive
computationally.  For example, if there are ten triggered rules
of the form $A \r A_{\a} A_1$ for ten different symbols $A_{\a}$ and
each $A_{\a}$ belongs to ten classes on average, then there
are roughly ${{10 \cdot 10}\choose 2} \approx 5000$ pairs of symbols
that we could consider classing.

To address this issue, we use heuristics to reduce the number of
classings we consider.  In particular, for a set of triggered rules
that only differ in a single position where the symbols occurring
in that position are $\{ A_{\a} \}$, we try to find a minimal
set of classes that all of the $A_{\a}$ recursively belong to,
and only try exhaustive pairwise classing among that minimal set.
We use a greedy algorithm to search for this set; we explain this
algorithm using an example.
Consider the case where we have triggered rules of
the form $A \r A_{\a} A_1$ for $A_{\a} = \{ A\su{\it Bob}, A\su{\it John},
	A\su{\it Mary}, A\su{\it the macaw}, A\su{\it a parrot},
	A\su{\it a frog} \}$.
Initially, we take the minimal covering set to
be just the set of all of the symbols:
$$ \{ A\su{\it Bob}, A\su{\it John}, A\su{\it Mary},
A\su{\it the macaw}, A\su{\it a parrot}, A\su{\it a frog} \} $$
Then, we try to find classes that multiple elements belong to.
Say we notice that there is an existing symbol $A\su{\it Bob$|$John}$
that expands to $A\su{\it Bob} | A\su{\it John}$.  We group
these two symbols by replacing the two symbols with the new symbol:
$$ \{ A\su{\it Bob$|$John}, A\su{\it Mary},
A\su{\it the macaw}, A\su{\it a parrot}, A\su{\it a frog} \} $$
Using this new set, we again try to find symbols that multiple elements
belong to.  Let's say we find an existing
symbol $A\su{\it the macaw$|$a parrot}$, giving us
$$ \{ A\su{\it Bob$|$John}, A\su{\it Mary},
	A\su{\it the macaw$|$a parrot}, A\su{\it a frog} \} $$
and a symbol $A\su{\it the macaw$|$a parrot$|$a frog}$ giving us
$$ \{ A\su{\it Bob$|$John}, A\su{\it Mary},
	A\su{\it the macaw$|$a parrot$|$a frog} \} $$
Then, if we can find no more classes that multiple elements
belong to, we take this to be the minimal covering set and
consider all possible pairwise classings of these three elements.

Notice that this algorithm attempts to find the natural groupings
of the elements in the list as expressed through existing classes,
and only tries to class together these higher-level classes.
This is a reasonable heuristic in selecting new classings to consider.

To constrain what groupings are performed, we only consider those
groupings that are profitable in terms of the objective function.
For instance, in the above example we only group together
the symbols $A\su{\it Bob}$ and $A\su{\it John}$ into the
symbol $A\su{\it Bob $|$ John}$ if creating the rule
$A \r A\su{\it Bob $|$ John} A_1$ is more profitable (or less unprofitable)
in terms of the objective function than creating the rules
$A \r A\su{\it Bob} A_1$ and $A' \r A\su{\it John} A_1$.

\sssec{Repetition Rules}

\begin{figure}
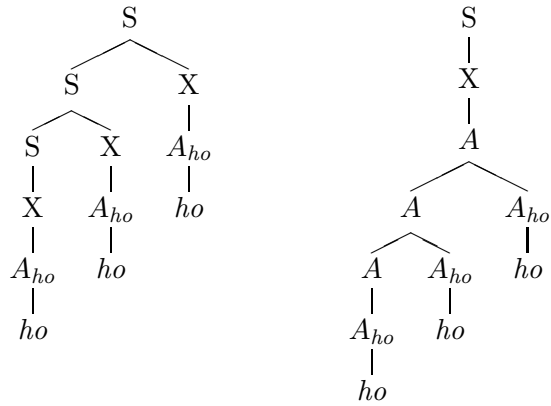


$$
\leaf{\it ho}
\branch{1}{$A\su{\it ho}$}
\branch{1}{X}
\branch{1}{S}
\leaf{\it ho}
\branch{1}{$A\su{\it ho}$}
\branch{1}{X}
\branch{2}{S}
\leaf{\it ho}
\branch{1}{$A\su{\it ho}$}
\branch{1}{X}
\branch{2}{S}
\vtop{\hbox{}\hbox{\tree}} \hspace{0.5in}
\leaf{\it ho}
\branch{1}{$A\su{\it ho}$}
\branch{1}{$A$}
\leaf{\it ho}
\branch{1}{$A\su{\it ho}$}
\branch{2}{$A$}
\leaf{\it ho}
\branch{1}{$A\su{\it ho}$}
\branch{2}{$A$}
\branch{1}{X}
\branch{1}{S}
\vtop{\hbox{}\hbox{\tree}} $$

\caption{Triggering and applying the repetition move} \label{fig:repeata}

\end{figure}

To make rules of the form $A \r AB | B$, we look for multiple
(\ie, at least three)
adjacent instances of the symbol X expanding to the symbol $B$.  For example,
the parse on the left of Figure \ref{fig:repeata} triggers a rule of
the form $A \r AA\su{\it ho} | A\su{\it ho}$.
To approximate how the best parse changes with the creation of a rule
$A \r AB | B$, we replace all chains of at least three consecutive X's
expanding to $B$'s with a single X expanding to $A$.
In this example, if the repetition rule is
actually created, then we would estimate the best parse to be the parse
on the right in Figure \ref{fig:repeata}.

\sssec{Smoothing Rules} \label{sssec:srules}

\begin{figure}

$$
\leaf{\it Bob}
\branch{1}{$A\su{\it Bob}$}
\leaf{\it talks}
\branch{1}{$A\su{\it talks}$}
\leaf{\it slowly}
\branch{1}{$A\su{\it slowly}$}
\branch{2}{$A_1$}
\branch{2}{$A_2$}
\branch{1}{X}
\branch{1}{S}
\vtop{\hbox{}\hbox{\tree}} \hspace{0.5in}
\leaf{\it Bob}
\branch{1}{$A\su{\it Bob}$}
\branch{1}{X}
\branch{1}{S}
\leaf{\it talks}
\branch{1}{$A\su{\it talks}$}
\branch{1}{X}
\branch{2}{S}
\leaf{\it quickly}
\branch{1}{$A\su{\it quickly}$}
\branch{1}{X}
\branch{2}{S}
\vtop{\hbox{}\hbox{\tree}} $$

\caption{Without smoothing rules} \label{fig:smootha}

$$
\leaf{\it Bob}
\branch{1}{$A\su{\it Bob}$}
\leaf{\it talks}
\branch{1}{$A\su{\it talks}$}
\leaf{\it slowly}
\branch{1}{$A\su{\it slowly}$}
\branch{2}{$A_1$}
\branch{2}{$A_2$}
\branch{1}{X}
\branch{1}{S}
\vtop{\hbox{}\hbox{\tree}} \hspace{0.5in}
\leaf{\it Bob}
\branch{1}{$A\su{\it Bob}$}
\leaf{\it talks}
\branch{1}{$A\su{\it talks}$}
\leaf{\it quickly}
\branch{1}{$A\su{\it quickly}$}
\branch{1}{$A\su{\it slowly}$}
\branch{2}{$A_1$}
\branch{2}{$A_2$}
\branch{1}{X}
\branch{1}{S}
\vtop{\hbox{}\hbox{\tree}} $$

\caption{With smoothing rules} \label{fig:smoothb}

\end{figure}

In this section, we describe the smoothing rules alluded to
earlier.  Just as smoothing improves the accuracy of $n$-gram
models, smoothing can improve grammatical models by
assigning nonzero probabilities to phenomena with no counts.
More importantly, they cause
text to be parsed in a way so as to provide informative triggers
for producing new rules.  For example, consider the case
where the only concatenation rules in the grammar are
$$ \begin{array}{ccl}
A_1 & \r & A\su{\it talks} A\su{\it slowly} \\
A_2 & \r & A\su{\it Bob} A_1
\end{array} $$
Then, if we see the sentences
{\it Bob talks slowly\/} and {\it Bob talks quickly}, they
will be parsed as in Figure \ref{fig:smootha}.  While the concatenation
rules can be used to parse the first sentence, they do
not apply to the second sentence.  However, on the surface
these sentences are very similar and it is desirable to be able to
capture this similarity.  Consider adding
the rule $A\su{\it slowly} \r A\su{\it quickly}$ to the grammar.
Then, we can parse the two sentences as in Figure \ref{fig:smoothb},
capturing the similarity in structure of the two sentences.

Furthermore, we can use the parse on the right as a trigger.  For example,
we might consider creating the rules
$$ \begin{array}{ccl}
A_1' & \r & A\su{\it talks} A\su{\it quickly} \\
A_2' & \r & A\su{\it Bob} A_1'
\end{array} $$
reflecting the similarity in structure between the two constructions.
The rule $A\su{\it slowly} \r A\su{\it quickly}$ helps us
capture the parallel nature of similar constructions in both
the best parse and the grammar.

In the grammar, we have a rule of the form $A \r B$ for
all nonterminal symbols $A, B \neq$ S, X, and we call
these {\it smoothing\/} rules.  They are implicitly created whenever
a new nonterminal symbol is created.  We assign them very
low probabilities so that they are used infrequently.
They are only used in the most probable parse if without them
few grammar rules can be applied to the given text, but with them
many rules can be applied, as in the above example.  This prevents
smoothing rules from indicating a parallel nature between overly
dissimilar constructions.

\begin{figure}

$$
\leaf{\it Bob}
\branch{1}{$A\su{\it Bob}$}
\leaf{\it talks}
\branch{1}{$A\su{\it talks}$}
\leaf{\it slowly}
\branch{1}{$A\su{\it slowly}$}
\branch{2}{$A_1$}
\branch{2}{$A_2$}
\branch{1}{X}
\branch{1}{S}
\vtop{\hbox{}\hbox{\tree}} \hspace{0.5in}
\leaf{\it Bob}
\branch{1}{$A\su{\it Bob}$}
\leaf{\it talks}
\branch{1}{$A\su{\it talks}$}
\leaf{\it $\e$}
\branch{1}{$A\su{\it slowly}$}
\branch{2}{$A_1$}
\branch{2}{$A_2$}
\branch{1}{X}
\branch{1}{S}
\vtop{\hbox{}\hbox{\tree}} $$

\caption{$\e$-smoothing rules} \label{fig:smoothc}
\end{figure}

In addition, we also have smoothing rules of the form $A \r \e$ for
every nonterminal symbol $A \neq$ S, X.  These can capture the
situation where two constructions are identical except that a word
is deleted in one.  We display possible parses of the
sentences {\it Bob talks slowly\/} and {\it Bob talks\/} in
Figure \ref{fig:smoothc}.

We assign probability $\frac{p_s}{2} p_G(B)$ to smoothing rules $A \r B$,
and probability $\frac{p_s}{2}$ to smoothing rules $A \r \e$.
We take the distribution
$p_G(B)$ to be different from the probabilities $p(B)$ associated
with rules of the form X $\r B$.  The probability $p(B)$ reflects
how frequently a symbol occurs in text, and it is unclear this is
an accurate reflection of how frequently a symbol occurs in a smoothing rule.
Instead, we guess that a better reflection of this frequency is
the frequency with which a symbol occurs in the {\it grammar}; as smoothing
rule occurrences trigger rule creations, these two quantities should
correlate.  Thus, we take
$$ p_G(B) = \frac{c_G(B)}{\sum_B c_G(B)} $$
where $c_G(B)$ is the number of times the symbol $B$ occurs in grammar rules.
The parameter $p_s$ is set arbitrarily; in most experiments, we used
the value 0.01.

\begin{figure}
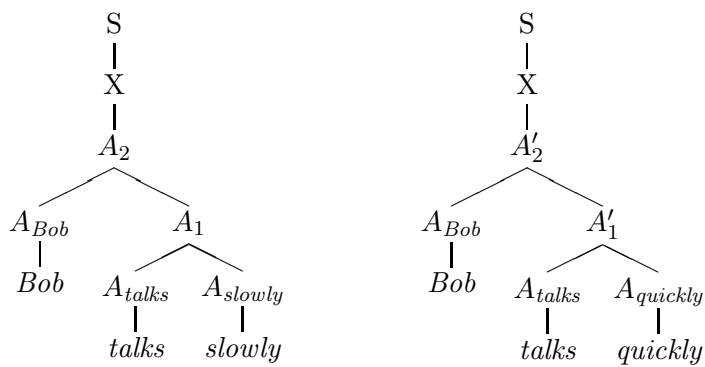


$$
\leaf{\it Bob}
\branch{1}{$A\su{\it Bob}$}
\leaf{\it talks}
\branch{1}{$A\su{\it talks}$}
\leaf{\it slowly}
\branch{1}{$A\su{\it slowly}$}
\branch{2}{$A_1$}
\branch{2}{$A_2$}
\branch{1}{X}
\branch{1}{S}
\vtop{\hbox{}\hbox{\tree}} \hspace{0.5in}
\leaf{\it Bob}
\branch{1}{$A\su{\it Bob}$}
\leaf{\it talks}
\branch{1}{$A\su{\it talks}$}
\leaf{\it quickly}
\branch{1}{$A\su{\it quickly}$}
\branch{2}{$A_1'$}
\branch{2}{$A_2'$}
\branch{1}{X}
\branch{1}{S}
\vtop{\hbox{}\hbox{\tree}} $$

\caption{After smoothing triggering} \label{fig:smoothd}
\end{figure}

When smoothing rules appear in the best parse $\P$, they trigger
rule creations in the way described in the examples given
earlier in this section.  In particular, we try to
build the smallest set of concatenation rules such that the given text can
be parsed without the smoothing rule.  Thus, for the
{\it Bob talks quickly/Bob talks slowly\/} example we try to
build the symbols $A_1'$ and $A_2'$ defined earlier.
To approximate how the best parse is affected, we just use
the heuristics given for concatenation rules; for this example,
this yields the parse in Figure \ref{fig:smoothd}.  In
Section \ref{ssec:extend}, we describe other types of moves
that we can trigger with smoothing rules.

Notice that in creating multiple rules, we pay quite a penalty
in the objective function from the term favoring smaller grammars.
To make this move more favorable, we have added an encoding
to our encoding scheme that describes these types of moves
compactly.  This is described in Section \ref{ssec:encode}.

\sssec{Specialization}

In the last section, we discussed a mechanism for handling
the case where a symbol is too specific.  For example, if
we have a symbol that expands only to a
string {\it Bob talks slowly} (ignoring smoothing rules),
by applying smoothing rules this symbol can expand to all strings
of the form {\it Bob talks $\a$}.  Furthermore, the application
of a smoothing rule triggers the creation of rules that expand to
these other strings.

\begin{figure}
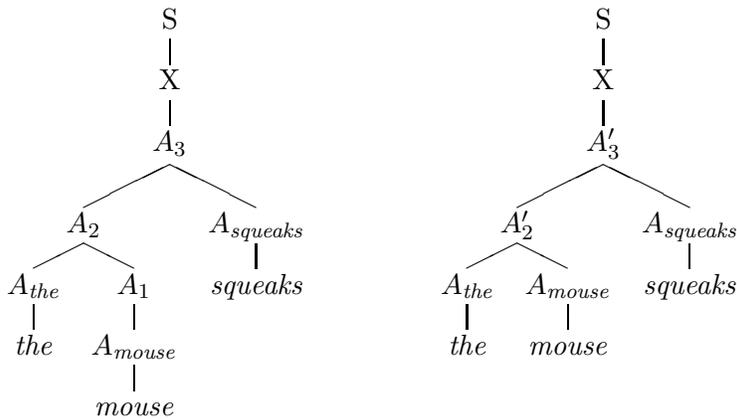


$$
\leaf{\it the}
\branch{1}{$A\su{\it the}$}
\leaf{\it mouse}
\branch{1}{$A\su{\it mouse}$}
\branch{1}{$A_1$}
\branch{2}{$A_2$}
\leaf{\it squeaks}
\branch{1}{$A\su{\it squeaks}$}
\branch{2}{$A_3$}
\branch{1}{X}
\branch{1}{S}
\vtop{\hbox{}\hbox{\tree}} \hspace{0.5in}
\leaf{\it the}
\branch{1}{$A\su{\it the}$}
\leaf{\it mouse}
\branch{1}{$A\su{\it mouse}$}
\branch{2}{$A_2'$}
\leaf{\it squeaks}
\branch{1}{$A\su{\it squeaks}$}
\branch{2}{$A_3'$}
\branch{1}{X}
\branch{1}{S}
\vtop{\hbox{}\hbox{\tree}} $$

\caption{Before and after specialization} \label{fig:speca}
\end{figure}

On the other hand, it may be possible that a symbol is too general.
For example, consider the rules
$$ \begin{array}{ccl}
A_1 & \r & A\su{\it mouse} | A\su{\it rat} \\
A_2 & \r & A\su{\it the} A_1 \\
A_3 & \r & A_2 A\su{\it squeaks}
\end{array} $$
The symbol $A_3$ expands to the strings {\it the mouse squeaks\/} and
{\it the rat squeaks}, but suppose only the former string appears in text.
We can create a {\it specialization\/} of the symbol $A_3$ by creating
the rules
$$ \begin{array}{ccl}
A_2' & \r & A\su{\it the} A\su{\it mouse} \\
A_3' & \r & A_2' A\su{\it squeaks}
\end{array} $$
The new symbol $A_3'$ expands only to the string {\it the mouse squeaks}.
Notice the similarity between this move and the move triggered
by the smoothing rule; the only difference is that instead of
being triggered whenever a smoothing rule occurs, this move is
triggered whenever a classing rule occurs.  Parses of {\it the mouse
squeaks\/} before and after the creation of these specialization rules
are displayed in Figure \ref{fig:speca}.  Like with the smoothing
rules, we try to create the minimal number of rules so that the
given text can be parsed without the class expansion.  Also like the
smoothing rules, there is a special encoding in
the encoding scheme to make these moves more favorable.

\begin{figure}
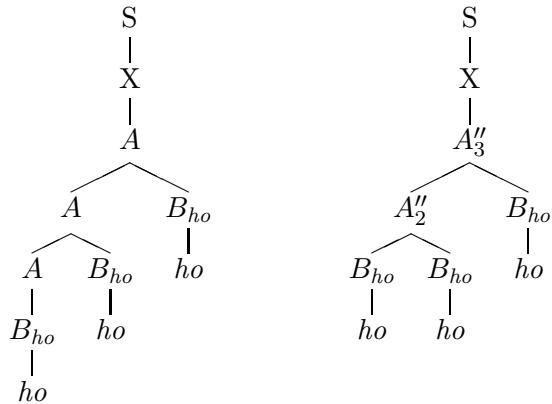


$$
\leaf{\it ho}
\branch{1}{$B\su{\it ho}$}
\branch{1}{$A$}
\leaf{\it ho}
\branch{1}{$B\su{\it ho}$}
\branch{2}{$A$}
\leaf{\it ho}
\branch{1}{$B\su{\it ho}$}
\branch{2}{$A$}
\branch{1}{X}
\branch{1}{S}
\vtop{\hbox{}\hbox{\tree}} \hspace{0.5in}
\leaf{\it ho}
\branch{1}{$B\su{\it ho}$}
\leaf{\it ho}
\branch{1}{$B\su{\it ho}$}
\branch{2}{$A_2''$}
\leaf{\it ho}
\branch{1}{$B\su{\it ho}$}
\branch{2}{$A_3''$}
\branch{1}{X}
\branch{1}{S}
\vtop{\hbox{}\hbox{\tree}} $$

\caption{Before and after repetition specialization} \label{fig:specb}
\end{figure}

It is also possible to specialize repetition rules.  For
example, consider the repetition rule $A \r AB\su{\it ho} | B\su{\it ho}$
expressing that the symbol $A$ can expand to any number of
$B\su{\it ho}$'s.  However, suppose that {\it ho\/} always occurs
exactly three times.  Then, it seems reasonable to create the rules
$$ \begin{array}{ccl}
A_2'' & \r & B\su{\it ho} B\su{\it ho} \\
A_3'' & \r & A_2'' B\su{\it ho}
\end{array} $$
so that we have a new symbol $A_3''$ that expands exactly to
the string {\it ho ho ho}.  Parses of {\it ho ho ho\/} before
and after the creation of these specialization rules are displayed
in Figure \ref{fig:specb}.  Such specializations are triggered
whenever a repetition symbol occurs in $\P$.  They involve the
creation of concatenation rules that generate
the repeated symbol the appropriate number of times, like above.

\sssec{Summary}

We have moves in our move set for building
concatenation, classing, and repetition grammar rules.
These operations are the building blocks for regular languages, and
as mentioned in Section \ref{ssec:constrain} our algorithm
can only create grammars that describe regular languages.
This is because whenever we create a new rule, we create a new
symbol to be placed on its left-hand side.  In addition, we have
no moves for modifying existing rules.  Thus, it is impossible
to introduce recursion into the grammar, except for the recursion present
in the repetition rule.

In addition, we have moves for generalizing and specializing
existing symbols.  Smoothing rules provide the trigger for
creating symbols that generalize the set of strings existing
symbols expand to.  Classing and repetition rules provide the trigger
for creating symbols that specialize the set of strings existing
symbols expand to.

While this forms a rich set of moves for constraining grammars,
there are some obvious shortcomings.  For example, there
are no moves for modifying existing rules or deleting rules,
or for changing the set of strings a symbol expands to.
Also, there are very few free parameters in the grammar; we may be
able to do better by allowing class expansions to have
probabilities that are trained.  In
Section \ref{ssec:extend}, we describe extensions such as these
that we have experimented with.

\ssec{Encoding} \label{ssec:encode}

In this section, we describe the encoding scheme that we use
to describe grammars.  This determines the length $l(G)$ of
a grammar $G$, which is used to calculate the prior $p(G) = 2^{-l(G)}$
on grammars, which is a term in our objective function.

While one can encode grammars using simple methods
such as textual description, we argue that it is important to use compact
encodings as touched on in Section \ref{ssec:desclen}.
First of all, we want the prior $p(G) = 2^{-l(G)}$ on grammars that
is associated with an encoding $l(G)$ to be an accurate prior; that is, $p(G)$
should model grammar frequencies accurately.
Just as good language models assign high probabilities to
training data, good priors should assign high probabilities
to typical or frequent grammars.  This corresponds to assigning
short lengths to typical grammars.

Furthermore, the compactness of the encoding dictates how
much data is needed as evidence to create a new grammar rule.  To clarify,
let us view the objective function from the MDL perspective, \ie,
as $l(G) + l(O|G)$, the length of the grammar added to the
length of the data given the grammar, recalling that $l(O|G)$ is
simply $\log_2 \frac{1}{p(O|G)}$.  Adding a grammar
rule increases the length $l(G)$ by some amount, say $\d$, so in
order for a new grammar rule to improve the objective function
its application must result in a decrease in the
length of the data of at least $\d$.
The more compactly we can encode grammar rules, the smaller
$\d$ will be, and the less a rule needs to
compress the training data in order to be profitable.
This corresponds to decreasing the amount
of evidence necessary to induce a grammar rule, \eg, decreasing
the number of times the symbols $B$ and $C$ need to occur
next to each other to make the creation of the rule $A \r BC$ profitable.

Before we describe the encoding proper, we first
describe how we encode positive integers with no upper limit,
such as the number of symbols in the grammar.
One option is to just set an arbitrary bound and to use
a fixed-length code, \eg, to code integers using 32 bits as
in a programming language.  However, it is inelegant to set
a bound, and this encoding is inefficient for
small integers.  Instead, we use the encoding associated with the
universal MDL prior
over the natural numbers $p\su{MDL}(n)$ \cite{Rissanen:89a}
mentioned in Section \ref{ssec:rules}, where
$$ p\su{MDL}(n) = \frac{6}{\pi^2} \frac{1}{n [\log_2 (n+1)]^2} $$
We take the length $l(n)$ of an integer $n$ to be
$\log_2 \frac{1}{p\su{MDL}(n)}$.  This assigns shorter
lengths to smaller integers as is intuitive; in addition, it assigns
as short lengths as possible asymptotically to large integers.

We now describe the encoding.
Recall that we are only concerned with description {\it lengths}, as opposed
to actual descriptions; thus, we only describe lengths here.  The encoding
is as follows:
\bi
\i First, we encode the list of all terminal symbols.  How this is
done is not important assuming the size of this list remains constant.
We are only concerned with {\it changes\/} in grammar size, as we are
only concerned with calculating {\it changes\/} in the objective function.

\i Then, we encode the number $n_s$ of nonterminal symbols excluding S and X
using the universal MDL prior.

\i For each nonterminal symbol $A \neq$ S, X, we code the following:
	\bi
	\i We code the count $c(A)$ (using the universal MDL prior) used
	to calculate $p(A) = \frac{c(A)}{\sum_A c(A)}$, the probability
	associated with the rule X $\r A$.
	\i We code the type of the symbol, \eg, whether it is a
	concatenation rule, a classing rule, or a repetition rule.  In all there
	are eight types (some of which we have yet to describe),
	so we use three bits to code this.
	\i For each rule type, we have a different way of
	coding the right-hand side of the rule, which will be
	described below.
	\ei

\ei
The symbol on the left-hand side of each rule is given implicitly
by the order in which the symbols are listed.  That is, the first
symbol listed is $A_1$, the second $A_2$, and so on up to $A_{n_s}$.
Notice that each symbol expands using exactly one of eight possible forms;
we do not have to consider listing multiple rules for a given symbol.

We do not have to list the rules expanding S or X
because they can be determined implicitly from the list of nonterminal symbols.
Likewise, all smoothing rules can be determined implicitly.
Furthermore, all probabilities associated with the grammar
can be determined from just the form of the grammar, except for
the smoothing probability $p_s$ and
the probabilities $p(A)$ associated with the rules X $\r A$.  The
probabilities $p(A)$ are coded explicitly above.  We assume
the probability $p_s$ is of constant size.

Below, we describe the eight different rule types and how we
code the right-hand side of each.
\begin{description}
\item[expansion to a terminal ($A \r a$)]  Since none of these rules
are created during the search process, it is not important how we
code these rules assuming their size is constant.
Recall that we are only concerned with {\it changes\/} in
grammar length.

\item[concatenation rule ($A \r BC$)]  We restrict concatenation rules to
have exactly two symbols on the right-hand side,
so we need not code concatenation length; we need only code
the identities of the two symbols.  We constrain these two symbols on
the right-hand side to be nonterminal symbols; this is not restrictive
since there is a nonterminal symbol expanding exclusively to each
terminal symbol.  As there are $n_s$ nonterminal symbols, we
can use $\log_2 n_s$ bits to code the identity of each of the two
symbols.  In all rule types, to code a symbol identity we
use $\log_2 n_s$ bits.

\item[classing rule ($A \r B|C$)]  Again, we need only code two
symbol identities, and we code each using $\log_2 n_s$ bits.

\item[repetition rule ($A \r AB|B$)]  We
need only code the identity of the $B$ symbol to uniquely identify
this type of rule, and we code this using $\log_2 n_s$ bits.

\begin{figure}
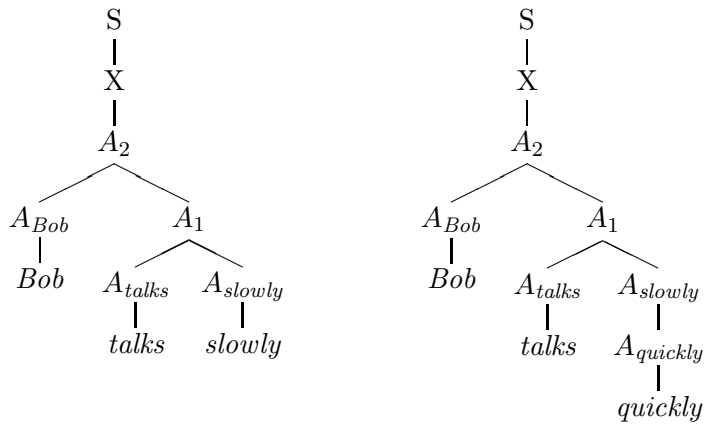


$$
\leaf{\it Bob}
\branch{1}{$A\su{\it Bob}$}
\leaf{\it talks}
\branch{1}{$A\su{\it talks}$}
\leaf{\it slowly}
\branch{1}{$A\su{\it slowly}$}
\branch{2}{$A_1$}
\branch{2}{$A_2$}
\branch{1}{X}
\branch{1}{S}
\vtop{\hbox{}\hbox{\tree}} \hspace{0.5in}
\leaf{\it Bob}
\branch{1}{$A\su{\it Bob}$}
\leaf{\it talks}
\branch{1}{$A\su{\it talks}$}
\leaf{\it quickly}
\branch{1}{$A\su{\it quickly}$}
\branch{1}{$A\su{\it slowly}$}
\branch{2}{$A_1$}
\branch{2}{$A_2$}
\branch{1}{X}
\branch{1}{S}
\vtop{\hbox{}\hbox{\tree}} $$

\caption{Smoothing rules} \label{fig:derived}
\end{figure}

\item[derived rule]  This is the encoding tailored to
the move triggered by the application of a smoothing rule, as described in
Section \ref{sssec:srules}.  We can
identify the set of rules that are created in such a move
with the following information: the
symbol underneath which the smoothing rule was applied, the
location of the application of the smoothing rule, and the
symbol the smoothing rule expanded to.  For instance, consider
the example given in Section \ref{ssec:moveset};
we re-display this in Figure \ref{fig:derived}.
Originally, the following rules exist:
$$ \begin{array}{ccl}
A_1 & \r & A\su{\it talks} A\su{\it slowly} \\
A_2 & \r & A\su{\it Bob} A_1
\end{array} $$
and the smoothing rule triggers the creation of the following rules:
$$ \begin{array}{ccl}
A_1' & \r & A\su{\it talks} A\su{\it quickly} \\
A_2' & \r & A\su{\it Bob} A_1'
\end{array} $$
We can describe the new symbol $A_2'$ as follows: it
is just like the symbol $A_2$, except where $A_2$ expands to
$A\su{\it slowly}$, $A_2'$ expands to $A\su{\it quickly}$ instead.
The definition
of $A_1'$ can implicitly be determined from this definition of $A_2'$.
Thus, to encode $A_2'$, we need to encode $A_2$,
the location of $A\su{\it slowly}$,
and the symbol $A\su{\it quickly}$.  To code the two symbols,
we use $\log_2 n_s$ bits for each as before.  To code the location,
in Figure \ref{fig:derived} we see
that there are five internal nodes in the parse tree headed by $A_2$ (if no
smoothing rules are applied).  A smoothing rule can be applied at any of
these five nodes.  Thus, we need $\log_2 5$ bits to code the location
of the application of the smoothing rule.  This size will
vary with the symbol under which the smoothing rule occurs.
In general, we need a total of $2 \log_2 n_s + \log_2
(\mbox{\it \# locations})$
bits to code the set of rules triggered by a smoothing rule.  We
call this type of encoding a {\it derived\/} rule, since it derives
the definition of one symbol from the definition of another.

\item[deletion-derived rule]  This is identical to
the derived rule just described, except that it corresponds
to the application of an $\e$-smoothing rule $A \r \e$ instead
of a regular smoothing rule $A \r B$.  In this case, we
define a new symbol as equal to an existing symbol except that
one of its subsymbols is deleted (\ie, is replaced with $\e$).
To code a deletion-derived rule, we just need to code
the original symbol ($\log_2 n_s$ bits) and the location
of the deletion ($\log_2 (\mbox{\it \# locations})$ bits); we do not
have to code a second symbol.

\begin{figure}
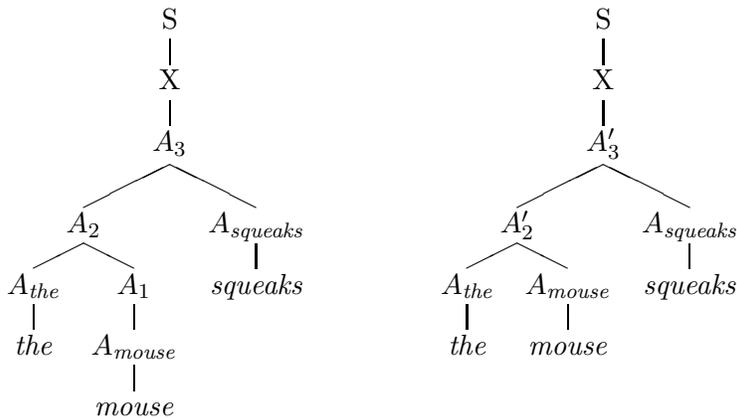


$$
\leaf{\it the}
\branch{1}{$A\su{\it the}$}
\leaf{\it mouse}
\branch{1}{$A\su{\it mouse}$}
\branch{1}{$A_1$}
\branch{2}{$A_2$}
\leaf{\it squeaks}
\branch{1}{$A\su{\it squeaks}$}
\branch{2}{$A_3$}
\branch{1}{X}
\branch{1}{S}
\vtop{\hbox{}\hbox{\tree}} \hspace{0.5in}
\leaf{\it the}
\branch{1}{$A\su{\it the}$}
\leaf{\it mouse}
\branch{1}{$A\su{\it mouse}$}
\branch{2}{$A_2'$}
\leaf{\it squeaks}
\branch{1}{$A\su{\it squeaks}$}
\branch{2}{$A_3'$}
\branch{1}{X}
\branch{1}{S}
\vtop{\hbox{}\hbox{\tree}} $$

\caption{Before and after specialization} \label{fig:specrule}
\end{figure}

\item[specialization rule]  This can be viewed as identical to the derived
rule, except that it corresponds to the application of
a classing rule as opposed to
a smoothing rule.  We will define a new symbol as
equal to an existing symbol, except that instead of replacing
a subsymbol with an arbitrary symbol as in a derived rule, we
replace the subsymbol with a symbol that the subsymbol expands to.
For instance, consider the example given in Section \ref{ssec:moveset};
we re-display this in Figure \ref{fig:specrule}.
We can define the symbol $A_3'$ to be equal to $A_3$,
except that the symbol $A_1$ is replaced with the symbol $A\su{\it mouse}$.
This rule can be coded in the same way as a derived rule.  However,
we have an additional constraint not present with derived rules:
we know that the symbol that is used to replace the original symbol
at a given location is a specialization of the original symbol.  That is,
the original symbol expands to the replacing symbol via some number of
classing rules.  We can code the replacing symbol taking
advantage of this observation; if the original symbol at the
given location expands to a total of $c$ different symbols via classing,
then we can code the replacing symbol using $\log_2 c$ bits.
Furthermore, we also have a constraint on the location not present
in derived rules: we know that the original symbol at that
location must be a classing symbol.  Thus, instead of coding the
location with $\log_2 (\mbox{\it \# locations})$ bits, we can code
it with $\log_2 (\mbox{\it \# class locations})$ bits.  In summary,
we can code specialization rules using
$\log_2 n_s + \log_2 (\mbox{\it \# class locations}) + \log_2 c$ bits.

\item[repetition specialization rule]  This is identical to
the specialization rule, except instead of dealing with classing
rules it is concerned with repetition rules.  Notice
that a repetition rule $A \r AB | B$ can just be viewed
as a classing rule of the form $A \r B | BB | BBB | \cdots$.
Thus, this rule can be coded in a similar manner to a
regular specialization rule.
Instead of coding the location using $\log_2 (\mbox{\it \# class locations})$
bits, we code it using $\log_2 (\mbox{\it \# repeat locations})$ bits.
Instead of coding the replacing symbol using $\log_2 c$ bits,
we code the number of repetitions using the universal MDL prior.

\end{description}

\ssec{Parsing}

To calculate the most probable parse of a sentence given
the current hypothesis grammar, we use a {\it probabilistic chart
parser} \cite{Younger:67a,Jelinek:92a}.
In chart parsing, one fills in a {\it chart\/}
composed of {\it cells}, where each cell represents a span
in the sentence to be parsed.  If the sentence
is composed of the words $w_1 \cdots w_m$, then there is a
cell for each $i$ and $j$ such that $1 \leq i \leq j \leq m$ corresponding
to the span $w_i \cdots w_j$.  Each cell is filled with
the set of symbols that can expand to the associated span $w_i \cdots w_j$.
For example, if the sentence is accepted under the grammar,
then the symbol S will occur in the cell corresponding to $w_1 \cdots w_m$.
The cells can be filled in an efficient manner with dynamic
programming \cite{Bellman:57a}.  Performing {\it probabilistic\/}
chart parsing just requires some extra bookkeeping; the algorithm
is essentially the same.\footnote{This is only true when trying
to calculate the most probable parse of a sentence.  In some
applications, one attempts to find the total probability of a sentence, which
involves summing the probabilities of all of its parses;
in this case probabilistic chart parsing is somewhat more involved
than normal chart parsing because of the difficulty in calculating
probabilities when some types of recursions are present in the grammar.}

However, straightforward parsing is not efficient given that
we have smoothing rules of the form $A \r B$ and $A \r \e$ for
all nonterminal symbols $A, B \neq$ S, X.  With these rules,
it is possible for any symbol to expand to any span of a sentence;
each cell in the chart will be filled with every symbol in
the grammar.  Consequently,
naive parsing with smoothing rules achieves the absolute worst-case
time bounds for chart parsing.  This is unacceptable in this application.

\begin{figure}
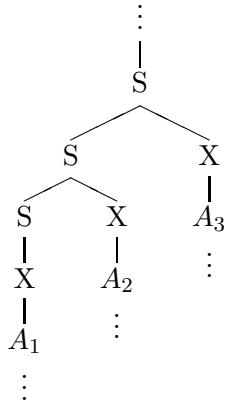


\leaf{$\begin{array}{c} A_1 \\ \vdots \end{array}$}
\branch{1}{X}
\branch{1}{S}
\leaf{$\begin{array}{c} A_2 \\ \vdots \end{array}$}
\branch{1}{X}
\branch{2}{S}
\leaf{$\begin{array}{c} A_3 \\ \vdots \end{array}$}
\branch{1}{X}
\branch{2}{S}
\branch{1}{$\vdots$}

$$ \tree $$

\caption{Typical parse-tree structure} \label{fig:parsea}
\end{figure}

Instead, we have heuristics for restricting the application of smoothing rules.
We first parse the sentence
without using any smoothing rules.  This yields a parse of
the form displayed in Figure \ref{fig:parsea}.  Then, we make the assumption
that applying smoothing rules to the structure below the $A_i$ in the
diagram is not profitable; smoothing rules improve the probability
of a parse the most if they enable grammar rules to apply in places
where none applied before.  We take the $A_i$ to
be the primitive units in the sentence, and only allow smoothing
rules to be applied immediately above these units.  We re-parse
the sequence $A_1 A_2 \cdots$ given these smoothing rules, yielding a new
best parse.  We repeat this process with this new best parse,
until the best parse is unchanging.  At some point,
smoothing rules will not affect the most probable parse.


\ssec{Extensions} \label{ssec:extend}

After completing the algorithm described in the previous part
of this section \cite{Chen:95c}, which we
will refer to as the {\it basic algorithm}, we experimented
with different extensions to arrive at what we refer to as the
{\it extended algorithm}.  In this section, we describe
the differences between the basic and extended algorithms.

\sssec{Concatenation}

In the basic algorithm, we restrict concatenation rules $A \r A_1 A_2$
to have exactly two symbols on the right-hand side.  In the
extended algorithm, we allow an arbitrary number of symbols
$A \r A_1 A_2 \cdots$.  However, we do not enhance the move
set by adding a move that concatenates arbitrary $k$-tuples instead
of just pairs, as the bookkeeping and computation required for this
would be very expensive.  Instead, we look for opportunities
to {\it unfold\/} shorter concatenation rules.  For example,
if we have two rules
$$ \begin{array}{ccl}
A_1 & \r & A_2 A_3 \\
A_3 & \r & A_4 A_5
\end{array} $$
and we notice that the symbol $A_3$ occurs nowhere else in the grammar,
then we can replace this pair of rules with the single rule
$$ \begin{array}{ccl}
A_1 & \r & A_2 A_4 A_5
\end{array} $$
We can {\it unfold\/} the definition of $A_3$ into the first
rule to yield the longer concatenation rule.  By allowing
long concatenations, we make it possible to express new
concatenations more compactly, which is advantageous
with respect to the objective function.

To handle this extension in our encoding scheme,
instead of just needing to code two symbol identities
as in the basic algorithm,
we first encode the number of symbols $k$ in the concatenation using
the universal MDL prior.  Then, we encode the $k$ symbol indentities.

\sssec{Classing}

In the basic algorithm, we restrict classing rules $A \r A_1 | A_2$
to have exactly two symbols on the right-hand side.  In the
extended algorithm, we allow an arbitrary number of symbols
$A \r A_1 | A_2 | \cdots$.  To support this representation,
we add a new move to the move set.  In the basic algorithm,
the only classing move is to create a new rule $A \r A_1 | A_2$.
In the extended algorithm, we also allow the move of adding
a new member to an existing class.  Both of these moves
fulfill the purpose of placing symbols
into a common class; the one that is chosen in
a particular situation is determined by which is preferred
by the objective function.

With this new move, it becomes
possible to build grammars that express arbitrary context-free languages,
instead of just regular languages.  In the basic algorithm,
it is impossible to create recursion (except in the repetition rule)
because symbols can only be defined in terms of symbols created
earlier in the search process.  Using this new
move, we can place into the definition of a symbol a symbol created
subsequently, thus enabling recursion.

Another difference in the extended
algorithm is that we train the probabilities associated
with classing rules.  In the basic algorithm, for a classing
rule $A \r A_1 | A_2$ we set the probabilities $p(A \r A_i)$ of $A$
expanding to each $A_i$ in a deterministic way depending only
on the form of the grammar as described in Section \ref{sssec:clrules}.
In the extended algorithm, we train the probabilities $p(A \r A_i)$
by counting the frequency
of each reduction $A \r A_i$ in the current best parse $\P$ of
the training data.  We take
$$ p(A \r A_i) = \frac{c_A(A_i)}{\sum_i c_A(A_i)} $$
(ignoring the factor of $1 - p_s$ for handling smoothing rules)
where $c_A(A_i)$ denotes the number of times
the reduction $A \r A_i$ is used in $\P$.
By training class probabilities, it should be possible to
build more accurate models of the training data.

In the encoding, instead of just needing to code two symbol identities
as in the basic algorithm,
we first encode the number of symbols $k$ in the class using
the universal MDL prior.  Then, we encode the $k$ symbol indentities.
We need also to encode the counts $c_A(A_i)$ for each $A_i$; we do this by
first coding the total number of counts $c_A = \sum_{i=1}^k c_A(A_i)$ using
the universal MDL prior.  There are
${{c_A + k - 1}\choose{k - 1}}$ possible ways to distribute $c_A$
counts among $k$ elements;\footnote{To see this, consider
a row of $c_A$ white balls.  By inserting $k-1$ black balls
into this row of white balls, we can represent a partitioning
of the $c_A$ white balls into $k$ bins: the black balls separate
each pair of adjacent bins.  Each of these partitionings correspond
to a different way of dividing $c_A$ counts among $k$ elements.
The total number of partitionings is equal to the number
of different ways of
placing the $k-1$ black balls among the total of $c_A + k - 1$ balls
in the row, or ${{c_A + k - 1}\choose{k-1}}$.
} thus, we need $\log_2 {{c_A + k - 1}\choose{k - 1}}$ bits to
specify the values $c_A(A_i)$ given $c_A$ and $k$.

\sssec{Smoothing Rules}

There are two modifications we make with respect to smoothing rules.
First, we add {\it insertion\/} smoothing
rules, which can be thought of as the complement of deletion or
$\e$-smoothing rules.  While deletion rules enable
a symbol that expands to the string {\it Bob talks
slowly\/} to also parse the string {\it Bob talks\/} as
described in Section \ref{sssec:srules}, insertion rules
allow the converse: they enable a symbol that expands to {\it Bob talks\/}
to also parse the string {\it Bob talks slowly}.

Insertion rules are of the form $A \r AB$
for all nonterminal symbols $A, B \neq$ S, X.  Such a rule ``inserts''
a $B$ immediately after an $A$.  The probability associated
with this rule is $\frac{p_s}{3} p_G(B)$ where $p_G(B)$ is
defined as in Section \ref{sssec:srules}, and the probability
of smoothing rules $A \r B$ and $A \r \e$ are reduced to
$\frac{p_s}{3} p_G(B)$ and $\frac{p_s}{3}$, respectively.
The occurrence of an insertion rule $A \r AB$ triggers the creation
of a concatenation rule concatenating $A$ and $B$.

\begin{figure}
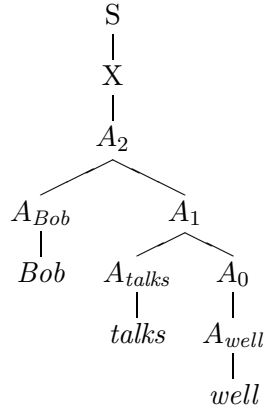


\leaf{\it Bob}
\branch{1}{$A\su{\it Bob}$}
\leaf{\it talks}
\branch{1}{$A\su{\it talks}$}
\leaf{\it well}
\branch{1}{$A\su{\it well}$}
\branch{1}{$A_0$}
\branch{2}{$A_1$}
\branch{2}{$A_2$}
\branch{1}{X}
\branch{1}{S}

$$ \tree $$

\caption{Generalization} \label{fig:realgen}
\end{figure}

The second modification deals with the moves that are triggered by
the occurrence of a smoothing rule.  Because the extended algorithm
contains a move for adding symbols to classes instead of just
a move for creating classes, in the extended algorithm
smoothing rules can trigger a move
we could not consider in the basic algorithm.  Consider the rules
$$ \begin{array}{ccl}
A_0 & \r & A\su{\it quickly} | A\su{\it slowly} \\
A_1 & \r & A\su{\it talks} A_0 \\
A_2 & \r & A\su{\it Bob} A_1
\end{array} $$
defining a symbol $A_2$ that expands to the strings {\it Bob talks quickly\/}
and {\it Bob talks slowly}.  Now, consider encountering
a new string {\it Bob talks well\/} that we parse using the
symbol $A_2$ and the smoothing rule $A_0 \r A\su{\it well}$,
as displayed in Figure \ref{fig:realgen}.  In the basic algorithm,
this triggers the possible creation of the rules
$$ \begin{array}{ccl}
A_1' & \r & A\su{\it talks} A\su{\it well} \\
A_2' & \r & A\su{\it Bob} A_1'
\end{array} $$
describing a symbol $A_2'$ that expands just to the string {\it Bob
talks well}.  However, with the ability to add new members to classes,
we can instead just trigger the addition of $A\su{\it well}$
to the class $A_0$ yielding
$$ \begin{array}{ccl}
A_0 & \r & A\su{\it quickly} | A\su{\it slowly} | A\su{\it well}
\end{array} $$
Instead of having a symbol $A_2$ expanding to {\it Bob talks quickly\/}
and {\it Bob talks slowly} and a symbol $A_2'$ expanding to {\it Bob
talks well}, this new move yields a single symbol $A_2$ expanding
to all three strings.  This is intuitively a more appropriate action, and
results in a more compact grammar.

\sssec{Grammar Compaction}

In the basic algorithm, the move set consists entirely of moves
that create new rules, \ie, moves that expand the grammar
(and hopefully compress the training data).  In the extended
algorithm, we consider moves that compact the grammar (and
leave the data the same size or perhaps even enlarge the training data).
These moves help correct for the greedy nature of the search
strategy, by providing a mechanism for reducing the number of grammar rules.
For example, if we have the following grammar rules
$$ \begin{array}{ccl}
A_1 & \r & A\su{\it talks} A\su{\it slowly} \\
A_2 & \r & A\su{\it Bob} A_1 \\
\mbox{} \\
A_1' & \r & A\su{\it talks} A\su{\it quickly} \\
A_2' & \r & A\su{\it Bob} A_1'
\end{array} $$
it may be profitable to replace the above rules with the rules
$$ \begin{array}{ccl}
A_0 & \r & A\su{\it quickly} | A\su{\it slowly} \\
A_1'' & \r & A\su{\it talks} A_0 \\
A_2'' & \r & A\su{\it Bob} A_1''
\end{array} $$
More generally, we search for rules
that differ by a single symbol, and attempt to merge these rules.

To constrain what rules we consider merging, we only consider merging rules
that expand symbols that occur in a common class.  This constraint
is necessary because there are many constructions that
are on the surface very similar that have different meanings.
For example, the strings {\it a can\/}
and {\it John can\/} differ by only one word, but are unrelated
in meaning.  We restrict rule merging to symbols that are in
a common class because hopefully symbols that have been placed
in the same class are semantically related.

In general, for any class $A \r A_1 | A_2 | \cdots$, we
attempt to create new symbols merging multiple $A_i$ whose definitions
differ by only a single symbol.  Whenever we make such a symbol,
we substitute it into the right-hand side of the classing rule.  If through
this merging we yield a classing rule $A \r B$ with
only a single symbol on the right-hand side, we merge
the two symbols $A$ and $B$ into a single symbol.

\sssec{Symbol Encoding}

In the basic algorithm, we encode symbol identities
using $\log_2 n_s$ bits, where $n_s$ is
the total number of nonterminal symbols.  However, recall that
coding theory states that a fixed-length coding such as
this is an optimal coding only if
all symbols are equiprobable; in general, a symbol with frequency
$p$ should be coded with $\log_2 \frac{1}{p}$ bits.  Intuitively,
it seems reasonable to code rules involving frequent symbols
such as $A\su{\it the}$ with fewer bits than rules involving
rare symbols such as $A\su{\it hippopotamus}$.

We calculate the frequency $p_G(A)$ of each symbol $A$ in the grammar
as $\frac{c_G(A)}{\sum_A c_G(A)}$, where $c_G(A)$ is the number
of times the symbol $A$ occurs in the grammar.  We code a symbol $A$
using $\log_2 \frac{1}{p_G(A)}$ bits.  However, notice that
we will not know $c_G(A)$ until the end of
the grammar description, but these values are needed to code the grammar.
To resolve this problem, we explicitly code the values of $c_G(A)$
for all $A$ before we code the grammar rules.  We first
code $c_G = \sum c_G(A)$, the total number of symbol occurrences
in the grammar, using the universal MDL prior.  Then, there are
${{c_G + n_s - 1}\choose{n_s - 1}}$ ways to distribute $c_G$ counts
among $n_s$ elements, so we just need
$\log_2 {{c_G + n_s - 1}\choose{n_s - 1}}$ bits to code the values
of $c_G(A)$ given $c_G$ and $n_s$.\footnote{Using an adaptive
coding method, it may be possible to encode symbols even more
compactly.  However, the gain in compactness probably does not
warrant the additional complexity of implementation.}

\sssec{Maintaining the Best Parse}

As the move set grows in complexity, it becomes more difficult
from an implementational standpoint to maintain an accurate estimate of
the most probable parse $\P$.  Furthermore, the amount of
memory needed to store $\P$ grows linearly in the training data size, so
for large training sets it may be impractical to store $\P$ entirely
in memory, as is necessary for good performance.  In the
extended algorithm, instead of keeping track of $\P$ explicitly,
we just estimate the {\it counts\/} of all salient
events in $\P$.

\begin{figure}
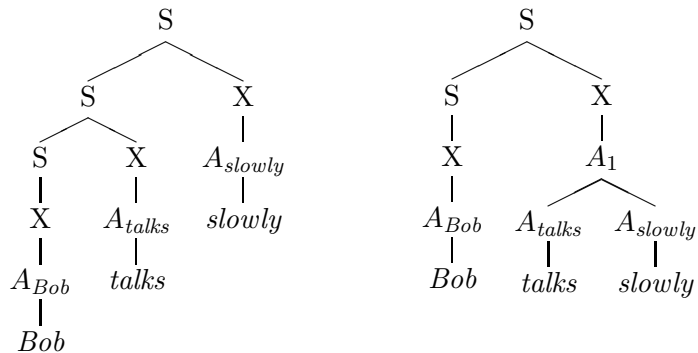


$$
\leaf{\it Bob}
\branch{1}{$A\su{\it Bob}$}
\branch{1}{X}
\branch{1}{S}
\leaf{\it talks}
\branch{1}{$A\su{\it talks}$}
\branch{1}{X}
\branch{2}{S}
\leaf{\it slowly}
\branch{1}{$A\su{\it slowly}$}
\branch{1}{X}
\branch{2}{S}
\vtop{\hbox{}\hbox{\tree}} \hspace{0.5in}
\leaf{\it Bob}
\branch{1}{$A\su{\it Bob}$}
\branch{1}{X}
\branch{1}{S}
\leaf{\it talks}
\branch{1}{$A\su{\it talks}$}
\leaf{\it slowly}
\branch{1}{$A\su{\it slowly}$}
\branch{2}{$A_1$}
\branch{1}{X}
\branch{2}{S}
\vtop{\hbox{}\hbox{\tree}} $$

\caption{Before and after concatenation} \label{fig:shady}
\end{figure}

For example, to calculate whether it is profitable to create
a concatenation rule $A \r BC$, we need to know the number of
times two adjacent X's expand to the symbols $B$ and $C$, respectively.
Let us call this quantity $c(B, C)$.
We need to keep track of counts $c(B, C)$ for all symbols $B$ and $C$.
Likewise, there are similar counts that we need to keep track of
for the other types of moves.  In the extended algorithm, we just
record these counts as opposed to the actual parse $\P$.

Unfortunately, while less expensive this system is also less accurate.
To update counts when
new sentences are parsed is straightforward; however, errors
are introduced whenever we apply a move to previous sentences.
For example, consider Figure \ref{fig:shady} depicting the parse
tree of {\it Bob talks slowly\/} before and after the creation
of the concatenation rule $A_1 \r A\su{\it talks} A\su{\it slowly}$.
After the move, we know to set $c(A\su{\it talks}, A\su{\it slowly})$
to zero as the concatenation
rule will be applied at each relevant location; however, it is
more difficult to update overlapping counts.
For example, for the sentence
in the example we should decrement $c(A\su{\it Bob}, A\su{\it talks})$
and increment $c(A\su{\it Bob}, A_1)$.  If we maintain $\P$
explicitly, this is straightforward; however, if we are only
maintaining counts on $\P$, the necessary information to update
these counts correctly is generally not available (unless enough
counts are available to completely reconstruct $\P$).  We use
heuristics to update counts as best we can given available information.

%
\sec{Results} \label{sec:gresults}
%

To evaluate our algorithm, we compare the performance of our algorithm
to that of $n$-gram models and the Lari and Young algorithm.

For $n$-gram models, we tried $n = 1, \ldots, 10$ for each domain.
To smooth the $n$-gram models, we use a popular version of
Jelinek-Mercer smoothing \cite{Jelinek:80a,Bahl:83a}, namely
the version that we refer to as {\tt interp-held-out} described in
Section \ref{sssec:jmimpl}.

In the Lari and Young algorithm, the initial grammar is taken to be
a probabilistic context-free grammar consisting
of all Chomsky normal form rules over $n$ nonterminal
symbols $\{X_1, \ldots X_n\}$ for some $n$, that is, all rules
$$ \begin{array}{cccll}
X_i & \r & X_j & X_k \hspace{0.3in} & i, j, k \in \{ 1, \ldots, n \} \\
X_i & \r & a & & i \in \{1, \ldots, n\}, a \in T
\end{array} $$
where $T$ denotes the set of terminal symbols in the domain.  All rule
probabilities
are initialized randomly.  From this starting point, the Inside-Outside
algorithm is run until the average entropy per word on the training
data changes less than a certain amount between iterations; in this
work, we take this amount to be 0.001 bits.

For smoothing the grammar yielded by
the Lari and Young algorithm, we interpolate the expansion distribution of each
symbol with a uniform distribution; that is,
for a grammar rule $A \r \a$ we take its smoothed probability
$p_s(A \r \a)$ to be
$$ p_s(A \r \a) = (1 - \l) p_b(A \r \a) + \l \frac{1}{n^3 + n|T|} $$
where $p_b(A \r \a)$ denotes its probability before smoothing.  The value
$n^3 + n|T|$ is the number of rules expanding a symbol under
the Lari and Young methodology.  The parameter $\l$ is trained through the
Inside-Outside algorithm on held-out data.  This smoothing is
also performed on the grammar yielded by the
Inside-Outside post-pass of our algorithm.
For each domain, we tried $n = 3, \ldots, 10$.

Because of the computational demands of our algorithm, it is
currently impractical to apply it to large vocabulary or
large training set problems.  However, we present the results of
our algorithm in three medium-sized domains.  In each case,
we use 4500 sentences for training, with 500 of these sentences held out
for smoothing.  We test on 500 sentences, and measure performance
by the entropy of the test data.

\begin{figure}

$$ \begin{tabular}{l}
\begin{tabular}{cccccc}
S & $\r$ & NP & VP & & (1.0) \\
NP & $\r$ & D & N & & (0.5) \\
	& $|$ & PN & & & (0.3) \\
	& $|$ & NP & PP & & (0.2) \\
VP & $\r$ & V0 & & & (0.2) \\
	& $|$ & V1 & NP & & (0.4) \\
	& $|$ & V2 & NP & NP & (0.2) \\
	& $|$ & VP & PP & & (0.2) \\
PP & $\r$ & P & NP & & (1.0)
\end{tabular} \\
\mbox{} \\
\begin{tabular}{ccl}
D & $\r$ & \t{a} $|$ \t{the} \\
N & $\r$ & \t{car} $|$ \t{bus} $|$ \t{boy} $|$ \t{girl} \\
PN & $\r$ & \t{Joe} $|$ \t{John} $|$ \t{Mary} \\
P & $\r$ & \t{on} $|$ \t{at} $|$ \t{in} $|$ \t{over} \\
V0 & $\r$ & \t{cried} $|$ \t{yelled} $|$ \t{ate} \\
V1 & $\r$ & \t{hit} $|$ \t{slapped} $|$ \t{hurt} \\
V2 & $\r$ & \t{gave} $|$ \t{presented}
\end{tabular}
\end{tabular} $$

\caption{Sample grammar used to generate data} \label{fig:handgram}
\end{figure}

In the first two domains, we created the training and test data
artificially so as to have an ideal grammar in hand to benchmark results.
In particular, we used a probabilistic context-free grammar to generate
the data.  In the first domain, we created this grammar by hand;
this simple English-like grammar is displayed in Figure \ref{fig:handgram}.
The numbers in parentheses are the probabilities associated with
each rule; rules without listed probabilities are equiprobable.
In the second domain, we derived the grammar from manually
parsed text.  From a million words of parsed Wall Street Journal
data from the Penn treebank, we extracted the 20 most
frequently occurring symbols, and the 10 most frequently occurring rules
expanding each of these symbols.  For each symbol that occurred
on the right-hand side of a rule that was not one of
the most frequent 20 symbols, we created a rule that expanded
that symbol to a unique terminal symbol.  After removing unreachable rules,
this yielded a grammar of roughly
30 nonterminals, 120 terminals, and 160 rules.
Parameters were set to reflect the frequency of the
corresponding rule in the parsed corpus.

For the third domain, we took English text and reduced the size of
the vocabulary by mapping each word to its part-of-speech tag.  We used
tagged Wall Street Journal text from the Penn treebank, which has
a tag set size of about fifty.  To reduce computation time, we
only used sentences with at most twenty words.

\begin{table}

$$ \begin{tabular}{|l|c|c|c|c|} \hline
& best & entropy & entropy relative \\
& $n$ & (bits/word) & to $n$-gram \\ \hline
ideal grammar & & 2.30 & $-$6.5\% \\ \hline
extended algorithm & 8 & 2.31 & $-$6.1\% \\ \hline
basic algorithm & 7 & 2.38 & $-$3.3\% \\ \hline
$n$-gram model & 4 & 2.46 & \\ \hline
Lari and Young & 9 & 2.60 & $+$5.7\% \\ \hline
\end{tabular} $$
\caption{English-like artificial grammar \label{tab:eng}}

\end{table}

\begin{table}

$$ \begin{tabular}{|l|c|c|c|c|} \hline
& best & entropy & entropy relative \\
& $n$ & (bits/word) & to $n$-gram \\ \hline
ideal grammar & & 4.13 & $-$10.4\% \\ \hline
extended algorithm & 7 & 4.41 & $-$4.3\% \\ \hline
basic algorithm & 9 & 4.41 & $-$4.3\% \\ \hline
$n$-gram model & 4 & 4.61 & \\ \hline
Lari and Young & 9 & 4.64 & $+$0.7\% \\ \hline
\end{tabular} $$
\caption{Wall Street Journal-like artificial grammar \label{tab:wsj}}

\end{table}

\begin{table}

$$ \begin{tabular}{|l|c|c|c|c|} \hline
& best & entropy & entropy relative \\
& $n$ & (bits/word) & to $n$-gram \\ \hline
$n$-gram model & 8 & 3.00 & \\ \hline
basic algorithm & 8 & 3.12 & $+$4.0\% \\ \hline
extended algorithm & 7 & 3.13 & $+$4.3\% \\ \hline
Lari and Young & 9 & 3.60 & $+$20.0\% \\ \hline
\end{tabular} $$
\caption{English sentence part-of-speech sequences \label{tab:pos}}

\end{table}

\begin{figure}
$$ \psfig{figure=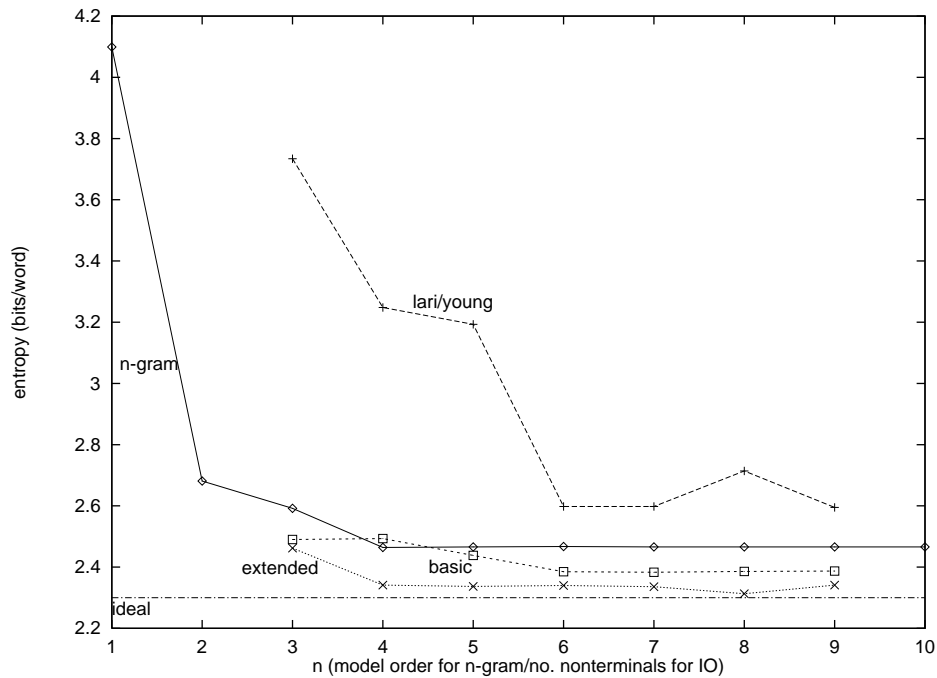,width=5in} $$
\caption{Performance versus model size, English-like artificial grammar}
	\label{fig:test1n}
\end{figure}

\begin{figure}
$$ \psfig{figure=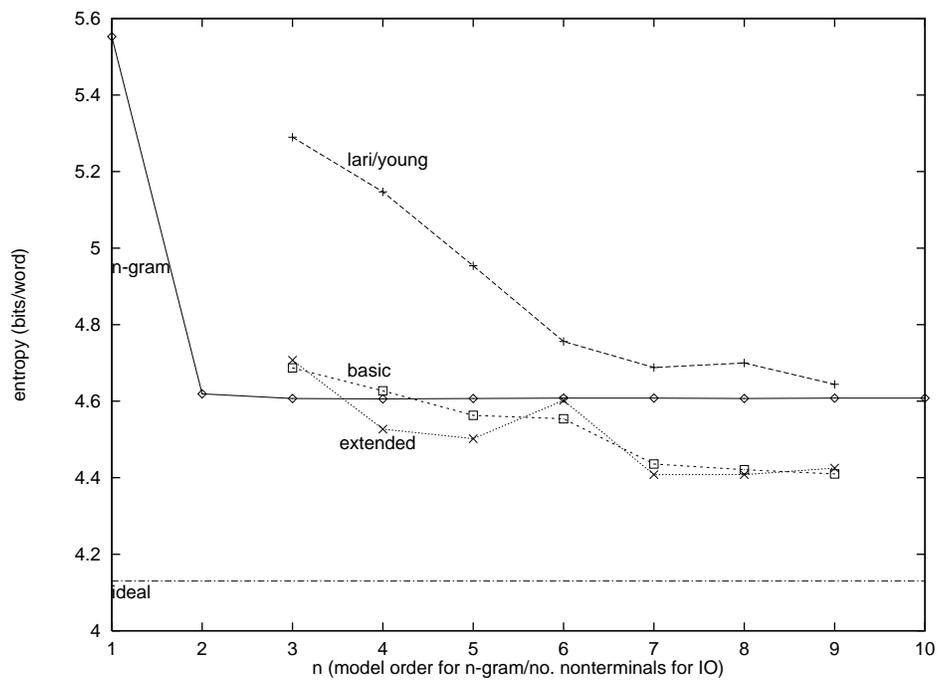,width=5in} $$
\caption{Performance versus model size, WSJ-like artificial grammar}
	\label{fig:wsj20n}
\end{figure}

\begin{figure}
$$ \psfig{figure=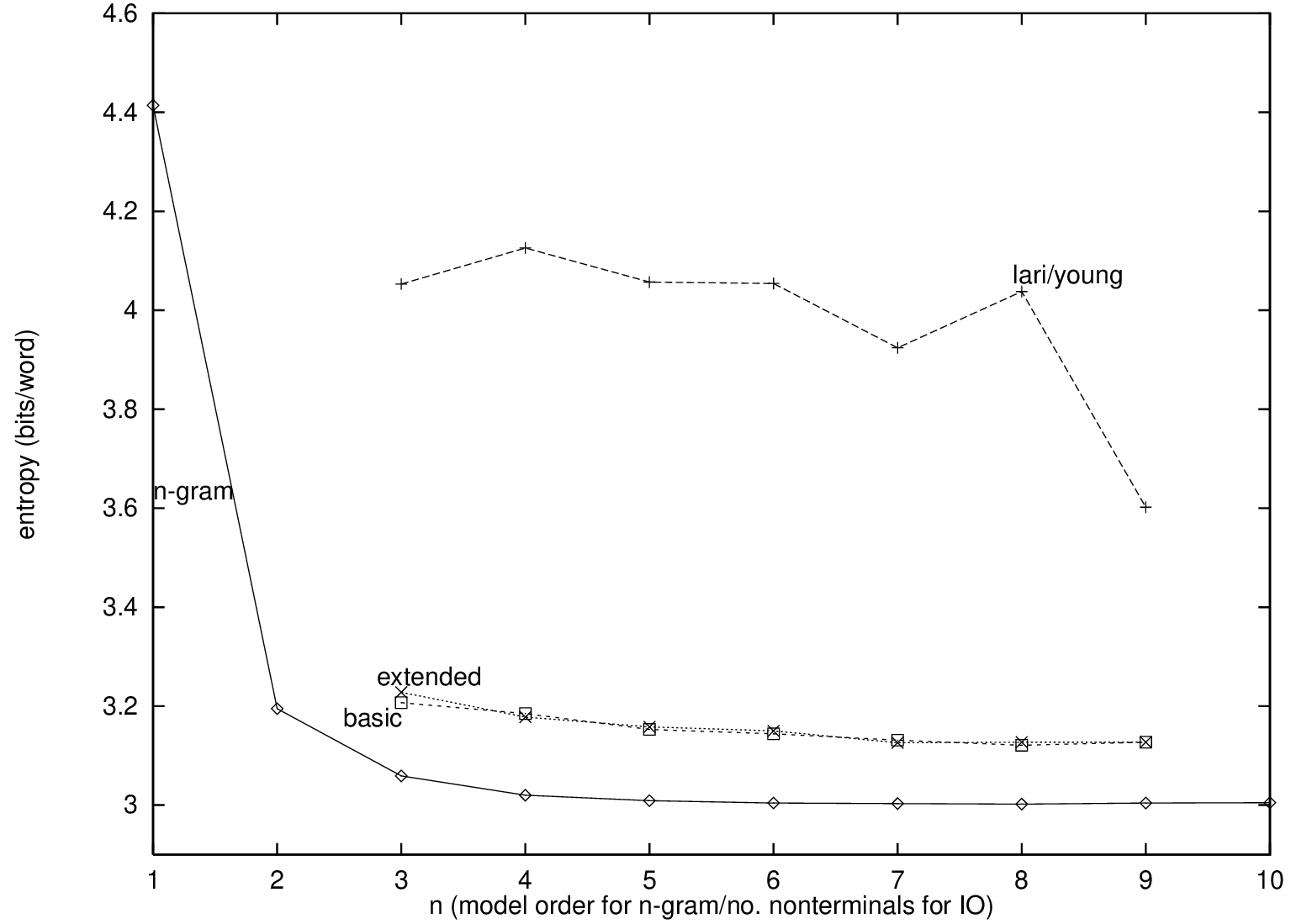,width=5in} $$
\caption{Performance versus model size, part-of-speech sequences}
	\label{fig:tagn}
\end{figure}

In Tables \ref{tab:eng}--\ref{tab:pos}, we summarize our results.
The {\it ideal grammar\/} denotes the grammar used to generate
the training and test data.  The rows {\it basic algorithm\/}
and {\it extended algorithm\/} describe the two versions of our algorithm.
For each algorithm, we list the best performance achieved over all $n$ tried,
and the {\it best $n$} column states which value realized
this performance.  For $n$-gram models, $n$ represents the order of
the $n$-gram model; for the other algorithms, $n$ represents
the number of nonterminal symbols used with the Inside-Outside
algorithm.\footnote{The data presented in Tables \ref{tab:eng}--\ref{tab:pos}
differ slightly from the corresponding data presented in an
earlier paper \cite{Chen:95c}.  Part of this difference
is due to the fact that we used a different random starting point
for the Inside-Outside post-pass of our algorithm.  In addition,
we used a different data set for the part-of-speech domain.}
In Figures \ref{fig:test1n}--\ref{fig:tagn}, we show the complete data,
the performance of each algorithm at each $n$ for each
of the three domains.

We achieve a moderate but significant improvement in performance over
$n$-gram models and the Lari and Young algorithm in the first two domains,
while in the part-of-speech domain we are outperformed by $n$-gram models
but we vastly outperform the Lari and Young algorithm.

Comparing the two versions of our algorithm, we find that the
extended algorithm performs significantly better than the basic
algorithm on the English-like artificial text, and performs
marginally better for most $n$ on the WSJ-like artificial text.
In the part-of-speech domain, the two algorithms perform almost
identically for most $n$.

\begin{figure}
{ \def\R{$\Rightarrow^*$} \it
$$ \begin{tabular}{ccl}
$A_1$ & \R & $($on$|$over$)$ John \\
$A_2$ & \R & in $($Mary$|$the girl$)$ \\
$A_3$ & \R & a bus \\
$A_4$ & \R & $($the boy$|$Joe$|$a car$|$Mary$|$a girl$|$a boy$)$ at \\
$A_5$ & \R & $($a girl$|$a boy$|$the bus$|$Mary$|$the girl$|$John$|$the car$)$
	$($hurt$|$slapped$|$hit$)$ the car \\
$A_6$ & \R & $($the bus$|$Mary$|$the girl$)$ cried \\
 & \vdots \\
 & \vdots & {\rm (13 prepositional phrases)} \\
 & \vdots \\
$A_{20}$ & \R & $($a girl$|$a boy$|$the bus$|$Mary$|$the
	girl$|$John$|$the car$)$ slapped $($the boy$|$a bus$)$ \\
$A_{21}$ & \R & at John \\
$A_{22}$ & \R & $($a girl$|$a boy$|$the bus$|$Mary$|$the girl$|$John$|$the
	car$)$ $($hurt$|$slapped$|$hit$)$ the bus
\end{tabular} $$ }
\caption{Expansions of symbols $A$ with highest frequency $p(A)$}
	\label{fig:freqsym}
\end{figure}

In Figure \ref{fig:freqsym}, we display a sample of the grammar induced
in the English-like artificial domain.  The figure displays the
expansions of the symbols $A$ with the highest probabilities $p(A)$,
which is proportional to how frequently the reduction X $\r A$ is used.
In some sense, these symbols are the most frequently occurring
symbols.  Each row corresponds to a different symbol,
listed in decreasing frequency starting from the most frequent symbol
in the grammar.  The expression displayed expresses all possible
strings the symbol expands to; it does not reflect the actual
grammar rules with that symbol on the left-hand side.  For example,
the most frequent symbol in the grammar expands to the strings
{\it on John\/} and {\it over John}, and the fifth-most frequent
symbol expands to the strings {\it a girl hurt the car}, {\it a boy hurt
the car}, etc.  Except for the fourth symbol, all
of these symbols expand to strings that are constituents according to
the original grammar.  Thus, in this domain our algorithm is
able to capture some of the structure present in the original grammar.

\begin{figure}
$$ \begin{tabular}{ccl}
$A_1$ & $\r$ & {\it hit} (0.08) $|$ {\it hurt} (0.08) $|$ {\it slapped} (0.09)
	$|$ {\it presented} (0.06) $|$ {\it gave} (0.06) \\
& $|$ & {\it over} (0.12) $|$ {\it on} (0.12) $|$ {\it in} (0.12)
	$|$ {\it at} (0.12) \\
& $|$ & $A_1$ $A_2$ (0.03) $|$ $A_1$ $A_8$ (0.04) $|$ $A_5$ $A_1$ (0.05) \\ 

$A_2$ & $\r$ & {\it Mary} (0.07) $|$ {\it Joe} (0.24) $|$ {\it John} (0.09) \\
& $|$ & $A_2$ $A_5$ (0.04) $|$ $A_4$ $A_3$ (0.12) $|$ $A_4$ $A_6$ (0.42) \\

$A_3$ & $\r$ & {\it girl} (0.03) $|$ {\it car} (0.42) $|$ {\it boy} (0.31)
	$|$ {\it bus} (0.15) \\
& $|$ & $A_3$ $A_5$ (0.03) $|$ $A_6$ $A_5$ (0.05) \\ 

$A_4$ & $\r$ & {\it the} (0.50) $|$ {\it a} (0.50) \\ 

$A_5$ & $\r$ & {\it yelled} (0.04) $|$ {\it cried} (0.04) $|$ {\it ate} (0.04)\\
& $|$ & $A_1$ $A_2$ (0.21) $|$ $A_1$ $A_7$ (0.19)
	$|$ $A_1$ $A_8$ (0.23) $|$ $A_1$ $A_9$ (0.21) $|$ $A_5$ $A_5$ (0.04) \\ 

$A_6$ & $\r$ & {\it girl} (0.26) $|$ {\it car} (0.23) $|$ {\it boy} (0.24)
	$|$ {\it bus} (0.26) \\ 

$A_7$ & $\r$ & {\it Mary} (0.09) $|$ {\it Joe} (0.06) $|$ {\it John} (0.19) \\
& $|$ & $A_2$ $A_5$ (0.02) $|$ $A_4$ $A_3$ (0.03) $|$ $A_4$ $A_6$ (0.55)
	$|$ $A_7$ $A_5$ (0.02) $|$ $A_8$ $A_5$ (0.04) \\ 

$A_8$ & $\r$ & {\it Mary} (0.17) $|$ {\it Joe} (0.07) $|$ {\it John} (0.09) \\
& $|$ & $A_4$ $A_3$ (0.04) $|$ $A_4$ $A_6$ (0.58) $|$ $A_8$ $A_5$ (0.05) \\ 

$A_9$ & $\r$ & $A_2$ $A_5$ (0.24) $|$ $A_7$ $A_5$ (0.27)
$|$ $A_8$ $A_5$ (0.37) $|$ $A_9$ $A_5$ (0.08) $|$ $A_9$ $A_7$ (0.03)
\end{tabular} $$

\caption{Grammar induced with Lari and Young algorithm} \label{fig:lygram}
\end{figure}

In Figure \ref{fig:lygram}, we display the grammar induced by
the Lari and Young algorithm with nine nonterminal symbols
in the English-like artificial domain.  We display only those
rules with probability above 0.01; rule probabilities are shown
in parentheses.  Unlike in Figure \ref{fig:freqsym} where we list
{\it all strings\/} a symbol expands to, in this figure we
list the {\it most frequent rules\/} a symbol expands with.
This grammar does a reasonable job of grouping together similar terminal
symbols.  However, it does less well at recognizing higher-level
structures in the grammar.  Most symbols in the induced grammar
do not match well with the symbols in the original grammar.  For
example, the symbol $A_3$ groups together nouns with
nouns followed by a verb taking no arguments.  Hence, we see
that our algorithm is clearly better than the Lari and Young
algorithm at capturing relevant structure.

\begin{table}
$$ \begin{tabular}{|l|c|r|r|r|} \hline
WSJ-like & $n$ & entropy & no. & time \\
artificial & & (bits/word) & params & (sec) \\ \hline
$n$-gram model & 3 & 4.61 & 15000 & 50 \\ \hline
Lari and Young & 9 & 4.64 & 2000 & 30000 \\ \hline
basic alg./first pass & & & 800 & 1000 \\ \hline
basic alg./post-pass & 5 & 4.56 & 4000 & 5000 \\ \hline
\end{tabular} $$
\caption{Number of parameters and training time of each algorithm
	\label{tab:param}}
\end{table}

In Table \ref{tab:param}, we display a sample of the number of parameters and
execution time (on a Decstation 5000/33) associated with each algorithm.
We choose $n$ to yield approximately equivalent performance for each
algorithm.  The {\it first pass\/} row refers to the main grammar induction
phase of our algorithm, and the {\it post-pass\/} row refers to
the Inside-Outside post-pass.

Notice that our algorithm produces a significantly more compact model
than the $n$-gram model, while running significantly faster than the
Lari and Young algorithm even though both algorithms employ the
Inside-Outside algorithm.
Part of this discrepancy is due to the fact that we require a smaller
number of new nonterminal symbols to achieve equivalent performance, but
we have also found that our post-pass converges more quickly even given
the same number of nonterminal symbols.

\begin{figure}
$$ \psfig{figure=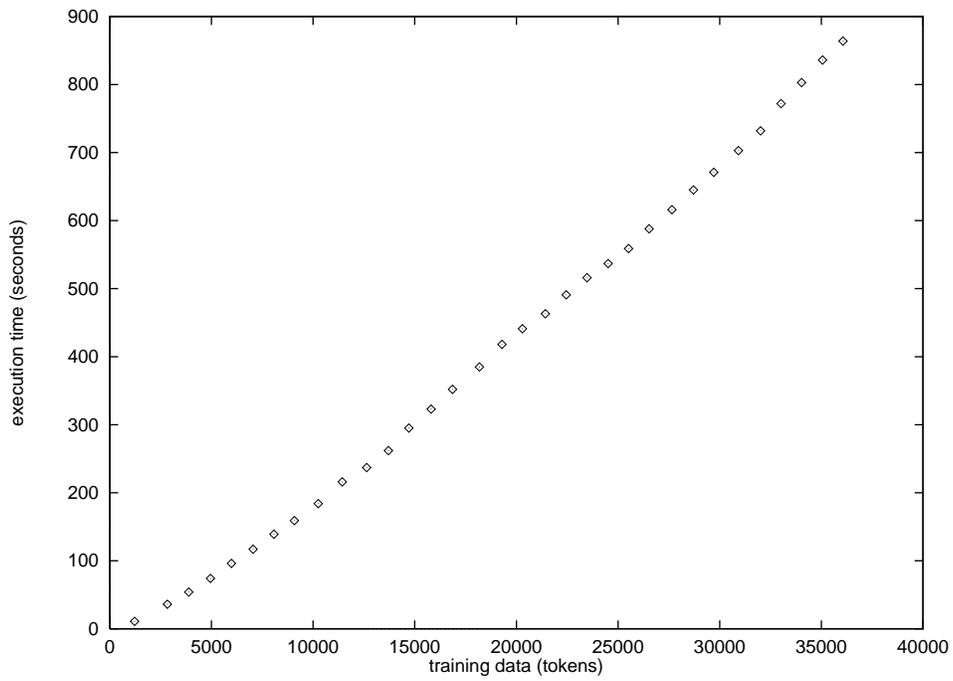,width=5in} $$
\caption{Execution time versus training data size} \label{fig:exectime}
\end{figure}

\begin{figure}
$$ \psfig{figure=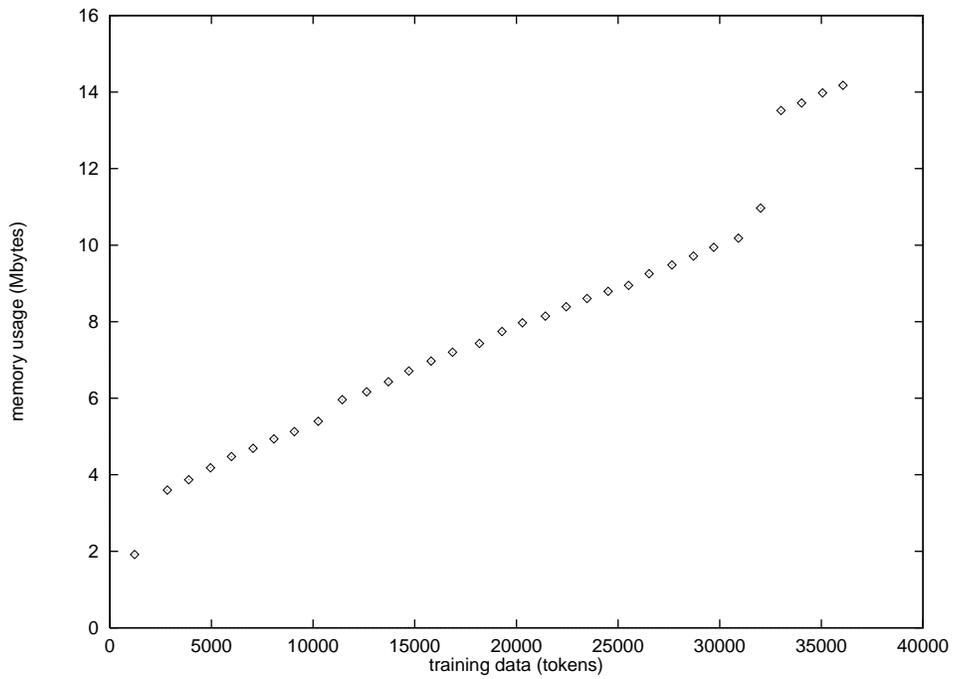,width=5in} $$
\caption{Memory usage versus training data size} \label{fig:execmem}
\end{figure}

In Figures \ref{fig:exectime} and \ref{fig:execmem}, we display
the execution time and memory usage of the main grammar induction
algorithm on various amounts of training data.  Both
of these graphs are nearly linear.\footnote{The jump in the
graph of memory usage betwen 30,000 and 35,000 tokens is an artifact of
the training data used to produce the graph.  At this point
in the training data, there are many long sequences of a single
token.  It turns out that
the number of possible ways to parse a long sequence of a single token
is very large.  For example, if we denote the
repeated token as $a$, then any symbol that can expand to $a^k$ for
some $k$ can be used to parse any substring of length $k$ in
the long sequence of $a$'s.  The jump in memory is due
to the additional memory needed to store the parse chart.}

%
\sec{Discussion}
%

Our algorithm consistently outperformed the Lari and Young algorithm
in these experiments.  One perspective of this result can
be taken noticing that both algorithms use the Inside-Outside
algorithm as a final step.  As the Inside-Outside algorithm is
a hill-climbing algorithm, it can be interpreted as finding the
nearest local minimum in the search space to the initial grammar.
In the Lari and Young framework, this initial grammar is chosen
essentially randomly.  In our framework, this initial grammar is
chosen intelligently using a Bayesian search.  Thus, it is not
surprising our algorithm outperforms the Lari and Young algorithm.

From a different perspective, the main mechanism of the Lari
and Young algorithm for selecting grammar rules is the hill-climbing
search of the Inside-Outside algorithm.  If we view this in
terms of a move set, the basic move of the Lari and Young algorithm
is to adjust rule probabilities.  In our algorithm, we use
a rich move set that corresponds to semantically meaningful
notions.  We have moves that can explicitly create new rules and
nonterminal symbols unlike Lari and Young, moves that express concepts
such as classing, specialization, and generalization.  While both
algorithms employ greedy searches and can thus be interpreted
as finding nearby local minima, our results demonstrate that
by using a richer move set this constraint is much less serious.
There have been several results demonstrating
the severity of the local minima problem for the
Inside-Outside algorithm \cite{Chen:91a,deMarcken:95a}.

In terms of efficiency, the algorithms differ significantly because
of the different ways rules are selected.
In the Lari and Young algorithm, one starts
with a large grammar and uses the Inside-Outside algorithm
to prune away unwanted rules by setting their probabilities to
near zero.  This approach scales poorly because the grammar
worked with is substantially larger than the desired target grammar,
and using large grammars has a great computational expense.
It is impractical to use more than tens of nonterminal
symbols with the Lari and Young approach because the number of
grammar rules is cubic in the number of nonterminal symbols.

In contrast, our algorithm begins with a small grammar and adds rules
incrementally.  The working grammar only contains
rules perceived as worthy and is thus not unnecessarily large.
In our algorithm, we can build grammars with hundreds of
nonterminal symbols, and still the execution time of the
algorithm is dominated by the Inside-Outside post-pass.

Outperforming $n$-gram models in the first two domains demonstrates
that our algorithm is able to take advantage of the grammatical
structure present in data.  For example, unlike $n$-gram models
our grammar can express classing and can handle long-distance dependencies.
However, the superiority of $n$-gram models in the part-of-speech
domain indicates that to be competitive in modeling naturally-occurring
data, it is necessary to model collocational information accurately.  We need
to modify our algorithm to more aggressively model $n$-gram information.

\ssec{Contribution}

This research represents a step forward in the quest for
developing grammar-based language models for natural language.
We consistently outperform the Lari and Young algorithm across domains
and outperform $n$-gram language models in medium-sized artificial domains.
The algorithm runs in near-linear time and space in terms of
the amount of training data and the grammars induced are relatively
compact, so the algorithm scales well.

We demonstrate the viability of the Bayesian approach for grammar
induction, and we show that the minimum description length
principle is a useful paradigm for building prior distributions for
grammars.  A minimum description length approach is crucial for
methods that do not limit the size of grammars; this approach favors
smaller grammars, which is necessary for preventing overfitting.

We describe an efficient framework for performing grammar
induction.  We present an algorithm that parses each
sentence only once, and we use the concept of {\it triggers\/}
to constrain the set of moves considered at each point.
We describe a rich move set for manipulating grammars
in intuitive ways, and this enables our search to be effective
despite its greedy nature.

Furthermore, this induction framework is not restricted to
probabilistic context-free grammars.  For example, notice
that we do not parameterize repetition rules strictly within the
PCFG paradigm.  More complex grammar formalisms can be
considered without changing the computational
complexity of the algorithm; we just need to enhance the move set.

%
%

\newcommand{\corptxt}[1]{\parbox[t]{2in}{\raggedright #1\rule[-1ex]{0cm}{1ex}}}
\newcommand{\corptxtb}[1]{\parbox[t]{1in}{\raggedright #1\rule[-1ex]{0cm}{1ex}}}
\newcommand{\bc}{$\backslash$SCM\{\}}
\newcommand{\ec}{$\backslash$ECM\{\}}
\def\bv{\bar{b}}
\def\bvv{\vec{b}}

\chapter{Aligning Sentences in Bilingual Text} \label{ch:align}

In this chapter, we describe an algorithm for aligning sentences with
their translations in a bilingual corpus \cite{Chen:93b}.
In experiments with the Hansard
Canadian parliament proceedings, our algorithm yields significantly better
accuracy than previous algorithms.  In addition, it is efficient,
robust, language-independent, and parallelizable.
Of the three structural levels at which we model language
in this thesis, this represents work at the sentence level.

%
\sec{Introduction}
%

A {\it bilingual corpus\/} is a corpus of text replicated in two
languages.  For example, the Hansard bilingual corpus contains
the Canadian parliament proceedings in both English and French.
Bilingual corpora have proven useful in
many tasks, including machine translation \cite{Brown:90b,Sadler:89a},
sense disambiguation \cite{Brown:91c,Dagan:91a,Gale:92a},
and bilingual lexicography \cite{Klavans:90a,Warwick:90a}.

For example, a bilingual corpus can be used
to automatically construct a bilingual dictionary.  A bilingual
dictionary can be expressed as a probabilistic
model $p(f|e)$ of how frequently a particular
word $e$ in one language, say English, translates to a particular
word $f$ in another language, say French.
Intuitively, one should be able to recover such a model from
a bilingual corpus.  For example, if a human translator were to mark
exactly which French words correspond to which English
words, we would be able to count how often a given English word $e$
translates to each French word $f$ and then normalize to get $p(f|e)$, \ie,
$$ p(f|e) = \frac{c(e,f)}{\sum_f c(e,f)} $$
where $c(e,f)$ denotes how often the word $e$ translates to $f$.
However, this word alignment information is not typically included
in a bilingual corpus.

Consider the case where instead of knowing which {\it words\/}
correspond to each other, we just know which {\it sentences\/}
correspond to each other.  Then, it is still possible to
build an approximate model of how frequently words translate
to each other by using an analogous equation to above:
$$ p'(f|e) = \frac{c'(e,f)}{\sum_f c'(e,f)} $$
where $c'(e,f)$ denotes how frequently the words $e$ and $f$ occur
in {\it aligned sentences}.  For words $e$ and $f$ that are
mutual translations, $c'(e, f)$ will be high since every time $e$
occurs in an English sentence $f$ will also occur in the corresponding
French sentence.  For words $e$ and $f$ that are not translations,
while $c'(e, f)$ may be above zero it will most probably not be
very high since it is very unlikely that unrelated words regularly co-occur
in aligned sentences.\footnote{This is not quite accurate.  For example,
the count $c'(e, \mbox{``le''})$ may be high for many English words $e$
just because the word {\it le\/} occurs in most French sentences.
However, this effect can be corrected for, such as described
in Section \ref{ssec:lexcor}.}
Thus, with sentence alignment information we can build an approximate
word translation model $p'(f|e)$.  Furthermore, we can use this
approximate model $p'(f|e)$ to bootstrap the construction of
an even more accurate model of word translation.
Given sentence alignment information and a model $p'(f|e)$ of which
words translate to each other, we can produce a relatively accurate word
alignment.  Then, using the procedure described in the last paragraph
of counting aligned word pairs we
can produce an improved word-to-word translation model.

Hence, sentence alignment is
a useful step in processing a bilingual corpus.
All of the applications mentioned earlier
such as machine translation and sense disambiguation
require bilingual corpora that are sentence-aligned.  As human
translators typically do not include sentence alignment information when
creating bilingual corpora, automatic algorithms for sentence
alignment are immensely useful.

\begin{figure}

{\it $$ \begin{tabular}[t]{ll}
& {\rm English ($\E$)} \\
$E_1$ & \corptxt{Hon.\ members opposite scoff at the freeze suggested
	by this party; to them it is laughable.}
\end{tabular} \rule[-1in]{0.1mm}{1in}
\begin{tabular}[t]{ll}
& {\rm French ($\F$)} \\
$F_1$ & \corptxt{Les d\'{e}put\'{e}s d'en face se moquent du gel que
	a propos\'{e} notre parti.} \\
$F_2$ & \corptxt{Pour eux, c'est une mesure risible.}
\end{tabular} $$}%

\caption{Two-to-one sentence alignment} \label{fig:manyone}
\end{figure}

In this work, we describe an accurate and efficient algorithm
for bilingual sentence alignment.
The task is difficult because sentences frequently do not align
one-to-one.  For example, in Figure \ref{fig:manyone} we
show an example of two-to-one alignment.  In addition,
there are often deletions in one of the supposedly parallel corpora
of a bilingual corpus.  These deletions can be substantial;
in the version of the Canadian Hansard corpus we worked with, there are
many deletions of several thousand sentences and one deletion
of over 90,000 sentences.  Such anomalies are not uncommon in
very large corpora.  Large corpora are often stored as as sizable set of
smaller files, some of which may be accidentally deleted or transposed.

\ssec{Previous Work}

The first sentence alignment algorithms successfully
applied to large bilingual corpora
are those of \newcite{Brown:91a} and Gale and Church
\shortcite{Gale:91a,Gale:93a}.
Brown \etal\ base alignment solely on the number of words in each
sentence; the actual identities of words are ignored.  The general
idea is that the closer in length two sentences are, the more likely
they align.  They construct a probabilistic model of alignment
and select the alignment with the highest probability under this model.
The parameters of their model include
$p(e^a f^b)$, the probability that $a$ English
sentences align with $b$ French sentences, and $p(l_f | l_e)$,
the probability that an English sentence (or sentences) containing $l_e$
words translates to a French sentence (or sentences)
containing $l_f$ words.  These parameters are estimated
statistically from bilingual text.  To search for the
most probable sentence alignment under this alignment model,
dynamic programming \cite{Bellman:57a}, an efficient and exhaustive
search method, is used.

Because dynamic programming
requires time quadratic in the number of sentences to be aligned and
a bilingual corpus can be many millions of sentences, it
is not practical to align a large corpus as a single unit.  The computation
required is drastically reduced if the bilingual corpus can be subdivided
into smaller chunks.  Brown \etal\ use {\it anchors\/} to
perform this subdivision.  An {\it anchor\/} is a piece of text of easily
recognizable form likely to
be present at the same location in both of the parallel corpora of a
bilingual corpus.  For example, Brown \etal\ notice that comments
such as {\it Author~=~Mr.\ Cossitt\/} and {\it Time~=~(1415)\/}
are interspersed in the English text of the Hansard corpus and
corresponding comments are present in the French text.
Dynamic programming is first used to determine which anchors align with
each other, and
then dynamic programming is used again to align the text between anchors.

The Gale and Church algorithm is similar
to the Brown algorithm except that instead
of basing alignment on the number of {\it words\/} in sentences,
alignment is based on the number of {\it characters\/} in sentences.
In addition, instead of using a probabilistic model and searching
for the alignment with the highest probability, they assign lengths
to different alignments and search for the alignment with the smallest
length.\footnote{This can be interpreted as just working
in description space instead of probability space.  As described
in Section \ref{ssec:mdl}, these two spaces are in some sense
equivalent.}
Dynamic programming is again used to search for the best alignment.
Large corpora are assumed to be already subdivided into smaller chunks.

\begin{figure}

{\it $$ \begin{tabular}[t]{ll}
& {\rm English ($\E$)} \\
$E_1$ & \corptxtb{Mr.\ McInnis?} \\
$E_2$ & \corptxtb{Yes.} \\
$E_3$ & \corptxtb{Mr.\ Saunders?} \\
$E_4$ & \corptxtb{No.} \\
$E_5$ & \corptxtb{Mr.\ Cossitt?} \\
$E_6$ & \corptxtb{Yes.}
\end{tabular} \rule[-1in]{0.1mm}{1in}
\begin{tabular}[t]{ll}
& {\rm French ($\F$)} \\
$F_1$ & \corptxtb{M.\ McInnis?} \\
$F_2$ & \corptxtb{Oui.} \\
$F_3$ & \corptxtb{M.\ Saunders?} \\
$F_4$ & \corptxtb{Non.} \\
$F_5$ & \corptxtb{M.\ Cossitt?} \\
$F_6$ & \corptxtb{Oui.}
\end{tabular} $$}%

\caption{A bilingual corpus fragment} \label{tab:frag}
\end{figure}

While these algorithms have achieved remarkably good performance, there
is definite room for improvement.  For example, consider the
excerpt from the Hansard corpus depicted in Figure \ref{tab:frag}.
Length-based algorithms do not particularly favor
aligning {\it Yes\/} with {\it Oui\/} over {\it Non\/} or
aligning {\it Mr.\ McInnis\/}
with {\it M.\ McInnis\/} over {\it M.\ Saunders}.  Thus, such algorithms
can easily misalign passages like these by an even number of
sentences if there are sentences missing in one of the languages.
Constructions in one language that translate to a very different
number of words in the other language may also cause errors.
In general, length-based
algorithms are not robust; they can align unrelated text because
word identities are ignored.

Alignment algorithms that take advantage of
lexical information offer a potential for
higher accuracy.  Previous work includes algorithms by
\newcite{Kay:93a} and \newcite{Catizone:89a}.  Kay and
R\"{o}scheisen perform alignment using a relaxation paradigm.
They keep track of all possible sentence pairs that may align
to each other.  Initially, this set is very large; it is just
constrained by the observation that a sentence in one language
is probably aligned with a sentence in the other language
with the same relative position in the corpus.  For example, an English
sentence halfway through the Hansard English corpus is probably
aligned to a French sentence near the midpoint of the Hansard
French corpus.  Given this set of possible alignment pairs,
word translations are induced based on distributional information.
Using these induced word translations, the set of possible alignment pairs
is pruned, which then yields new word translations, etc.
This process is repeated until convergence.
However, previous lexically-based algorithms have not proved efficient
enough to be suitable for large corpora.  The largest corpus
aligned by Kay and R\"{o}scheisen contains 1,000 sentences in
each language; existing bilingual corpora have many millions of
sentences.

\ssec{Algorithm Overview}

We describe a fast and accurate
algorithm for sentence alignment that
uses lexical information.  Like Brown \etal, we build
a sentence-based translation model and find the alignment
with the highest probability given the model.  However, unlike
Brown \etal\ the translation
model makes use of a word-to-word translation model.
We bootstrap these models using a small amount of pre-aligned text;
the models then refine themselves on the fly during the alignment process.
The search strategy used is dynamic programming with thresholding.  Because
of thresholding, the search is linear in the length of the corpus so that
a corpus need not be subdivided into smaller chunks.

In addition, the search strategy includes a separate mechanism for handling
large deletions in one of the corpora
of a bilingual corpus.  When a deletion is present, thresholding
is not effective and dynamic programming requires time quadratic 
in the length of the deletion to identify its extent, which is
unacceptable for large deletions.  Instead, we have a mechanism
that keys off of rare words to locate the bounds of a deletion in
time linear in the length of the deletion.  This deletion
recovery mechanism can also be used to subdivide a corpus into small
chunks; this enables the parallelization of our algorithm.

%
\sec{The Alignment Model}
%

\ssec{The Alignment Framework}

In this section, we present the general framework of our algorithm,
that of building a probabilistic translation model and
finding the alignment that yields the highest probability under this
model.

More specifically, we try to find the alignment $\A$ with
the highest probability given the bilingual corpus, \ie,
\begin{equation}
\A = \argmax_{\A} p(\A | \E, \F) \label{eqn:abasis}
\end{equation}
where $\E$ and $\F$ denote the English corpus and French corpus,
respectively.  (In this paper, we assume the two languages being
aligned are English and French; however, none of the discussion
is specific to this language pair, except for the discussion
of {\it cognates\/} in Section \ref{ssec:cognate}.)  For now,
we take an alignment $\A$ to be a list of integers
representing which sentence in the French corpus is the first to align
with each successive sentence in the English corpus.  For example,
the alignment $\A = (1, 2, 4, 5, \ldots)$ aligns the first
English sentence with the first French sentence, the
second English sentence with the second and third French
sentences, the third English sentence with the fourth French sentence, etc.

We manipulate equation (\ref{eqn:abasis}) into a more intuitive form:
\begin{eqnarray*}
\A & = & \argmax_{\A} p(\A | \E, \F) \\
	& = & \argmax_{\A} \frac{p(\A, \E, \F)}{\sum_{\A} p(\A, \E, \F)} \\
	& = & \argmax_{\A} p(\A, \E, \F) \\
	& = & \argmax_{\A} p(\A, \F | \E) p(\E) \\
	& = & \argmax_{\A} p(\A, \F | \E)
\end{eqnarray*}
The probability $p(\A, \F|\E)$ represents the probability that
the English corpus $\E$ translates to the French corpus $\F$ with
alignment $\A$.  Making the assumption that successive sentences
translate independently of each other, we can re-express $p(\A, \F | \E)$ as
\begin{equation}
p(\A, \F | \E) = \prod_{i=1}^{l(\E)} p(F_{A_i}^{A_{i+1}-1} | E_i)
	\label{eqn:adecomp}
\end{equation}
where $E_i$ denotes the $i$th sentence in the
English corpus, $F_i^j$ denotes the $i$th through $j$th sentences in the
French corpus, $A_i$ denotes the index of the first French
sentence aligning to the $i$th English sentence in alignment $\A$,
and $l(\E)$ denotes the number of sentences in the English corpus.\footnote{
In this discussion and future discussion, we only consider alignments
$\A$ that are {\it consistent\/} with $\E$ and $\F$.
For example, we do not consider alignments of incorrect length or
alignments that refer to sentences beyond the
end of the French corpus.  Obviously, for inconsistent alignments $\A$
we have $p(\A, \F|\E) = 0$.  Furthermore, in equation (\ref{eqn:adecomp})
$A_{l(\E) + 1}$ is implicitly taken to be $l(\F) + 1$; this enforces
the constraint that the last English sentence aligns with French sentences
ending in the last French sentence.}
We refer to the distribution $p(F_i^j|E)$ as a {\it translation model}, as it
describes how likely an English sentence $E$ translates to
the French sentences $F_i^j$.

\begin{figure}
{\it $$ \begin{tabular}[t]{ll}
& {\rm English ($\E$)} \\
$E_1$ & \corptxt{That is what the consumers are interested in and
	that is what the party is interested in.} \\
$E_2$ & \corptxt{Hon.\ members opposite scoff at the freeze suggested
	by this party; to them it is laughable.}
\end{tabular} \rule[-1.35in]{0.1mm}{1.35in}
\begin{tabular}[t]{ll}
& {\rm French ($\F$)} \\
$F_1$ & \corptxt{Voil\`{a} ce qui int\'{e}resse le consommateur et
	voil\`{a} ce qui int\'{e}resse notre parti.} \\
$F_2$ & \corptxt{Les d\'{e}put\'{e}s d'en face se moquent du gel que
	a propos\'{e} notre parti.} \\
$F_3$ & \corptxt{Pour eux, c'est une mesure risible.}
\end{tabular} $$}%
\caption{A bilingual corpus} \label{tab:corp}
\end{figure}

With an accurate translation model $p(F_i^j|E)$, we can use the relation
$$ \A = \argmax_{\A} p(\A, \F | \E) =
	\argmax_{\A} \prod_{i=1}^{l(\E)} p(F_{A_i}^{A_{i+1}-1} | E_i) $$
to perform accurate sentence alignment.  To give an example,
consider the bilingual corpus $(\E, \F)$ displayed in Figure \ref{tab:corp}.
Now, consider the alignment $\A_1 = (1, 2)$
aligning sentence $E_1$ to sentence $F_1$ and sentence
$E_2$ to sentences $F_2$ and $F_3$.  We have
$$ p(\A_1, \F | \E) = p(F_1 | E_1) p(F_2^3 | E_2), $$
This value should be relatively large, since $F_1$ is a good translation
of $E_1$ and $F_2^3$ is a good translation of $E_2$.
Another possible alignment $\A_2 = (1, 1)$ aligns
sentence $E_1$ to nothing and sentence
$E_2$ to $F_1$, $F_2$, and $F_3$.  We get
$$ p(\A_2, \F | \E) = p(\e | E_1) p(F_1^3 | E_2) $$
This value should be fairly low, as $\e$ is a poor translation of $E_1$
and $F_1^3$ is a poor translation of $E_2$.
Hence, if our translation model $p(F_i^j|E)$ is accurate we will have
$$ p(\A_1, \F | \E) \gg p(\A_2, \F | \E) $$
In general, the more sentences that are mapped to their translations in an
alignment $\A$, the higher the value of $p(\A, \F | \E)$.

However, because our translation model is expressed in terms of a conditional
distribution $p(F_i^j | E)$,
the above framework is not amenable to the situation where
a French sentence corresponds to multiple English sentences.  Hence,
we use a slightly different framework.  We view a bilingual corpus
as a sequence of {\it sentence beads\/} \cite{Brown:91a}, where
a sentence bead corresponds to an irreducible group of sentences that
align with each other.  For example, the correct alignment $\A_1$ of
the bilingual corpus in Figure \ref{tab:corp} consists of the sentence bead
$[E_1, F_1]$ followed by the sentence bead $[E_2, F_2^3]$.
Instead of expressing an alignment $\A$ as a list of sentence indices in
the French corpus, we express an alignment $\A$ as a list
of pair of indices $((A_1^e, A_1^f), (A_2^e, A_2^f), \ldots)$,
the indices representing which English and
French sentence begin each successive sentence bead.  Under this
new convention, we have that $\A_1 = ((1, 1), (2, 2))$.
Unlike the previous framework, this framework is symmetric
and can handle the case where a French sentence aligns with
zero or multiple English sentences.

In this framework, instead of taking
$$ \A = \argmax_{\A} p(\A, \F | \E) =
	\argmax_{\A} \prod_{i=1}^{l(\E)} p(F_{A_i}^{A_{i+1}-1} | E_i) $$
we take
\begin{equation}
\A = \argmax_{\A} p(\A, \E, \F) = \argmax_{\A} p\su{A-len}(l(\A))
	\prod_{i=1}^{l(\A)} p([E_{A_i^e}^{A_{i+1}^e-1},
		F_{A_i^f}^{A_{i+1}^f-1}]) \label{eqn:akey}
\end{equation}
where $l(\A)$ denotes the number of sentence beads in $\A$ and
$p\su{A-len}(l)$ denotes the probability that an alignment contains exactly $l$
sentence beads.  The term $p\su{A-len}(l(\A))$ is necessary for normalization
purposes; otherwise we would not have
$\sum_{\A, \E, \F} p(\A, \E, \F) = 1$.\footnote{To show this,
notice that for any $l$ we have that
$$ \sum_{x_1, \ldots, x_l} \prod_{i=1}^l p(x_i) =
	\sum_{x_1} p(x_1) \sum_{x_2} p(x_2) \cdots \sum_{x_l} p(x_l) =
	1 \cdot 1 \cdot \cdots \cdot 1 = 1 $$
Applying this relation to equation (\ref{eqn:akey}), we have that
\begin{eqnarray*}
\sum_{\A, \E, \F} p(\A, \E, \F)
	& = & \sum_{\A, \E, \F} p\su{A-len}(l(\A)) \prod_{i=1}^{l(\A)}
		p([E_{A_i^e}^{A_{i+1}^e-1}, F_{A_i^f}^{A_{i+1}^f-1}]) \\
	& = & \sum_{l} p\su{A-len}(l) \sum_{l(\A) = l, \E, \F} \prod_{i=1}^{l}
		p([E_{A_i^e}^{A_{i+1}^e-1}, F_{A_i^f}^{A_{i+1}^f-1}]) \\
	& = & \sum_{l} p\su{A-len}(l) \\
	& = & 1
\end{eqnarray*}
Without the term $p\su{A-len}(l(\A))$ the sum is infinite.}
Instead of having a conditional translation model $p(F_i^j | E)$,
our translation model is now expressed as a distribution
$p([E_i^j, F_k^l])$ representing
the frequencies of sentence beads $[E_i^j, F_k^l]$.

\ssec{The Basic Translation Model} \label{ssec:btm}

As we can see from equation (\ref{eqn:akey}), the key to accurate
alignment in our framework is coming up with an accurate
translation model $p([E_i^j, F_k^l])$.  For this translation model,
we desire the simplest model that incorporates lexical information
effectively.  We describe our model in terms of a series
of increasingly complex models.  In this section, we
consider only sentence beads $[E, F]$ containing a single
English sentence $E = e_1 \cdots e_{l(E)}$ and single French sentence
$F = f_1 \cdots f_{l(F)}$.  As a starting point, consider
a model that assumes that all individual words are independent, \ie,
a model where the probability of some text is the product of
the probabilities of each word in the text.  More specifically, we can take
$$ p'([E, F]) = p\su{e-len}(l(E)) p\su{f-len}(l(F))
	\prod_{i=1}^{l(E)} p_e(e_i) \prod_{j=1}^{l(F)} p_f(f_j) $$
where $p\su{e-len}(l)$ is the probability that an English sentence is
$l$ words long,
$p\su{f-len}(l)$ is the probability that a French sentence is $l$ words long,
$p_e(e_i)$ is the frequency of the word $e_i$ in English, and
$p_f(f_j)$ is the frequency of the word $f_j$ in French.  The terms
$p\su{e-len}(l(E))$ and $p\su{f-len}(l(F))$ are necessary
to make $\sum_{E,F} p'([E, F]) = 1$ (as in
equation (\ref{eqn:akey})).\footnote{This model can
be considered somewhat similar to the alignment model used by Brown \etal.
As the probability of words are taken to be independent, these
probabilities do not depend on sentence alignment and can be
ignored, as in Brown.  However, unlike Brown we also take sentence
lengths to be independent; a model closer to Brown would express
sentence lengths as a joint probability $p\su{len}(l(E), l(F))$
that assigned higher probabilities to sentence pairs with similar lengths.
} For example, we have that
\begin{eqnarray*}
\lefteqn{p'([\t{do you speak French}, \t{parlez-vous fran\c{c}ais}]) = } \\
	& & p\su{e-len}(4) p\su{f-len}(3)
	p_e(\t{do}) p_e(\t{you}) p_e(\t{speak}) p_e(\t{French})
	p_f(\t{parlez}) p_f(\t{vous}) p_f(\t{fran\c{c}ais})
\end{eqnarray*}
Clearly, this is a poor translation model because it takes the
English sentence and French sentence to be independent,
so it does not assign higher probabilities
to sentence pairs that are translations.  For example,
it would assign about the same probabilities to the following two
sentence beads:
$$\begin{array}{l}
{[} \t{do you speak French}, \t{parlez-vous fran\c{c}ais} {]} \\
{[} \t{did they eat German}, \t{parlez-vous fran\c{c}ais} {]}
\end{array}$$

To capture the dependence between individual English words and
individual French words, we assign probabilities to word pairs
in addition to just single words.  For two words $e$ and $f$ that are mutual
translations, instead of having the two terms $p_e(e)$ and $p_f(f)$ in the
above equation we would like a single term $p(e, f)$ that is
substantially larger than $p_e(e)p_f(f)$.  To this end, we introduce the
concept of a {\it word bead}.  A word bead is either a single English
word, a single French word, or a single English word and a single
French word.  We refer to these as 1:0, 0:1, and 1:1 word beads, respectively.
Instead of modeling a pair of sentences as a list of independent words,
we model sentences as a list of word beads,
using the 1:1 word beads to capture the dependence
between English and French words.  To address the issue that corresponding
English and French words may not occur in identical positions in
their respective sentences,
we abstract over the positions of words in sentences;
we consider sentences to be unordered multisets\footnote{A
{\it multiset\/} is a set in which a given element can occur more than once.}
of words, so that 1:1 word beads can pair words from arbitrary positions
in sentences.

As a first cut at this behavior, consider the following ``model'':
$$ p''(\bv) = p\su{b-len}(l(\bv)) \prod_{i=1}^{l(\bv)} p_b(b_i) $$
where $\bv = \{ b_1, \ldots, b_{l(\bv)} \}$ is a multiset of word beads,
$p\su{b-len}(l)$ is the probability that an English
sentence and a French sentence contain $l$ word beads, and $p_b(b_i)$
denotes the frequency of the word bead $b_i$.
This simple model captures lexical dependencies between
English and French sentences.  For example, we might express
the sentence bead
$$ [\t{do you speak French}, \t{parlez-vous fran\c{c}ais}] $$
with the word beading
$$ \bv = \{ [\t{do}], [\t{you}, \t{vous}], [\t{speak}, \t{parlez}],
	[\t{French}, \t{fran\c{c}ais}] \} $$
with probability
$$ p''(\bv) = p\su{b-len}(4)
	p_b([\t{do}]) p_b([\t{you}, \t{vous}]) p_b([\t{speak}, \t{parlez}])
	p_b([\t{French}, \t{fran\c{c}ais}]) $$
If $p_b([e,f]) \gg p_b([e]) p_b([f])$ for words $e$ and $f$ that are
mutual translations, then word beadings of sentence pairs that
contain many words that are mutual translations can have much
higher probability than word beadings of unrelated sentence pairs.

However, this ``model'' $p''(\bv)$ does not satisfy the constraint
that $\sum_{\bv} p''(\bv) = 1$.  To see this, consider the case that
the ordering of beads is significant so that instead of having
a multiset $\bv = \{ b_1, \ldots, b_{l(\bv)} \}$, we have a list
$\bvv = ( b_1, \ldots, b_{l(\bvv)} )$.  For this case, we have
$\sum_{\bvv} p''(\bvv) = 1$ (as for equation (\ref{eqn:akey})).
Then, because our beadings $\bv$ are actually unordered,
multiple terms in this last sum that are just different orderings of
the same multiset will be collapsed to a single term in our actual
summation.  Hence, the true $\sum_{\bv} p''(\bv)$ will be substantially
less than one.  To force this model to sum to one, we simply
normalize to retain the qualitative aspects of the model.  We take
$$ p(\bv) = \frac{p\su{b-len}(l(\bv))}{N_{l(\bv)}}
	\prod_{i=1}^{l(\bv)} p_b(b_i) $$
where
\begin{equation}
N_l = \sum_{l(\bv) = l} \prod_{i=1}^l p_b(b_i) \label{eqn:aefnorm}
\end{equation}

To derive the probabilities of sentence beads $p([E, F])$ from
the probabilities of word beadings $p(\bv)$, we need to consider
the issue of word ordering.  A beading $\bv$ describes an
{\it unordered\/} multiset of English and French words, while sentences
are {\it ordered\/} sequences of words.
We need to model word ordering, and ideally the probability of
a sentence bead should depend on the ordering of its component
words.  For example, the sentence {\it John ate Fido\/} should
have a higher probability of aligning with the sentence
{\it Jean a mang\'{e} Fido\/} than with the sentence
{\it Fido a mang\'{e} Jean}.  However, modeling how
word order mutates under translation is
notoriously difficult \cite{Brown:91g},
and it is unclear how much improvement in accuracy an accurate model of
word order would provide.  Hence, we ignore this issue and
take all word orderings to be equiprobable.  Let $O(E)$ denote
the number of distinct ways of ordering the words in a sentence $E$.\footnote{
For a sentence $E$ containing words that are all distinct, we just
have $O(E) = l(E)!$.  More generally, we have that
$O(E) = {{l(E)}\choose{\{m(e_i)\}}}$, where $\{m(e_i)\}$ denotes
the multiplicities of the different words $e_i$ in the sentence.}
Then, we take
\begin{equation}
p([E, F] | \bv) = \frac{1}{O(E) O(F)} \label{eqn:aorder}
\end{equation}
for those sentence beads $[E, F]$ {\it consistent\/} with $\bv$, \ie,
those sentence beads containing the same words as $\bv$.  (For inconsistent
beads $[E, F]$, we have $p([E, F] | \bv) = 0$.)

This gives us
$$ p([E, F], \bv) = p(\bv) p([E, F] | \bv) =
	\frac{p\su{b-len}(l(\bv))}{N_{l(\bv)} O(E) O(F)}
	\prod_{i=1}^{l(\bv)} p_b(b_i) $$
To get the total probability $p([E, F])$ of a sentence bead, we
need to sum over all beadings $\bv$ consistent with $[E, F]$, giving us
\begin{equation}
p([E, F]) = \sum_{\bv \sim [E, F]} p([E, F], \bv) =
	\sum_{\bv \sim [E, F]} \frac{p\su{b-len}(l(\bv))}{N_{l(\bv)} O(E) O(F)}
	\prod_{i=1}^{l(\bv)} p_b(b_i) \label{eqn:aprobefa}
\end{equation}
where $\bv \sim [E, F]$ denotes $\bv$ being consistent with $[E, F]$.

\ssec{The Complete Translation Model} \label{ssec:ctm}

In this section, we extend the translation model to other types
of sentence beads besides beads that contain a single
English and French sentence.  Like Brown \etal, we only consider sentence
beads consisting of one English sentence, one French sentence, one
English sentence and one French sentence, two English sentences and
one French sentence, and one English sentence and two French sentences.
We refer to these as 1:0, 0:1, 1:1, 2:1, and 1:2 sentence beads,
respectively.

For 1:1 sentence beads, we take
\begin{equation}
p([E, F]) = p(1:1) \sum_{\bv \sim [E, F]}
	\frac{p_{1:1}(l(\bv))}{N_{l(\bv)}^{1:1} O(E) O(F)}
	\prod_{i=1}^{l(\bv)} p_b(b_i) \label{eqn:aprobef}
\end{equation}
This differs from equation (\ref{eqn:aprobefa}) in that we have added
the term $p(1:1)$ representing the probability or frequency of 1:1
sentence beads; this term is necessary for normalization purposes
now that we consider other types of sentence beads.  In addition,
we now refer to $p\su{b-len}$ as $p_{1:1}$ and to $N_l$ as $N_l^{1:1}$
as there will be analogous terms for the other types of sentence beads.

To model 1:0 sentence beads, we use a similar equation except that
we only need to consider 1:0 word beads (\ie, individual English words),
and we do not need to sum over beadings since there is only one word beading
consistent with a 1:0 sentence bead.  We take
\begin{equation}
p([E]) = p(1:0) \frac{p_{1:0}(l(E))}{N_{l(E)}^{1:0} O(E)}
	\prod_{i=1}^{l(E)} p_e(e_i) \label{eqn:aprobe}
\end{equation}
Instead of the distribution $p_b(b_i)$ of word bead frequencies,
we have the distribution $p_e(e_i)$ of English word frequencies.
We use an analogous equation for 0:1 sentence beads.

For 2:1 sentence beads, we take
\begin{equation}
p([E_i^{i+1}, F]) = p(2:1) \sum_{\bv \sim [E_i^{i+1}, F]}
	\frac{p_{2:1}(l(\bv))}{N_{l(\bv)}^{2:1} O(E_i) O(E_{i+1}) O(F)}
	\prod_{i=1}^{l(\bv)} p_b(b_i) \label{eqn:aprobeef}
\end{equation}
We use an analogous equation for 1:2 sentence beads.\footnote{To be
more consistent with 1:1 sentence beads, in equation (\ref{eqn:aprobeef})
instead of the expression $O(E_i) O(E_{i+1})$ we should have the
expression $O(E_i^{i+1}) (l+1)$ where $l = l(E_i) + l(E_{i+1})$.  There
are $O(E_i^{i+1})$ different ways to order the $l$ English words in $\bv$,
and there are $l+1$ different places to divide the list
of $l$ English words into two sentences (assuming we allow sentences
of length zero).  Thus, instead of equation
(\ref{eqn:aorder}) as for 1:1 sentence beads, we have
$$ p([E_i^{i+1}, F] | \bv) = \frac{1}{O(E_i^{i+1}) O(F) (l+1)} $$
if we take all of these possibilities to be equiprobable.  However,
we choose the expression $O(E_i) O(E_{i+1})$ because then the
contribution of this word ordering factor is independent of alignment and can
be ignored.  Mathematically, we account for this choice
through the normalization constants $N_l^{2:1}$; see
Section \ref{ssec:anorm}.}

%
\sec{Implementation}
%

In this section, we describe our implementation of the alignment
algorithm.  We describe how we train the parameters or
probabilities associated with the translation model, and
how we perform the search for the best alignment.  In addition,
we describe approximations that we use to make the algorithm
computationally tractable.  We hypothesize that because our translation model
incorporates lexical information strongly,
correct alignments are tremendously more probable than incorrect
alignments so that moderate errors in calculation will
not greatly affect results.

\ssec{Evaluating the Probability of a Sentence Bead} \label{ssec:aevalsb}

The probability of a 0:1 or 1:0 sentence bead can be
calculated efficiently using equation (\ref{eqn:aprobe})
in Section \ref{ssec:ctm}.  To evaluate the
probabilities of other types of sentence beads exactly requires
a sum over a vast number of possible word beadings.
We make the gross approximation that
this sum is roughly equal to the maximum term in the sum.  For
example, for 1:1 sentence beads we have
\begin{eqnarray*}
p([E,F]) & = & p(1:1) \sum_{\bv \sim [E, F]}
	\frac{p_{1:1}(l(\bv))}{N_{l(\bv)}^{1:1} O(E) O(F)}
	\prod_{i=1}^{l(\bv)} p_b(b_i) \\
	& \approx & p(1:1) \max_{\bv \sim [E, F]} \{
	\frac{p_{1:1}(l(\bv))}{N_{l(\bv)}^{1:1} O(E) O(F)}
	\prod_{i=1}^{l(\bv)} p_b(b_i) \}
\end{eqnarray*}

Even with this approximation, the calculation of $p([E,F])$ is
still expensive since it requires a search for the most probable beading.
We use a greedy heuristic to perform this search; the heuristic is not
guaranteed to find the most probable beading.
We begin with every word in its own bead.  We then find the 0:1 bead
and 1:0 bead that, when replaced with a 1:1 word bead, results in the
greatest increase in the probability of the beading.
We repeat this process until
we can no longer find a 0:1 and 1:0 bead pair that when replaced
would increase the beading's probability.

For example, consider the sentence bead
$$ [\t{do you speak French}, \t{parlez-vous fran\c{c}ais}] $$
To search for the most probable word beading of this sentence bead,
we begin with each word in its own word bead:
$$ \bv = \{ [\t{do}], [\t{you}], [\t{speak}], [\t{French}],
	[\t{parlez}], [\t{vous}], [\t{fran\c{c}ais}] \} $$
Then, we find the pair of word beads that when replaced with a 1:1 bead
results in the largest increase in $p(\bv)$; suppose this pair
is $[\t{French}]$ and $[\t{fran\c{c}ais}]$.  We substitute in the 1:1 bead
yielding
$$ \bv = \{ [\t{do}], [\t{you}], [\t{speak}], [\t{French}, \t{fran\c{c}ais}],
	[\t{parlez}], [\t{vous}] \} $$
We repeat this process until there are no more pairs that are profitable
to replace; this is apt to yield a beading such as
$$ \bv = \{ [\t{do}], [\t{you}, \t{vous}], [\t{speak}, \t{parlez}],
	[\t{French}, \t{fran\c{c}ais}] \} $$

This greedy search for the most probable beading
can be performed in time roughly linear in
the number of words in the involved sentences, as long as
the probability distribution $p_b([e,f])$ is fairly sparse,
that is, as long as
for most words $e$, there are few words $f$ such that $p_b([e, f]) > 0$.
To perform this search efficiently, for each English word $e$
we maintain a list of all French words $f$ such that $p_b([e, f]) > 0$.
In addition, we maintain a list for each French word
storing information about the current sentence.  (For
expository purposes, we assume the sentence bead contains a single
English and French sentence.)  Then, for a given sentence bead we
do as follows:
\bi
\i For each word $e$ in the English sentence, we take all associated
beads $[e, f]$ with $p_b([e, f]) > 0$ and append these beads to
the list associated with the word $f$.
\i We take the lists associated with each word $f$ in the French sentence,
and merge them into a single list.  We sort the resulting list according to
the increase in probability associated with each bead if substituted
into the beading.
\ei
With this procedure, we can find all applicable beads $[e, f]$ with nonzero
probability in near-linear time in sentence length.
With the sorted list of beads, performing
the greedy search efficiently is fairly straightforward.

\ssec{Normalization} \label{ssec:anorm}

The exact evaluation of the normalization constants $N_l$ is
very expensive.  For example, for 1:0 sentence beads we have that
$$ N_l^{1:0} =
	\sum_{\bar{e} = \{ e_1, \ldots, e_l \}} \prod_{i=1}^l p_e(e_i) $$
This is identical to the normalization for 1:1 sentence beads
given in equation (\ref{eqn:aefnorm}), except that we restrict word beads
to be single English words as 1:0 sentence beads only contain English.
It is impractical to sum over all sets of words $\{ e_1, \ldots, e_l \}$.
Furthermore, we continually re-estimate the parameters $p_e(e_i)$
during the alignment process, so the exact value of $N_l^{1:0}$
is constantly changing.  Hence, we only approximate
the normalization constants $N_l$.

Let us first consider the constants $N_l^{1:0}$.
Notice that when we sum over ordered {\it lists\/} $\vec{e}$
instead of unordered {\it sets\/} $\bar{e}$, we have the relation
$$ \sum_{\vec{e} = (e_1, \ldots, e_l)} \prod_{i=1}^l p_e(e_i) = 1 $$
Let $O(\bar{e})$ be the number of distinct orderings
of the elements in the (multi-)set $\bar{e} = \{b_1, \ldots, b_l\}$;
this is equal to the number of different lists $\vec{e}$ that can be
formed using all of the words in $\bar{e}$.  Then, we have
$$ \sum_{\bar{e} = \{e_1, \ldots, e_l\}} O(\bar{e})
	\prod_{i=1}^l p_e(e_i) = 1 $$
We make the approximation that $O(\bar{e}) = l!$ for all $\bar{e}$.
This approximation is exact for all sets $E$ containing no duplicate words.
This gives us
$$ N_l^{1:0}
	= \sum_{\bar{e} = \{ e_1, \ldots, e_l \}} \prod_{i=1}^l p_e(e_i)
	= \frac{1}{l!} \sum_{\bar{e} = \{ e_1, \ldots, e_l \}} l!
		\prod_{i=1}^l p_e(e_i)
	\approx \frac{1}{l!} \sum_{\bar{e} = \{ e_1, \ldots, e_l \}} O(\bar{e})
		\prod_{i=1}^l p_e(e_i)
	= \frac{1}{l!} $$ 
We use analogous approximations for $N_l^{0:1}$ and $N_l^{1:1}$.

Now, let us consider the constants $N_l^{2:1}$.  These need to
be calculated differently from the above constants.  To show this,
we contrast the 2:1 sentence bead case with the 1:1 sentence bead case given
in Section \ref{ssec:btm}.  For 1:1 sentence beads, we have
$$ p([E, F] | \bv) = \frac{1}{O(E) O(F)} $$
and this results in the expression $O(E) O(F)$ in equation (\ref{eqn:aprobef}).
The analogous expression for 2:1 sentence beads in
equation (\ref{eqn:aprobeef}) is $O(E_i) O(E_{i+1}) O(F)$.  However,
the distribution
$$ p([E_i^{i+1}, F] | \bv) = \frac{1}{O(E_i) O(E_{i+1}) O(F)} $$
is not proper in that $\sum_{E_i^{i+1}, F} p([E_i^{i+1}, F] | \bv) \neq 1$;
as described in Section \ref{ssec:ctm}, the expression
$O(E_i) O(E_{i+1}) O(F)$ in not equal to
the number of different 2:1 sentence beads corresponding to $\bv$.
Thus, the derivation for 1:1 sentence beads does not hold for 2:1 sentence
beads and we need to calculate the normalization constants differently.

Instead, we choose $N_l^{2:1}$ so that the probabilities
$p([E_i^{i+1}, F], \bv)$ sum correctly.  In particular,
we want to choose $N_l^{2:1}$ such that
$$ \sum_{E_i^{i+1}, F, l(\bv) = l} p([E_i^{i+1}, F], \bv) = p(2:1)p_{2:1}(l) $$
as $p(2:1) p_{2:1}(l)$ is the amount of probability allocated by the model
for word beadings of length $l$ of 2:1 sentence beads.  Substituting
in equation (\ref{eqn:aprobeef}), we get
$$ \sum_{\bv \sim [E_i^{i+1}, F], l(\bv) = l}
	p(2:1) \frac{p_{2:1}(l)}{N_l^{2:1} O(E_i) O(E_{i+1}) O(F)}
	\prod_{i=1}^l p_b(b_i) = p(2:1)p_{2:1}(l) $$
Rearranging, we get
$$ N_l^{2:1} = \sum_{\bv \sim [E_i^{i+1}, F], l(\bv) = l}
	\frac{1}{O(E_i) O(E_{i+1}) O(F)} \prod_{i=1}^l p_b(b_i) $$

Now, we re-express the sum over $E_i$, $E_{i+1}$, and $F$ as a sum over
unordered sets $\bar{e}_i = \{ e_1, \ldots, e_{l(E_i)} \}$,
$\bar{e}_{i+1} = \{ e'_1, \ldots, e'_{l(E_{i+1})} \}$,
and $\bar{f} = \{ f_1, \ldots, f_{l(F)} \}$, giving us
$$ N_l^{2:1}
	= \sum_{\bv \sim [\bar{e}_i, \bar{e}_{i+1}, \bar{f}], l(\bv) = l}
		\frac{O(\bar{e}_i) O(\bar{e}_{i+1}) O(\bar{f})}{O(\bar{e}_i)
		O(\bar{e}_{i+1}) O(\bar{f})} \prod_{i=1}^l p_b(b_i)
	= \sum_{\bv \sim [\bar{e}_i, \bar{e}_{i+1}, \bar{f}], l(\bv) = l}
		\prod_{i=1}^l p_b(b_i) $$
Then, let us consider how many different $[\bar{e}_i, \bar{e}_{i+1},
\bar{f}]$ are consistent with a given word beading $\bv$.  There
is only a single way to allocate the French words in $\bv$ to
get $\bar{f}$; however, there are many ways of dividing the English
words in $\bv$ to get $\bar{e}_i$ and $\bar{e}_{i+1}$.  In particular,
there are $2^{n_e(\bv)}$ ways of doing this, where $n_e(\bv)$ denotes
the number of English words in $\bv$.  Each of the $n_e(\bv)$ English
words in $\bv$ can be placed in either of the two English sets.  Using
this observation, we have that
$$ N_l^{2:1} = \sum_{l(\bv) = l} 2^{n_e(\bv)} \prod_{i=1}^l p_b(b_i) $$

To evaluate this equation, we use several approximations.
The first approximation we make is that $p_b(b_i)$ is a uniform
distribution, \ie, $p_b(b) = \frac{1}{B}$ for all $b$ where $B$
is the total number of different word beads.  This gives us
$$ N_l^{2:1} \approx \sum_{l(\bv) = l} 2^{n_e(\bv)} \prod_{i=1}^l \frac{1}{B}
	= \frac{1}{B^l} \sum_{l(\bv) = l} 2^{n_e(\bv)} $$
Then, let $b_l(n)$ be the number of bead sets $\bv$ of size $l$ containing
exactly $n$ English words.  We can rewrite the preceding sum as
\begin{equation}
N_l^{2:1} \approx \frac{1}{B_l} \sum_{n=0}^l b_l(n) 2^n \label{eqn:anormeefa}
\end{equation}

To approximate $b_l(n)$, let $B_{+e}$ be the number of
different word beads containing English words (\ie, 1:0 and 1:1 beads), and
let $B_{-e}$ be the number of different word beads not containing
English words (\ie, 0:1 beads), so that
$B = B_{+e} + B_{-e}$.  Notice that the number of bead {\it lists}
(as opposed to sets) $b'_l(n)$ of length $l$ containing $n$ English
words is
$$ b'_l(n) = {l\choose n} B_{+e}^n\ B_{-e}^{l-n} $$
To estimate the number of bead {\it sets\/} $b_l(n)$, we use the same
approximation used in calculating $N_l^{1:0}$ and simply divide by $l!$.
This gives us
$$ b_l(n) \approx \frac{1}{l!} {l\choose n} B_{+e}^n\ B_{-e}^{l-n} $$
Substituting this into equation (\ref{eqn:anormeefa}), we get
$$ N_l^{2:1} \approx \frac{1}{B^l l!} \sum_{n=0}^l {l\choose n}
		B_{+e}^n B_{-e}^{l-n} 2^n
	= \frac{1}{B^l l!} \sum_{n=0}^l {l\choose n}
		(2B_{+e})^n B_{-e}^{l-n} $$
Using the binomial identity $(x+y)^l = \sum_{n=0}^l {l\choose n} x^n y^{l-n}$,
we get
$$ N_l^{2:1} \approx \frac{1}{B^l l!}\ (2 B_{+e} + B_{-e})^l
	= \frac{1}{l!}\ (\frac{2 B_{+e} + B_{-e}}{B})^l
	= \frac{1}{l!}\ (\frac{B_{+e} + B}{B})^l
	= \frac{(1 + \frac{B_{+e}}{B})^l}{l!} $$
We use an analogous approximation for $N_l^{1:2}$.

\ssec{Parameterization}

To model the parameters $p\su{A-len}(L)$ in equation (\ref{eqn:akey})
representing the probability that
a bilingual corpus is $L$ sentence beads in length, we
assume a uniform distribution;\footnote{To be precise, we assume
a uniform distribution over some arbitrarily large finite range,
as one cannot have a uniform distribution over a countably infinite set.}
it is unclear what {\it a priori\/} information we have on the
length of a corpus.
This allows us to ignore the term, since this length will not affect the
probability of an alignment.

We model sentence length (in beads) using a Poisson distribution, \ie,
\begin{equation}
p_{1:0}(l) = \frac{\l_{1:0}^l}{l!\ e^{\l_{1:0}}} \label{eqn:asentlen}
\end{equation}
for some $\l_{1:0}$, and we have analogous equations
for the other types of sentence beads.
To prevent the possibility of some of the $\l$'s being assigned
unnaturally small or large values during the training process
to specifically model very short or very long sentences,
we tie together the $\l$ values for the
different types of sentence beads.  We take
\begin{equation}
\l_{1:0} = \l_{0:1} = \frac{\l_{1:1}}{2} = \frac{\l_{2:1}}{3}
	= \frac{\l_{1:2}}{3} \label{eqn:lambda}
\end{equation}

In modeling the frequency of word beads, there are
three distinct distributions we need to model: the distribution $p_e(e_i)$
of 1:0 word beads in 1:0 sentence beads, the distribution $p_f(f_i)$
of 0:1 word beads in 0:1 sentence beads, and the
distribution of all word beads $p_b(b_i)$ in
1:1, 2:1, and 1:2 sentence beads.\footnote{Conceivably, we could
consider using three different distributions $p_b(b_i)$ for
1:1, 2:1, and 1:2 sentence beads.  However, we assume these
distributions are identical to reduce the number of parameters.}
To reduce the number of independent
parameters we need to estimate, we tie these distributions together.
We take $p_e(e_i)$ and $p_f(f_i)$ to be identical to $p_b(b_i)$,
except restricted to the relevant subset of word beads and normalized
appropriately, \ie,
$$ p_e(e) = \frac{p_b([e])}{ \sum_{e} p_b([e]) } $$
and
$$ p_f(f) = \frac{p_b([f])}{ \sum_{f} p_b([f]) } $$
where $[e]$ and $[f]$ denote 1:0 and 0:1 word beads,
respectively.

To further reduce the number of parameters, we convert all words
to lowercase.  For example, we consider the words {\it May\/} and {\it may\/}
to be identical.

\ssec{Parameter Estimation Framework} \label{ssec:aparesta}

The basic method we use for estimating the parameters or probabilities
of our model is to just take counts on previously aligned data and
to normalize.  For example,
to estimate $p(1:0)$, the probability or frequency of 1:0 sentence
beads, we count the total number of 1:0 sentence beads in
previously aligned data and divide by the total number of sentence beads.
To bootstrap the model, we first take counts on a small amount of data that
has been aligned by hand or by some other algorithm.  Once the
model has been bootstrapped, it can align sentences by itself,
and we can take counts on the data already aligned by the algorithm to improve
the parameter estimates for aligning future data.
For the Hansard corpus, we have found that one hundred sentence pairs
are sufficient to bootstrap the alignment model.

This method can be considered to be a
variation of the Viterbi version of the
{\it expectation-maximization\/} (EM) algorithm \cite{Dempster:77a}.
In the EM algorithm, an {\it expectation\/} phase, where counts on the corpus
are taken using the current estimates of the parameters, is alternated
with a {\it maximization\/} phase, where parameters are re-estimated
based on the counts just taken.  Improved parameters lead to
improved counts, which lead to even more accurate parameters.
In the incremental version of the EM algorithm we use, instead
of re-estimating parameters after each complete pass through the
corpus, we re-estimate parameters after each sentence.  By
re-estimating parameters continually as we take counts on the
corpus, we can align later sections of the corpus more reliably
based on the alignment of earlier sections.  We can align a corpus
with only a single pass, simultaneously producing alignments and
updating the model as we proceed.

However, to align a corpus in a single pass, the model must be
fairly accurate before starting or else the beginning of the corpus
will be poorly aligned.  Hence, after bootstrapping the translation model
on one hundred sentence pairs and before starting the one pass through
the entire corpus to produce our final alignment, we first
refine the translation model by using the algorithm to align a chunk
of the unaligned target bilingual corpus.
In experiments with the Hansard corpus, we
train on 20,000 unaligned sentence pairs before performing the final
alignment.

Because the search algorithm considers many partial
alignments simultaneously, it is not obvious how to determine
when it is certain that a particular sentence bead will be part of the
final alignment and thus can be trained on.  To elaborate, our
search algorithm maintains a set of partial alignments,
each partial alignment representing a possible alignment between some
prefix of the English corpus and some prefix of the French corpus.
These hypothesis alignments are extended incrementally during the
search process.  To address the problem of determining which
sentence beads can be trained on,
we keep track of the longest partial alignment common to all
partial alignments currently being considered.  It is assured that
this common partial alignment will be part of the final alignment.
Hence, whenever a sentence bead is added to this common alignment we use
it to train on.  The point
in the corpus at the end of this common alignment is called
the {\it confluence point}.

\ssec{Parameter Estimation Details} \label{ssec:aparestb}

One issue with the framework described in the last section
is that in a straightforward implementation,
probabilities that are initialized to zero will remain zero during
the training process.  An object with zero probability will never
occur in an alignment, and thus it will never receive any counts.
The probability of an object is just its count normalized, so
such an object will always have probability zero.  Thus, it
is important to initialize all probabilities to nonzero values.
Unless otherwise specified, we achieve this by setting all
counts initially to 1.

We now describe in detail how we estimate specific parameters.
To estimate word bead frequencies $p_b(b)$,
we maintain a count $c(b)$ for each word bead $b$ reflecting
the number of times that word bead has occurred in the most probable
word beading of a sentence bead.  More specifically, given
some aligned data to train on, we first find the most
probable word beading of each sentence bead in the alignment using
the current model, and we then use these most probable word
beadings to update word bead counts.  We take
$$ p_b(b) = \frac{c(b)}{\sum_{b} c(b)} $$
For 0:1 and 1:0 word beads, we initialize the counts $c(b)$ to 1.
For 1:1 word beads, we initialize these counts to zero;
this is because our algorithm for
searching for the most probable word beading of a sentence bead
will not be efficient unless $p_b([e, f])$ is sparse, as described in
Section \ref{ssec:aevalsb}.  Instead, we use a heuristic
for initializing particular $c([e, f])$ to nonzero values during the
training process; whenever we see a 0:1 and a 1:0 word bead occur
in the most probable beading of a sentence bead, we initialize
the count of the corresponding 1:1 word bead to a small value.\footnote{
The particular value we use is $\frac{n_e + n_f}{2 n_e n_f}$,
where $n_e$ denotes the number of 1:0 word beads in the beading,
and $n_f$ denotes the number of 0:1 word beads.  This can be
thought of as dividing $\frac{n_e + n_f}{2}$ counts evenly among all
$n_e n_f$ bead pairs.}  This heuristic is effective for
constraining the number of 1:1 word beads with nonzero probability.

To estimate the sentence length parameters $\l$, we divide the
number of word beads in the most probable beadings of the previously
aligned sentences by the total number of sentences.  This
gives us the mean number of word beads per sentence.  In a
Poisson distribution, the mean coincides with the value of the $\l$ parameter.
(Recall that we model sentence length with a Poisson distribution
as in equation (\ref{eqn:asentlen}).)
We take $\l_{1:0}$ to be this mean value,
and the other $\l$ parameters
can be calculated using equation (\ref{eqn:lambda}).
For the situation before we have any counts,
we set $\l_{1:0}$ to an arbitrary constant; we chose the value 7.

To estimate the probabilities $p(1:0)$, $p(0:1)$, $p(1:1)$, $p(2:1)$,
and $p(1:2)$ of each type of sentence bead,
we count the number of times each type of bead occurred in the
previously aligned data and divide by the total number of sentence beads.
These counts are initialized to 1.

\ssec{Cognates} \label{ssec:cognate}

There are many words that possess the same spelling in two different
languages.  For example, punctuation, numbers, and proper names generally have
the same spellings in English and French.
Such words are members of a class called {\it cognates} \cite{Simard:92a}.
Because identically spelled words can be recognized automatically and
are frequently translations of each other,
it is sensible to use this {\it a priori\/} information in
initializing word bead frequencies.  To this end,
we initialize to 1 the count of all 1:1 word beads that contain
words that are spelled identically.

\ssec{Search} \label{ssec:asearch}

It is natural to use dynamic programming to search for the best
alignment; one can find the most probable of an exponential number
of alignments using quadratic time and memory.
Using the same perspective as \newcite{Gale:93a},
we view alignment as a ``shortest path'' problem.
Recall that we try to find the alignment $\A$ such that
$$ \A = \argmax_{\A} p(\A, \E, \F) = \argmax_{\A} p\su{A-len}(l(\A))
	\prod_{i=1}^{l(\A)} p([E_{A_i^e}^{A_{i+1}^e-1},
		F_{A_i^f}^{A_{i+1}^f-1}]) $$
Manipulating this equation, we get
\begin{eqnarray*}
\A & = & \argmax_{\A} p(\A, \E, \F) \\
	& = & \argmin_{\A} -\ln p(\A, \E, \F) \\
	& = & \argmin_{\A} -\ln \{ p\su{A-len}(l(\A))
		\prod_{i=1}^{l(\A)} p([E_{A_i^e}^{A_{i+1}^e-1},
		F_{A_i^f}^{A_{i+1}^f-1}]) \} \\
	& = & \argmin_{\A} \{ -\ln p\su{A-len}(l(\A)) +
		\sum_{i=1}^{l(\A)} -\ln p([E_{A_i^e}^{A_{i+1}^e-1},
		F_{A_i^f}^{A_{i+1}^f-1}]) \} \\
	& = & \argmin_{\A} \sum_{i=1}^{l(\A)} -\ln p([E_{A_i^e}^{A_{i+1}^e-1},
		F_{A_i^f}^{A_{i+1}^f-1}])
\end{eqnarray*}
(Recall that we take $p\su{A-len}(l)$ to be a uniform distribution.)
In other words, finding the alignment with highest probability
is equivalent to finding the alignment that minimizes
the sum of the negative logarithms of the probabilities of the
sentence beads that compose the alignment.\footnote{This transformation
is equivalent to the transformation between probability space
and length space given in Section \ref{ssec:mdl}.  (Recall
that $-\ln p = \ln \frac{1}{p}$.) }
Thus, by assigning to each sentence bead a
``length'' equal to the negative logarithm of its probability,
the sentence alignment problem is reduced to finding the shortest
path from the beginning of a corpus to its end.

The shortest path problem has a well-known dynamic programming
solution.  We evaluate a lattice $D(i, j)$ representing the
shortest distance from the beginning of the corpus to the
$i$th English sentence and $j$th French sentence.  The distance $D(i, j)$
is equal to the length of the most probable alignment aligning
the first $i$ and $j$ sentences of the English and
French corpora, respectively.  This lattice
can be calculated efficiently using a simple recurrence relation;
in this case, the recurrence relation is:
\begin{equation}
D(i, j) = \min \left\{ \begin{array}{lll}
	D(i - 1, j) & + & -\ln p([E_i]) \\
	D(i, j - 1) & + & -\ln p([F_j]) \\
	D(i - 1, j - 1) & + & -\ln p([E_i, F_j]) \\
	D(i - 2, j - 1) & + & -\ln p([E_{i-1}^i, F_j]) \\
	D(i - 1, j - 2) & + & -\ln p([E_i, F_{j-1}^j])
	\end{array} \right. \label{eqn:arecur}
\end{equation}
In other words, the most probable alignment of the first $i$ and $j$
English and French sentences can be expressed in terms of
the most probable alignment of some prefix of these sentences
extended by a single sentence bead.
The rows in the equation correspond to 1:0, 0:1, 1:1, 2:1, and 1:2
sentence beads, respectively.  The value $D(0,0)$ is taken to be zero.
The value $D(l(\E), l(\F))$ represents
the shortest distance through the whole corpus, and it is possible
to recover the alignment corresponding to this shortest path through
some simple bookkeeping.

Intuitively, this search can be viewed as maintaining a
set of partial alignments and extending them incrementally.
We fill in the lattice in increasing diagonals, where
the $k$th diagonal consists of all cells $D(i, j)$ such that $i +j = k$;
the $k$th diagonal corresponds to alignments containing a total of $k$
sentences.  Each cell $D(i, j)$ corresponds to the most probable alignment
ending at the $i$th and $j$th sentence in the English and French corpora.
We can consider the alignments corresponding to the $D(i, j)$ in the current
diagonal $i + j = k$ to be the set of current partial alignments.
Filling in the lattice in increasing diagonals can be considered
as extending the current partial alignments incrementally.

Notice that this algorithm is quadratic in the number of sentences
in the bilingual corpora, as we need to fill in a lattice with
$l(\E) l(\F)$ cells.
Given the size of existing bilingual corpora and the computation
necessary to evaluate the probability of a sentence bead, a quadratic
algorithm is too profligate.  However,
we can reap great savings in computation through intelligent
thresholding.  Instead of evaluating the entire lattice $D(i, j)$,
we ignore parts of the lattice that look as if they correspond
to poor alignments.
By considering only a subset of all possible alignments,
we reduce the computation to a linear one.

More specifically, we notice that the length $D(i, j)$ of an alignment prefix
is proportional to the number of sentences in the alignment
prefix $i+j$.  Hence, it is reasonable to compare the lengths of
two partial alignments if they contain the same number of sentences.
We prune all alignment prefixes that have a substantially
lower probability than the most probable alignment prefix of the same length.
That is, whenever $D(i, j) > D(i', j') + c$ for
some $i', j'$ and constant $c$ where
$i + j = i' + j'$, we set $D(i, j)$ to $\infty$.
This discards from consideration all alignments that
begin by aligning the first $i$ English sentences with the first
$j$ French sentences.  We evaluate the array $D(i, j)$ diagonal by
diagonal, so that $i + j$ increases monotonically.  For $c$, we
use the value 500, and with the Hansard corpus
this resulted in an average search beam width through
the dynamic programming lattice of about thirty; that is, on average
we evaluated thirty different $D(i, j)$ such that $i +j = k$
for each value $k$.

\ssec{Deletion Identification} \label{ssec:delid}

The dynamic programming framework described above can handle the case when
there is a small deletion in one of the corpora of a bilingual
corpus.  However, the framework is ineffective for deletions
larger than hundreds of sentences; the thresholding
mechanism is unreliable in this situation.  When the deletion point
is reached, the search algorithm will attempt to extend the current
partial alignments with sentence beads that align unrelated English
and French sentences.  There will be no correct alignment with
significantly higher probability to provide a meaningful standard
with which to threshold against.  Any thresholding that occurs
will be due to the random variation in alignment probabilities.

One solution is to not threshold when a deletion is detected.  However,
this is also impractical since dynamic programming is quadratic
without thresholding and deletions can be many thousands of
sentences long.  Thus, we handle long deletions outside of the
dynamic programming framework.

To detect the beginning of a deletion, we use the {\it confluence
point\/} mentioned in Section \ref{ssec:aparesta} as an indicator.
Recall that the confluence point is the point at the end
of the longest partial alignment common to all current hypothesis
alignments.\footnote{To calculate the confluence point, we keep
track of all sentence beads currently belonging to an active
partial alignment.  Whenever a sentence bead becomes the only
active bead crossing a particular diagonal in the distance
lattice $D$, \ie, the only active sentence bead $[E_{i_1}^{i_2}, F_{j_1}^{j_2}]$
such that $i_1 + j_1 \leq k$ and $i_2 + j_2 > k$ for some $k$,
then we know all active partial alignments include that sentence
bead and we can move the confluence point ahead of that sentence bead.}
The distance (in sentences) from the confluence point to
the diagonal in the lattice $D(i, j)$ currently being filled in
can be thought of as representing the uncertainty in alignment
at the current location in the lattice.  In the usual case, there is one
correct alignment that receives vastly greater probability
than other alignments, and thresholding is very aggressive so
this distance is small.  However, when there
is a large deletion in one of the parallel corpora,
consistent lexical correspondences disappear so
no one alignment has a much higher probability than the others.
Thus, there will be little thresholding and the distance from
the confluence point to the frontier of the lattice will become
large.  When this distance reaches a certain value, we take
this to indicate the beginning of a deletion.

In thresholding with $c=500$ on the Hansard corpus, we have found
that the confluence point is typically
about thirty sentences away from the frontier of $D(i, j)$.  Whenever the
confluence point lags 400 sentences behind the frontier,
we assume a deletion is present.

To identify the end of a deletion, we search for the occurrence of
infrequent words that are mutual translations.
We search linearly through both
corpora simultaneously.  All occurrences of words whose
frequency is below a certain value are recorded in a hash table;
with the Hansard corpus we logged words occurring ten or fewer times
previously.  Whenever we notice the occurrence of a rare word $e$ in one
corpus and its translation $f$ in the other (\ie, $p_b([e, f]) > 0$),
we take this as a candidate location for the end of the deletion.

To give an example, assume that the current confluence point
is located after the $i$th English sentence and $j$th French
sentence, and that we are currently calculating the diagonal in $D$
consisting of alignments containing a total of $i+j+400$ sentences.
Since this frontier is 400 sentences away from the confluence point,
we assume a deletion is present.  We iterate
through the English and French corpora simultaneously starting
from the $i$th and $j$th sentences, respectively, logging all
rare words.  Suppose the following rare words occur:
{\it $$ \begin{tabular}{|l|l|l|} \hline
{\rm language} & {\rm sentence \#} & {\rm word} \\ \hline \hline
English & $i + 7$ & Socratic \\ \hline
English & $i + 63$ & epidermis \\ \hline
French & $j + 127$ & indemnisation \\ \hline
English & $i + 388$ & Topeka \\ \hline
French & $j + 416$ & gypsophile \\ \hline
English & $i + 472$ & solecism \\ \hline
French & $j + 513$ & socratique \\ \hline
$\vdots$ & $\vdots$ & $\vdots$ \\ \hline
\end{tabular} $$}%
When we reach the 513th French sentence after the confluence point,
we observe the word {\it socratique\/} which is a translation of
the word {\it Socratic\/} found in the 7th English sentence
after the confluence point.  (We assume that we
have $p_b([\t{Socratic}, \t{socratique}]) > 0$.)  We then take
the $(i+7)$th and $(j+513)$th English and French sentences to
be a candidate location for the end of the deletion.

We test the correctness of a candidate location using a two-stage
process for efficiency.  First, we calculate
the probability of the sentence bead composed
of the two sentences containing the two rare words.  If this is ``sufficiently
high,'' we then examine the forty sentences following the occurrence of
the rare word in each of the two parallel corpora.  We use dynamic programming
to find the probability of the best alignment of these
two blocks of sentences.  If this probability is also
``sufficiently high'' we take the candidate location to be the end of
the deletion.
Because it is extremely unlikely that there are two very similar
sets of forty sentences in a corpus, this deletion identification algorithm
is robust.  In addition, because we key off of rare words
in searching for the end of a deletion, deletion identification requires time
linear in the length of the deletion.

We consider the probability $p\su{actual}$ of an alignment
to be ``sufficiently high''
if its score is a certain fraction $f$ of the highest possible
score given just the English sentences in the segment.
More specifically, we use the following equation to calculate $f$:
$$ f = \frac{(-\ln p\su{actual}) - (-\ln p\su{min})}{(-\ln p\su{max}) -
	(-\ln p\su{min})} $$
To calculate $p\su{max}$, we
calculate the French sentences that would yield the highest possible
alignment score given the English sentences in the alignment;
these sentences can be constructed
by just taking the most probable word-to-word translation for each word in
the English sentences.  The probability of the alignment of
these optimal French sentences with the given English sentences is
$p\su{max}$.  The probability $p\su{min}$ is taken to be the probability
assigned to the alignment where the sentences are aligned
entirely with 0:1 and 1:0 sentence beads; this approximates the
lowest possible achievable score.

We take this quotient $f$ to represent the quality of an alignment.
For the initial sentence-to-sentence
comparison, we took the fraction 0.57 to be ``sufficiently high.''  For
the alignment of the next forty sentences, we took the fraction 0.4
to be ``sufficiently high.''  These values were arrived at empirically,
by trying several values on several deletion points in the
Hansard corpus and choosing the value with the best subjective performance.
Reasonable performance can be achieved with values within a couple
tenths of these values; higher or lower values may be used to improve
alignment precision or recall.\footnote{We use {\it precision\/}
to describe the fraction of sentence beads returned by the
algorithm that represent correct alignments; {\it recall\/}
describes the fraction of all correct sentence beads
that are found by the algorithm.  In these measures, we only consider
sentence beads containing both English and
French sentences as these are the beads most useful in applications.}

Because we key off of rare words in recovering from deletions, it
is possible to overshoot the true recovery point by a significant
amount.  To correct for this, after we find a location for
the end of a deletion using the mechanisms described previously,
we backtrack through the corpus.  We take the ten preceding
sentences in each corpus from the recovery point, and find the
probability of their alignment.  If this probability is ``sufficiently high,''
we move the recovery point back and repeat the process.  We take
the fraction 0.4 using the measure described in the last paragraph
to be ``sufficiently high.''

\ssec{Subdividing a Corpus for Parallelization} \label{ssec:divide}

Sentence alignment is a task that seems well-suited to parallelization,
since the alignment of different sections of a bilingual corpus are
basically independent.  However, to parallelize sentence alignment
it is necessary to be
able to divide a bilingual corpus into many sections accurately.  That is,
division points in the two corpora of a bilingual corpus must
correspond to identical points in the text.
Our deletion recovery mechanism can be used for this purpose.  We start
at the beginning of each corpus in the bilingual corpus, and
skip some number of sentences in each corpus.  The number of sentences
we skip is the number of sentences we want in each subdivision of
the bilingual corpus.  We then employ the deletion recovery mechanism
to find a subsequent point in the two corpora that align, and we take this to
be the end of the subdivision.  We repeat this process to divide
the whole corpus into small sections.

\ssec{Algorithm Summary}

We summarize the algorithm below, not including parallelization.

{ \def\r#1{{\bf #1}} \singlespace
\noindent\rule{5in}{0.1pt}
\begin{tabbing}
mm \= mm \= mm \= \hspace{2in} \= \kill
initialize all counts and parameters \\
\mbox{} \\
; {\it bootstrap the model by training on manually-aligned data} \\
; {\it $( [E_{A_1^e}^{A_2^e-1}, F_{A_1^f}^{A_2^f - 1}], \ldots,
	[E_{A_L^e}^{A_{L+1}^e-1}, F_{A_L^f}^{A_{L+1}^f - 1}]$ ) } \\
\r{for} $i=1$ \r{to} $L$ \r{do} \\
\> \r{begin} \\
\> $\bv$ := most probable word beading of
	$[E_{A_i^e}^{A_{i+1}^e-1}, F_{A_i^f}^{A_{i+1}^f - 1}]$ \\
\> update counts and parameters based on $\bv$ \\
\> \r{end} \\
\mbox{} \\
; {\it this is the main loop where we fill the $D(i, j)$ lattice} \\
; {\it $(i\su{conf}, j\su{conf})$ holds indices of the
	English and French sentences at confluence point} \\
; {\it $k$ holds value of current diagonal being filled;
	diagonal is all $D(i, j)$ with $k = i + j$} \\
$i\su{conf} := 1$ \\
$j\su{conf} := 1$ \\
$D(0, 0) := 0$ \\
\r{for} $k=2$ \r{to} $l(\E) + l(\F)$ \r{do} \\
\> \r{begin} \\
\> ; {\it check for long deletion} \\
\> \r{if} $k - (i\su{conf} + j\su{conf}) > 400$ \r{then} \\
\> \> \r{begin} \\
\> \> $(i\su{end}, j\su{end})$ := location of end of deletion (see
	Section \ref{ssec:delid}) \\
\> \> $k := i\su{end} + j\su{end}$ \\
\> \> $(i\su{conf}, j\su{conf}) := (i\su{end}, j\su{end})$ \\
\> \> \r{end} \\
\mbox{} \\
\> ; {\it fill in diagonal in $D$ array} \\
\> ; {\it $D\su{best}$ holds the best score in the diagonal, used for
	thresholding purposes} \\
\> $D\su{best} := \infty$ \\
\> ; {\it only fill in cells in diagonal corresponding to extending a
	nonthresholded alignment} \\
\> \r{for} all $i, j \geq 1$ such that $i + j = k$ and \\
\> \> $\exists i', j'$ with $i - 2 \leq i' \leq i$,
	$j - 2 \leq j' \leq j$, $D(i', j') < \infty$ \r{do} \\
\> \> \r{begin} \\
\> \> $D(i, j)$ := {\it expression given in equation (\ref{eqn:arecur})} \\
\> \> \r{if} $D(i, j) < D\su{best}$ \r{then} \\
\> \> \> $D\su{best} := D(i, j)$ \\
\> \> \r{end} \\
\mbox{} \\
\> ; {\it threshold items in diagonal} \\
\> \r{for} all $i, j \geq 1$ such that $i + j = k$ and
	$D(i, j) < \infty$ \r{do} \\
\> \> \r{if} $D(i, j) < D\su{best} - 500$ \r{then} \\
\> \> \> $D(i, j) := \infty$ \\
\mbox{} \\
\> ; {\it update confluence point (see Section \ref{ssec:delid})} \\
\> \r{if} confluence point has moved \r{then} \\
\> \> \r{begin} \\
\> \> ($i\su{conf}$, $j\su{conf}$) := new location of confluence point \\
\> \> \r{for} each sentence bead
	$[E_{A_i^e}^{A_{i+1}^e-1}, F_{A_i^f}^{A_{i+1}^f - 1}]$
	moved behind confluence point \r{do} \\
\> \> \> \r{begin} \\
\> \> \> $\bv$ := most probable word beading of
	$[E_{A_i^e}^{A_{i+1}^e-1}, F_{A_i^f}^{A_{i+1}^f - 1}]$ \\
\> \> \> update counts and parameters based on $\bv$ \\
\> \> \> \r{end} \\
\> \> \r{end} \\
\> \r{end}
\end{tabbing} }

%
\sec{Results}
%

Using this algorithm, we have aligned three large English/French corpora.
We have aligned a corpus of 3,000,000 sentences (of both English and French)
of the Canadian Hansards, a corpus of 1,000,000 sentences
of newer Hansard proceedings, and a corpus of 2,000,000 sentences
of proceedings from the European Economic Community.
In each case, we first bootstrapped the translation model by training
on 100 previously aligned sentence pairs.  We then trained the model further
on 20,000 (unaligned) sentences of the target corpus.

Because of the very low error rates involved, instead of direct
sampling we decided to
estimate our error on the old Hansard corpus through comparison with
the alignment found by Brown \etal\ on the same corpus.  We manually
inspected over 500 locations where the two alignments differed to
estimate our error rate on the alignments disagreed upon.  Taking
the error rate of the Brown alignment to be 0.6\%, we estimated
the overall error rate of our alignment to be 0.4\%.

In addition, in the Brown alignment approximately 10\% of the corpus
was discarded because of indications that it would be difficult
to align.  Their error rate of 0.6\% holds on the remaining sentences.
Our error rate of 0.4\% holds on the entire corpus.
Gale reports an approximate error rate of 2\% on a different body
of Hansard data with no discarding, and an error rate of 0.4\% if 20\% of the
sentences can be discarded.

Hence, with our algorithm we can achieve at least as high accuracy
as the Brown and Gale algorithms {\it without\/} discarding any data.
This is especially significant since, presumably,
the sentences discarded by the Brown and Gale
algorithms are those sentences most difficult to align.

To give an idea of the nature of the errors our algorithm makes,
we randomly sampled 300 alignments
from the newer Hansard corpus.  The two errors we found are
displayed in Figures \ref{tab:err1} and \ref{tab:err2}.
\begin{figure}
{\it $$ \begin{tabular}[t]{ll}
$E_1$ & \corptxt{If there is some evidence that it
	\ldots\ and I will see that it does.} \\
$E_2$ & \corptxt{\bc\ Translation \ec}
\end{tabular} \rule[-0.7in]{0.1mm}{0.7in}
\begin{tabular}[t]{ll}
$F_1$ & \corptxt{Si on peut prouver que elle \ldots\ je verrais \`{a} ce que
	elle se y conforme.
	\bc\ Language = French \ec} \\
$F_2$ & \corptxt{\bc\ Paragraph \ec}
\end{tabular} $$}%
\caption{An alignment error} \label{tab:err1}
\end{figure}
\begin{figure}
{\it $$ \begin{tabular}[t]{ll}
$E_1$ & \corptxt{Motion No.\ 22 that Bill C-84 be amended in
	\ldots\ and substituting the following
	therefor :\ second anniversary of.}
\end{tabular} \rule[-0.7in]{0.1mm}{0.7in}
\begin{tabular}[t]{ll}
$F_1$ & \corptxt{Motion No 22 que on modifie le projet de loi C-84
	\ldots\ et en la rempla\c{c}ant par ce qui suit :\ ` 18.} \\
$F_2$ & \corptxt{Deux ans apr\`{e}s :\ '.}
\end{tabular} $$}%
\caption{Another alignment error} \label{tab:err2}
\end{figure}
In the first error, $E_1$ was aligned with $F_1$ and $E_2$ was aligned
with $F_2$.  The correct alignment maps $E_1$ and $E_2$ to $F_1$ and
$F_2$ to nothing.  In the second error, $E_1$ was aligned with $F_1$
and $F_2$ was aligned to nothing.  The correct alignment maps $E_1$
to both $F_1$ and $F_2$.  Both of these errors could have
been avoided with improved sentence boundary detection.

The rate of alignment ranged from 2,000 to 5,000 sentences of both
English and French per hour on
an IBM RS/6000 530H workstation.  Using the technique described
in section \ref{ssec:divide}, we subdivided corpora into
small sections (20,000 sentences) and aligned sections in parallel.
While it required on the order of 500 machine-hours to align the
newer Hansard corpus, it took only 1.5 days of real time to complete the
job on fifteen machines.

\ssec{Lexical Correspondences} \label{ssec:lexcor}

One of the by-products of alignment is the distribution $p_b(b)$
of word bead frequencies.  For 1:1 word beads $b = [e, f]$, the
probability $p_b(b)$ can be interpreted as a measure of how strongly
the words $e$ and $f$ translate to each other.  Hence, $p_b(b)$
in some sense represents a probabilistic word-to-word bilingual
dictionary.

For example, we can use the following measure $t(e, f)$ as
an indication of how strong the words $e$ and $f$ translate to each
other:\footnote{This measure is closely related to the measure
$$ {\rm MI}_F(x, y) = \frac{p_{X,Y}(x, y)}{p_X(x) p_Y(y)} $$
referred to as {\it mutual information\/} by \newcite{Magerman:90a}.
This is not to be confused with the more common definition of
mutual information:
$$ I(X; Y) = \sum_{x,y} p(x, y) \log \frac{p(x, y)}{p(x)p(y)} $$ }
$$ t(e, f) = \ln \frac{p_b([e, f])}{p_e(e) p_f(f)} $$
We divide $p_b([e, f])$ by the frequencies of the individual words to
correct for the effect that $p_b([e, f])$ will tend to be higher
for higher frequency words $e$ and $f$.  Thus, $t(e, f)$ should
not be skewed by the frequency of the individual words.  Notice
that $t(e, f)$ is roughly equal to the gain in the (logarithm of the)
probability of a word beading if word
beads $[e]$ and $[f]$ are replaced with the bead $[e, f]$.  Thus,
$t(e, f)$ dictates the order in which 1:1 word beads are applied
in the search for the most probable word beading of a sentence
bead described in Section \ref{ssec:aevalsb}.

In Appendix \ref{app:lex}, for a group of randomly sampled
English words we list the French words that
translate most strongly to them according to this measure.
In general, the correspondences are fairly accurate.
However, for some common prepositions
the correspondences are rather poor; examples of this are
also listed in Appendix \ref{app:lex}.
Prepositions sometimes occur in
situations in which they have no good translation, and many prepositions
have numerous translations.\footnote{It has been suggested that
removing closed-class words before alignment may improve performance;
these words do not provide much reliable alignment information
and are often assigned spurious translations by our
algorithm \cite{Shieber:96a}.}

In many lists, the ``French'' word with the exact same spelling as
the English word occurs near
the top of the list.  (A word is considered to be ``French'' if it occurs
at any point in the French corpus.)  This is due to the initialization
described in Section \ref{ssec:cognate} we perform for cognates.

Notice that we are capable of acquiring strong lexical correspondences
between non-cognates.  Preliminary experiments indicate that
cognate initialization does not significantly affect
alignment accuracy.  Hence, our alignment algorithm is applicable
to languages with differing alphabets.

%
\sec{Discussion}
%

We have described an accurate, robust, and efficient algorithm for
sentence alignment.  The algorithm can handle large deletions in
text, it is language independent, and it is parallelizable.
It requires a minimum of human intervention; for each language pair
100 sentences need to be aligned by hand to bootstrap the translation
model.  Unlike previous algorithms,
our algorithm does not require that the bilingual corpus
be predivided into small chunks or that one can
identify markers in the text that make this subdivision easier.
Our algorithm produces a probabilistic bilingual dictionary, and
it can take advantage of cognate correspondences, if present.

The use of lexical information requires a great computational cost.
Even with numerous approximations, this algorithm is tens of times slower
than the length based algorithms of Brown \etal\ and Gale and Church.
This is acceptable given available computing power and given
that alignment is a one-time cost.
It is unclear, though, whether more powerful models are worth
pursuing.

One limitation of the algorithm is that it only considers
aligning a single sentence to zero, one, or two sentences in the
other language.  It may be useful to extend the set of sentence
beads to include 2:2 or 1:3 alignments, for example.  In addition,
we do not consider sentence ordering transpositions between
the two corpora.  For example, the case where the first sentence in an English
corpus translates to the second sentence in a French corpus
and the second English sentence translates to the first French
sentence cannot be handled correctly by our algorithm.  At best,
our algorithm will align one pair of sentences correctly and
align each of the remaining two sentences to nothing.
However, while extending the algorithm in these ways can potentially
reduce the error rate by allowing the algorithm to express a
wider range of alignments, it may actually increase error rate
because the algorithm must consider a larger set of possible
alignments and thus becomes more susceptible to random error.

Thus, before adding extensions like these it may be wise to strengthen
the translation model, which should improve performance in
general anyway.  For example,
one natural extension to the translation model is to
account for word ordering.  \newcite{Brown:91g} describe several
such possible models.
However, substantially greater computing power is required before these
approaches can become practical, and there is not much room for
further improvements in accuracy.  In addition, parameter estimation
becomes more difficult with larger models.

%
%

\chapter{Conclusion} \label{ch:concl}

In this thesis, we have presented techniques for modeling
language at the word level, the constituent level, and the sentence level.
At each level, we have developed methods that surpass the performance
of existing methods.

At the word level, we examined the task of smoothing
$n$-gram language models.  While smoothing is a fundamental
technique in statistical modeling, the literature lacks any sort
of systematic comparison of smoothing techniques for language
tasks.  We present an extensive empirical comparison
of the most widely-used smoothing algorithms for $n$-gram
language models, the current standard in language modeling.
We considered several issues not considered in previous work,
such as how training data size, $n$-gram order, and parameter
optimization affect performance.  In addition, we introduced
two novel smoothing techniques that surpass all previous
techniques on trigram models and that perform well on
bigram models.  We provide some detailed analysis that
helps explain the relative performance of different algorithms.

At the constituent level, we investigated grammar induction
for language modeling.  While yet to achieve comparable performance,
grammar-based models are a promising
alternative to $n$-gram models as they can express
both short and long-distance dependencies and they
have the potential to be more compact than equivalent
$n$-gram models.  We introduced a probabilistic context-free grammar
induction algorithm that uses the Bayesian framework and the
minimum description length principle.  By using a rich move
set and the technique of triggering, our search algorithm
is efficient and effective.  We demonstrated
that our algorithm significantly outperforms the Lari
and Young induction algorithm, the most widely-used
algorithm for probabilistic grammar induction.  In addition, we were able to
surpass the performance of $n$-gram models on artificially-generated data.

At the sentence level, we examined bilingual sentence alignment.
Bilingual sentence alignment is a necessary step in processing
a bilingual corpus for use in many applications.  Previous
algorithms suitable for large corpora ignore word identity
information and just consider sentence length; we introduce
an algorithm that uses lexical information efficient enough
for large bilingual corpora.  Furthermore, our algorithm
is robust, language-independent, and parallelizable.  We
surpass all previously reported accuracy rates on the
Hansard corpus, the most widely-used corpus in machine translation
research.

It is interesting to note that for these three tasks,
we use three very different frameworks.  In the next
section, we explain what these frameworks are and how they relate
to the Bayesian framework.  We argue that each framework used
is appropriate for the associated problem.  Finally,
we show how our work on these three problems address
two central issues in probabilistic modeling: the sparse
data problem and the problem of inducing hidden structure.

%
\sec{Bayesian Modeling}
%

In our work on grammar induction, we use the Bayesian framework.
We attempt to find the grammar $G$ with the largest probability
given the training data, or observations, $O$.  Applying
Bayes' rule we get
\begin{equation}
G = \argmax_G p(G|O) = \argmax_G \frac{p(G)p(O|G)}{p(O)} =
	\argmax_G p(G)p(O|G) \label{eqn:cbayes}
\end{equation}
We search for the grammar $G$ that maximizes the objective function
$p(G) p(O|G)$.

The Bayesian framework is a very elegant and general framework.
In the objective function, the term describing how well a grammar
models the data, $p(O|G)$, is separate from the term describing our
{\it a priori\/} notion of how likely a grammar is, $p(G)$.  Furthermore,
because we express the target grammar in a static manner
instead of an algorithmic manner, we have a separation between
the objective function and the search strategy.
Thus, we can switch around prior distributions or search strategies
without changing other parts of an algorithm.
The Bayesian framework modularizes search problems in a general
and logical way.

However, notice that we could have framed both $n$-gram smoothing
and sentence alignment in the Bayesian framework as well, but we chose not to.
To express smoothing in the Bayesian framework,
we can use an analogous equation to equation (\ref{eqn:cbayes}), \eg,
$$ M = \argmax_M p(M|O) = \argmax_M \frac{p(M)p(O|M)}{p(O)} =
	\argmax_M p(M)p(O|M) $$
where $M$ denotes a smoothed $n$-gram model.  We can design
a prior $p(M)$ over smoothed $n$-gram models and search for the model $M$ that
maximizes $p(M)p(O|M)$.\footnote{Actually, a slightly different
Bayesian formulation is more appropriate for smoothing, as will
be mentioned in Section \ref{ssec:bayessmooth}.}
Instead, most existing smoothing algorithms
as well as our novel algorithms involve a straightforward mapping from
training data to a smoothed model.\footnote{This is not quite true;
Jelinek-Mercer smoothing uses the Baum-Welch algorithm to perform
a maximum likelihood search for $\l$ values.  In addition, we performed
automated parameter optimization in a maximum likelihood manner.}

In sentence alignment, from aligned data $(\E, \F)$ we build
a model $p_t(E_i^j; F_k^l)$ of the frequency with which sentences $E_i^j$
and sentences $F_k^l$ occur as mutual translations in a bilingual corpus.
To frame this in the Bayesian framework, we can use the equation
$$ p_t = \argmax_{p_t} p(p_t|\E, \F) =
	\argmax_{p_t} \frac{p(p_t)p(\E, \F|p_t)}{p(\E, \F)} =
	\argmax_{p_t} p(p_t)p(\E, \F|p_t) $$
We could devise a prior distribution $p(p_t)$ over possible translation
models and attempt to find the model $p_t$ that maximizes
$p(p_t) p(\E, \F|p_t)$.  Instead, we use a variation of
the Expectation-Maximization algorithm to perform a
deterministic maximum-likelihood search for the model $p_t$.

Thus, while we could have used the Bayesian framework in each
of the three problems we addressed, we instead used three different
approaches.  Below, we examine each problem in turn and
argue why the chosen approach was appropriate for the
given problem.

\ssec{Smoothing $n$-Gram Models} \label{ssec:bayessmooth}

First, we note that the Bayesian formulation given previously
of finding the most probable model
$$ M\su{best} = \argmax_M p(M|O) $$
is not appropriate for the smoothing problem.  The reason
the most probable model $M\su{best}$ is significant is because we expect
it to be a good model of data that will be seen in the future, \eg,
data that is seen during the actual use of an application.
In other words, we find $M\su{best}$ because we expect that $p(O_f|M\su{best})$
is a good model of future data $O_f$.  However, notice that
finding $M\su{best}$ is actually superfluous; what we are really
trying to find is just $p(O_f|O)$,
a model of what future data will be like given our training data.
This is a more accurate Bayesian formulation of the smoothing problem (and
of modeling problems in general); the identity of $M\su{best}$ is not
important in itself.

Using this perspective, we can explain why smoothing is
generally necessary in $n$-gram modeling and other types of
statistical modeling.  Expressing $p(O_f|O)$ in terms of models $M$, we get
$$ p(O_f|O) = \sum_M p(O_f,M|O) = \sum_M p(O_f|M,O) p(M|O)
	= \sum_M p(O_f|M) p(M|O) $$
According to this perspective, instead of predicting future
data using $p(O_f|M\su{best})$ for just the most probable model $M\su{best}$,
we should actually sum $p(O_f|M)$ over all models $M$, weighing
each model by its probability given the training data.  However,
performing a sum over all models is generally impractical.
Instead, one might consider trying to approximate this sum with
its maximal term $\max_M p(O_f|M) p(M|O)$.  However, the identity
of this term depends on $O_f$, data that has not been seen yet.
Thus, just using $p(O_f|M\su{good})$ for some good model $M\su{good}$, such as
the most probable model $M\su{best}$, may be
the best we can do in practice.  Smoothing can be
interpreted as a method for correcting this gap
between theory and reality, between $p(O_f|O)$ and
$p(O_f|M\su{good})$.  Viewed from this perspective,
most smoothing algorithms for $n$-gram models do not even
use an intermediate model $M\su{good}$, but
just estimate the distribution $p(O_f|O)$ directly from counts in
the training data.\footnote{Alternatively, we can just
view $M\su{good}$ as being the maximum likelihood $n$-gram model.}

We argue that the Bayesian approach
is not attractive for the smoothing problem from a methodological
perspective.  In the Bayesian framework,
the nature of the prior probability $p(M)$ is the largest factor in
determining performance.  The distribution $p(M)$
should reflect how frequently smoothed $n$-gram models $M$ occur
in the real world.  However, it is unclear what {\it a priori\/}
information we have pertaining to the frequency of different
smoothed $n$-gram models; we have little or no intuition on this topic.
In addition, adjusting $p(M)$ to try to yield
an accurate distribution $p(O_f|O)$ is a rather indirect process.

For smoothing, it is much easier to estimate $p(O_f|O)$ directly.
We have insight into what a smoothed distribution $p(O_f|O)$
should look like given the counts in the training data.
For example, we know that an $n$-gram with zero counts should
be given some small nonzero corrected count, and that an $n$-gram
with $r > 0$ counts should be given a corrected count slightly less
than $r$, so that there will be counts available for zero-count $n$-grams.
These corrected counts lead directly to a model $p(O_f|O)$.
In addition, detailed performance analyses such as the analysis
described in Section \ref{sec:sresults} lend themselves
well to improving algorithms that estimate $p(O_f|O)$ directly.

There are several existing Bayesian smoothing methods
\cite{Nadas:84a,MacKay:95a,Ristad:95a}, but none perform
particularly well or are in wide use.

\ssec{Bayesian Grammar Induction}

In grammar induction, we have a very different situation from that found
in smoothing.  In smoothing, we have intuitions on how to
estimate $p(O_f|O)$, but little intuition of how to estimate $p(M)$.
In grammar induction, we have the opposite situation.  In this
case, the distribution $p(O_f|O)$ has a very complex nature.
In Section \ref{ssec:mdl}, we give examples that hint at the complexity of
this distribution.  For example, if we see a sequence of numbers in some text
that happen to be consecutive prime numbers, people know how to
predict future numbers in the sequence with high probability.
In Section \ref{ssec:mdl}, we explain how complex behaviors such
as this can be modeled by using the Bayesian abstraction.  We
take the probability $p(O)$ of some data $O$ to be
$$ p(O) = \sum_{G_p} p(O, G_p) = \sum_{G_p} p(G_p) p(O|G_p)
	= \sum\su{output($G_p$) = $O$} p(G_p) $$
where $G_p$ represents a program; that is, we view data as being
the output of some program.  By assigning higher probabilities
to shorter programs, we get the desirable behavior that complex patterns
in data can be modeled.  In the grammar induction task, we just restrict
programs $G_p$ to those that correspond to grammars $G$.
In this domain, we have insight into the nature of the
prior probability $p(G)$, which takes a fairly simple form,
while $p(O_f|O)$ takes a very complicated form.  Thus, unlike
smoothing, we find that the Bayesian perspective is appropriate
for grammar induction.

\ssec{Bilingual Sentence Alignment}

In sentence alignment, the situation is more similar to grammar
induction than smoothing.  We have a complex distribution
$p(O_f|O)$, or using our alignment notation, $p(\E_f, \F_f|\E, \F)$.
Instead of modeling $p(\E_f, \F_f|\E, \F)$ directly, it is more
reasonable to use the Bayesian framework and to design a prior
on translation models using the minimum description length
principle.  This would yield some desirable behaviors;
for example, in the description of the full translation
model we would include the description of the word-to-word
translation model $p_b(b)$, which we can describe by listing
all word beads $b$ with nonzero probability.  A minimum description
length objective function would favor models with fewer nonzero-probability
word beads, thus discouraging words from having
superfluous translations in the model.

In actuality, we decided to use a maximum likelihood approach,
which is equivalent to using the Bayesian approach with a
uniform prior over translation models.  Specifically, we
used a variation of the Expectation-Maximization (EM)
algorithm, which is a hill-climbing search on the likelihood
of the training data.  However, we used an incremental variation
of the algorithm.  Typically, the EM algorithm is an iterative
algorithm, where in each iteration the entire training data is
processed.  At the end of each iteration, parameters are
re-estimated so that the likelihood of the data increases (or
does not decrease, at least).  In our incremental version,
we make a single pass through
the training data, and we re-estimate parameters after
each sentence bead.  Thus, we do not perform a true maximum likelihood
search; we just do something roughly equivalent to a single iteration
in a conventional EM search.

While we make this choice partially because
a full EM search is too expensive computationally, we also believe that
a full EM search would yield poorer results.
Notice that with each iteration of EM performed, the
current hypothesis model fits closer to the training data.
Recall that maximum likelihood models tend to
overfit training data, as the uniform prior assigns too much probability to
large models.  Thus, additional EM iterations
will eventually lead to more and more overfitting.  By only
doing something akin to a single iteration of EM, we avoid
the overfitting problem.\footnote{Better performance may
be achieved by using held-out data to determine when overfitting
begins, and stopping the EM search at that point.  However, additional
EM passes are expensive computationally and our single-pass approach
performs well.}

One way of viewing this is that we express a prior distribution over models
{\it procedurally}.  Instead of having an explicit prior that
prefers smaller models, we avoid overfitting through heuristics
in our search strategy.  This violates the separation
between the objective function and the search strategy found in the
Bayesian framework.  On the other hand,
it significantly decreases the complexity of the implementation.
In the Bayesian framework, one needs to
design an explicit prior over models and to perform a search for
the most probable model; these tasks are expensive both from
a design perspective and computationally.
Furthermore, in sentence alignment it is not necessary to have a tremendously
accurate translation model, as lexical information usually
provides a great deal of distinction between correct and incorrect
alignments.  The use of {\it ad hoc\/} methods is not likely
to decrease performance significantly.  Thus, we argue that
the use of {\it ad hoc\/} methods such as procedural priors
is justified for sentence alignment as it saves a great deal
of effort and computation without loss in performance.

In summary, while the Bayesian framework provides a principled
and effective approach to many tasks, it is best to adjust
the framework to the exigencies of the particular problem.
In grammar induction, an explicit Bayesian implementation proved
worthwhile, while in sentence alignment less rigorous methods
performed well.  Finally, for smoothing we argue that non-Bayesian
methods can be more effective.

%
\sec{Sparse Data and Inducing Hidden Structure}
%

As mentioned in Chapter \ref{ch:intro}, perhaps the two most
important issues
in probabilistic modeling are the sparse data problem and the problem
of inducing hidden structure.  The sparse data problem describes
the situation of having insufficient data to train one's model accurately.
The problem of inducing hidden structure refers to the task of
building models that capture structure not explicitly present
in the training data, \eg, the grammatical structure that we
capture in our work in grammar induction.  It is widely thought
that models that capture hidden structure can ultimately
outperform the shallow models that currently offer the best performance.
In this thesis, we present several techniques that address these
two central issues in probabilistic modeling.

\ssec{Sparse Data}

Approaches to the sparse data problem can be grouped into
two general categories.  The first type of approach is
{\it smoothing}, where one takes existing models and
investigates techniques for more accurately assigning
probabilities in the presence of sparse data.  Our work on
smoothing $n$-gram models greatly forwards the literature
on smoothing for the most frequently used probabilistic models
for language.  We clarify the relative performance of different
smoothing algorithms on a variety of data sets, facilitating
the selection of smoothing algorithms for different applications;
no thorough empirical study existed before.  In addition, we
provide two new smoothing techniques that outperform existing
techniques.

The second general approach to the sparse data problem is
the use of compact models.  Compact models contain fewer
probabilities that need to be estimated and hence require
less data to train.  While $n$-gram models are yet to be
outperformed by more compact models, this approach to the
sparse data problem seems to be the most promising as smoothing
most likely will yield limited gains.\footnote{There has been
some success in combining the two approaches.  For
example, \newcite{Brown:90a} show
how {\it class-based\/} $n$-gram models can achieve performance
near to that of $n$-gram models using much fewer parameters.  They
then show that by linearly interpolating or {\it smoothing\/}
conventional $n$-gram models with class-based $n$-gram models,
it is possible to achieve performance slightly superior to that of
conventional $n$-gram models alone.}
As mentioned in Section \ref{ssec:pcfgngram}, probabilistic
grammars offer the potential for achieving
performance equivalent to that of $n$-gram models with much smaller models.
In our work on Bayesian grammar induction, we introduce a novel
algorithm for grammar induction that employs the minimum
description length principle, under which the objective
function used in the grammar search process explicitly favors
compact models.  In experiments on artificially-generated data,
we can achieve performance equivalent to that of $n$-gram models
using a probabilistic grammar that is many times smaller, as
shown in Section \ref{sec:gresults}.  While unable to outperform $n$-gram
models on naturally-occurring data, this work represents very real progress
in constructing compact models.

\ssec{Inducing Hidden Structure}

Models that capture the hidden structure underlying language have
the potential to outperform shallow models such
as $n$-gram models.  Not only does our work in grammar induction
forward research in building compact models as described above, it
also demonstrates techniques for inducing hidden structure.  (Clearly,
the problems of sparse data and inducing hidden structure
are not completely orthogonal, as taking advantage of hidden structure
can lead to more compact models.)  In Section \ref{sec:gresults},
we show how on artificially-generated text our grammar induction algorithm is
able to capture much of the structure present in the grammar
used to generate the text, demonstrating that our algorithm can
effectively extract hidden structure from data.

Our work in bilingual sentence alignment also addresses
the problem of inducing hidden structure.  Given a
raw bilingual corpus, sentence alignment involves
recovering the hidden mapping between the two texts that
specifies the sentence(s) in one language that translate
to each sentence in the other language.  To this end,
our alignment algorithm also calculates a rough mapping
between individual words in sentences in the two languages.
Unlike in grammar induction, the model used to induce this
hidden structure is a fairly shallow model;
the model is just used to annotate data with
the extracted structural information.  This
annotated data can then be used to train structured models,
as in work by \newcite{Brown:90b}.

In this work on hidden structure, we place an emphasis
on the use of {\it efficient\/} algorithms.  A major issue
in inducing hidden structure is constraining the search process.
Because the structure is {\it hidden}, it is difficult to
select which structures to consider creating, and as a result
many algorithms dealing with hidden structure induction are
inefficient because they do not adequately constrain the search
space.  Our algorithms for grammar induction and bilingual
sentence alignment are both near-linear; both are far more
efficient than all other algorithms (involving hidden
structure induction) that offer comparable performance.

We achieve this efficiency through data-driven heuristics
that constrain the set of hypotheses considered in the search process
and through heuristics that allow hypotheses to be evaluated
very quickly.  In grammar induction, we
introduce the concept of {\it triggers}, or particular patterns
in the data that indicate that the creation of certain rules may be
favorable.  Triggers reduce the number of grammars considered
to a manageable amount.  In addition, to evaluate the
objective function efficiently, we use sensible heuristics to
estimate the most probable parse of the training data given the current
grammar and to estimate the optimal values of rule probabilities.

In sentence alignment, we use thresholding to reduce the computation
of the dynamic programming lattice from quadratic to linear in data size.
To enable efficient evaluation of hypothesis alignments,
we use heuristics to constrain which word beads have nonzero probability.
As mentioned in Section \ref{ssec:aparestb}, limiting the number of
such word beads greatly simplifies the search for the most
probable beading of a sentence bead.  We believe that data-driven
heuristics such as the ones that we have employed are crucial
for making hidden structure induction efficient enough for
large data sets.

In conclusion, this thesis represents a very significant step
forward towards addressing two central issues in probabilistic
modeling: the sparse data problem and the problem of inducing
hidden structure.  We introduce novel techniques for smoothing
and for constructing compact models, as well as novel and efficient
techniques for inducing hidden structure.

\appendix

%
\chapter{Sample of Lexical Correspondences Acquired During
Bilingual Sentence Alignment} \label{app:lex}
%

In this section, we list randomly-sampled lexical correspondences acquired
during the alignment of 20,000 sentences pairs.
We only list words occurring at least ten times.
The numbers adjacent to the English words are the number
of times those English words occurred in the corpus.  The number next to
each French word $f$ is the value
$$ t(e,f) = \ln \frac{p_b([e, f])}{p_e(e) p_f(f)} $$
for the English word $e$ above; this is a measure of
how strongly the English word and French word translate to each other.

{
\parskip=0pt
\begin{tabular}[t]{p{0.3in}p{1.1in}l}
\multicolumn{3}{l}{quality (27)} \\
	& qualit\'{e}     & 11.69\\
	& qualitatives     & 11.46\\
	& eaux             &  9.52
\end{tabular} \hspace{0.7in}
\begin{tabular}[t]{p{0.3in}p{1.1in}l}
\multicolumn{3}{l}{keeps (16)} \\
	& engag\'{e}e     &  9.18\\
	& continue         &  8.61\\
	& que              &  4.66
\end{tabular}

\begin{tabular}[t]{p{0.3in}p{1.1in}l}
\multicolumn{3}{l}{form (70)} \\
	& forme            & 10.11\\
	& trouveraient     &  8.72\\
	& sorte            &  7.18\\
	& obligations      &  6.69\\
	& ajouter          &  6.49\\
	& sous             &  5.03\\
	& une              &  4.96\\
	& avons            &  3.97
\end{tabular} \hspace{0.7in}
\begin{tabular}[t]{p{0.3in}p{1.1in}l}
\multicolumn{3}{l}{houses (20)} \\
	& maisons          & 11.47\\
	& chambres         & 10.77\\
	& habitations      &  9.41\\
	& maison           &  9.34\\
	& domiciliaire     &  9.17\\
	& logements        &  9.14\\
	& parlementaires   &  8.01\\
	& acheter          &  7.69
\end{tabular}

\begin{tabular}[t]{p{0.3in}p{1.1in}l}
\multicolumn{3}{l}{throughout (33)} \\
	& travers          &  9.51\\
	& agriculteurs     &  8.58\\
	& long             &  8.56\\
	& toute            &  7.87\\
	& dans             &  7.34\\
	& organismes       &  7.28\\
	& o\`{u}          &  6.83\\
	& durant           &  6.55\\
	& moyens           &  5.98\\
	& tout             &  5.53
\end{tabular} \hspace{0.7in}
\begin{tabular}[t]{p{0.3in}p{1.1in}l}
\multicolumn{3}{l}{minimum (27)} \\
	& minimum          & 12.44\\
	& minimal          & 12.43\\
	& minimale         & 11.80\\
	& minimaux         & 11.42\\
	& minimums         & 11.19\\
	& minimales        & 11.03\\
	& \'{e}tudi\'{e}es &  8.95\\
	& moins            &  6.61\\
	& jusque           &  5.87\\
	& avec             &  4.87
\end{tabular}

\begin{tabular}[t]{p{0.3in}p{1.1in}l}
\multicolumn{3}{l}{enterprises (21)} \\
	& entreprises      & 10.21\\
	& poursuite        &  8.92\\
	& faciliter        &  8.87\\
	& importance       &  6.78\\
	& m\^{e}me         &  4.94\\
	& une              &  4.48
\end{tabular} \hspace{0.7in}
\begin{tabular}[t]{p{0.3in}p{1.1in}l}
\multicolumn{3}{l}{appear (35)} \\
	& compara\^{i}tre  &  9.66\\
	& semblent         &  9.36\\
	& semble           &  8.92\\
	& voulons          &  7.36\\
	& frais            &  6.73\\
	& -t-              &  5.26
\end{tabular}

\begin{tabular}[t]{p{0.3in}p{1.1in}l}
\multicolumn{3}{l}{delivery (17)} \\
	& livraison        & 11.75\\
	& livraisons       & 10.57\\
	& modifi\'{e}es   &  9.57\\
	& cependant        &  7.72\\
	& avant            &  7.28\\
	& ;                &  6.37
\end{tabular} \hspace{0.7in}
\begin{tabular}[t]{p{0.3in}p{1.1in}l}
\multicolumn{3}{l}{stocks (17)} \\
	& stocks           & 11.75\\
	& r\'{e}serves    &  9.54\\
	& bancs            &  9.51\\
	& valeurs          &  8.75\\
	& actions          &  8.19\\
	& exercer          &  7.78
\end{tabular}

\begin{tabular}[t]{p{0.3in}p{1.1in}l}
\multicolumn{3}{l}{steadily (14)} \\
	& constamment      &  8.21\\
	& cesse            &  8.18
\end{tabular} \hspace{0.7in}
\begin{tabular}[t]{p{0.3in}p{1.1in}l}
\multicolumn{3}{l}{combined (13)} \\
	& joints           &  9.91\\
	& deux             &  7.67
\end{tabular}

\begin{tabular}[t]{p{0.3in}p{1.1in}l}
\multicolumn{3}{l}{especially (25)} \\
	& surtout          & 11.20\\
	& particuli\`{e}rement & 10.99\\
	& sp\'{e}cialement & 10.18\\
	& particulier      &  9.55\\
	& notamment        &  9.43\\
	& pr\'{e}cis\'{e}ment &  7.65\\
	& assurer          &  6.17\\
	& qui              &  4.74
\end{tabular} \hspace{0.7in}
\begin{tabular}[t]{p{0.3in}p{1.1in}l}
\multicolumn{3}{l}{floor (30)} \\
	& plancher         & 11.61\\
	& parole           &  9.42\\
	& locataires       &  8.66\\
	& chambre          &  7.45\\
\multicolumn{3}{l}{losing (19)} \\
	& perdons          & 10.54\\
	& perd             &  9.92\\
	& perdre           &  9.43
\end{tabular}

\begin{tabular}[t]{p{0.3in}p{1.1in}l}
\multicolumn{3}{l}{new (134)} \\
	& new              & 11.19\\
	& nouveaux         & 11.15\\
	& nouvelles        & 10.74\\
	& nouvelle         & 10.70\\
	& nouveau          & 10.54\\
	& nouvel           & 10.30\\
	& d\'{e}mocrate   &  9.59\\
	& neuves           &  9.46\\
	& d\'{e}mocrates  &  9.17\\
	& neuf             &  8.64
\end{tabular} \hspace{0.7in}
\begin{tabular}[t]{p{0.3in}p{1.1in}l}
\multicolumn{3}{l}{manner (43)} \\
	& fa\c{c}on        &  8.59\\
	& mani\`{e}re     &  8.27\\
	& toucher          &  7.06\\
	& quoi             &  6.07\\
	& avaient          &  5.77\\
	& m.               &  5.76\\
	& avec             &  5.75\\
	& destin\'{e}es   &  5.70\\
	& -t-              &  5.43\\
	& aussi            &  4.99
\end{tabular}

\vskip\baselineskip
}

\section{Poor Correspondences}

In this section, we list some selected English words for which the
acquired lexical correspondences are not overly appropriate.

{
\parskip=0pt
\begin{tabular}[t]{p{0.3in}p{1.1in}l}
\multicolumn{3}{l}{the (19379)} \\
	& the              &  7.42\\
	& la               &  6.56\\
	& \`{a}           &  5.65\\
	& au               &  5.23\\
	& encourag\'{e}   &  5.17\\
	& cette            &  4.96\\
	& prescrit         &  4.93\\
	& profitable       &  4.92\\
	& parenchymes      &  4.90\\
	& tenu             &  4.86
\end{tabular} \hspace{0.7in}
\begin{tabular}[t]{p{0.3in}p{1.1in}l}
\multicolumn{3}{l}{at (2292)} \\
	& at               &  7.66\\
	& heures           &  6.89\\
	& entrepos\'{e}s  &  6.60\\
	& heure            &  6.39\\
	& conformit\'{e}  &  5.98\\
	& rapatri\'{e}    &  5.95\\
	& immobilis\'{e}s &  5.93\\
	& lors             &  5.89\\
	& voyant           &  5.89\\
	& respectifs       &  5.88
\end{tabular}

\begin{tabular}[t]{p{0.3in}p{1.1in}l}
\multicolumn{3}{l}{on (1577)} \\
	& commenter        &  6.96\\
	& au               &  6.84\\
	& devrai           &  6.70\\
	& visites          &  6.63\\
	& affreusement     &  6.43\\
	& soul\`{e}ve     &  6.15\\
	& attaquant        &  5.98\\
	& vinicole         &  5.97\\
	& victime          &  5.87\\
	& ensuite          &  5.84
\end{tabular} \hspace{0.7in}
\begin{tabular}[t]{p{0.3in}p{1.1in}l}
\multicolumn{3}{l}{of (23032)} \\
	& of               &  7.91\\
	& rappel           &  6.63\\
	& fermeture        &  5.56\\
	& historiques      &  5.51\\
	& des              &  5.12\\
	& demanderais      &  4.87\\
	& ordre            &  4.77\\
	& entendu          &  4.77\\
	& pr\'{e}sider    &  4.64\\
	& r\`{e}gne       &  4.63
\end{tabular}

}

%
%

\end{document}